\documentclass{article} 
\usepackage{iclr2017_conference,times}
\usepackage{amsmath}
\usepackage{booktabs}
\usepackage{hyperref}
\usepackage{url}
\usepackage{epsfig}
\usepackage{subcaption}
\usepackage{etoolbox}
\usepackage{xcolor}

\title{A Learned Representation for Artistic Style}

\author{%
Vincent Dumoulin \& Jonathon Shlens \& Manjunath Kudlur \\
Google Brain, Mountain View, CA \\
\texttt{vi.dumoulin@gmail.com, shlens@google.com, keveman@google.com}
}

\iclrfinalcopy 

\begin{document}

\maketitle

\begin{abstract}
The diversity of painting styles represents a rich visual vocabulary for the
construction of an image. The degree to which one may learn and parsimoniously
capture this visual vocabulary measures our understanding of the higher level
features of paintings, if not images in general. In this work we investigate the
construction of a single, scalable deep network that can parsimoniously capture
the artistic style of a diversity of paintings. We demonstrate that such
a network generalizes across a diversity of artistic styles by reducing a
painting to a point in an embedding space. Importantly, this model
permits a user to explore new painting styles by arbitrarily combining the
styles learned from individual paintings. We hope that this work provides a
useful step towards building rich models of paintings and offers a window on to
the structure of the learned representation of artistic style.
\end{abstract}

\section{Introduction}

A pastiche is an artistic work that imitates the style of another one. Computer
vision and more recently machine learning have a history of trying to automate
pastiche, that is, render an image in the style of another one. This task is
called {\em style transfer}, and is closely related to the texture synthesis
task. While the latter tries to capture the statistical relationship between the
pixels of a source image which is assumed to have a stationary distribution at
some scale, the former does so while also attempting to preserve some notion of
content.

On the computer vision side, \cite{efros1999texture} and \cite{wei2000fast}
attempt to ``grow'' textures one pixel at a time using non-parametric sampling
of pixels in an examplar image. \cite{efros2001image} and \cite{liang2001real}
extend this idea to ``growing'' textures one patch at a time, and
\cite{efros2001image} uses the approach to implement ``texture transfer'', i.e.
transfering the texture of an object onto another one. \cite{kwatra2005texture}
approaches the texture synthesis problem from an energy minimization
perspective, progressively refining the texture using an EM-like algorithm.
\cite{hertzmann2001image} introduces the concept of ``image analogies'': given a
pair of ``unfiltered'' and ``filtered'' versions of an examplar image, a target
image is processed to create an analogous ``filtered'' result. More recently,
\cite{frigo2016split} treats style transfer as a local texture transfer (using
an adaptive patch partition) followed by a global color transfer, and
\cite{elad2016style} extends Kwatra's energy-based method into a style transfer
algorithm by taking content similarity into account.

On the machine learning side, it has been shown that a trained classifier can
be used as a feature extractor to drive texture synthesis and style transfer.
\cite{gatys2015texture} uses the VGG-19 network \citep{simonyan2014very} to
extract features from a texture image and a synthesized texture. The two sets of
features are compared and the synthesized texture is modified by gradient
descent so that the two sets of features are as close as possible.
\cite{gatys2015neural} extends this idea to style transfer by adding the
constraint that the synthesized image also be close to a content image with
respect to another set of features extracted by the trained VGG-19 classifier.

\begin{figure}[t]
    \begin{subfigure}[t]{\linewidth}
    \begin{center}
        \includegraphics[width=0.15\linewidth]{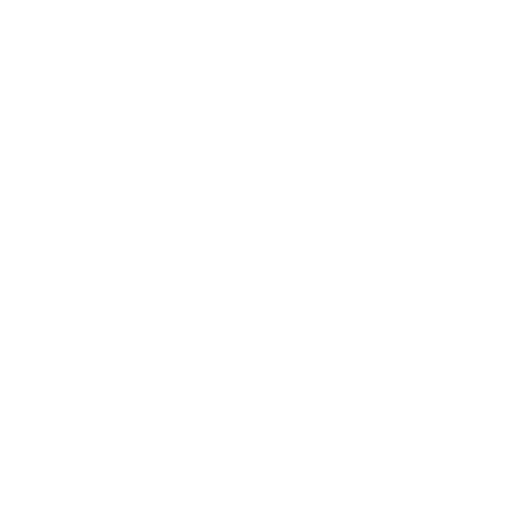}
        \includegraphics[width=0.15\linewidth]{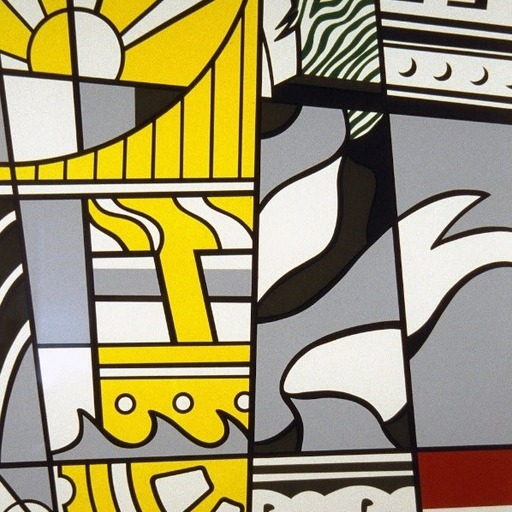}
        \includegraphics[width=0.15\linewidth]{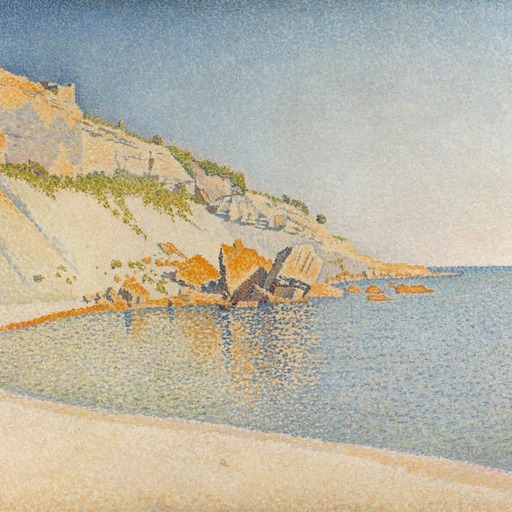}
        \includegraphics[width=0.15\linewidth]{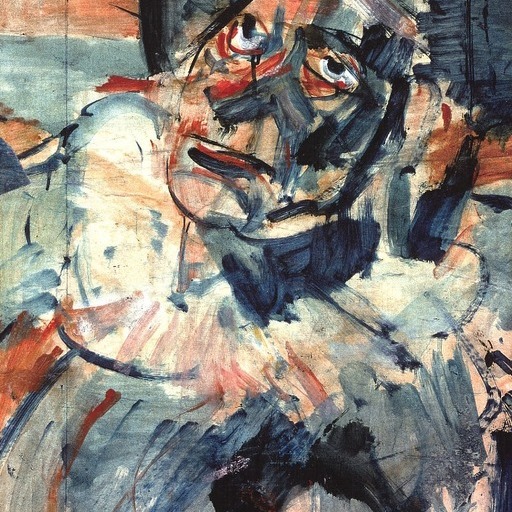}
        \includegraphics[width=0.15\linewidth]{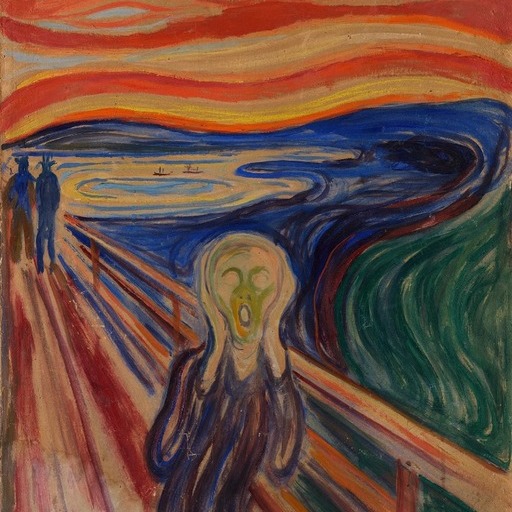}
        \includegraphics[width=0.15\linewidth]{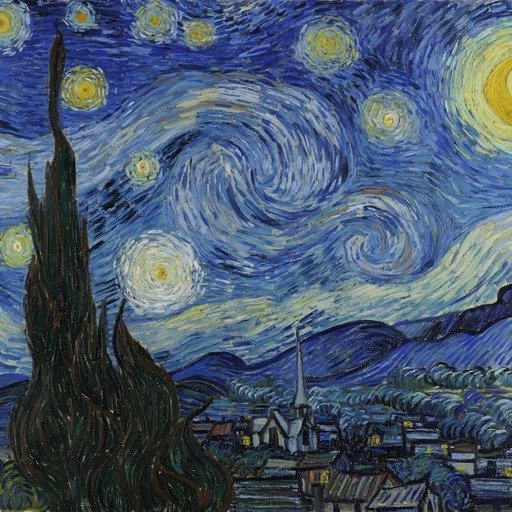} \\
        \includegraphics[width=0.15\linewidth]{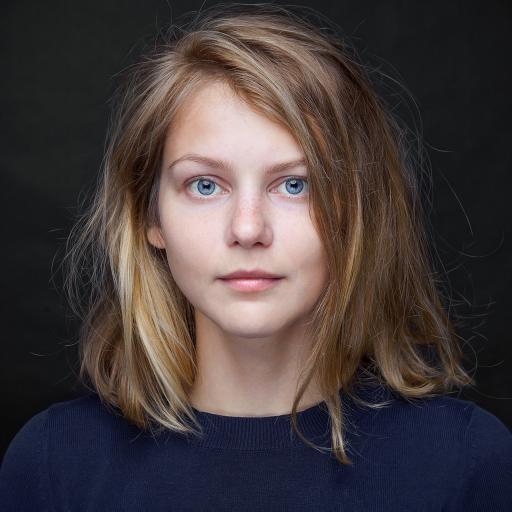}
        \includegraphics[width=0.15\linewidth]{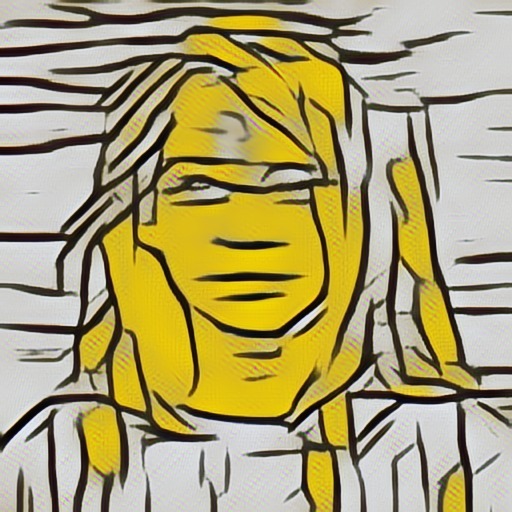}
        \includegraphics[width=0.15\linewidth]{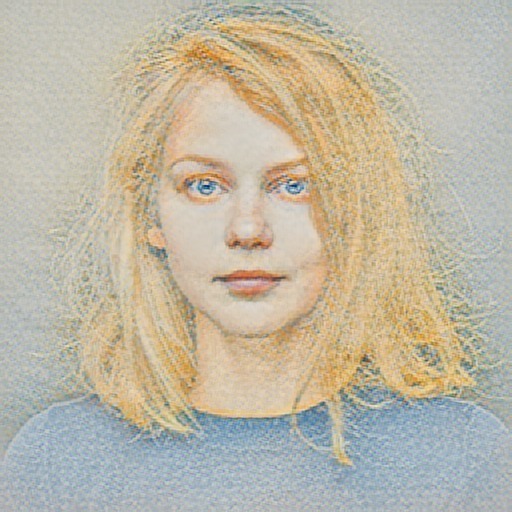}
        \includegraphics[width=0.15\linewidth]{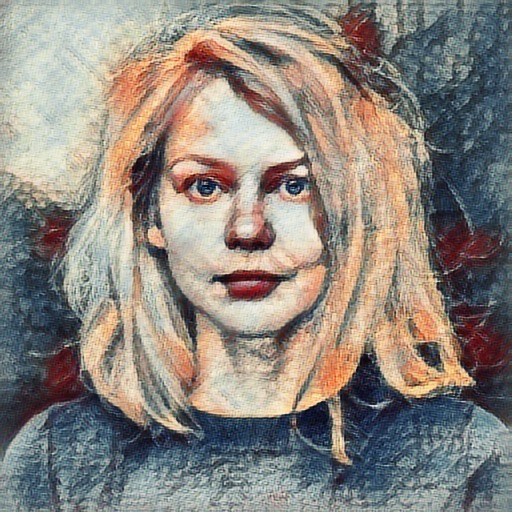}
        \includegraphics[width=0.15\linewidth]{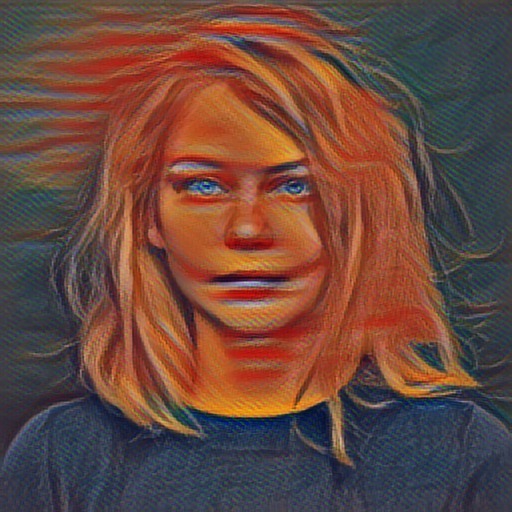}
        \includegraphics[width=0.15\linewidth]{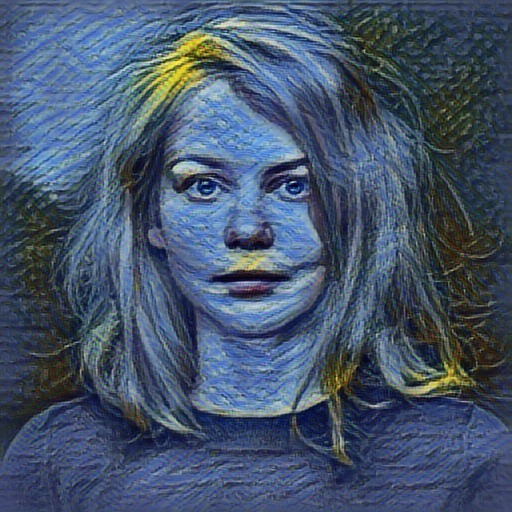} \\
        \includegraphics[width=0.15\linewidth]{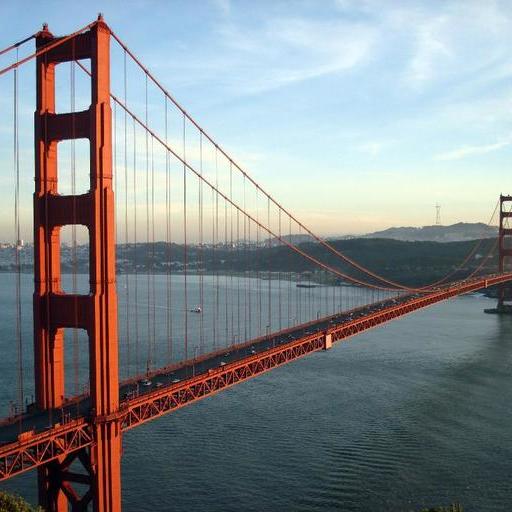}
        \includegraphics[width=0.15\linewidth]{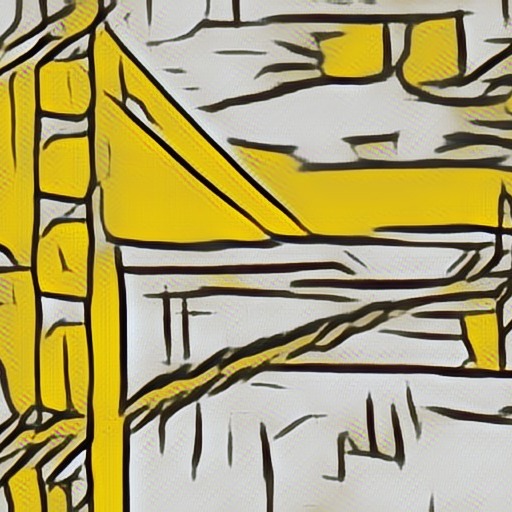}
        \includegraphics[width=0.15\linewidth]{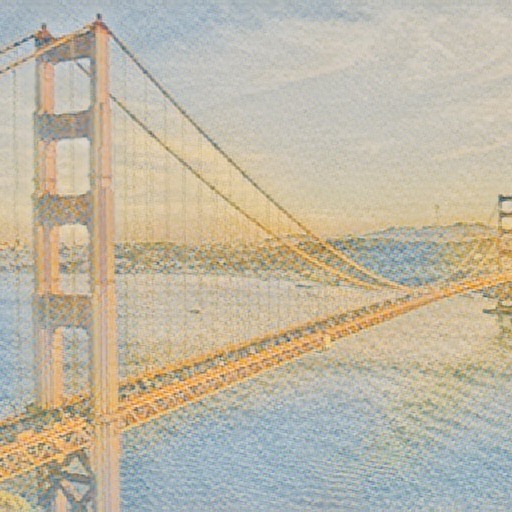}
        \includegraphics[width=0.15\linewidth]{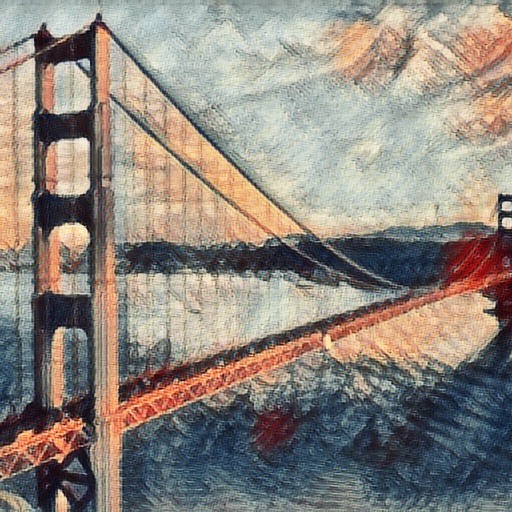}
        \includegraphics[width=0.15\linewidth]{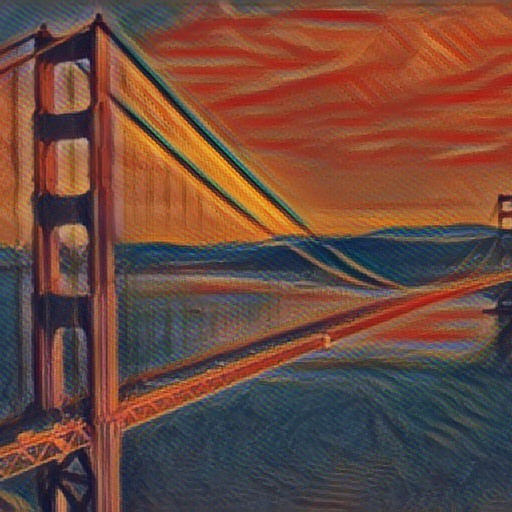}
        \includegraphics[width=0.15\linewidth]{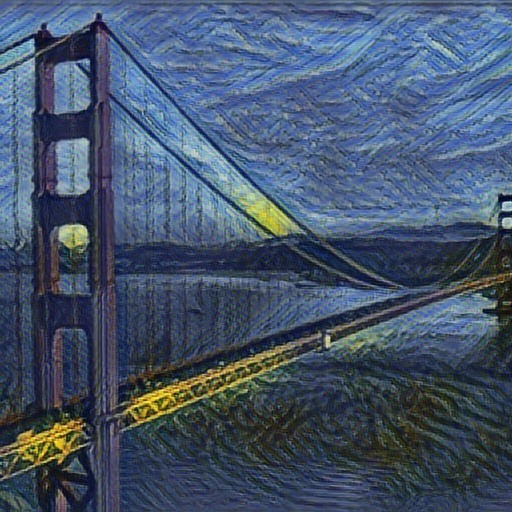} \\
        \includegraphics[width=0.15\linewidth]{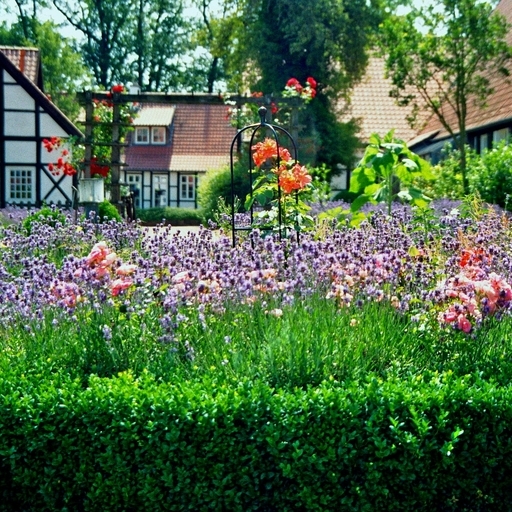}
        \includegraphics[width=0.15\linewidth]{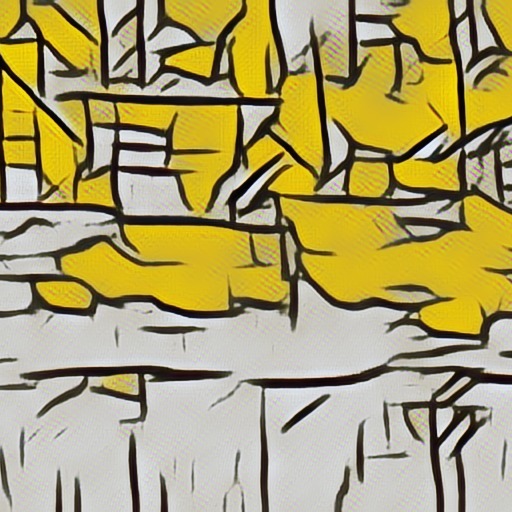}
        \includegraphics[width=0.15\linewidth]{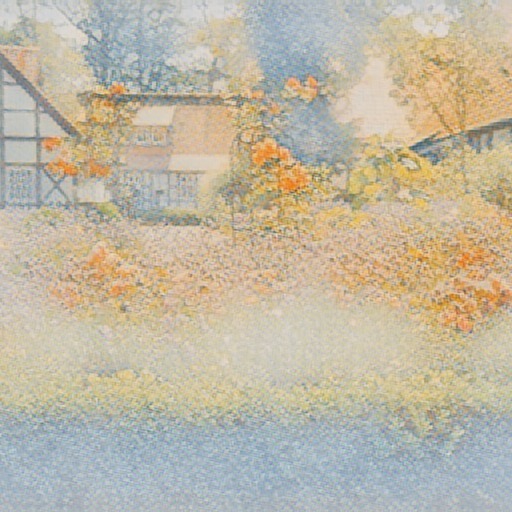}
        \includegraphics[width=0.15\linewidth]{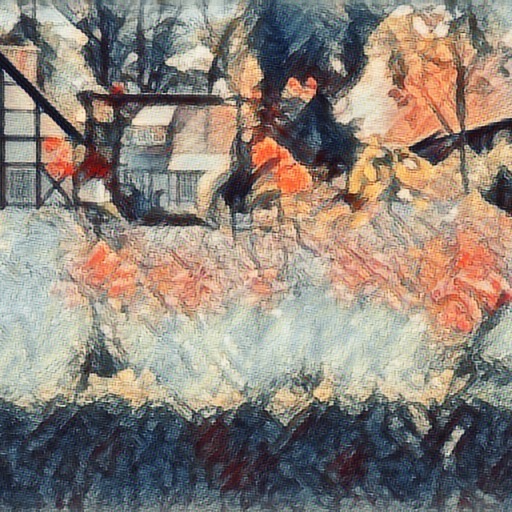}
        \includegraphics[width=0.15\linewidth]{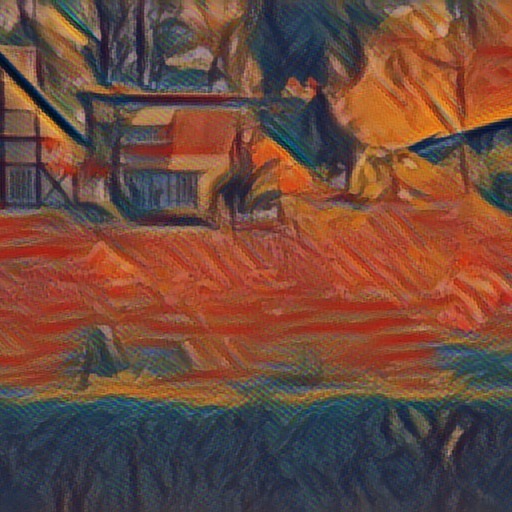}
        \includegraphics[width=0.15\linewidth]{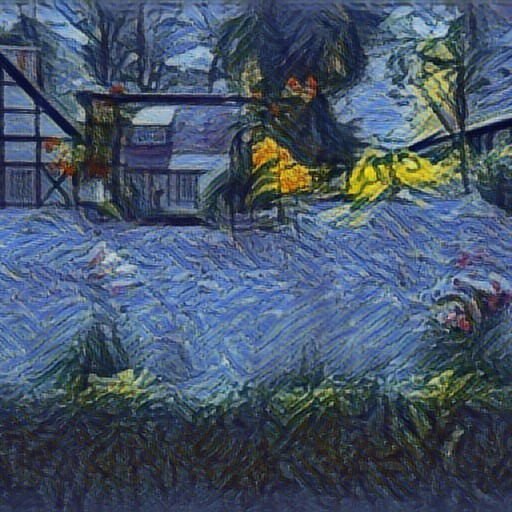}
        \caption{\label{fig:very_varied_partial_stylizations} With conditional
            instance normalization, a single style transfer network can capture
            32 styles at the same time, five of which are shown here. All 32
            styles in this single model are in the Appendix. Golden Gate Bridge
            photograph by Rich Niewiroski Jr.}
    \end{center}
    \end{subfigure}\\
    \begin{subfigure}[t]{\linewidth}
    \vspace{0.2cm}
    \begin{center}
        \includegraphics[width=0.15\linewidth]{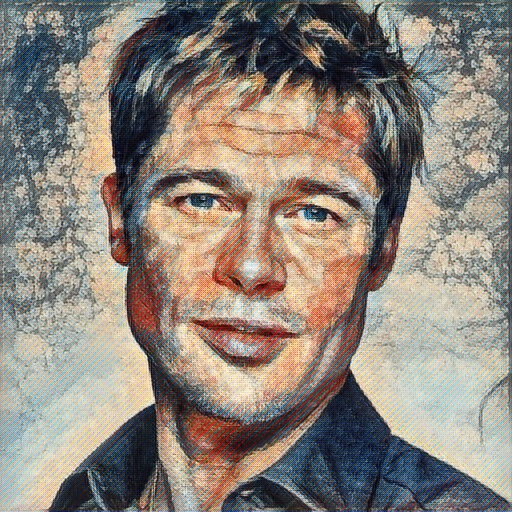}
        \includegraphics[width=0.15\linewidth]{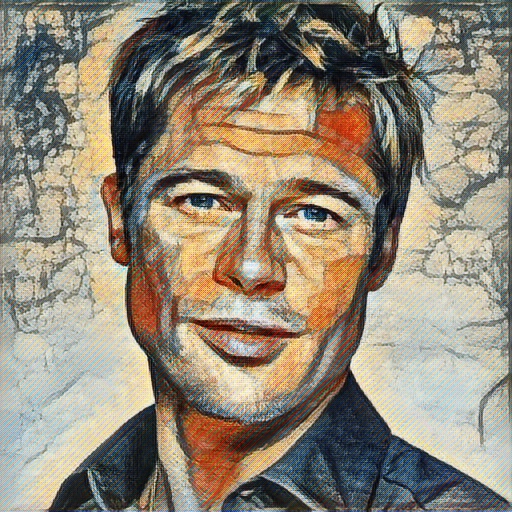}
        \includegraphics[width=0.15\linewidth]{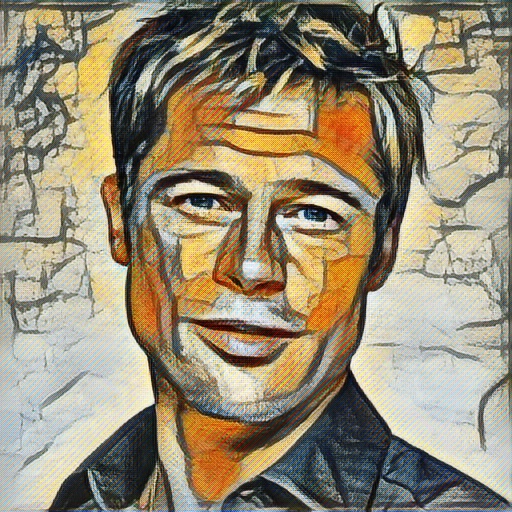}
        \includegraphics[width=0.15\linewidth]{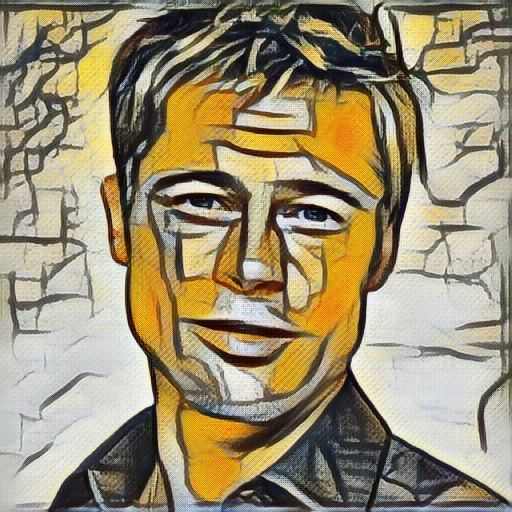}
        \includegraphics[width=0.15\linewidth]{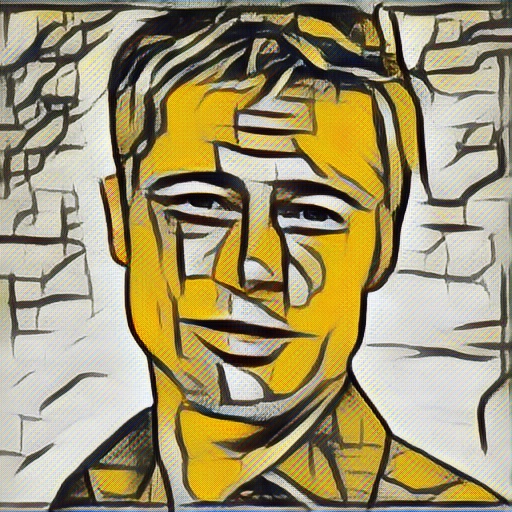}
        \includegraphics[width=0.15\linewidth]{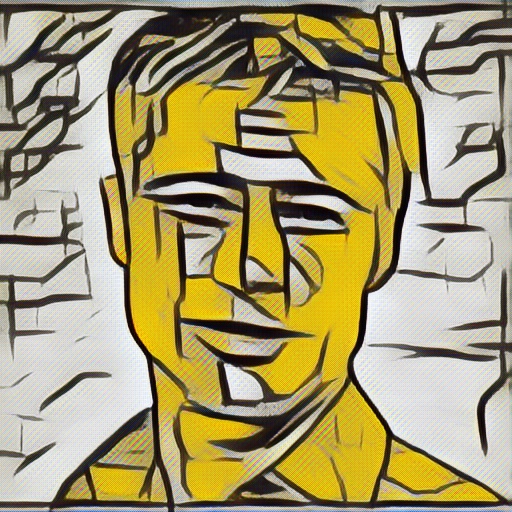}
        \caption{\label{fig:very_varied_interpolations} The style representation
            learned via conditional instance normalization permits the
            arbitrary combination of
            artistic styles. Each pastiche in the sequence corresponds
            to a different step in interpolating between the $\gamma$ and
            $\beta$ values associated with two styles the model was trained on.}
    \end{center}
    \end{subfigure}
    \caption{\label{fig:very_varied} Pastiches produced by a style transfer
        network trained on 32 styles chosen for their variety.}
\end{figure}

While very flexible, this algorithm is expensive to run due to the optimization
loop being carried. \cite{ulyanov2016texture}, \cite{liwand2016}
and \cite{johnson2016perceptual}
tackle this problem by introducing a {\em feedforward style transfer network},
which is trained to go from content to pastiche image in one pass. However, in
doing so some of the flexibility of the original algorithm is lost: the style
transfer network is tied to a single style, which means that separate networks
have to be trained for every style being modeled. Subsequent work has brought
some performance improvements to style transfer networks, e.g. with respect to
color preservation \citep{gatys2016preserving} or style transfer quality
\citep{ulyanov2016instance}, but to our knowledge the problem of the
single-purpose nature of style transfer networks remains untackled.

We think this is an important problem that, if solved, would have both
scientific and practical importance. First, style transfer has already found
use in mobile applications,
for which on-device processing is contingent upon the models having a reasonable
memory footprint.
More broadly, building a separate network for each style ignores the fact that
individual paintings share many common visual elements and a true model that
captures
artistic style would be able to exploit and learn from such regularities.
Furthermore, the degree to which an artistic styling model might generalize
across painting styles would directly measure our ability to build systems that
parsimoniously capture the higher level features and statistics of photographs
and images~\citep{simoncelli2001review}.

In this work, we show that a simple modification of the style transfer network,
namely the introduction of {\em conditional instance normalization}, allows it
to learn multiple styles (\autoref{fig:very_varied_partial_stylizations}).We
demonstrate that this approach is flexible yet comparable to single-purpose
style transfer networks, both qualitatively and in terms of convergence
properties.  This model reduces each style image into a point in an embedding
space. Furthermore, this model provides a generic representation for artistic
styles that seems flexible enough to capture new artistic styles much faster
than a single-purpose network. Finally, we show that the embeddding space
representation permits one to arbitrarily combine artistic styles in novel ways
not previously observed (\autoref{fig:very_varied_interpolations}). 

\begin{figure}[t]
\begin{center}
\includegraphics[width=0.9\linewidth]{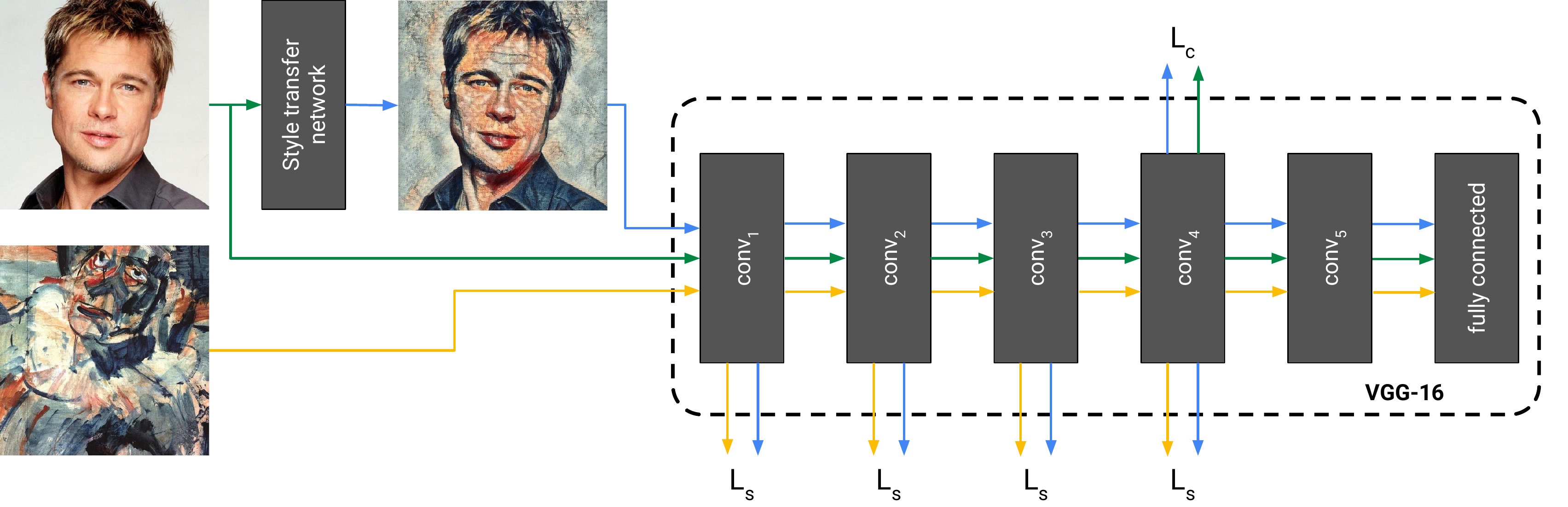}
\end{center}
\caption{\label{fig:style_transfer_network} Style transfer network training
    diagram \citep{johnson2016perceptual, ulyanov2016texture}.
    A pastiche image is produced by feeding a content image through the
    style transfer network. The two images, along with a style image, are passed
    through a trained classifier, and the resulting intermediate representations
    are used to compute the content loss $\mathcal{L}_c$ and style loss
    $\mathcal{L}_s$. The parameters of the classifier are kept fixed throughout
    training.}
\end{figure}

\section{Style transfer with deep networks}

Style transfer can be defined as finding a pastiche image $p$ whose content is
similar to that of a content image $c$ but whose style is similar to that of a
style image $s$. This objective is by nature vaguely defined, because similarity
in content and style are themselves vaguely defined.

The neural algorithm of artistic style proposes the following definitions:
\begin{itemize}
    \item Two images are similar in content if their high-level features as
        extracted by a trained classifier are close in Euclidian distance.
    \item Two images are similar in style if their low-level features as
        extracted by a trained classifier share the same statistics or, more
        concretely, if the difference between the features' Gram matrices has a
        small Frobenius norm.
\end{itemize}

The first point is motivated by the empirical observation that high-level
features in classifiers tend to correspond to higher levels of abstractions (see
\cite{zeiler2014visualizing} for visualizations; see
\cite{johnson2016perceptual} for style transfer features). The second point is
motivated by the observation that the artistic style of a painting may be
interpreted as a visual texture~\citep{gatys2015texture}. A visual texture is
conjectured to be spatially homogenous and consist of repeated structural motifs
whose minimal sufficient statistics are captured by lower order statistical
measurements ~\citep{julesz1962,portilla1999}.

In its original formulation, the neural algorithm of artistic style proceeds as
follows: starting from some initialization of $p$ (e.g. $c$, or some random
initialization), the algorithm adapts $p$ to minimize the loss function
\begin{equation}
    \mathcal{L}(s, c, p) = \lambda_s\mathcal{L}_s(p) + \lambda_c\mathcal{L}_c(p),
\end{equation}
where $\mathcal{L}_s(p)$ is the style loss, $\mathcal{L}_c(p)$ is the content
loss and $\lambda_s, \lambda_c$ are scaling hyperparameters. Given a set of
``style layers'' $\mathcal{S}$ and a set of ``content layers'' $\mathcal{C}$,
the style and content losses are themselves defined as
\begin{equation}
    \mathcal{L}_s(p) = \sum_{i \in \mathcal{S}}
        \frac{1}{U_i} \mid\mid G(\phi_i(p)) - G(\phi_i(s)) \mid\mid_F^2
\end{equation}
\begin{equation}
    \mathcal{L}_c(p) = \sum_{j \in \mathcal{C}}
        \frac{1}{U_j} \mid\mid \phi_j(p) - \phi_j(c) \mid\mid_2^2
\end{equation}
where $\phi_l(x)$ are the classifier activations at layer $l$, $U_l$ is the
total number of units at layer $l$ and $G(\phi_l(x))$ is the Gram matrix
associated with the layer $l$ activations. In practice, we set
$\lambda_c = 1.0$ and and leave $\lambda_s$ as a free hyper-parameter.

In order to speed up the procedure outlined above, a feed-forward
convolutional network, termed a style transfer network $T$, is introduced
to learn the transformation
\citep{johnson2016perceptual, liwand2016, ulyanov2016texture}.
It takes as input a content image $c$ and outputs the pastiche image $p$
directly (Figure \ref{fig:style_transfer_network}). The network is
trained on many content images \citep{imagenet} using the same loss
function as above, i.e.
\begin{equation}
\mathcal{L}(s, c) = \lambda_s\mathcal{L}_s(T(c)) + \lambda_c\mathcal{L}_c(T(c)).
\end{equation}

While feedforward style transfer networks solve the problem of speed at
test-time, they also suffer from the fact that the network $T$ is tied
to one specific painting style.
This means that a separate network $T$ has to be trained for {\em every}
style to be imitated. The real-world impact of this limitation is that it
becomes prohibitive to implement a style transfer application on a
memory-limited device, such as a smartphone.

\subsection{N-styles feedforward style transfer networks}

Our work stems from the intuition that many styles probably share some degree of
computation, and that this sharing is thrown away by training $N$ networks from
scratch when building an $N$-styles style transfer system. For instance, many
impressionist paintings share similar paint strokes but differ in the color
palette being used. In that case, it seems very wasteful to treat a set of $N$
impressionist paintings as completely separate styles.

To take this into account, we propose to train a single conditional style
transfer network $T(c, s)$ for $N$ styles. The conditional network is given both
a content image and the identity of the style to apply and produces a pastiche
corresponding to that style. While the idea is straightforward on paper, there
remains the open question of how conditioning should be done. In exploring this
question, we found a very surprising fact about the role of normalization in
style transfer networks: to model a style, it is sufficient to specialize
scaling and shifting parameters after normalization to each specific style. In
other words, all convolutional weights of a style transfer network can be shared
across many styles, and it is sufficient to tune parameters for an affine
transformation after normalization for each style.

\begin{figure}[t]
\begin{center}
\includegraphics[width=0.6\linewidth]{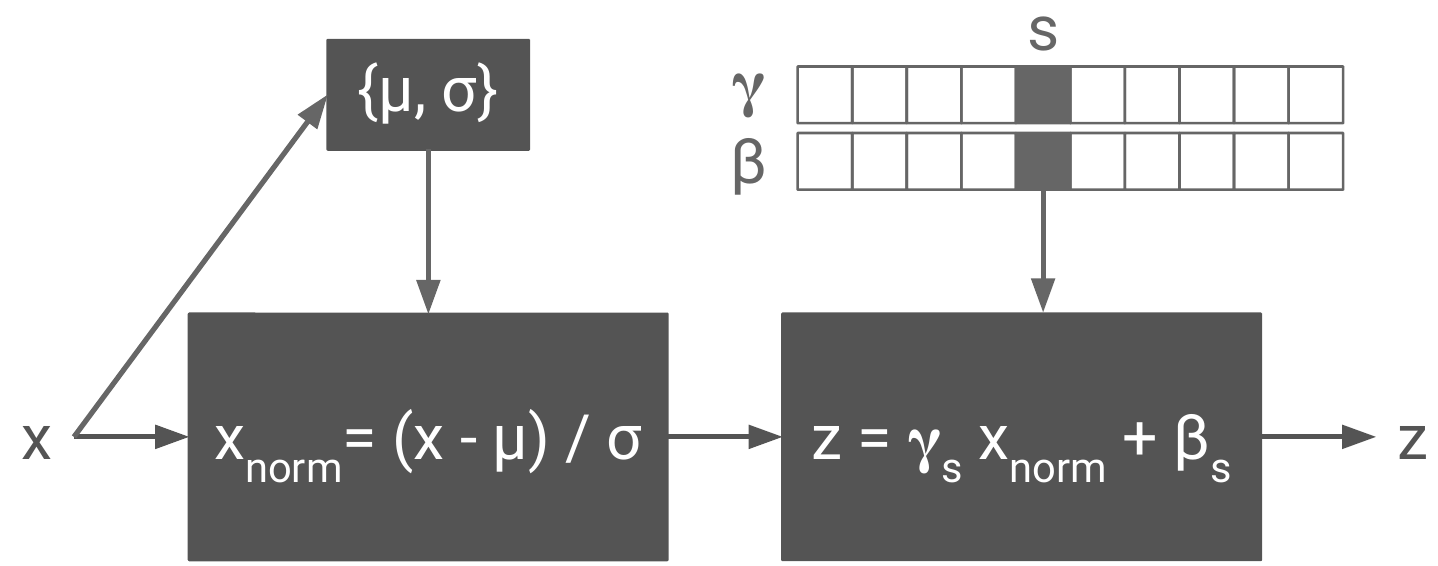}
\end{center}
\caption{\label{fig:conditional_instance_norm} Conditional instance
    normalization. The input activation $x$ is normalized across both 
    spatial dimensions and subsequently 
    scaled and shifted using style-dependent parameter vectors $\gamma_s, \beta_s$
    where $s$ indexes the style label.}
\end{figure}

We call this approach {\em conditional instance normalization}.
The goal of the procedure is transform a layer's activations $x$ into
a normalized activation $z$ specific to painting style $s$.
Building off the
instance normalization technique proposed in \cite{ulyanov2016instance}, we
augment the $\gamma$ and $\beta$ parameters so that they're $N \times C$
matrices, where $N$ is the number of styles being modeled and $C$ is the number
of output feature maps. Conditioning on a style is achieved as follows:
\begin{equation}
    z = \gamma_s \left (\frac{x - \mu}{\sigma} \right) + \beta_s
\end{equation}
where $\mu$ and $\sigma$ are $x$'s mean and standard deviation taken across
spatial axes and $\gamma_s$ and $\beta_s$ are obtained by selecting the row
corresponding to $s$ in the $\gamma$ and $\beta$ matrices
(\autoref{fig:conditional_instance_norm}). One added benefit of this approach
is that one can stylize a single image into $N$ painting styles with a single
feed forward pass of the network with a batch size of $N$.
In constrast, a single-style network requires
$N$ feed forward passes to perform $N$ style
transfers~\citep{johnson2016perceptual,liwand2016,ulyanov2016texture}.

Because conditional instance normalization only acts on the scaling and shifting
parameters, training a style transfer network on $N$ styles requires fewer
parameters than the naive approach of training $N$ separate networks. In a
typical network setup, the model consists of roughly 1.6M parameters, only
around 3K (or $0.2\%$) of which specify individual artistic styles. In fact,
because the size of $\gamma$ and $\beta$ grows linearly with respect to the
number of feature maps in the network, this approach requires $O(N \times L)$
parameters, where $L$ is the total number of feature maps in the network.

In addition, as is discussed in \autoref{sec:n_plus_1}, conditional instance
normalization presents the advantage that integrating an $N+1^{th}$ style to the
network is cheap because of the very small number of parameters to train.

\section{Experimental results}

\subsection{Methodology}

Unless noted otherwise, all style transfer networks were trained using the
hyperparameters outlined in the Appendix's \autoref{tab:architecture}.

We used the same network architecture as in \cite{johnson2016perceptual}, except
for two key details: zero-padding is replaced with mirror-padding, and
transposed convolutions (also sometimes called {\em deconvolutions}) are
replaced with nearest-neighbor upsampling followed by a convolution. The use of
mirror-padding avoids border patterns sometimes caused by zero-padding in
SAME-padded convolutions, while the replacement for transposed convolutions
avoids checkerboard patterning, as discussed in in \cite{odena2016avoiding}. We
find that with these two improvements training the network no longer requires a
total variation loss that was previously employed to remove high frequency noise
as proposed in ~\cite{johnson2016perceptual}.

Our training procedure follows \cite{johnson2016perceptual}. Briefly, we employ
the ImageNet dataset \citep{imagenet} as a corpus of training content images.
We train the $N$-style network with stochastic gradient descent
using the Adam optimizer \citep{kingma2014adam}. Details of the model
architecture are in the Appendix.
A complete implementation of the model in TensorFlow \citep{tensorflow}
as well as a pretrained model are available for
download \footnote{\texttt{https://github.com/tensorflow/magenta}}.
The evaluation images used for this work were resized such that their smaller
side has size 512. Their stylized versions were then center-cropped to
512x512 pixels for display.

\subsection{Training a single network on N styles produces stylizations comparable to independently-trained models}

\begin{figure}[t]
\begin{center}
    \includegraphics[width=0.15\linewidth]{figures/blank.jpg}
    \includegraphics[width=0.15\linewidth]{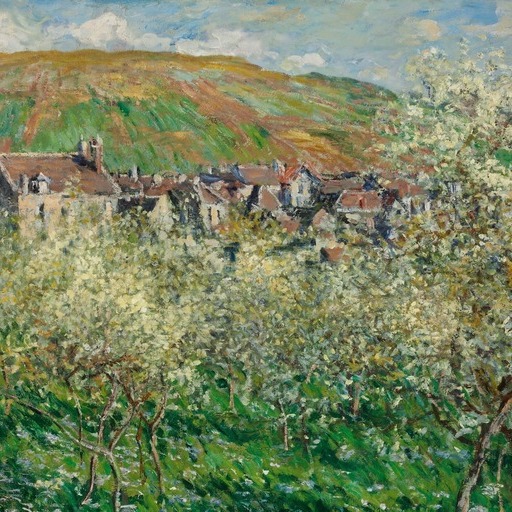}
    \includegraphics[width=0.15\linewidth]{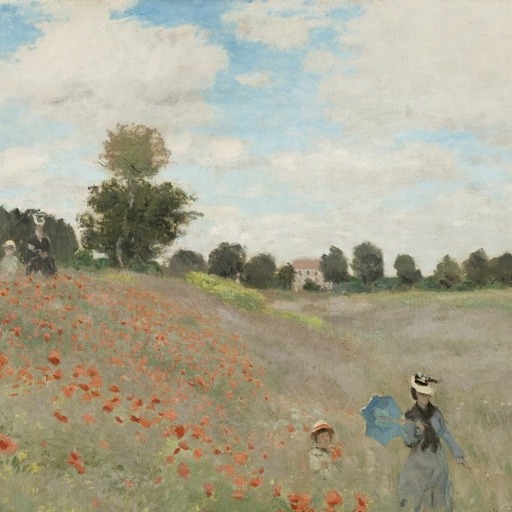}
    \includegraphics[width=0.15\linewidth]{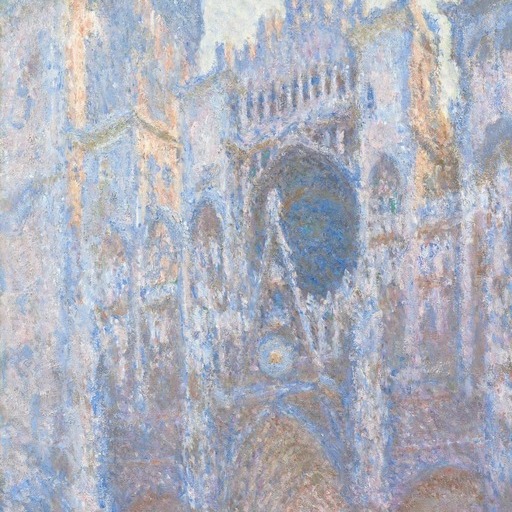}
    \includegraphics[width=0.15\linewidth]{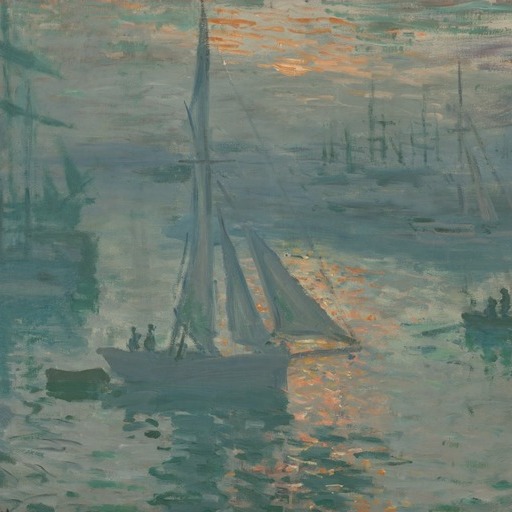}
    \includegraphics[width=0.15\linewidth]{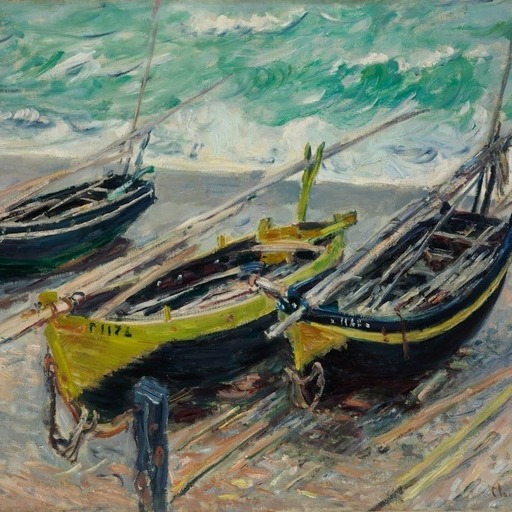} \\
    \includegraphics[width=0.15\linewidth]{figures/karya.jpg}
    \includegraphics[width=0.15\linewidth]{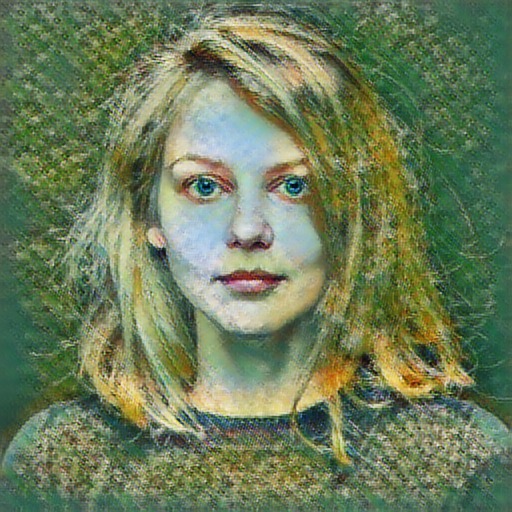}
    \includegraphics[width=0.15\linewidth]{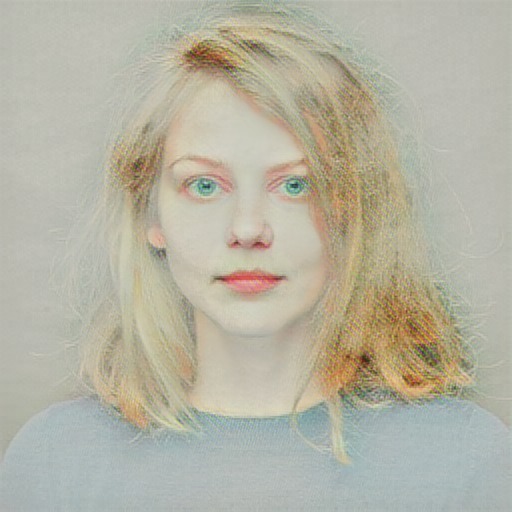}
    \includegraphics[width=0.15\linewidth]{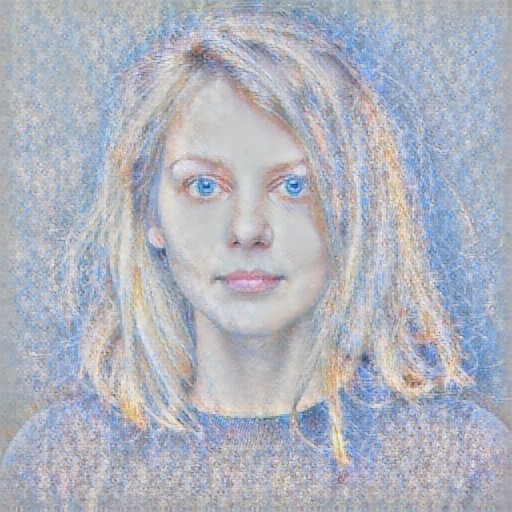}
    \includegraphics[width=0.15\linewidth]{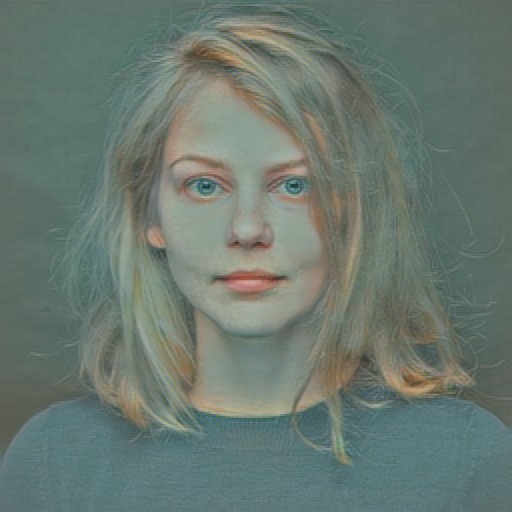}
    \includegraphics[width=0.15\linewidth]{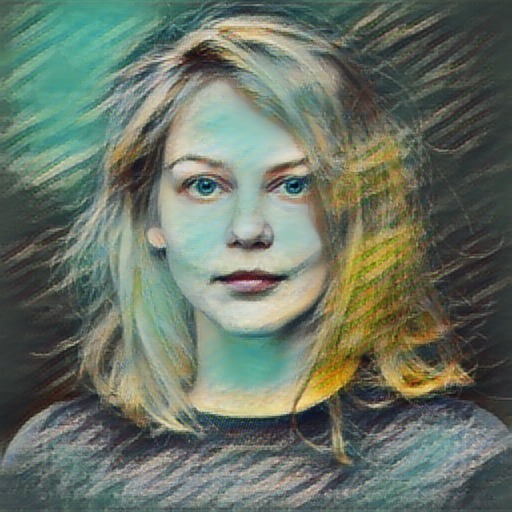} \\
    \includegraphics[width=0.15\linewidth]{figures/golden_gate.jpg}
    \includegraphics[width=0.15\linewidth]{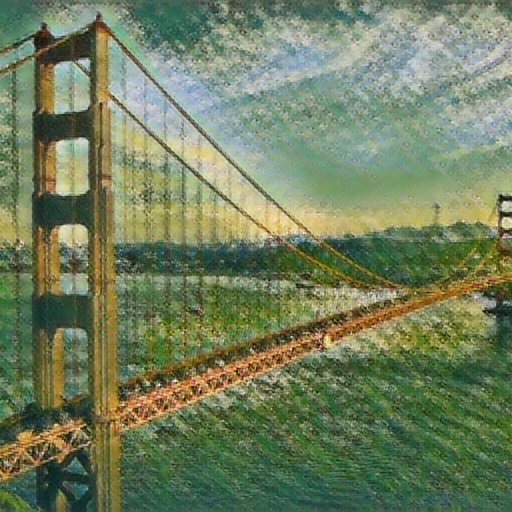}
    \includegraphics[width=0.15\linewidth]{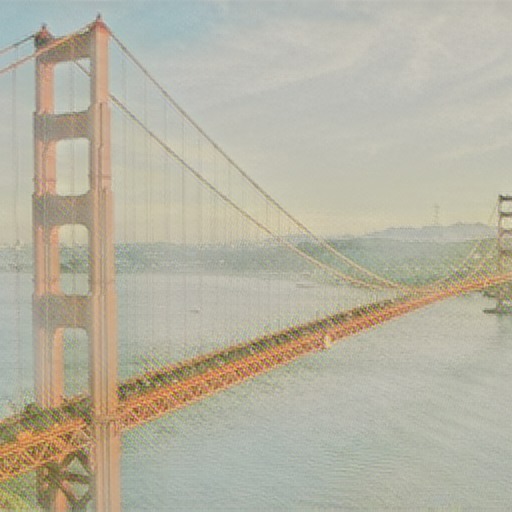}
    \includegraphics[width=0.15\linewidth]{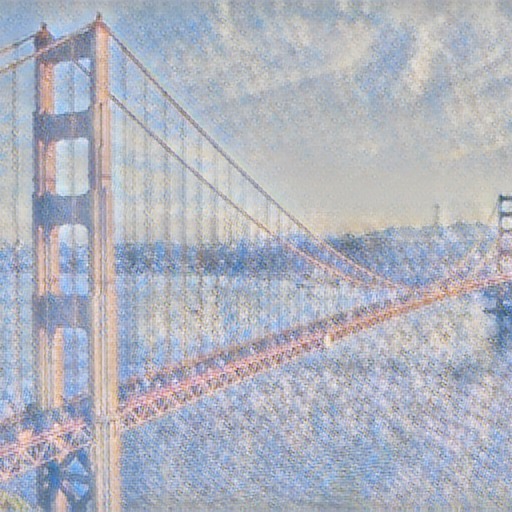}
    \includegraphics[width=0.15\linewidth]{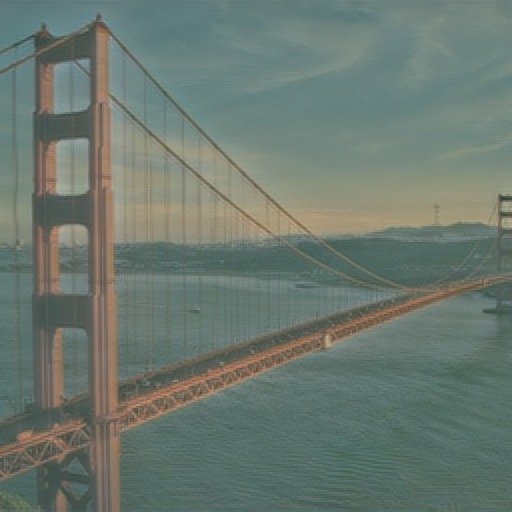}
    \includegraphics[width=0.15\linewidth]{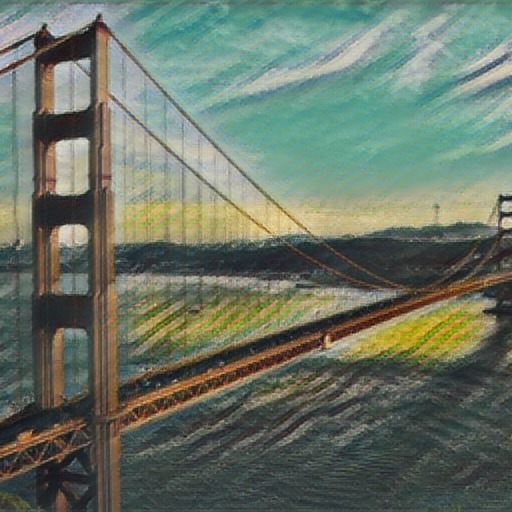} \\
    \includegraphics[width=0.15\linewidth]{figures/schultenhof_mettingen_bauerngarten.jpg}
    \includegraphics[width=0.15\linewidth]{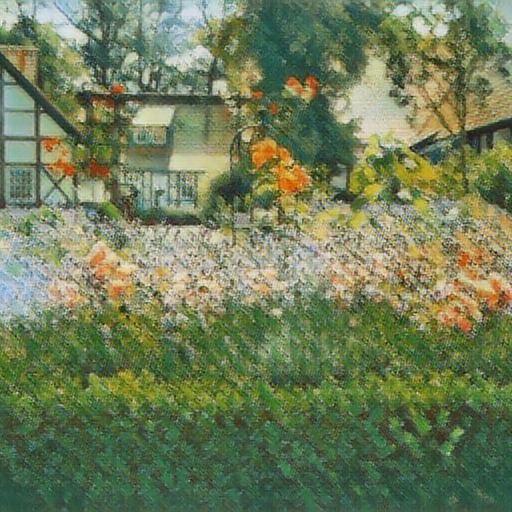}
    \includegraphics[width=0.15\linewidth]{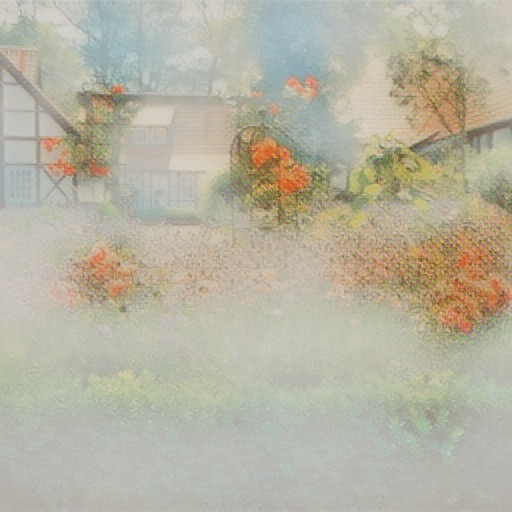}
    \includegraphics[width=0.15\linewidth]{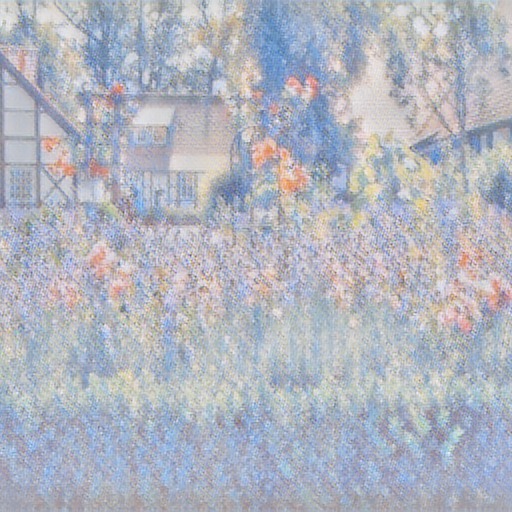}
    \includegraphics[width=0.15\linewidth]{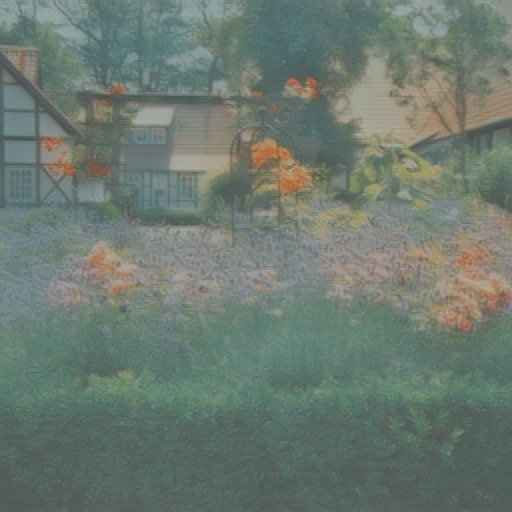}
    \includegraphics[width=0.15\linewidth]{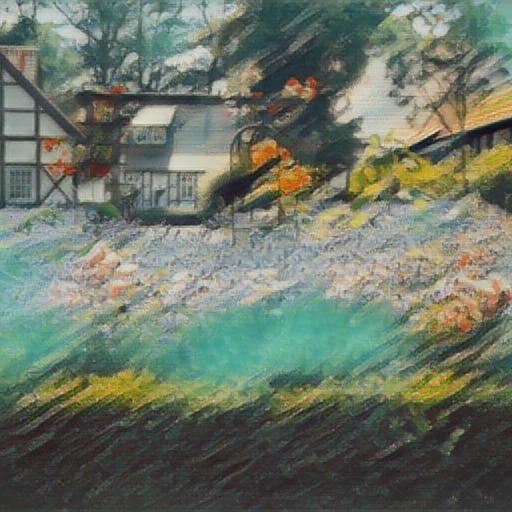}
\end{center}
\caption{\label{fig:monet_partial_stylizations} A single style transfer network 
    was trained to capture the style of 10 Monet paintings, five of which are
    shown here. All 10 styles in this single model are in the Appendix. Golden
    Gate Bridge photograph by Rich Niewiroski Jr.}
\end{figure}

As a first test, we trained a 10-styles model on stylistically similar images,
namely 10 impressionist paintings from Claude Monet.
\autoref{fig:monet_partial_stylizations} shows the result of applying the
trained network on evaluation images for a subset of the styles, with the full
results being displayed in the Appendix. The model captures different color
palettes and textures.  We emphasize that $99.8\%$ of the parameters are shared
across all styles in contrast to $0.2\%$ of the parameters which are unique to
each painting style.

\begin{figure}[t]
    \begin{minipage}{0.71\linewidth}
    \begin{center}
        \includegraphics[width=\linewidth]{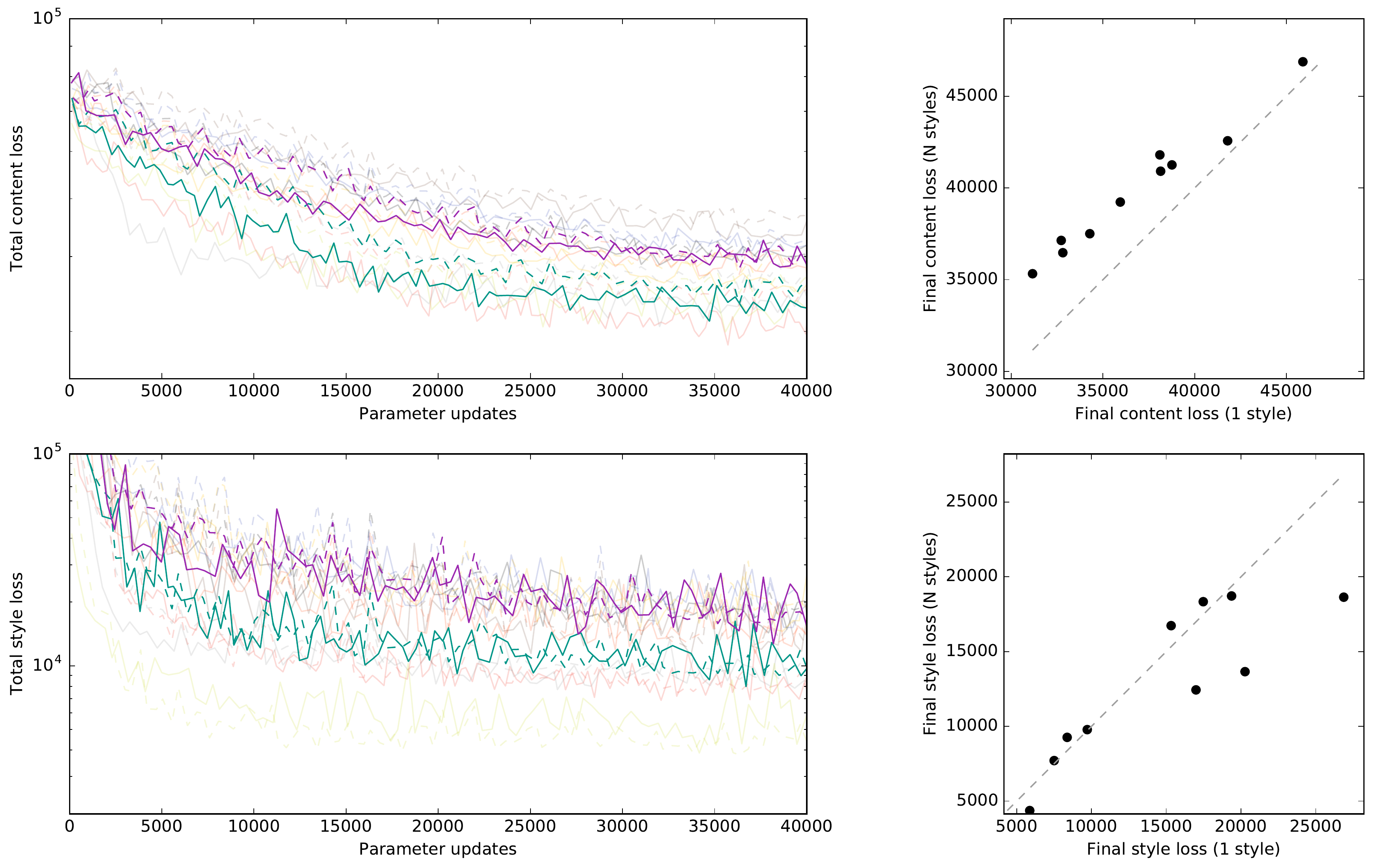}
    \end{center}
    \end{minipage}%
    \hspace{0.1cm}
    \begin{minipage}{0.24\linewidth}
    \begin{subfigure}[t]{0.49\linewidth}
    \begin{center}
        \includegraphics[width=0.8\linewidth]{figures/karya_plum_trees.jpg}
        \includegraphics[width=0.8\linewidth]{figures/karya_poppy_field.jpg}
        \includegraphics[width=0.8\linewidth]{figures/karya_rouen_cathedral.jpg}
        \includegraphics[width=0.8\linewidth]{figures/karya_sunrise_marine.jpg}
        \includegraphics[width=0.8\linewidth]{figures/karya_three_fishing_boats.jpg}
    \caption*{$N$ styles}
    \end{center}
    \end{subfigure}
    \begin{subfigure}[t]{0.49\linewidth}
    \begin{center}
        \includegraphics[width=0.8\linewidth]{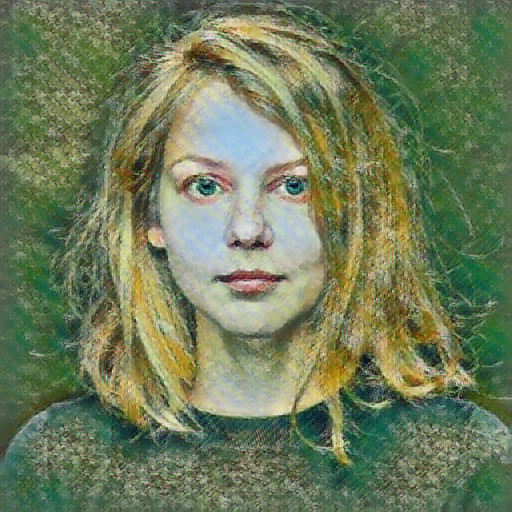}
        \includegraphics[width=0.8\linewidth]{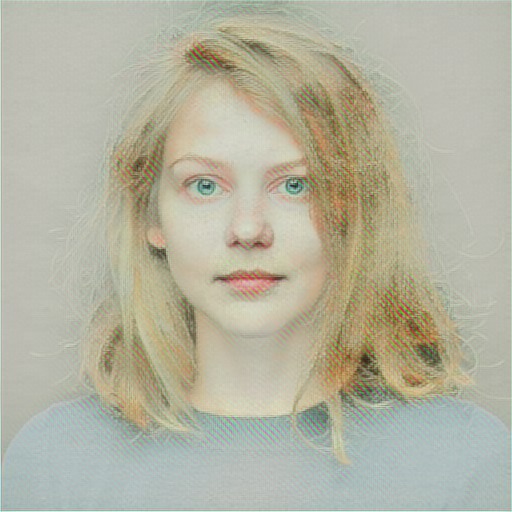}
        \includegraphics[width=0.8\linewidth]{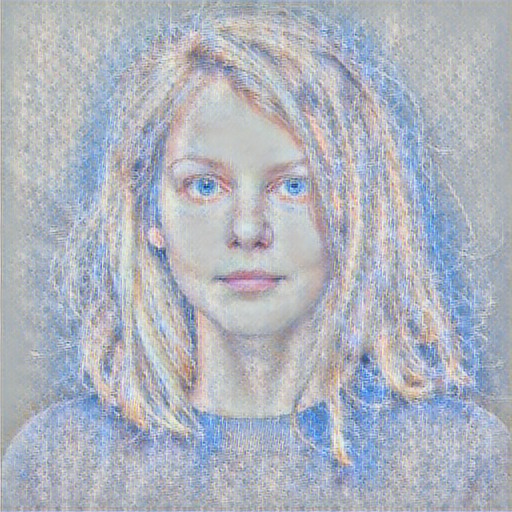}
        \includegraphics[width=0.8\linewidth]{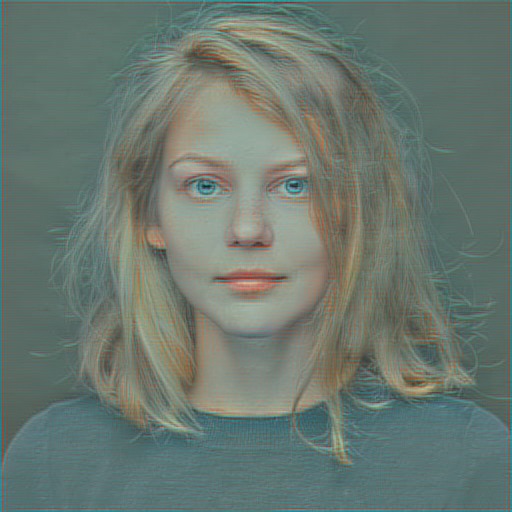}
        \includegraphics[width=0.8\linewidth]{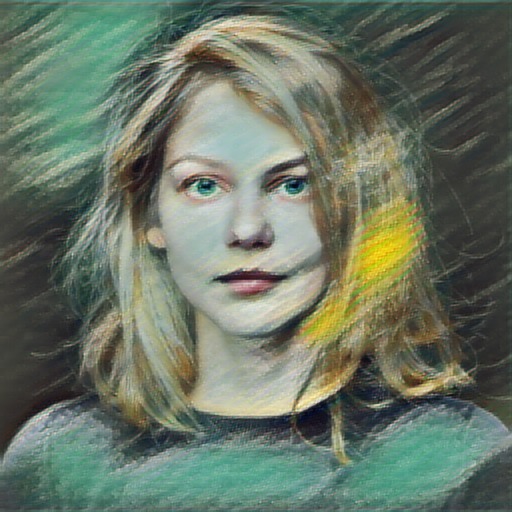}
    \caption*{1 style}
    \end{center}
    \end{subfigure}
    \end{minipage}
    \caption{\label{fig:monet_learning} The $N$-styles model exhibits
        learning dynamics comparable to individual models. (Left column) The
        N-styles model converges slightly slower in terms of content loss (top)
        and as fast in terms of style loss (bottom) than individual models.
        Training on a single Monet painting is represented by two curves
        with the same color.
        The dashed curve represents the $N$-styles model, and the
        full curves represent individual models.
        Emphasis has been added on the styles for
        \textit{Vetheuil (1902)} (teal) and \textit{Water Lilies} (purple) for
        visualization purposes; remaining colors correspond to other Monet
        paintings (see Appendix).  (Center column) The N-styles model reaches a
        slightly higher final content loss than (top, $8.7 \pm 3.9\%$ increase)
        and a final style loss comparable to (bottom, $8.9 \pm 16.5\%$ decrease)
        individual models. (Right column) Pastiches produced by the $N$-styles
        network are qualitatively comparable to those produced by individual
        networks.}
\end{figure}

To get a sense of what is being traded off by folding 10 styles into a single
network, we trained a separate, single-style network on each style and compared
them to the 10-styles network in terms of style transfer quality and training
speed (\autoref{fig:monet_learning}).

The left column compares the learning curves for style and content losses
between the single-style networks and the 10-styles network. The losses were
averaged over 32 random batches of content images. By visual inspection, we
observe that the 10-styles network converges as quickly as the single-style
networks in terms of style loss, but lags slightly behind in terms of
content loss.

In order to quantify this observation, we compare the final losses for 10-styles
and single-style models (center column). The 10-styles network's content loss is
around $8.7 \pm 3.9\%$ higher than its single-style counterparts, while the
difference in style losses ($8.9 \pm 16.5\%$ lower) is insignificant. While the
$N$-styles network suffers from a slight decrease in content loss convergence
speed, this may not be a fair comparison, given that it takes $N$ times more
parameter updates to train $N$ single-style networks separately than to train
them with an $N$-styles network.

The right column shows a comparison between the pastiches produced by the
10-styles network and the ones produced by the single-style networks. We see
that both results are qualitatively similar.

\subsection{The N-styles model is flexible enough to capture very different styles}

We evaluated the flexibility of the $N$-styles model by training a style
transfer network on 32 works of art chosen for their diversity.
\autoref{fig:very_varied_partial_stylizations} shows the result of applying the
trained network on evaluation images for a subset of the styles. Once again, the
full results are displayed in the Appendix. The model appears to be
capable of modeling all 32 styles in spite of the tremendous variation in color
palette and the spatial scale of the painting styles.

\subsection{The trained network generalizes across painting styles}
\label{sec:n_plus_1}

\begin{figure}[t]
\begin{center}
\includegraphics[width=0.9\linewidth]{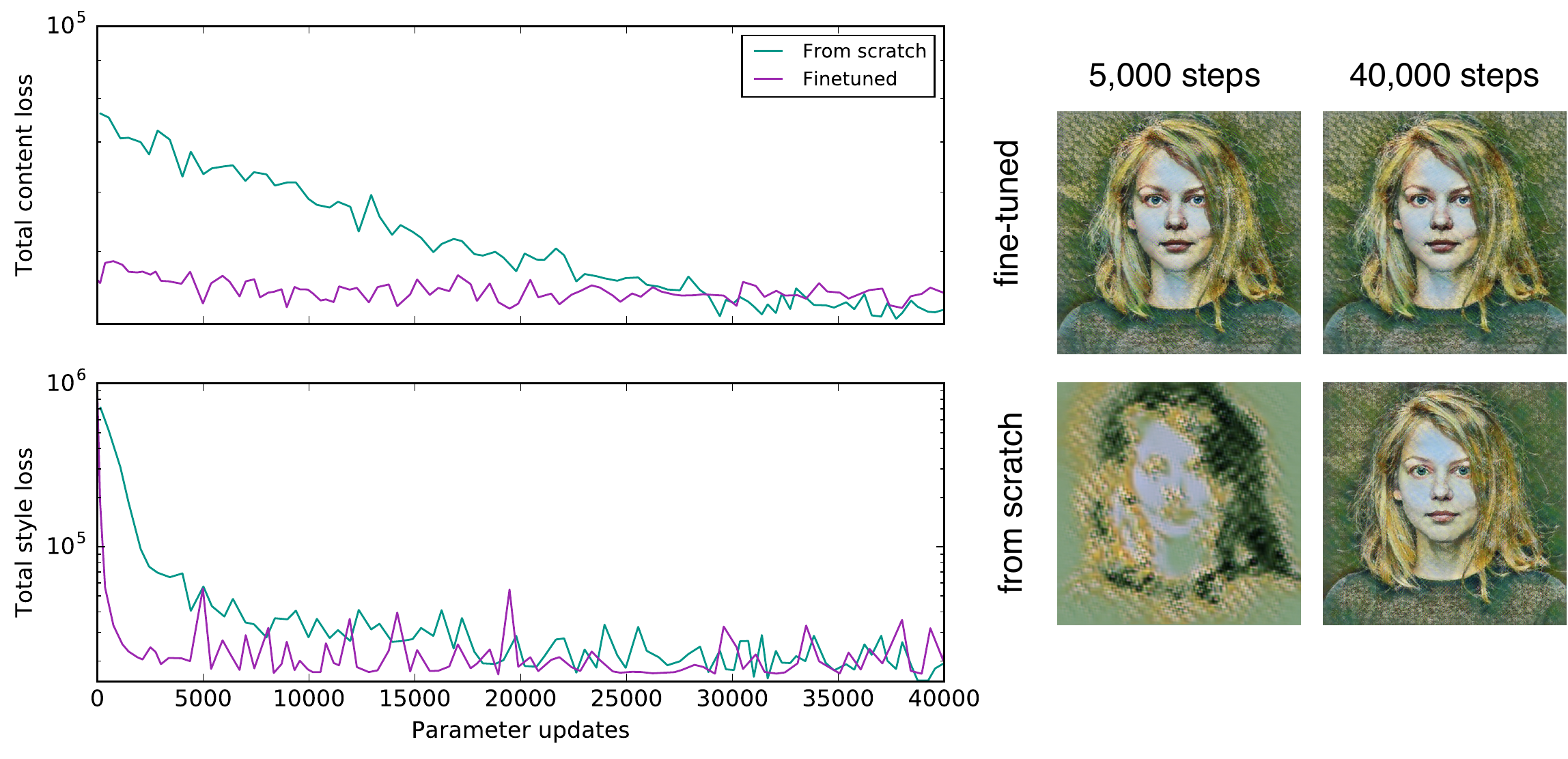}
\end{center}
\caption{\label{fig:monet_finetune} The trained network is efficient at learning
    new styles. (Left column) Learning $\gamma$ and $\beta$ from a trained style
    transfer network converges much faster than training a model from scratch.
    (Right) Learning $\gamma$ and $\beta$ for 5,000 steps from a trained style
    transfer network produces pastiches comparable to that of a single
    network trained from scratch for 40,000 steps. Conversely, 5,000 step of
    training from scratch produces leads to a poor pastiche.}
\end{figure}

Since all weights in the transformer network are shared between styles, one way
to incorporate a new style to a trained network is to keep the trained weights
fixed and learn a new set of $\gamma$ and $\beta$ parameters. To test the
efficiency of this approach, we used it to incrementally
incorporate Monet's {\em Plum Trees
in Blossom} painting to the network trained on 32 varied styles.
\autoref{fig:monet_finetune} shows that doing so is much faster than training a
new network from scratch (left) while yielding comparable pastiches: even after
eight times fewer parameter updates than its single-style counterpart, the
fine-tuned model produces comparable pastiches (right).

\subsection{The trained network can arbitrarily combine painting styles}

The conditional instance normalization approach raises some interesting
questions about style representation. In learning a different set of $\gamma$
and $\beta$ parameters for every style, we are in some sense learning an
embedding of styles.

\begin{figure}[t]
\begin{subfigure}[t]{0.6\linewidth}
\begin{center}
    \includegraphics[width=\linewidth]{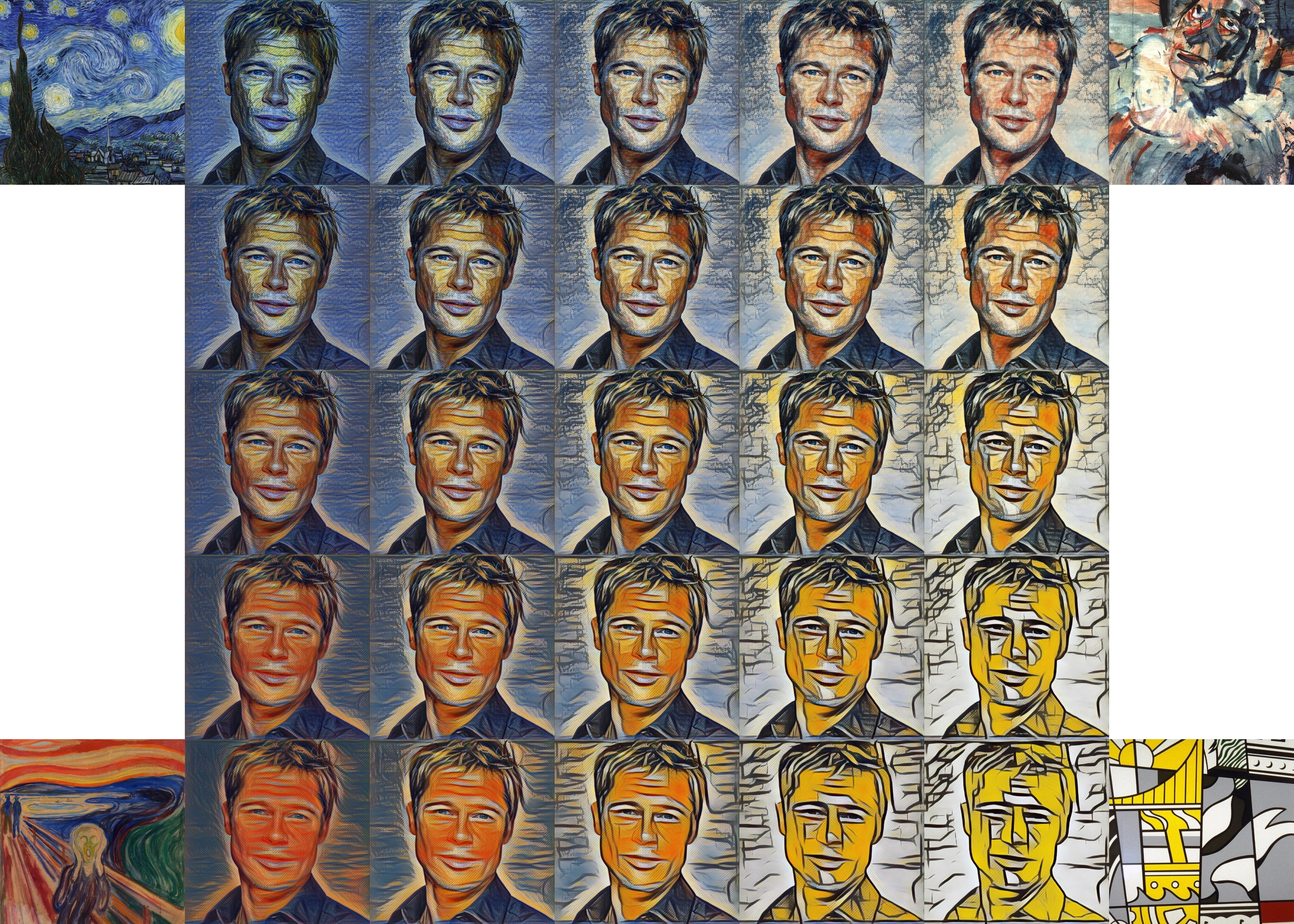}
\end{center}
\end{subfigure}%
~
\begin{subfigure}[t]{0.39\linewidth}
\begin{center}
    \includegraphics[width=\linewidth]{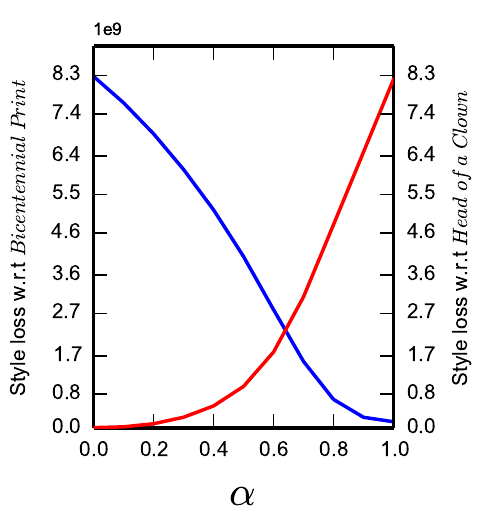}
\end{center}
\end{subfigure}
\caption{\label{fig:style_interpolation} The $N$-styles network can arbitrarily
    combine artistic styles. (Left) Combining four styles, shown in the corners.
    Each pastiche corresponds to a different convex combination of the four
    styles' $\gamma$ and $\beta$ values. (Right) As we transition from one style
    to another ({\em Bicentennial Print} and {\em Head of a Clown} in this
    case), the style losses vary monotonically.}
\end{figure}

Previous work suggested that cleverly balancing optimization
strategies offers an opportunity to blend painting styles
\footnote{For instance, \texttt{https://github.com/jcjohnson/neural-style}}.
To probe the utility of this embedding, we tried convex combinations of the
$\gamma$ and $\beta$ values to blend very distinct painting styles
(\autoref{fig:very_varied_interpolations}; \autoref{fig:style_interpolation},
left column). Employing a single convex combination produces a smooth transition
from one style to the other. Suppose ($\gamma_1, \beta_1$) and ($\gamma_2,
\beta_2$) are the parameters corresponding to two different styles. We use
$\gamma = \alpha \times \gamma_1 + (1 - \alpha) \times \gamma_2$ and $\beta =
\alpha \times \beta_1 + (1 - \alpha) \times \beta_2$ to stylize an image.
Employing convex combinations may be extended to an arbitrary number of
styles \footnote{Please see the code repository for real-time, interactive
demonstration. A screen capture is available at
\texttt{https://www.youtube.com/watch?v=6ZHiARZmiUI}.}.
\autoref{fig:style_interpolation} (right column) shows the style loss from the
transformer network for a given source image, with respect to the {\em
Bicentennial Print} and {\em Head of a Clown} paintings, as we vary $\alpha$
from $0$ to $1$. As $\alpha$ increases, the style loss with respect to {\em
Bicentennial Print} increases, which explains the smooth fading out of that
style's artifact in the transformed image.

\section{Discussion}

It seems surprising that such a small proportion of the network's
parameters can have such an impact on the overall process of style transfer.
A similar intuition has been observed in auto-regressive models of
images \citep{pixelcnn} and audio \citep{wavenet} where the conditioning process
is mediated by adjusting the biases for subsequent samples from the model.
That said, in the case of art stylization when posed as a feedforward
network, it could be that the specific network architecture is unable to take
full advantage of its capacity. We see evidence for this behavior in that
pruning the architecture leads to qualitatively similar results. Another
interpretation could be that the convolutional weights of the style transfer
network encode transformations that represent ``elements of style''. The scaling
and shifting factors would then provide a way for each style to inhibit or
enhance the expression of various elements of style to form a global identity of
style. While this work does not attempt to verify this hypothesis, we think that
this would constitute a very promising direction of research in understanding
the computation behind style transfer networks as well as the representation
of images in general.

Concurrent to this work, \cite{gatys2} demonstrated exciting new methods
for revising the loss to selectively adjust the spatial scale, color
information and spatial localization of the artistic style information.
These methods are complementary to the results in this paper and present an
interesting direction for exploring how spatial and color information uniquely
factor into artistic style representation.

The question of how predictive each style image is of its corresponding style
representation is also of great interest. If it is the case that the style
representation can easily be predicted from a style image, one could imagine
building a transformer network which skips learning an individual conditional
embedding and instead learn to produce a pastiche directly from a style and a
content image, much like in the original neural algorithm of artistic style,
but without any optimization loop at test time.

Finally, the learned style representation opens the door to generative models of
style: by modeling enough paintings of a given artistic movement (e.g.
impressionism), one could build a collection of style embeddings upon which a
generative model could be trained. At test time, a style representation would be
sampled from the generative model and used in conjunction with the style
transfer network to produce a random pastiche of that artistic movement.

In summary, we demonstrated that conditional instance normalization constitutes
a simple,
efficient and scalable modification of style transfer networks that allows them
to model multiple styles at the same time.
A practical consequence of this approach is that a new painting style may be
transmitted to and stored on a mobile device with a small number of parameters.
We showed that despite its
simplicity, the method is flexible enough to capture very different styles while
having very little impact on training time and final performance of the trained
network. Finally, we showed that the learned representation of style is useful
in arbitrarily combining artistic styles.
This work suggests the existence of a learned representation for
artistic styles whose vocabulary
is flexible enough to capture a diversity of the painted world.

\subsubsection*{Acknowledgments}

We would like to thank Fred Bertsch, Douglas Eck, Cinjon Resnick and the rest of
the Google Magenta team for their feedback; Peyman Milanfar, Michael Elad, Feng
Yang, Jon Barron, Bhavik Singh, Jennifer Daniel as well as the the Google Brain
team for their crucial suggestions and advice; an anonymous reviewer for
helpful suggestions about applying this model in a mobile domain.
Finally, we would like to thank
the Google Cultural Institute, whose curated collection of art photographs was
very helpful in finding exciting style images to train on.

\clearpage
\bibliography{iclr2017_conference}

\begin{thebibliography}{27}
\providecommand{\natexlab}[1]{#1}
\providecommand{\url}[1]{\texttt{#1}}
\expandafter\ifx\csname urlstyle\endcsname\relax
  \providecommand{\doi}[1]{doi: #1}\else
  \providecommand{\doi}{doi: \begingroup \urlstyle{rm}\Url}\fi

\bibitem[Abadi et~al.(2016)Abadi, Agarwal, Barham, Brevdo, Chen, Citro,
  Corrado, Davis, Dean, Devin, et~al.]{tensorflow}
Mart{\i}n Abadi, Ashish Agarwal, Paul Barham, Eugene Brevdo, Zhifeng Chen,
  Craig Citro, Greg~S Corrado, Andy Davis, Jeffrey Dean, Matthieu Devin, et~al.
\newblock Tensorflow: Large-scale machine learning on heterogeneous distributed
  systems.
\newblock \emph{arXiv preprint arXiv:1603.04467}, 2016.

\bibitem[Deng et~al.(2009)Deng, Dong, Socher, Li, Li, and Fei-Fei]{imagenet}
Jia Deng, Wei Dong, Richard Socher, Li-Jia Li, Kai Li, and Li~Fei-Fei.
\newblock Imagenet: A large-scale hierarchical image database.
\newblock In \emph{Computer Vision and Pattern Recognition, 2009. CVPR 2009.
  IEEE Conference on}, pp.\  248--255. IEEE, 2009.

\bibitem[Efros \& Freeman(2001)Efros and Freeman]{efros2001image}
Alexei~A Efros and William~T Freeman.
\newblock Image quilting for texture synthesis and transfer.
\newblock In \emph{Proceedings of the 28th annual conference on Computer
  graphics and interactive techniques}, pp.\  341--346. ACM, 2001.

\bibitem[Efros \& Leung(1999)Efros and Leung]{efros1999texture}
Alexei~A Efros and Thomas~K Leung.
\newblock Texture synthesis by non-parametric sampling.
\newblock In \emph{Computer Vision, 1999. The Proceedings of the Seventh IEEE
  International Conference on}, volume~2, pp.\  1033--1038. IEEE, 1999.

\bibitem[Elad \& Milanfar(2016)Elad and Milanfar]{elad2016style}
Michael Elad and Peyman Milanfar.
\newblock Style-transfer via texture-synthesis.
\newblock \emph{arXiv preprint arXiv:1609.03057}, 2016.

\bibitem[Frigo et~al.(2016)Frigo, Sabater, Delon, and Hellier]{frigo2016split}
Oriel Frigo, Neus Sabater, Julie Delon, and Pierre Hellier.
\newblock Split and match: Example-based adaptive patch sampling for
  unsupervised style transfer.
\newblock 2016.

\bibitem[Gatys et~al.(2015{\natexlab{a}})Gatys, Ecker, and
  Bethge]{gatys2015texture}
Leon Gatys, Alexander~S Ecker, and Matthias Bethge.
\newblock Texture synthesis using convolutional neural networks.
\newblock In \emph{Advances in Neural Information Processing Systems}, pp.\
  262--270, 2015{\natexlab{a}}.

\bibitem[Gatys et~al.(2015{\natexlab{b}})Gatys, Ecker, and
  Bethge]{gatys2015neural}
Leon~A Gatys, Alexander~S Ecker, and Matthias Bethge.
\newblock A neural algorithm of artistic style.
\newblock \emph{arXiv preprint arXiv:1508.06576}, 2015{\natexlab{b}}.

\bibitem[Gatys et~al.(2016{\natexlab{a}})Gatys, Bethge, Hertzmann, and
  Shechtman]{gatys2016preserving}
Leon~A Gatys, Matthias Bethge, Aaron Hertzmann, and Eli Shechtman.
\newblock Preserving color in neural artistic style transfer.
\newblock \emph{arXiv preprint arXiv:1606.05897}, 2016{\natexlab{a}}.

\bibitem[Gatys et~al.(2016{\natexlab{b}})Gatys, Ecker, Bethge, Hertzmann, and
  Shechtman]{gatys2}
Leon~A. Gatys, Alexander~S. Ecker, Matthias Bethge, Aaron Hertzmann, and Eli
  Shechtman.
\newblock Controlling perceptual factors in neural style transfer.
\newblock \emph{CoRR}, abs/1611.07865, 2016{\natexlab{b}}.
\newblock URL \url{http://arxiv.org/abs/1611.07865}.

\bibitem[Hertzmann et~al.(2001)Hertzmann, Jacobs, Oliver, Curless, and
  Salesin]{hertzmann2001image}
Aaron Hertzmann, Charles~E Jacobs, Nuria Oliver, Brian Curless, and David~H
  Salesin.
\newblock Image analogies.
\newblock In \emph{Proceedings of the 28th annual conference on Computer
  graphics and interactive techniques}, pp.\  327--340. ACM, 2001.

\bibitem[Johnson et~al.(2016)Johnson, Alahi, and
  Fei-Fei]{johnson2016perceptual}
Justin Johnson, Alexandre Alahi, and Li~Fei-Fei.
\newblock Perceptual losses for real-time style transfer and super-resolution.
\newblock \emph{arXiv preprint arXiv:1603.08155}, 2016.

\bibitem[Julesz(1962)]{julesz1962}
Bela Julesz.
\newblock Visual pattern discrimination.
\newblock \emph{IRE Trans. Info Theory}, 8:\penalty0 84--92, 1962.

\bibitem[Kingma \& Ba(2014)Kingma and Ba]{kingma2014adam}
Diederik Kingma and Jimmy Ba.
\newblock Adam: A method for stochastic optimization.
\newblock \emph{arXiv preprint arXiv:1412.6980}, 2014.

\bibitem[Kwatra et~al.(2005)Kwatra, Essa, Bobick, and
  Kwatra]{kwatra2005texture}
Vivek Kwatra, Irfan Essa, Aaron Bobick, and Nipun Kwatra.
\newblock Texture optimization for example-based synthesis.
\newblock \emph{ACM Transactions on Graphics (ToG)}, 24\penalty0 (3):\penalty0
  795--802, 2005.

\bibitem[Li \& Wand(2016)Li and Wand]{liwand2016}
Chuan Li and Michael Wand.
\newblock Precomputed real-time texture synthesis with markovian generative
  adversarial networks.
\newblock \emph{ECCV}, 2016.
\newblock URL \url{http://arxiv.org/abs/1604.04382}.

\bibitem[Liang et~al.(2001)Liang, Liu, Xu, Guo, and Shum]{liang2001real}
Lin Liang, Ce~Liu, Ying-Qing Xu, Baining Guo, and Heung-Yeung Shum.
\newblock Real-time texture synthesis by patch-based sampling.
\newblock \emph{ACM Transactions on Graphics (ToG)}, 20\penalty0 (3):\penalty0
  127--150, 2001.

\bibitem[Odena et~al.(2016)Odena, Olah, and Dumoulin]{odena2016avoiding}
Augustus Odena, Christopher Olah, and Vincent Dumoulin.
\newblock Avoiding checkerboard artifacts in neural networks.
\newblock \emph{Distill}, 2016.

\bibitem[Portilla \& Simoncelli(1999)Portilla and Simoncelli]{portilla1999}
Javier Portilla and Eero Simoncelli.
\newblock A parametric texture model based on joint statistics of complex
  wavelet coefficients.
\newblock \emph{International Journal of Computer Vision}, 40:\penalty0 49--71,
  1999.

\bibitem[Simoncelli \& Olshausen(2001)Simoncelli and
  Olshausen]{simoncelli2001review}
Eero Simoncelli and Bruno Olshausen.
\newblock Natural image statistics and neural representation.
\newblock \emph{Annual Review of Neuroscience}, 24:\penalty0 1193--1216, 2001.

\bibitem[Simonyan \& Zisserman(2014)Simonyan and Zisserman]{simonyan2014very}
Karen Simonyan and Andrew Zisserman.
\newblock Very deep convolutional networks for large-scale image recognition.
\newblock \emph{arXiv preprint arXiv:1409.1556}, 2014.

\bibitem[Ulyanov et~al.(2016{\natexlab{a}})Ulyanov, Lebedev, Vedaldi, and
  Lempitsky]{ulyanov2016texture}
Dmitry Ulyanov, Vadim Lebedev, Andrea Vedaldi, and Victor Lempitsky.
\newblock Texture networks: Feed-forward synthesis of textures and stylized
  images.
\newblock \emph{arXiv preprint arXiv:1603.03417}, 2016{\natexlab{a}}.

\bibitem[Ulyanov et~al.(2016{\natexlab{b}})Ulyanov, Vedaldi, and
  Lempitsky]{ulyanov2016instance}
Dmitry Ulyanov, Andrea Vedaldi, and Victor Lempitsky.
\newblock Instance normalization: The missing ingredient for fast stylization.
\newblock \emph{arXiv preprint arXiv:1607.08022}, 2016{\natexlab{b}}.

\bibitem[van~den Oord et~al.(2016{\natexlab{a}})van~den Oord, Dieleman, Zen,
  Simonyan, Vinyals, Graves, Kalchbrenner, Senior, and Kavukcuoglu]{wavenet}
A{\"{a}}ron van~den Oord, Sander Dieleman, Heiga Zen, Karen Simonyan, Oriol
  Vinyals, Alex Graves, Nal Kalchbrenner, Andrew~W. Senior, and Koray
  Kavukcuoglu.
\newblock Wavenet: {A} generative model for raw audio.
\newblock \emph{CoRR}, abs/1609.03499, 2016{\natexlab{a}}.
\newblock URL \url{http://arxiv.org/abs/1609.03499}.

\bibitem[van~den Oord et~al.(2016{\natexlab{b}})van~den Oord, Kalchbrenner,
  Vinyals, Espeholt, Graves, and Kavukcuoglu]{pixelcnn}
A{\"{a}}ron van~den Oord, Nal Kalchbrenner, Oriol Vinyals, Lasse Espeholt, Alex
  Graves, and Koray Kavukcuoglu.
\newblock Conditional image generation with pixelcnn decoders.
\newblock \emph{CoRR}, abs/1606.05328, 2016{\natexlab{b}}.
\newblock URL \url{http://arxiv.org/abs/1606.05328}.

\bibitem[Wei \& Levoy(2000)Wei and Levoy]{wei2000fast}
Li-Yi Wei and Marc Levoy.
\newblock Fast texture synthesis using tree-structured vector quantization.
\newblock In \emph{Proceedings of the 27th annual conference on Computer
  graphics and interactive techniques}, pp.\  479--488. ACM
  Press/Addison-Wesley Publishing Co., 2000.

\bibitem[Zeiler \& Fergus(2014)Zeiler and Fergus]{zeiler2014visualizing}
Matthew~D Zeiler and Rob Fergus.
\newblock Visualizing and understanding convolutional networks.
\newblock In \emph{European Conference on Computer Vision}, pp.\  818--833.
  Springer, 2014.

\end{thebibliography}
\bibliographystyle{iclr2017_conference}

\clearpage
\section*{Appendix}

\subsection*{Hyperparameters}

\begin{table}[h]
\centering
\begin{tabular}{@{}rllllll@{}} \toprule
Operation      & Kernel size & Stride & Feature maps & Padding & Nonlinearity \\ \midrule
{\bf Network} -- $256 \times 256 \times 3$ input                              \\
Convolution    & $9$         & $1$    & $32$         &  SAME   & ReLU         \\
Convolution    & $3$         & $2$    & $64$         &  SAME   & ReLU         \\
Convolution    & $3$         & $2$    & $128$        &  SAME   & ReLU         \\
Residual block &             &        & $128$        &         &              \\
Residual block &             &        & $128$        &         &              \\
Residual block &             &        & $128$        &         &              \\
Residual block &             &        & $128$        &         &              \\
Residual block &             &        & $128$        &         &              \\
Upsampling     &             &        & $64$         &         &              \\
Upsampling     &             &        & $32$         &         &              \\
Convolution    & $9$         & $1$    & $3$          &  SAME   & Sigmoid      \\
{\bf Residual block} -- $C$ feature maps                                      \\
Convolution    & $3$         & $1$    & $C$          &  SAME   & ReLU         \\
Convolution    & $3$         & $1$    & $C$          &  SAME   & Linear       \\
               & \multicolumn{6}{@{}l@{}}{\em Add the input and the output}   \\
{\bf Upsampling} -- $C$ feature maps                                          \\
               & \multicolumn{6}{@{}l@{}}{\em Nearest-neighbor interpolation, factor 2} \\
Convolution    & $3$         & $1$    & $C$        &  SAME   & ReLU           \\ \midrule
Padding mode           & \multicolumn{6}{@{}l@{}}{REFLECT}                    \\
Normalization          & \multicolumn{6}{@{}l@{}}{Conditional instance normalization after every convolution} \\
Optimizer              & \multicolumn{6}{@{}l@{}}{Adam \citep{kingma2014adam} ($\alpha = 0.001$, $\beta_1 = 0.9$, $\beta_2 = 0.999$)}  \\
Parameter updates      & \multicolumn{6}{@{}l@{}}{40,000}                     \\
Batch size             & \multicolumn{6}{@{}l@{}}{16}                         \\
Weight initialization  & \multicolumn{6}{@{}l@{}}{Isotropic gaussian ($\mu = 0$, $\sigma = 0.01$)}  \\ \bottomrule
\end{tabular}
\vspace{0.2cm}
\caption{\label{tab:architecture} Style transfer network hyperparameters.}
\end{table}

\clearpage
\subsection*{Monet pastiches}

\begin{figure}[ht]
\begin{center}
    \includegraphics[width=0.18\linewidth]{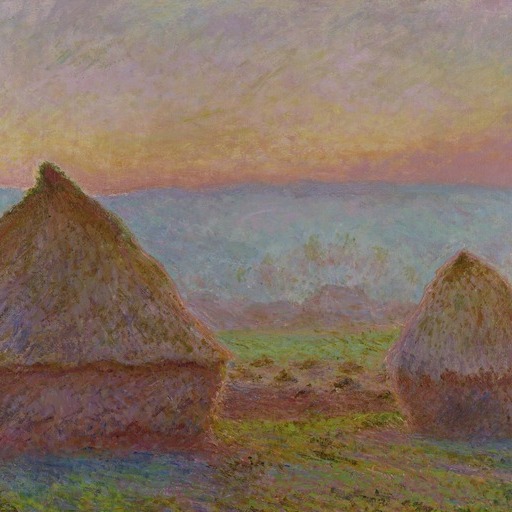}
    \includegraphics[width=0.18\linewidth]{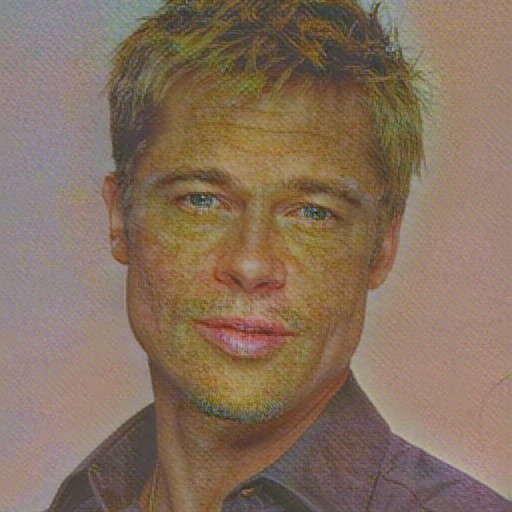}
    \includegraphics[width=0.18\linewidth]{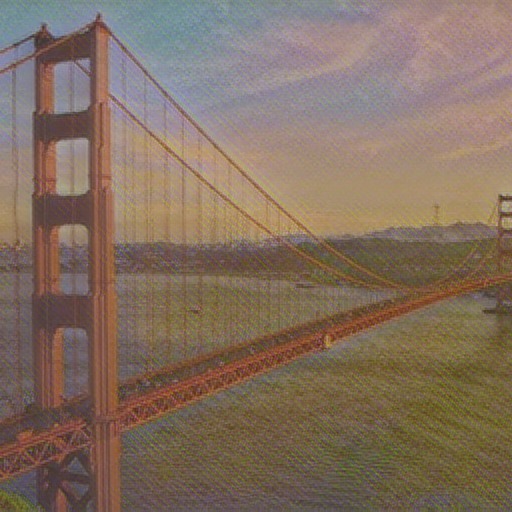}
    \includegraphics[width=0.18\linewidth]{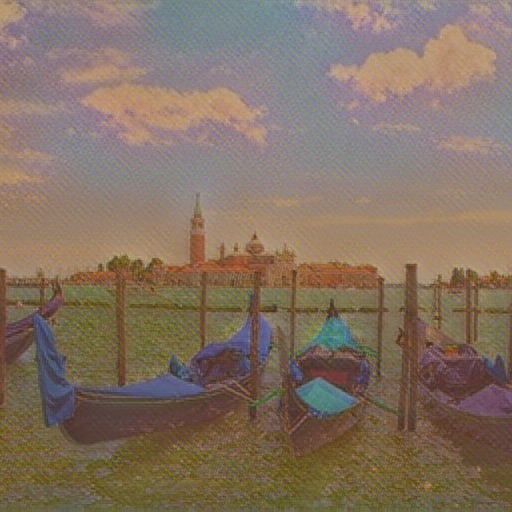}
    \includegraphics[width=0.18\linewidth]{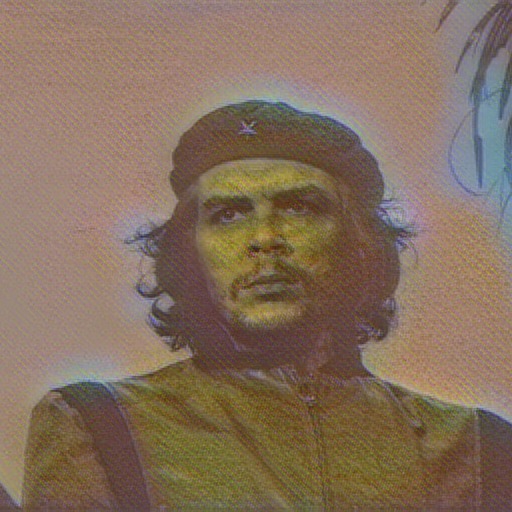} \\
    \includegraphics[width=0.18\linewidth]{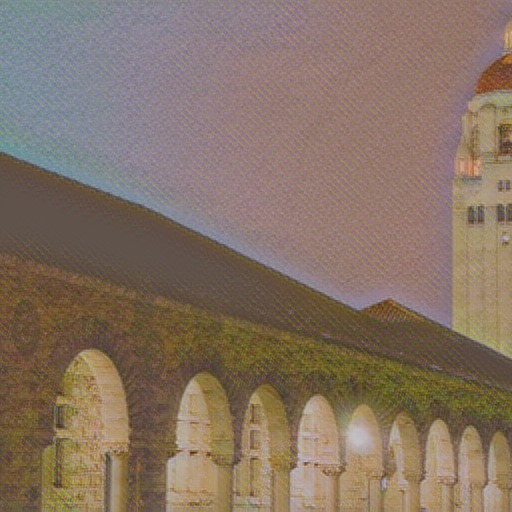}
    \includegraphics[width=0.18\linewidth]{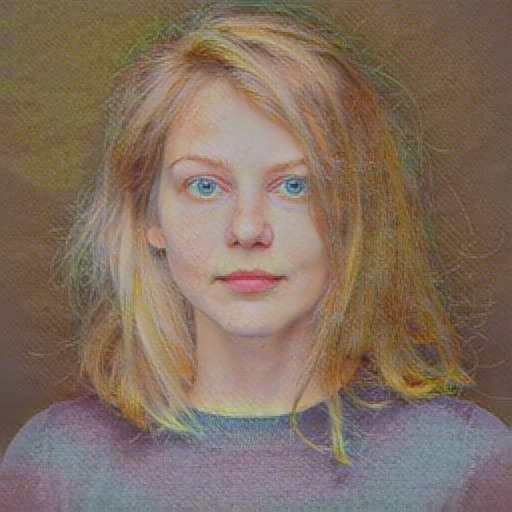}
    \includegraphics[width=0.18\linewidth]{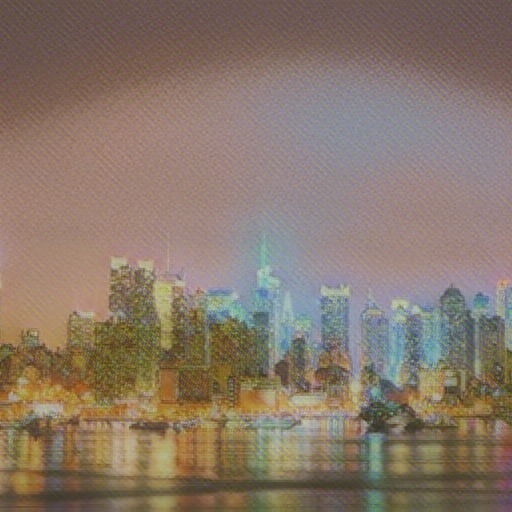}
    \includegraphics[width=0.18\linewidth]{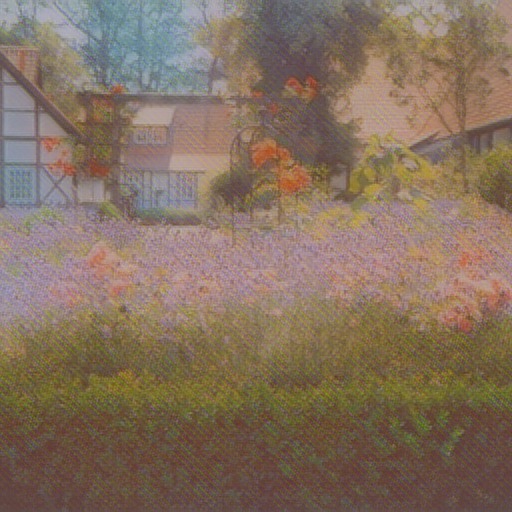}
    \includegraphics[width=0.18\linewidth]{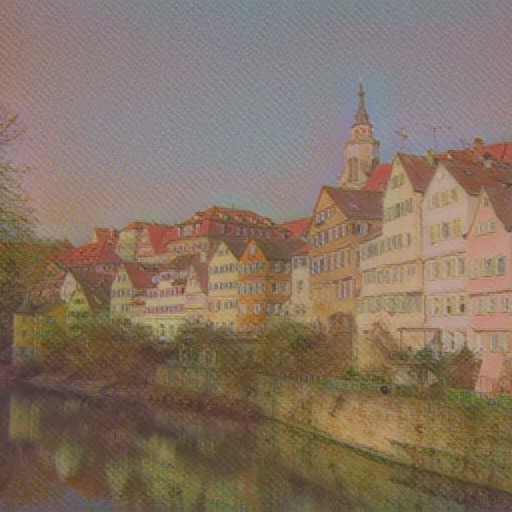}
\end{center}
\caption*{Claude Monet, {\em Grainstacks at Giverny; the Evening Sun}
    (1888/1889).}
\end{figure}

\begin{figure}[ht]
\begin{center}
    \includegraphics[width=0.18\linewidth]{figures/plum_trees_in_blossom.jpg}
    \includegraphics[width=0.18\linewidth]{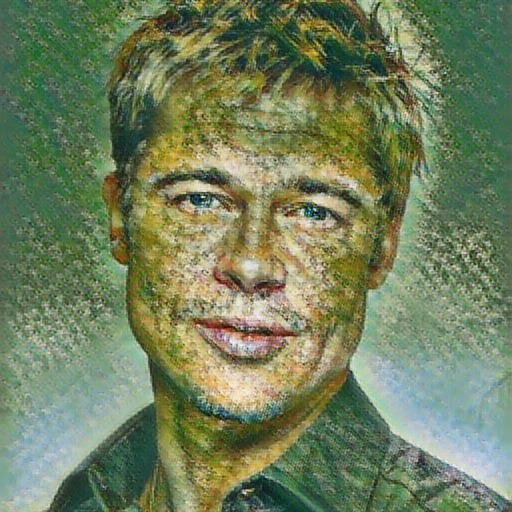}
    \includegraphics[width=0.18\linewidth]{figures/golden_gate_plum_trees.jpg}
    \includegraphics[width=0.18\linewidth]{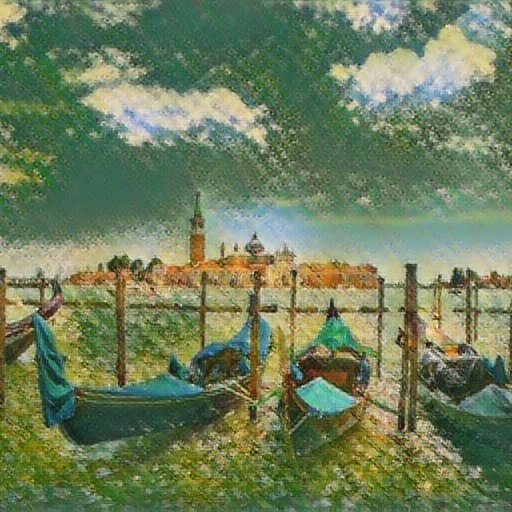}
    \includegraphics[width=0.18\linewidth]{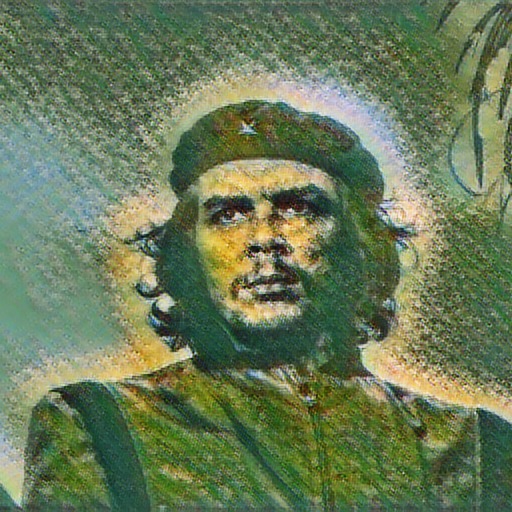} \\
    \includegraphics[width=0.18\linewidth]{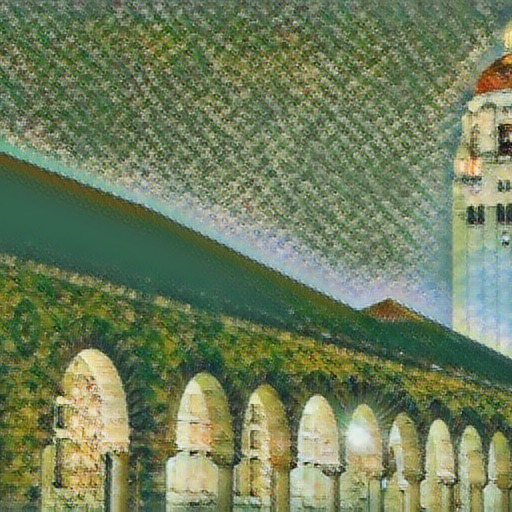}
    \includegraphics[width=0.18\linewidth]{figures/karya_plum_trees.jpg}
    \includegraphics[width=0.18\linewidth]{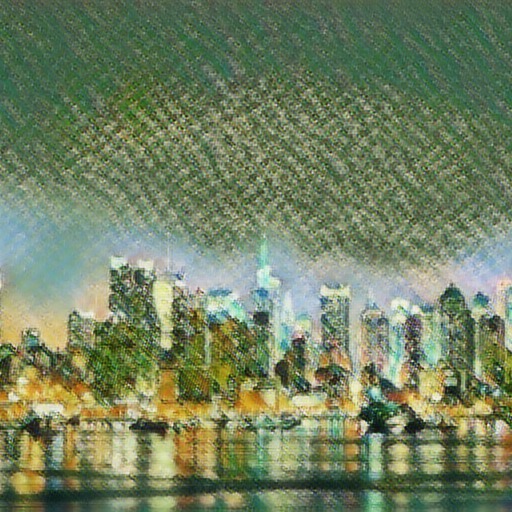}
    \includegraphics[width=0.18\linewidth]{figures/schultenhof_mettingen_bauerngarten_plum_trees.jpg}
    \includegraphics[width=0.18\linewidth]{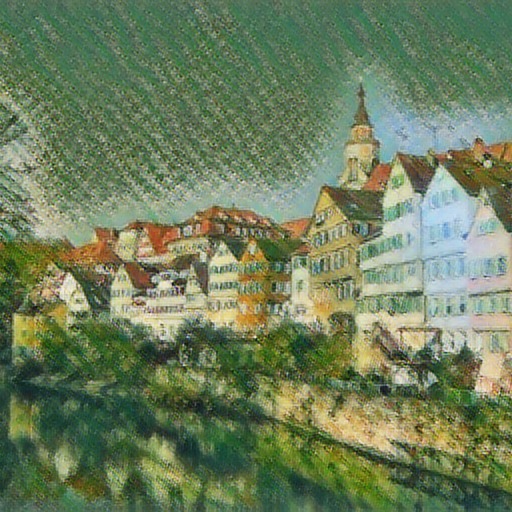}
\end{center}
\caption*{Claude Monet, {\em Plum Trees in Blossom} (1879).}
\end{figure}

\begin{figure}[ht]
\begin{center}
    \includegraphics[width=0.18\linewidth]{figures/poppy_field.jpg}
    \includegraphics[width=0.18\linewidth]{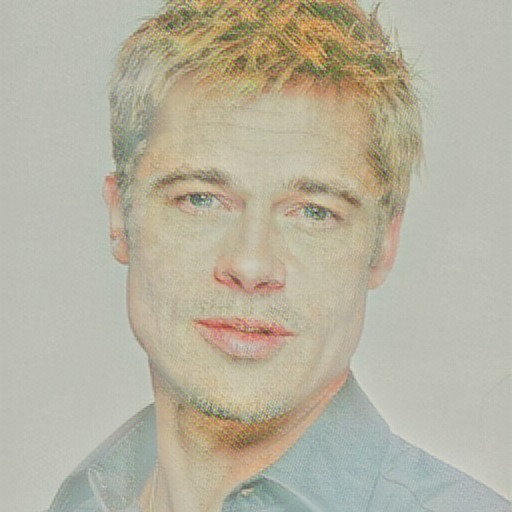}
    \includegraphics[width=0.18\linewidth]{figures/golden_gate_poppy_field.jpg}
    \includegraphics[width=0.18\linewidth]{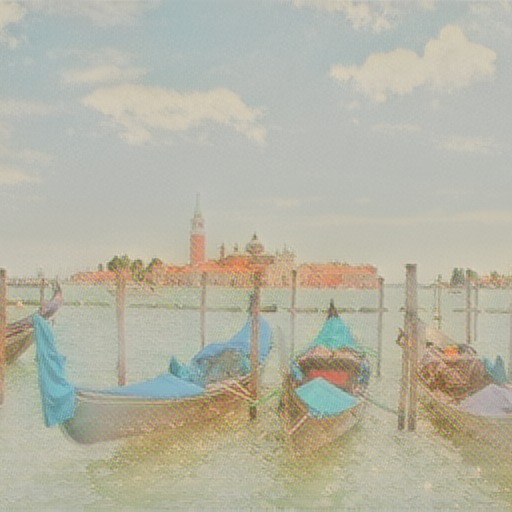}
    \includegraphics[width=0.18\linewidth]{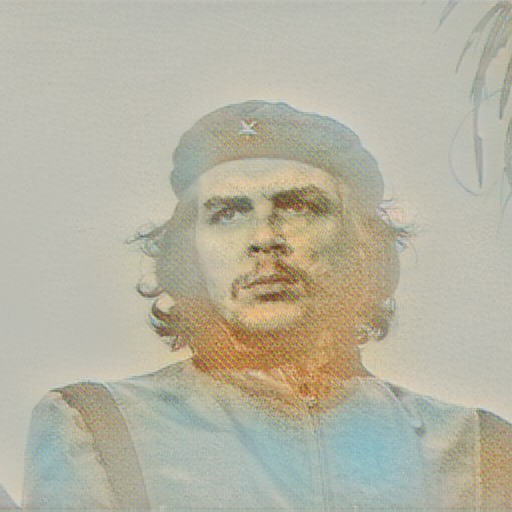} \\
    \includegraphics[width=0.18\linewidth]{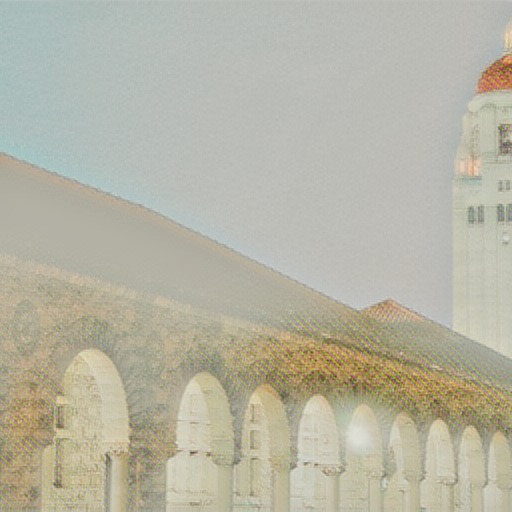}
    \includegraphics[width=0.18\linewidth]{figures/karya_poppy_field.jpg}
    \includegraphics[width=0.18\linewidth]{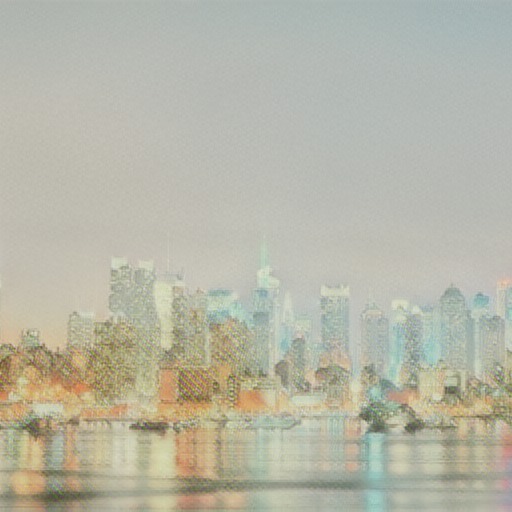}
    \includegraphics[width=0.18\linewidth]{figures/schultenhof_mettingen_bauerngarten_poppy_field.jpg}
    \includegraphics[width=0.18\linewidth]{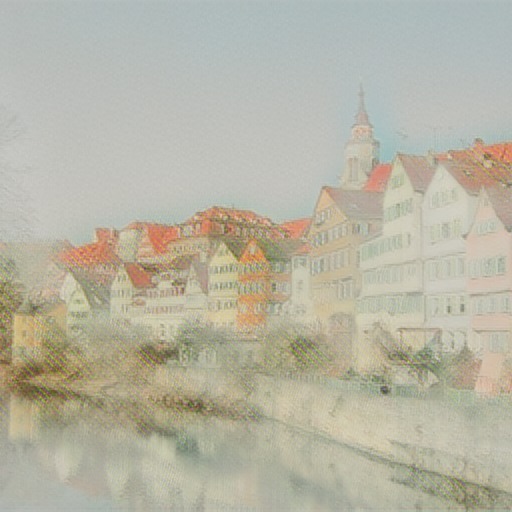}
\end{center}
\caption*{Claude Monet, {\em Poppy Field} (1873).}
\end{figure}

\clearpage
\begin{figure}[ht]
\begin{center}
    \includegraphics[width=0.18\linewidth]{figures/rouen_cathedral_west_facade.jpg}
    \includegraphics[width=0.18\linewidth]{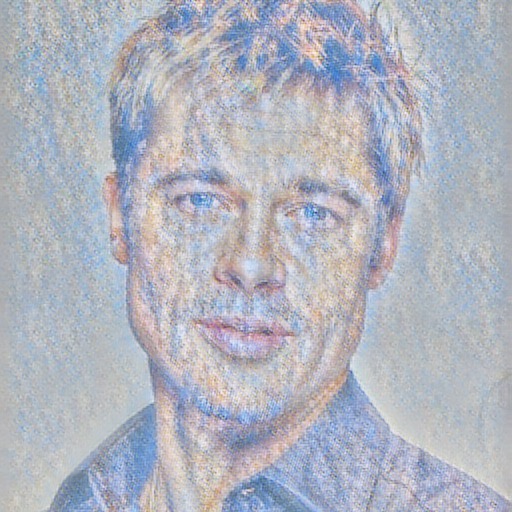}
    \includegraphics[width=0.18\linewidth]{figures/golden_gate_rouen_cathedral.jpg}
    \includegraphics[width=0.18\linewidth]{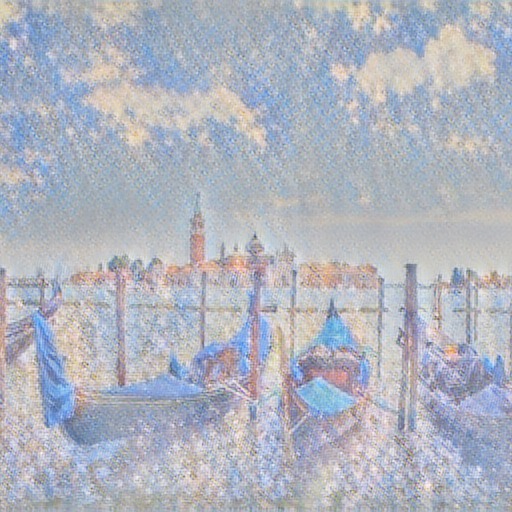}
    \includegraphics[width=0.18\linewidth]{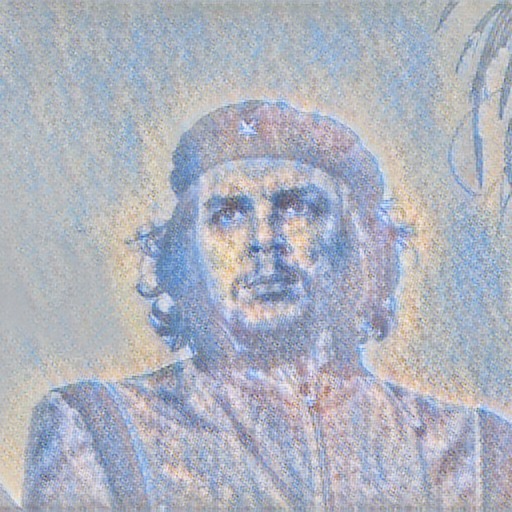} \\
    \includegraphics[width=0.18\linewidth]{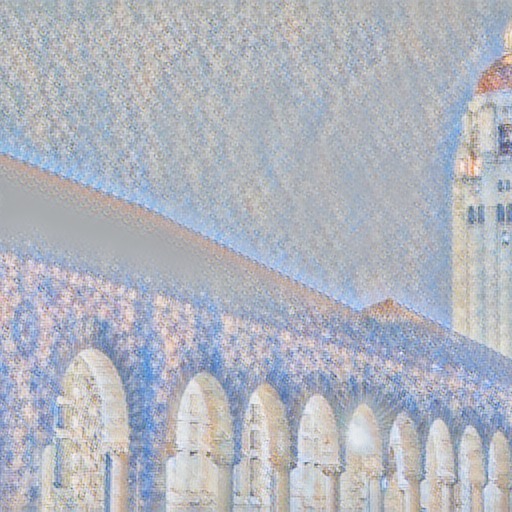}
    \includegraphics[width=0.18\linewidth]{figures/karya_rouen_cathedral.jpg}
    \includegraphics[width=0.18\linewidth]{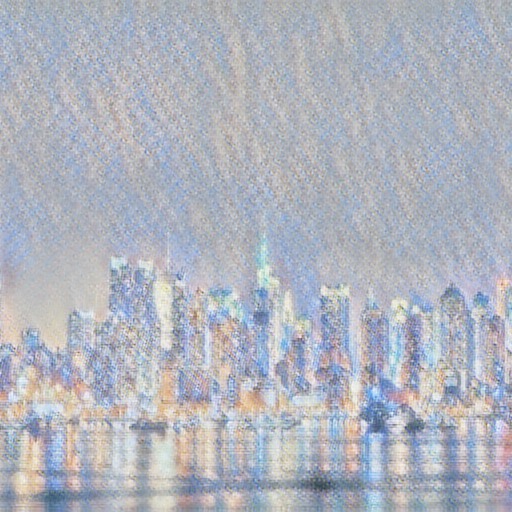}
    \includegraphics[width=0.18\linewidth]{figures/schultenhof_mettingen_bauerngarten_rouen_cathedral.jpg}
    \includegraphics[width=0.18\linewidth]{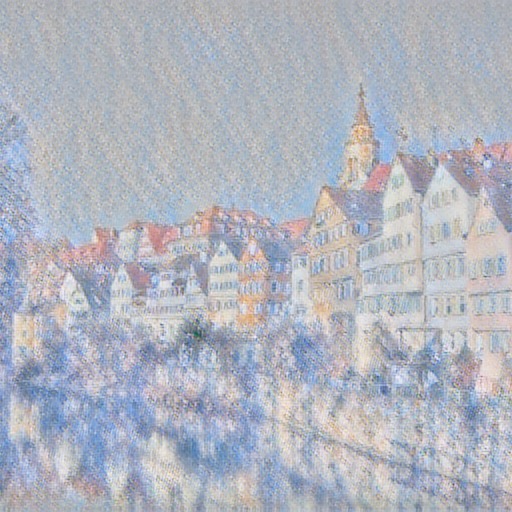}
\end{center}
\caption*{Claude Monet, {\em Rouen Cathedral, West Fa\c{c}ade} (1894).}
\end{figure}

\begin{figure}[ht]
\begin{center}
    \includegraphics[width=0.18\linewidth]{figures/sunrise_marine.jpg}
    \includegraphics[width=0.18\linewidth]{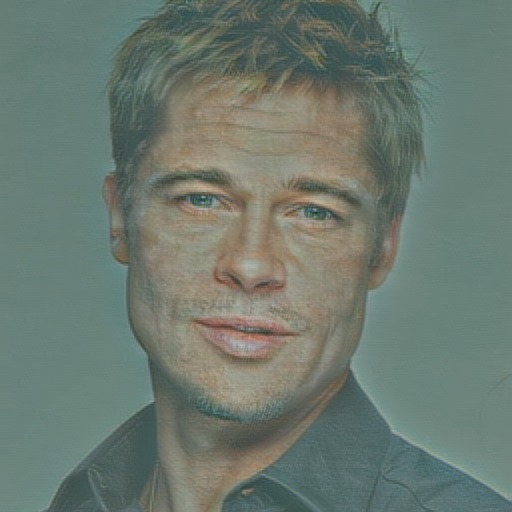}
    \includegraphics[width=0.18\linewidth]{figures/golden_gate_sunrise_marine.jpg}
    \includegraphics[width=0.18\linewidth]{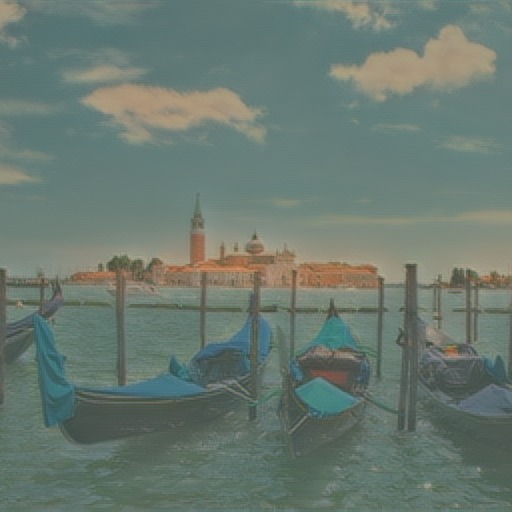}
    \includegraphics[width=0.18\linewidth]{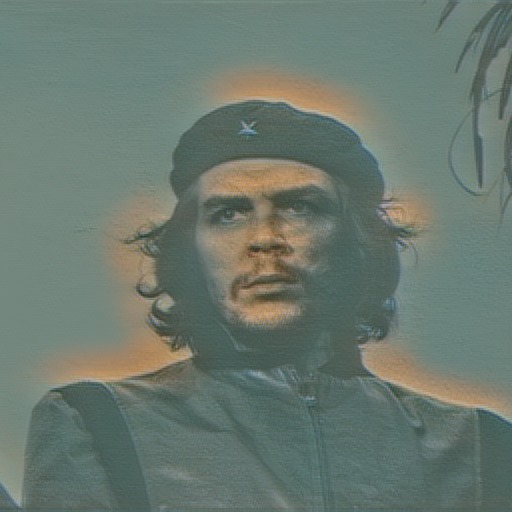} \\
    \includegraphics[width=0.18\linewidth]{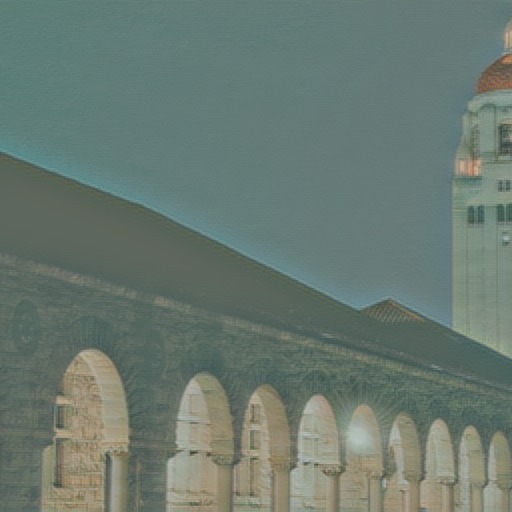}
    \includegraphics[width=0.18\linewidth]{figures/karya_sunrise_marine.jpg}
    \includegraphics[width=0.18\linewidth]{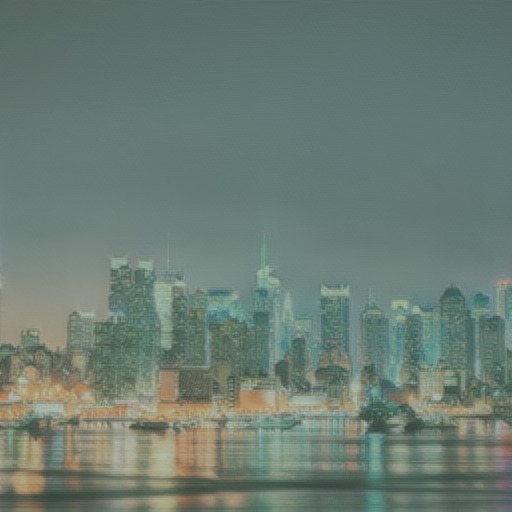}
    \includegraphics[width=0.18\linewidth]{figures/schultenhof_mettingen_bauerngarten_sunrise_marine.jpg}
    \includegraphics[width=0.18\linewidth]{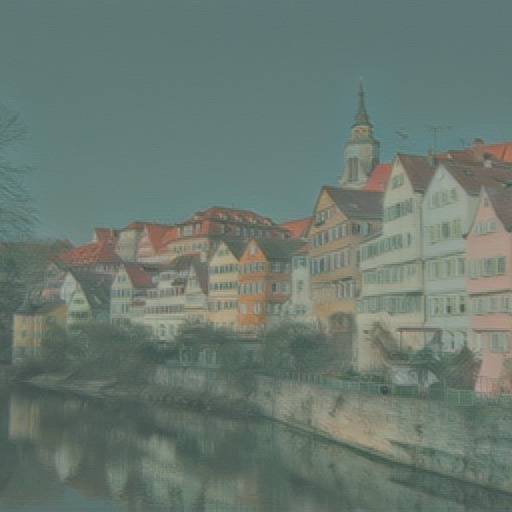}
\end{center}
\caption*{Claude Monet, {\em Sunrise (Marine)} (1873).}
\end{figure}

\begin{figure}[ht]
\begin{center}
    \includegraphics[width=0.18\linewidth]{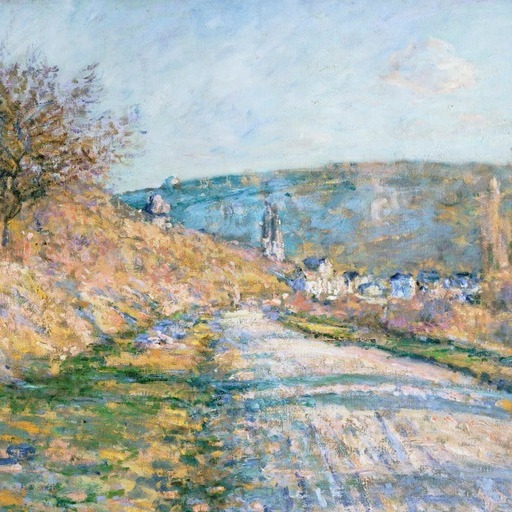}
    \includegraphics[width=0.18\linewidth]{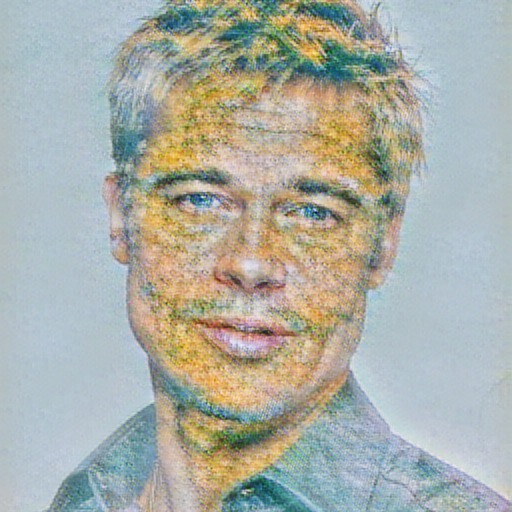}
    \includegraphics[width=0.18\linewidth]{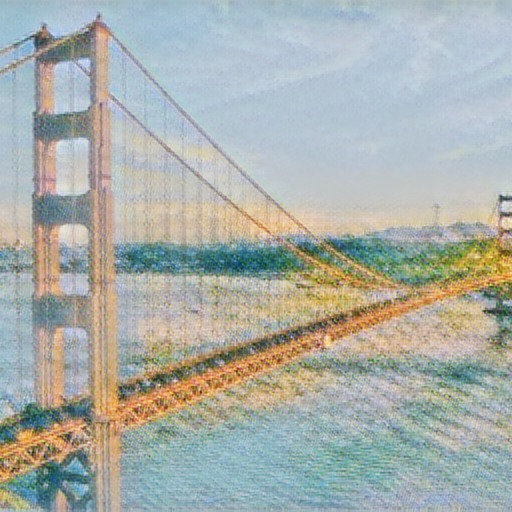}
    \includegraphics[width=0.18\linewidth]{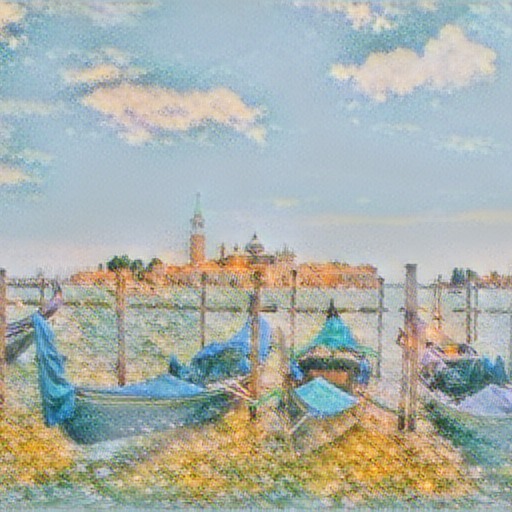}
    \includegraphics[width=0.18\linewidth]{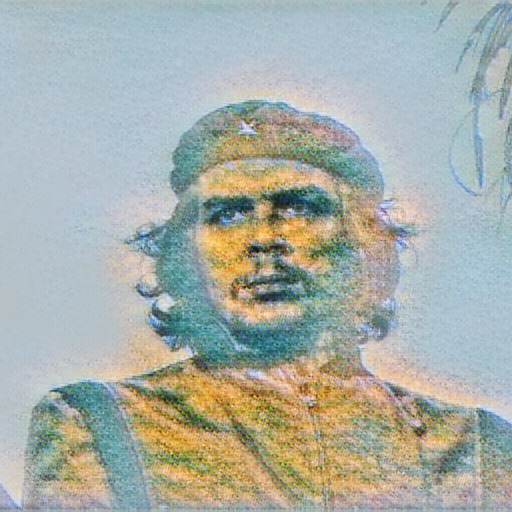} \\
    \includegraphics[width=0.18\linewidth]{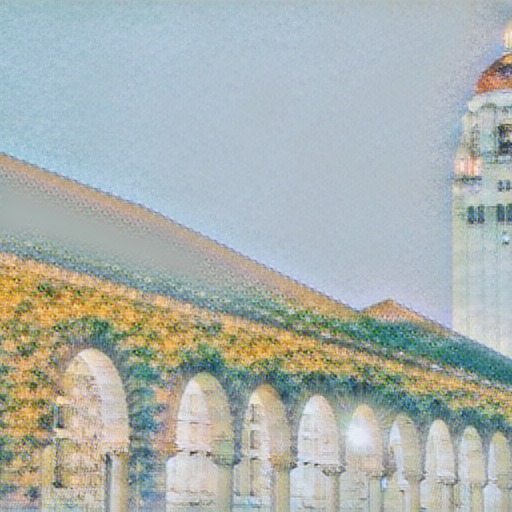}
    \includegraphics[width=0.18\linewidth]{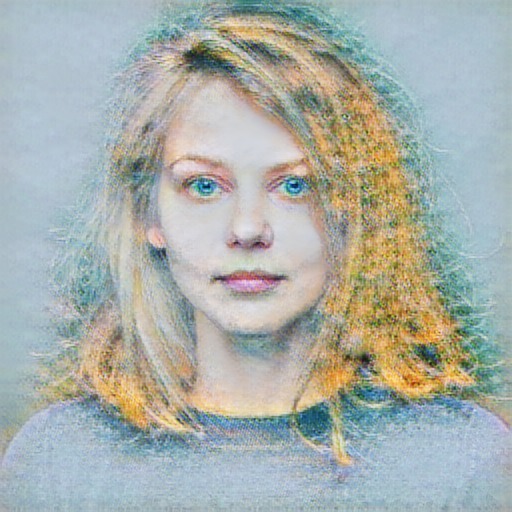}
    \includegraphics[width=0.18\linewidth]{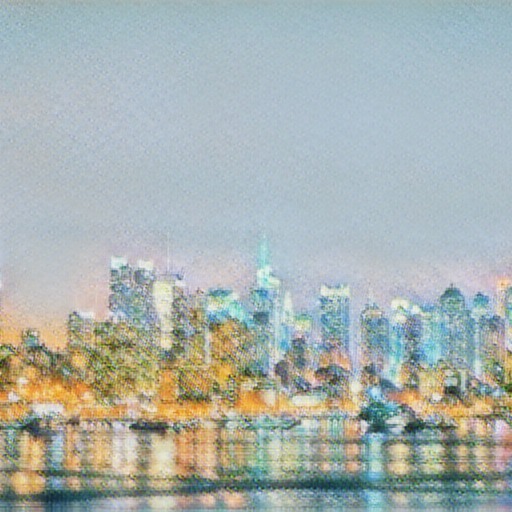}
    \includegraphics[width=0.18\linewidth]{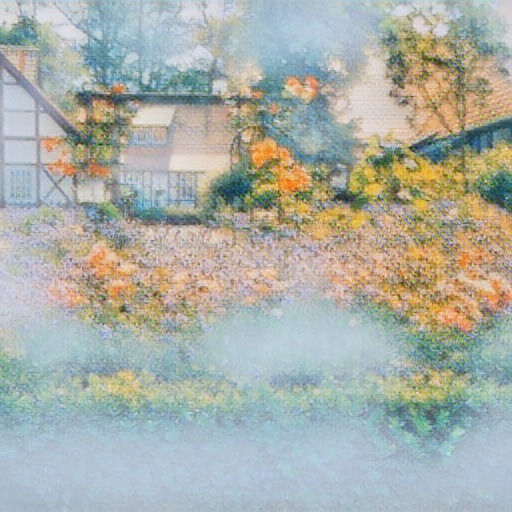}
    \includegraphics[width=0.18\linewidth]{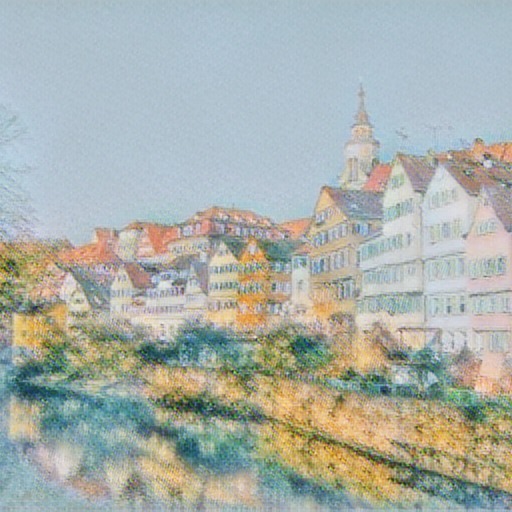}
\end{center}
\caption*{Claude Monet, {\em The Road to V\'{e}theuil} (1879).}
\end{figure}

\clearpage
\begin{figure}[ht]
\begin{center}
    \includegraphics[width=0.18\linewidth]{figures/three_fishing_boats.jpg}
    \includegraphics[width=0.18\linewidth]{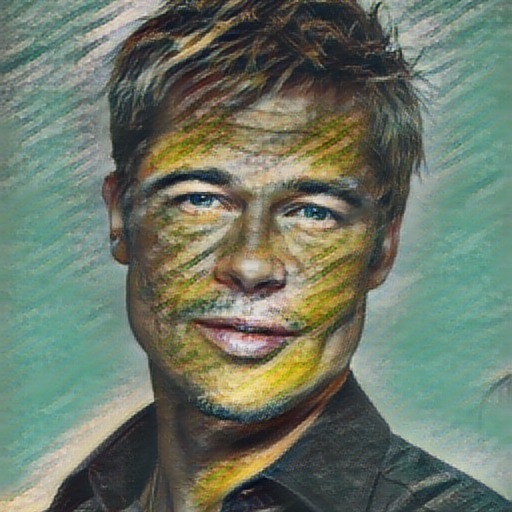}
    \includegraphics[width=0.18\linewidth]{figures/golden_gate_three_fishing_boats.jpg}
    \includegraphics[width=0.18\linewidth]{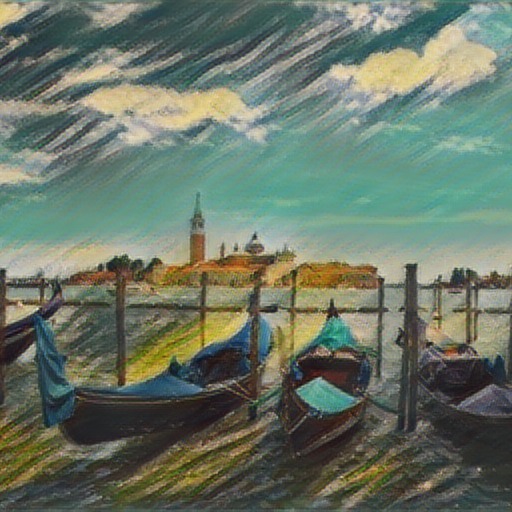}
    \includegraphics[width=0.18\linewidth]{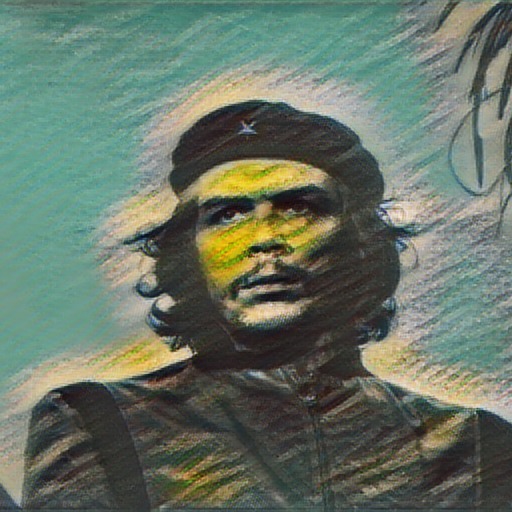} \\
    \includegraphics[width=0.18\linewidth]{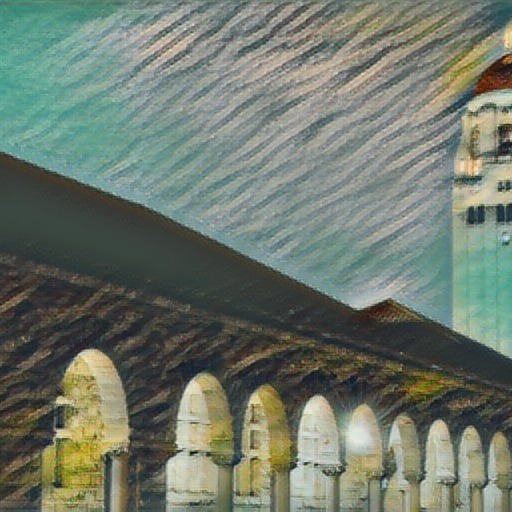}
    \includegraphics[width=0.18\linewidth]{figures/karya_three_fishing_boats.jpg}
    \includegraphics[width=0.18\linewidth]{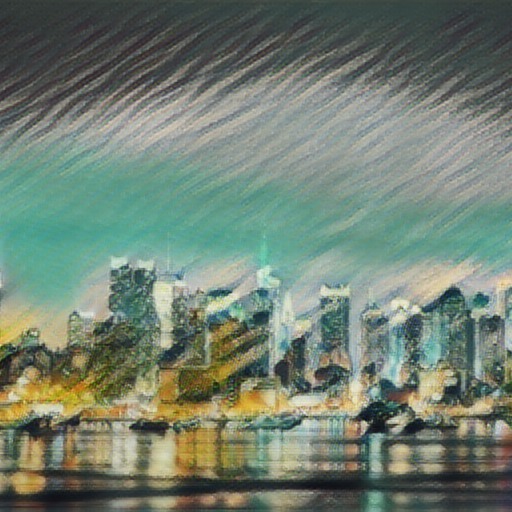}
    \includegraphics[width=0.18\linewidth]{figures/schultenhof_mettingen_bauerngarten_three_fishing_boats.jpg}
    \includegraphics[width=0.18\linewidth]{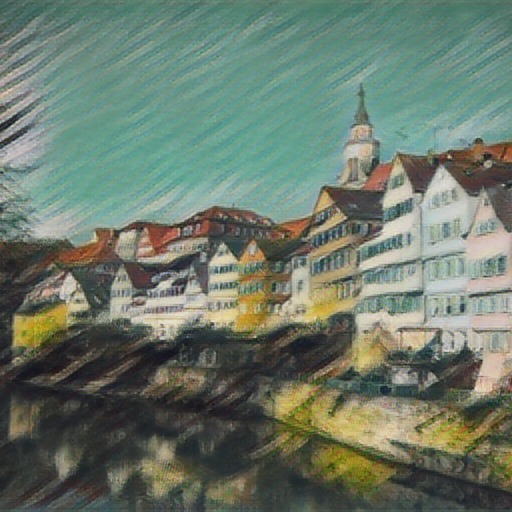}
\end{center}
\caption*{Claude Monet, {\em Three Fishing Boats} (1886).}
\end{figure}

\begin{figure}[ht]
\begin{center}
    \includegraphics[width=0.18\linewidth]{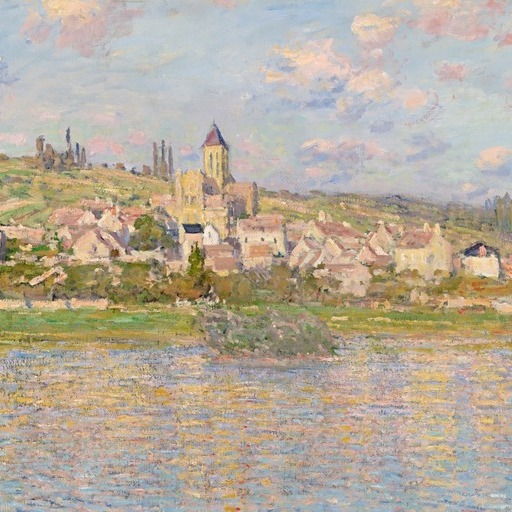}
    \includegraphics[width=0.18\linewidth]{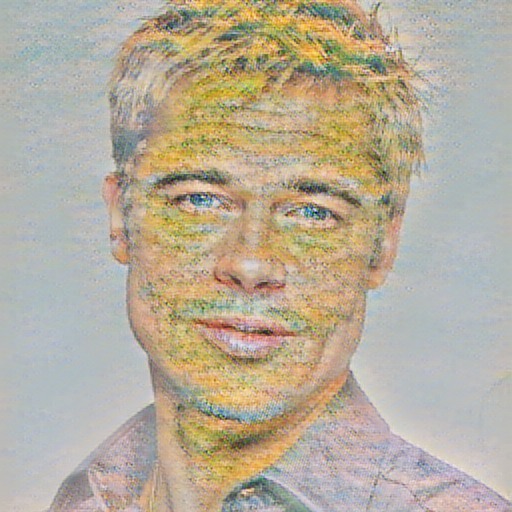}
    \includegraphics[width=0.18\linewidth]{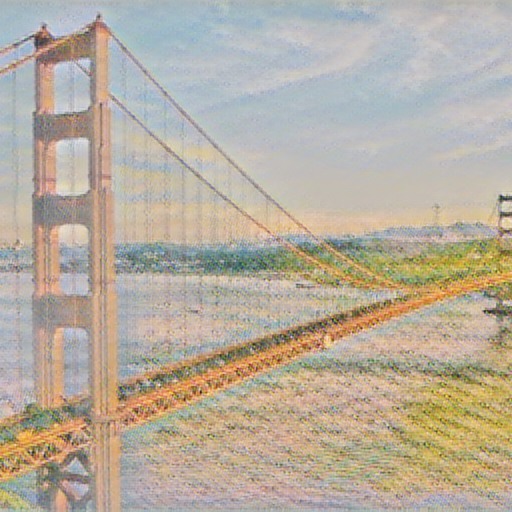}
    \includegraphics[width=0.18\linewidth]{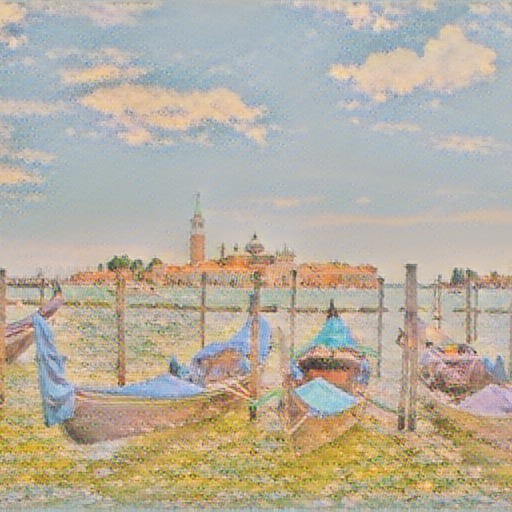}
    \includegraphics[width=0.18\linewidth]{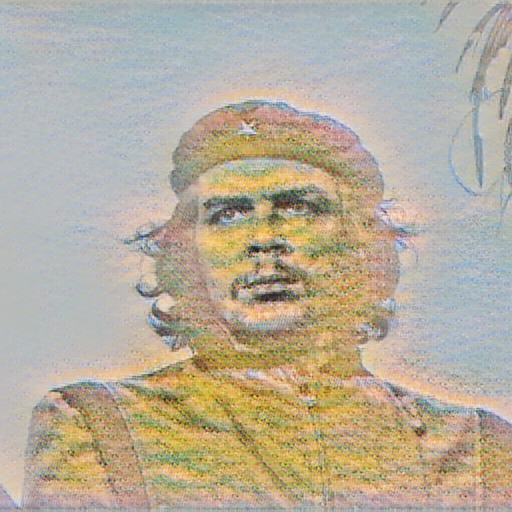} \\
    \includegraphics[width=0.18\linewidth]{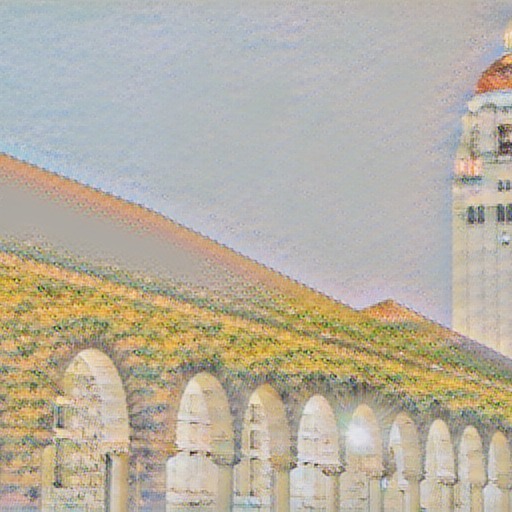}
    \includegraphics[width=0.18\linewidth]{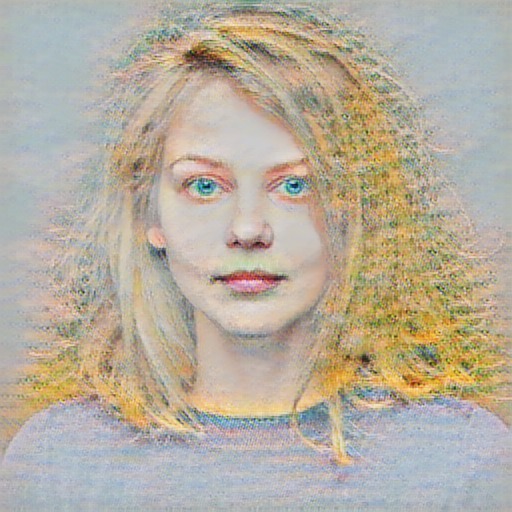}
    \includegraphics[width=0.18\linewidth]{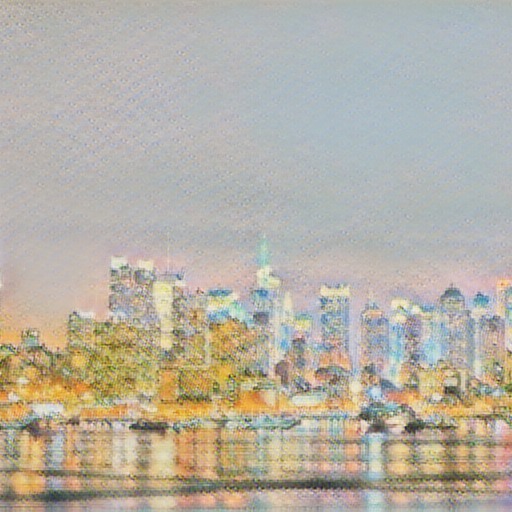}
    \includegraphics[width=0.18\linewidth]{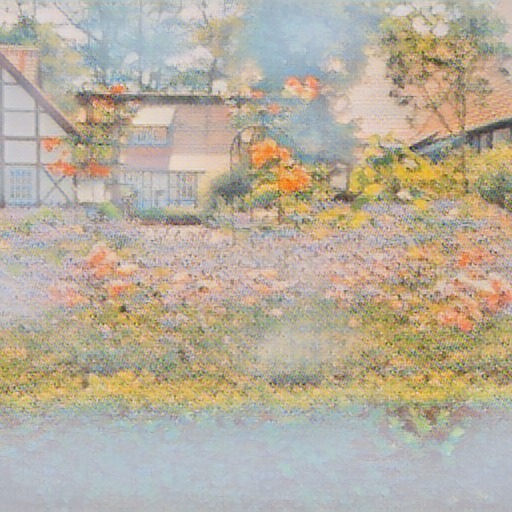}
    \includegraphics[width=0.18\linewidth]{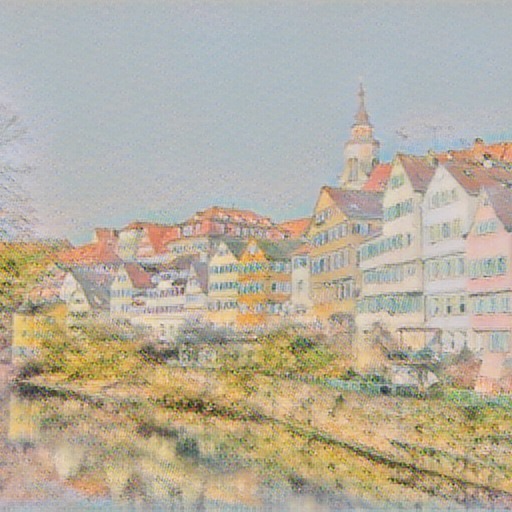}
\end{center}
\caption*{Claude Monet, {\em V\'{e}theuil} (1879).}
\end{figure}

\begin{figure}[ht]
\begin{center}
    \includegraphics[width=0.18\linewidth]{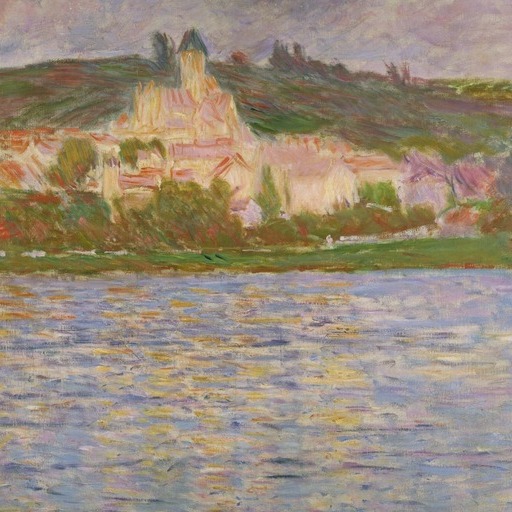}
    \includegraphics[width=0.18\linewidth]{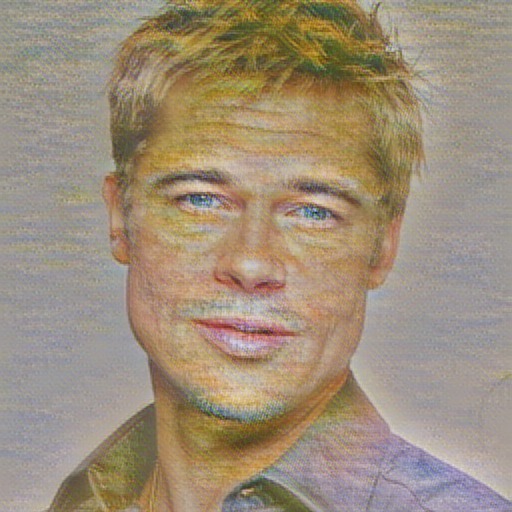}
    \includegraphics[width=0.18\linewidth]{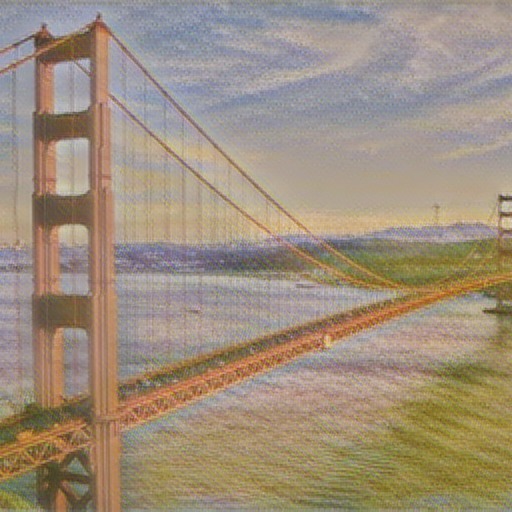}
    \includegraphics[width=0.18\linewidth]{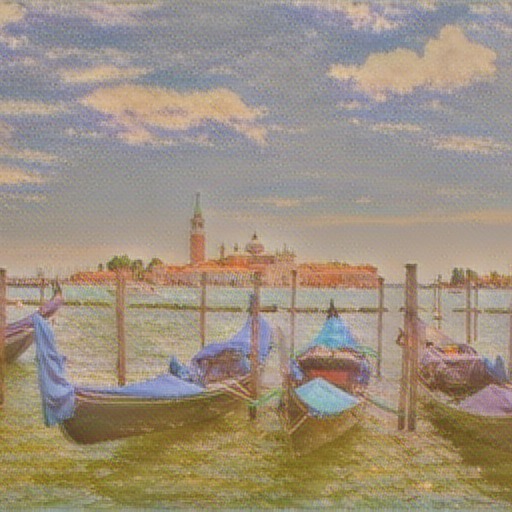}
    \includegraphics[width=0.18\linewidth]{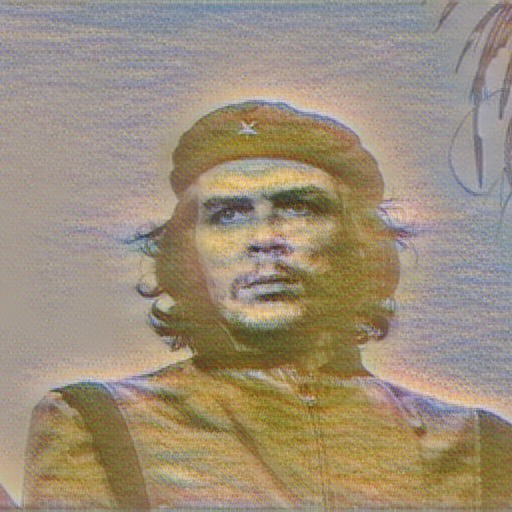} \\
    \includegraphics[width=0.18\linewidth]{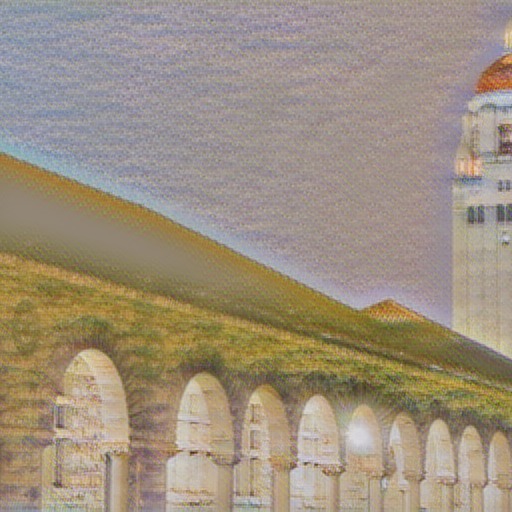}
    \includegraphics[width=0.18\linewidth]{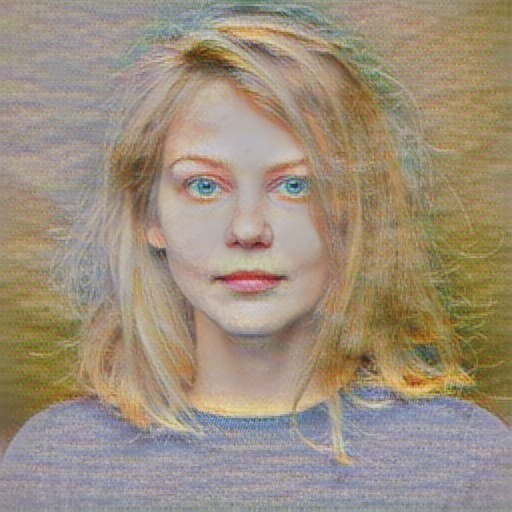}
    \includegraphics[width=0.18\linewidth]{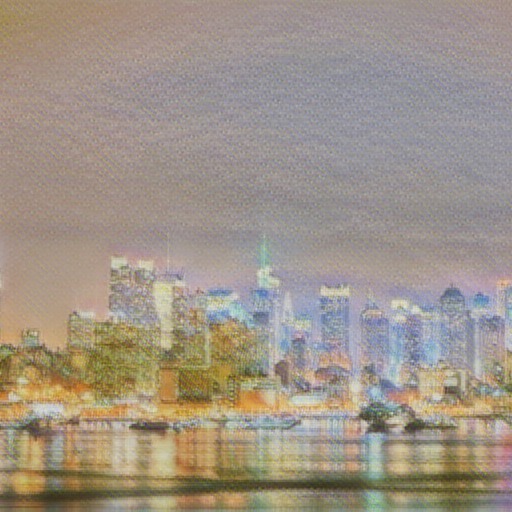}
    \includegraphics[width=0.18\linewidth]{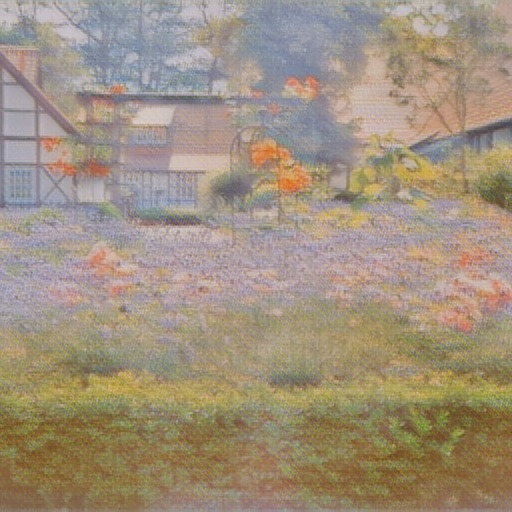}
    \includegraphics[width=0.18\linewidth]{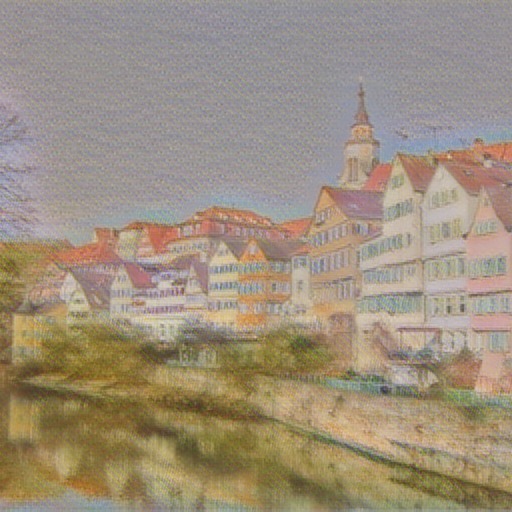}
\end{center}
\caption*{Claude Monet, {\em V\'{e}theuil} (1902).}
\end{figure}

\clearpage
\begin{figure}[ht]
\begin{center}
    \includegraphics[width=0.18\linewidth]{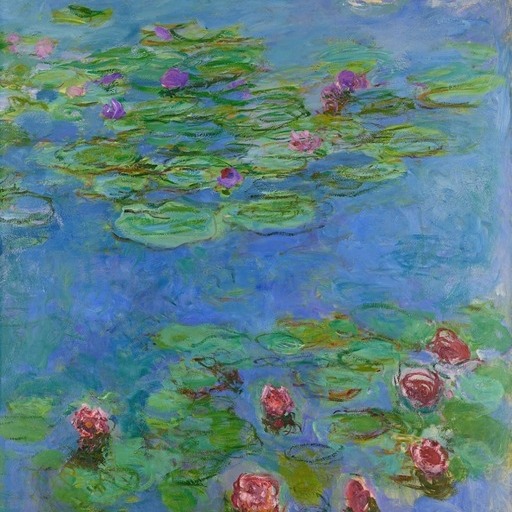}
    \includegraphics[width=0.18\linewidth]{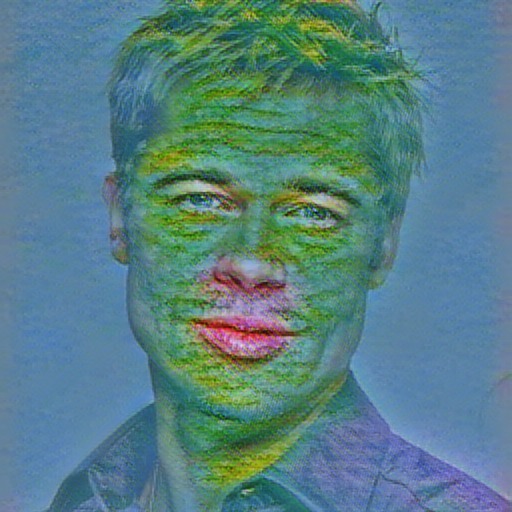}
    \includegraphics[width=0.18\linewidth]{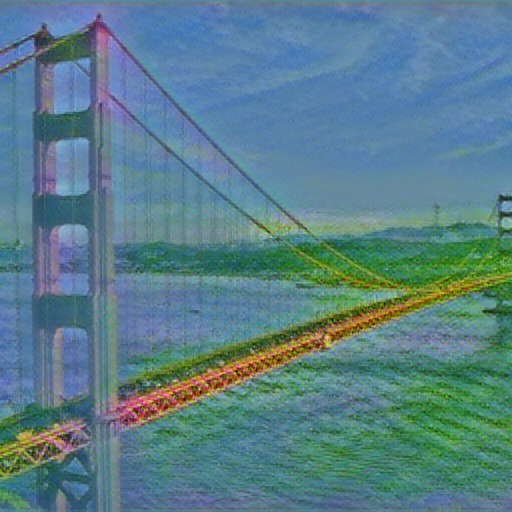}
    \includegraphics[width=0.18\linewidth]{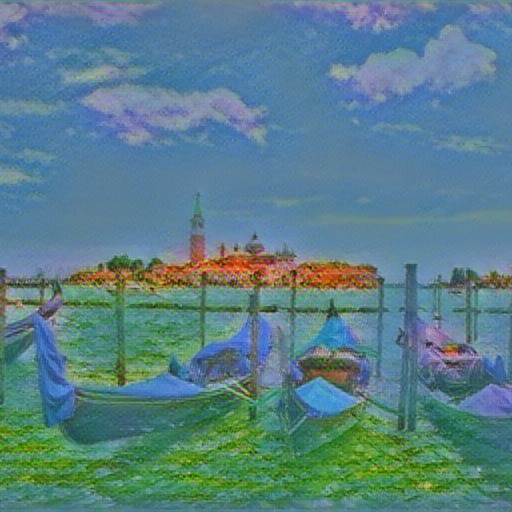}
    \includegraphics[width=0.18\linewidth]{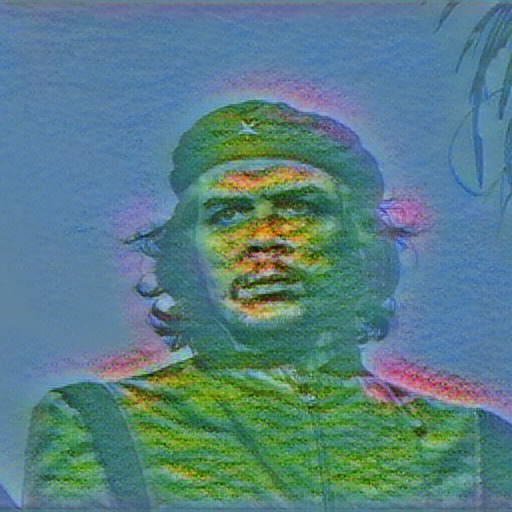} \\
    \includegraphics[width=0.18\linewidth]{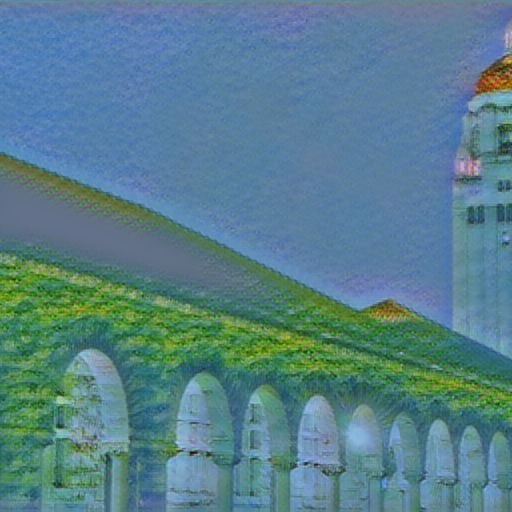}    \includegraphics[width=0.18\linewidth]{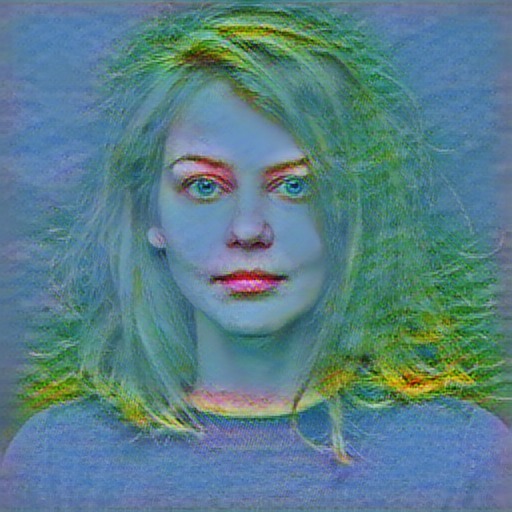}
    \includegraphics[width=0.18\linewidth]{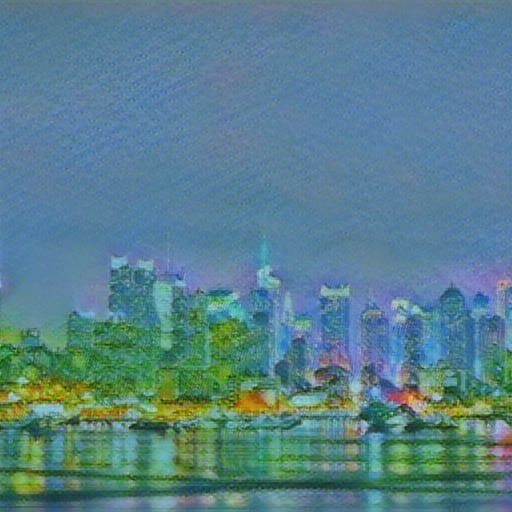}
    \includegraphics[width=0.18\linewidth]{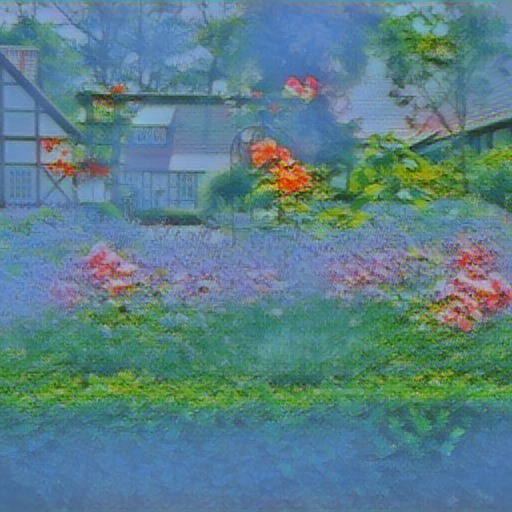}
    \includegraphics[width=0.18\linewidth]{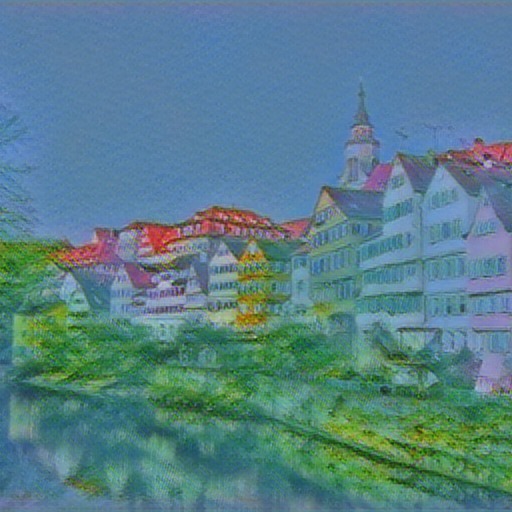}
\end{center}
\caption*{Claude Monet, {\em Water Lilies} (ca. 1914-1917).}
\end{figure}

\subsection*{Varied pastiches}

\begin{figure}[ht]
\begin{center}
    \includegraphics[width=0.18\linewidth]{figures/bicentennial_print.jpg}
    \includegraphics[width=0.18\linewidth]{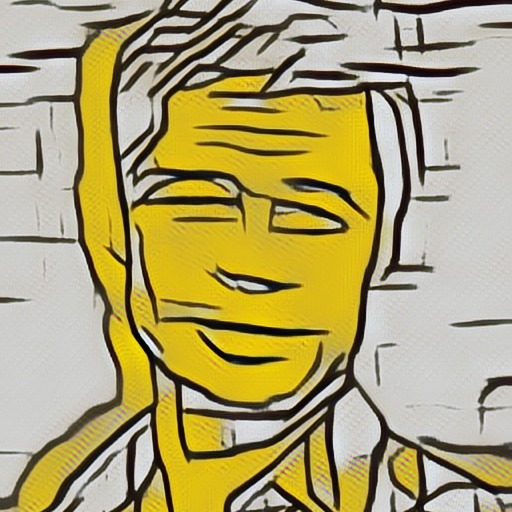}
    \includegraphics[width=0.18\linewidth]{figures/golden_gate_bicentennial_print.jpg}
    \includegraphics[width=0.18\linewidth]{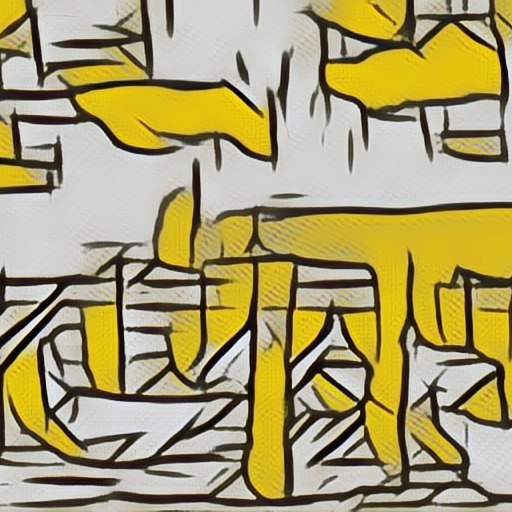}
    \includegraphics[width=0.18\linewidth]{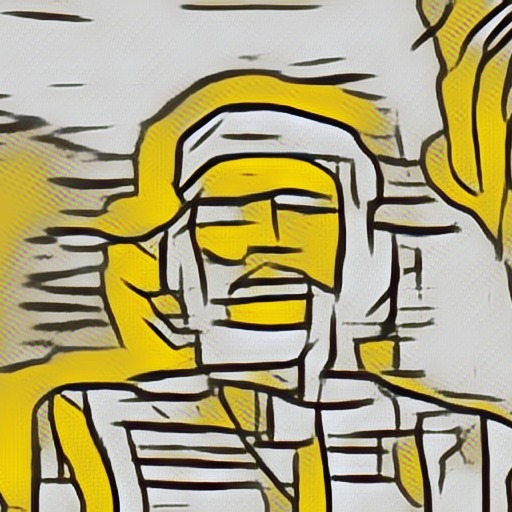} \\
    \includegraphics[width=0.18\linewidth]{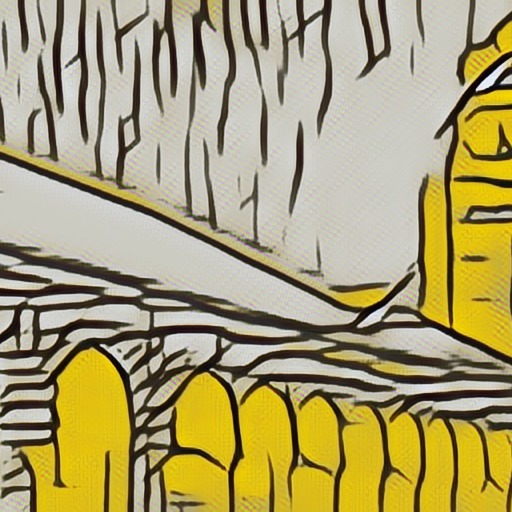}
    \includegraphics[width=0.18\linewidth]{figures/karya_bicentennial_print.jpg}
    \includegraphics[width=0.18\linewidth]{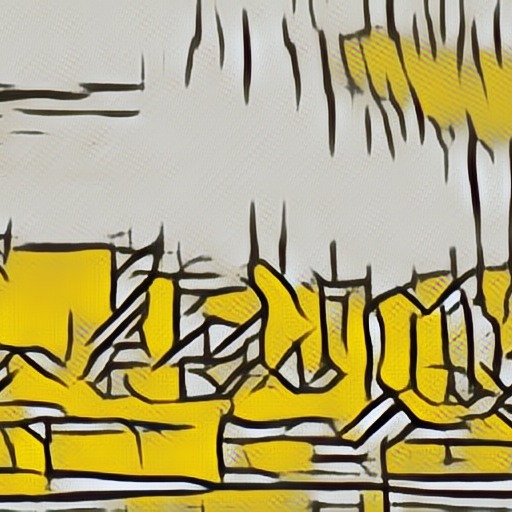}
    \includegraphics[width=0.18\linewidth]{figures/schultenhof_mettingen_bauerngarten_bicentennial_print.jpg}
    \includegraphics[width=0.18\linewidth]{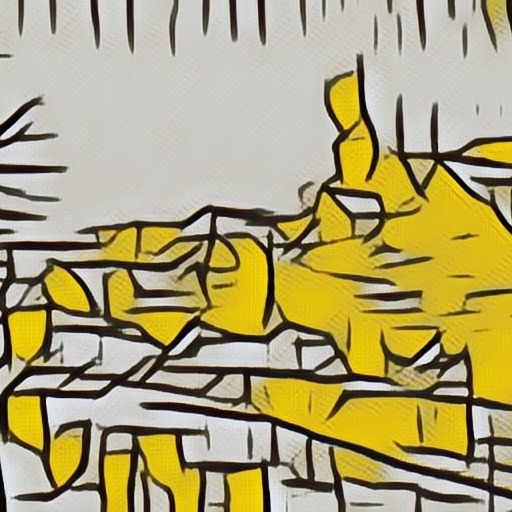}
\end{center}
\caption*{Roy Lichtenstein, {\em Bicentennial Print} (1975).}
\end{figure}

\begin{figure}[ht]
\begin{center}
    \includegraphics[width=0.18\linewidth]{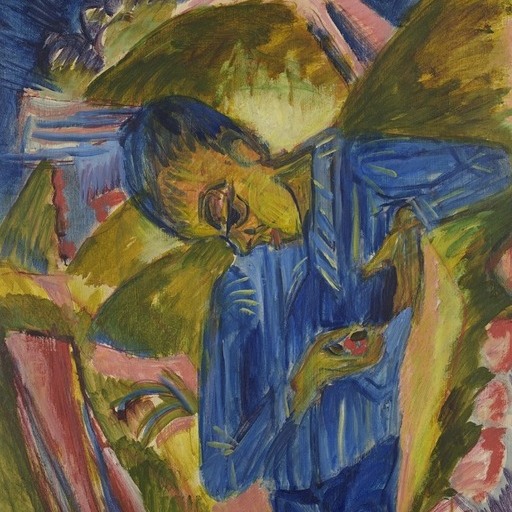}
    \includegraphics[width=0.18\linewidth]{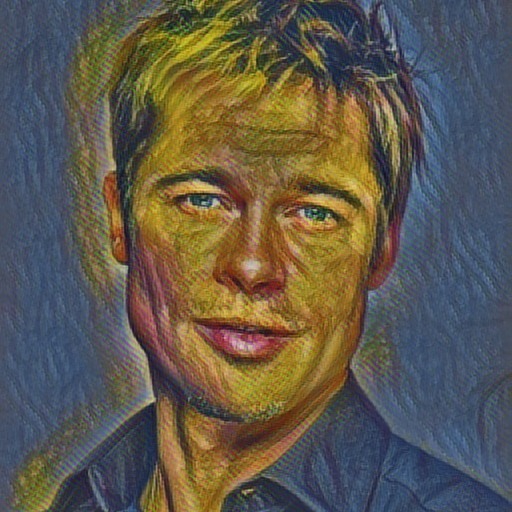}
    \includegraphics[width=0.18\linewidth]{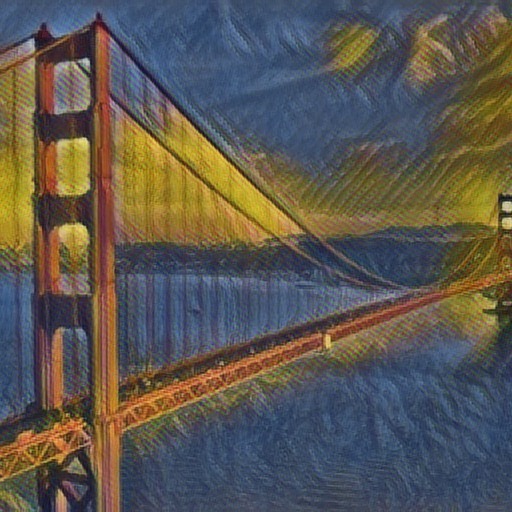}
    \includegraphics[width=0.18\linewidth]{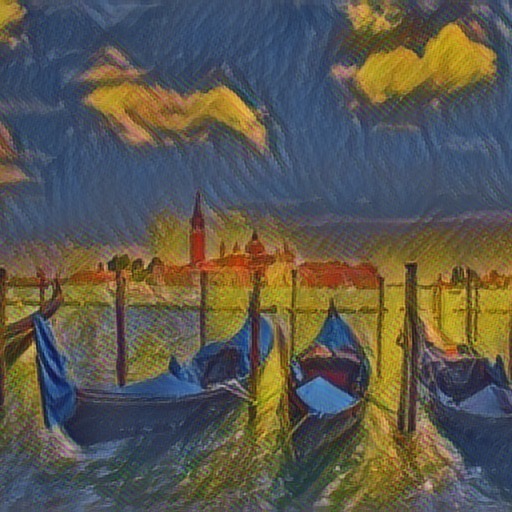}
    \includegraphics[width=0.18\linewidth]{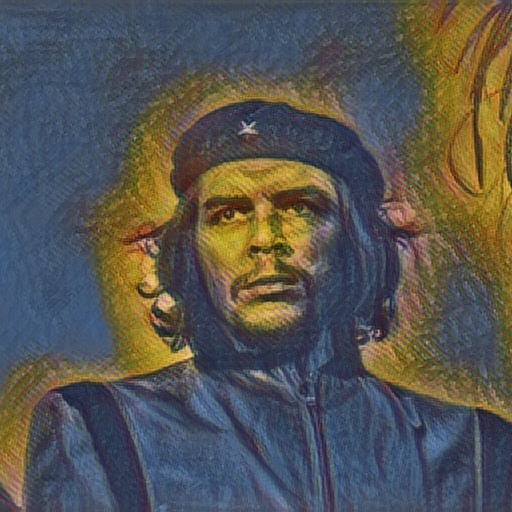} \\
    \includegraphics[width=0.18\linewidth]{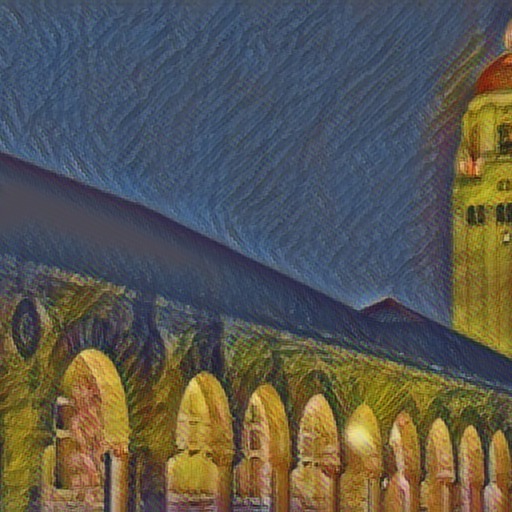}
    \includegraphics[width=0.18\linewidth]{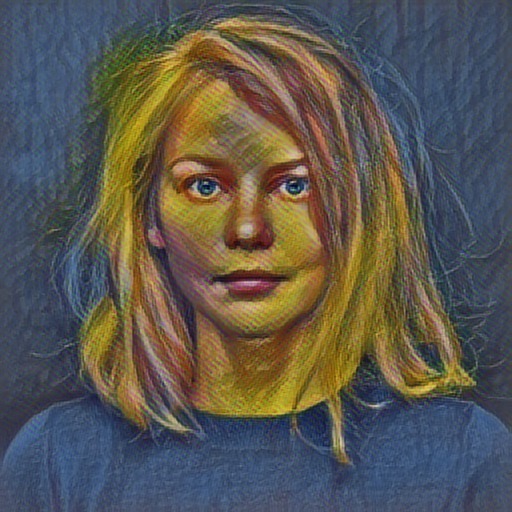}
    \includegraphics[width=0.18\linewidth]{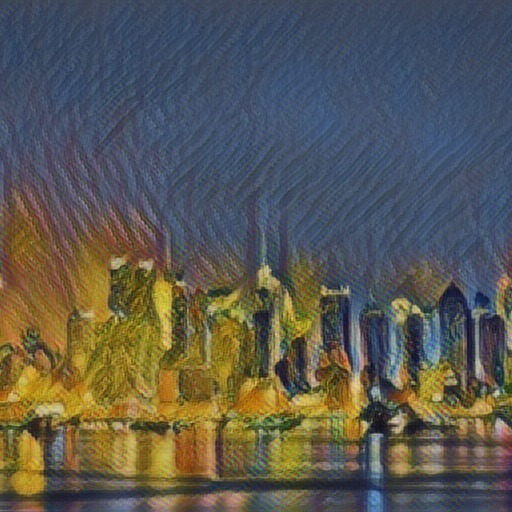}
    \includegraphics[width=0.18\linewidth]{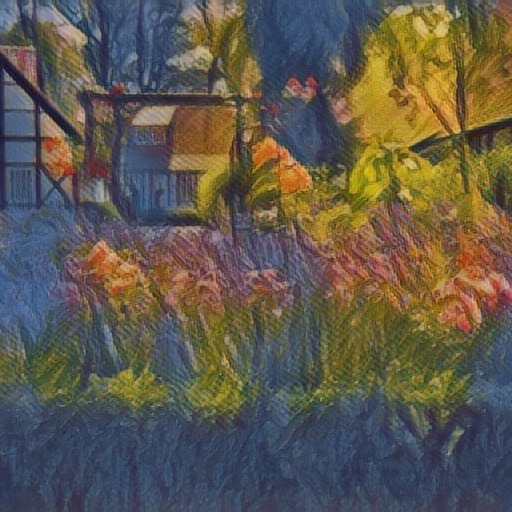}
    \includegraphics[width=0.18\linewidth]{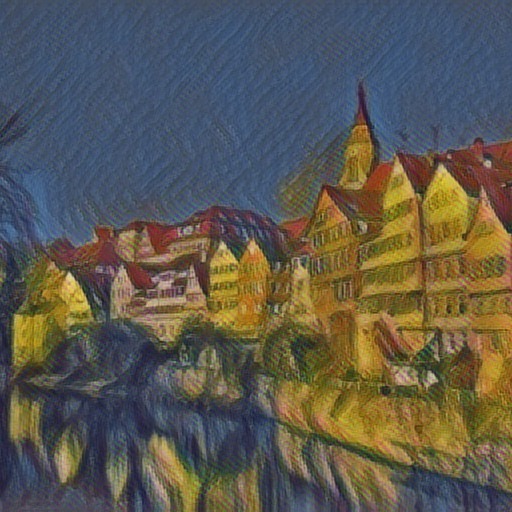}
\end{center}
\caption*{Ernst Ludwig Kirchner, {\em Boy with Sweets} (1918).}
\end{figure}

\clearpage
\begin{figure}[ht]
\begin{center}
    \includegraphics[width=0.18\linewidth]{figures/cassis_cap_lombard_opus_196.jpg}
    \includegraphics[width=0.18\linewidth]{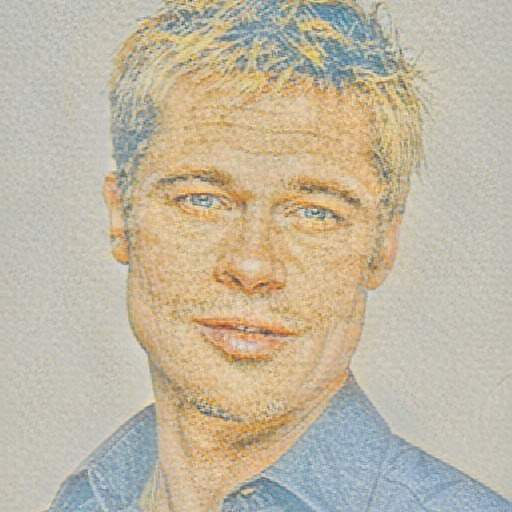}
    \includegraphics[width=0.18\linewidth]{figures/golden_gate_cassis_cap_lombard_opus_196.jpg}
    \includegraphics[width=0.18\linewidth]{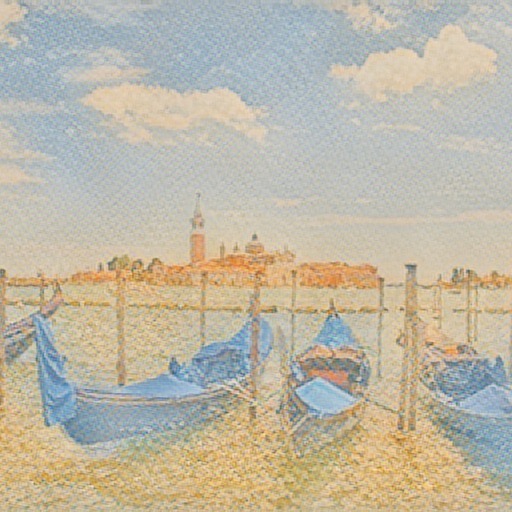}
    \includegraphics[width=0.18\linewidth]{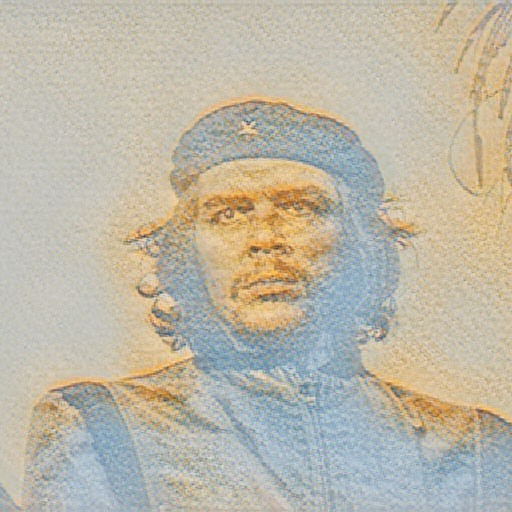} \\
    \includegraphics[width=0.18\linewidth]{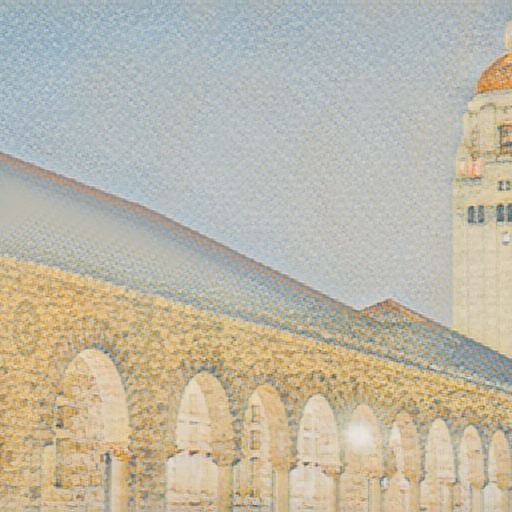}
    \includegraphics[width=0.18\linewidth]{figures/karya_cassis_cap_lombard_opus_196.jpg}
    \includegraphics[width=0.18\linewidth]{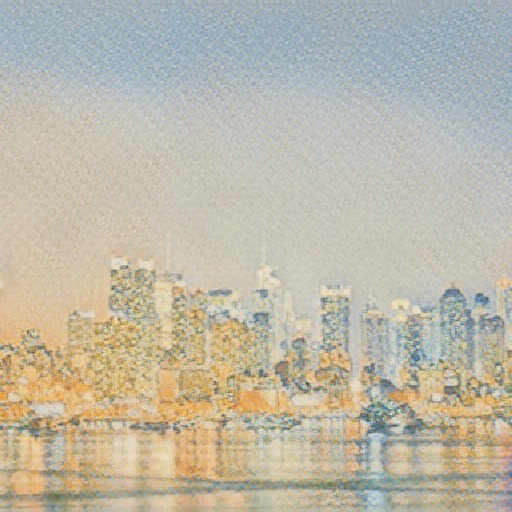}
    \includegraphics[width=0.18\linewidth]{figures/schultenhof_mettingen_bauerngarten_cassis_cap_lombard_opus_196.jpg}
    \includegraphics[width=0.18\linewidth]{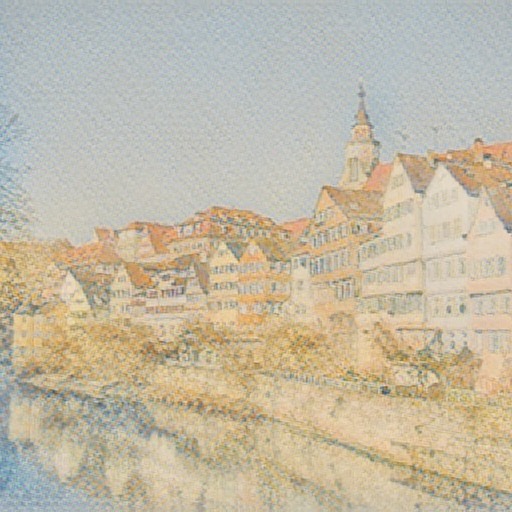}
\end{center}
\caption*{Paul Signac, {\em Cassis, Cap Lombard, Opus 196} (1889).}
\end{figure}

\begin{figure}[ht]
\begin{center}
    \includegraphics[width=0.18\linewidth]{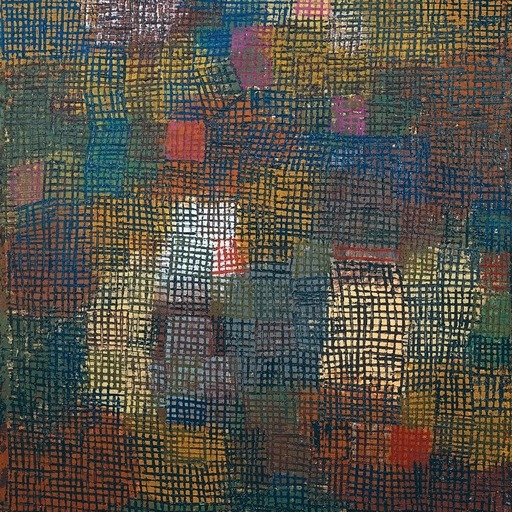}
    \includegraphics[width=0.18\linewidth]{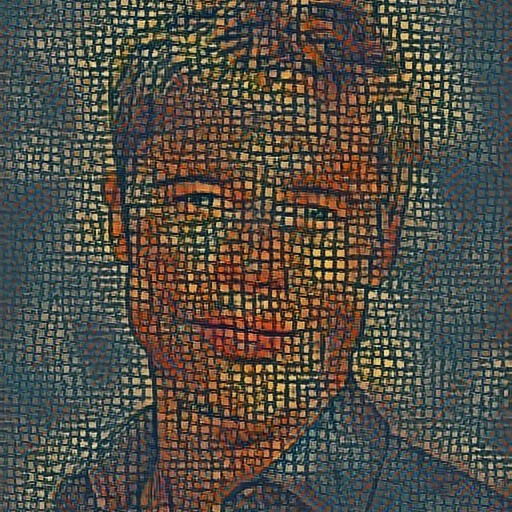}
    \includegraphics[width=0.18\linewidth]{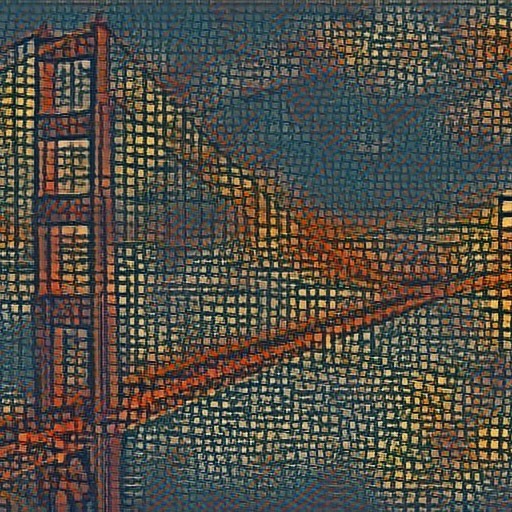}
    \includegraphics[width=0.18\linewidth]{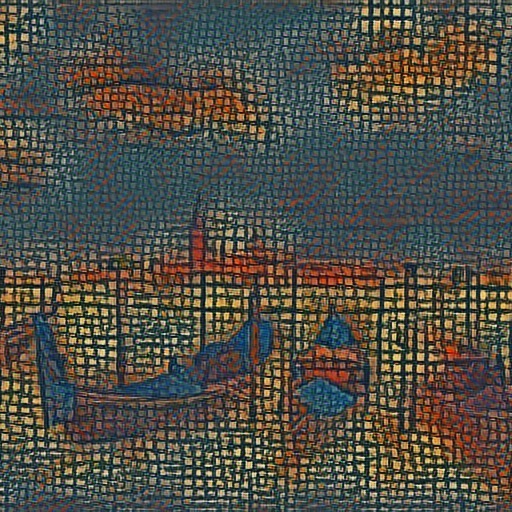}
    \includegraphics[width=0.18\linewidth]{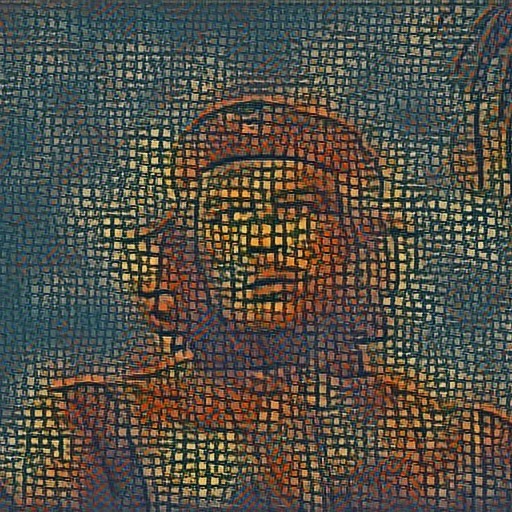} \\
    \includegraphics[width=0.18\linewidth]{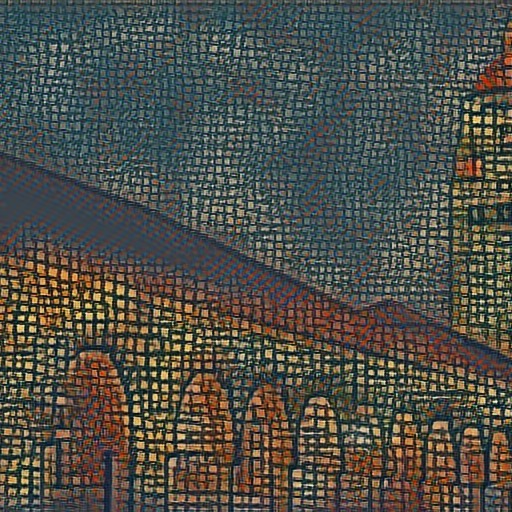}
    \includegraphics[width=0.18\linewidth]{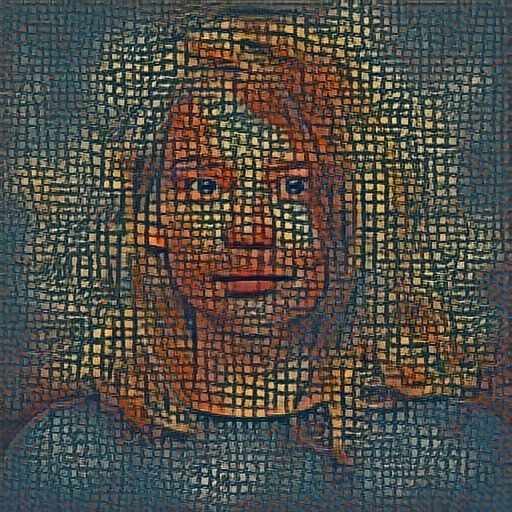}
    \includegraphics[width=0.18\linewidth]{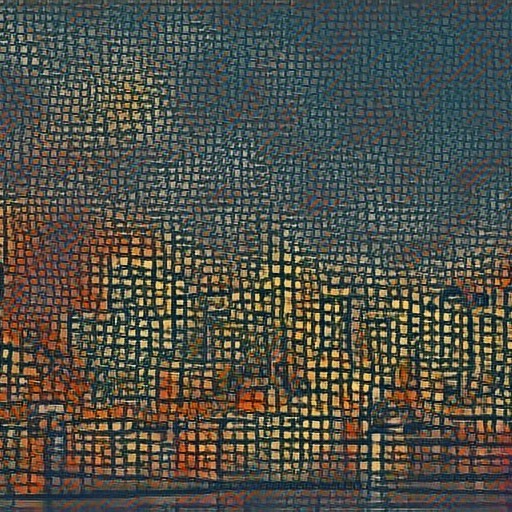}
    \includegraphics[width=0.18\linewidth]{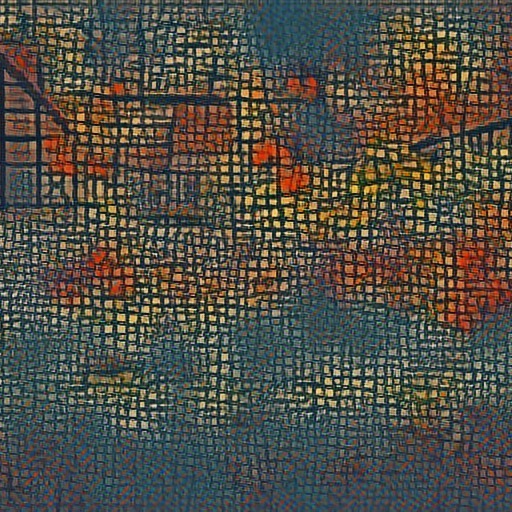}
    \includegraphics[width=0.18\linewidth]{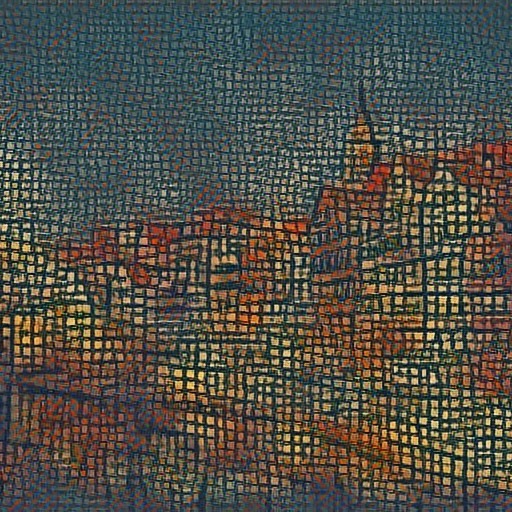}
\end{center}
\caption*{Paul Klee, {\em Colors from a Distance} (1932).}
\end{figure}

\begin{figure}[ht]
\begin{center}
    \includegraphics[width=0.18\linewidth]{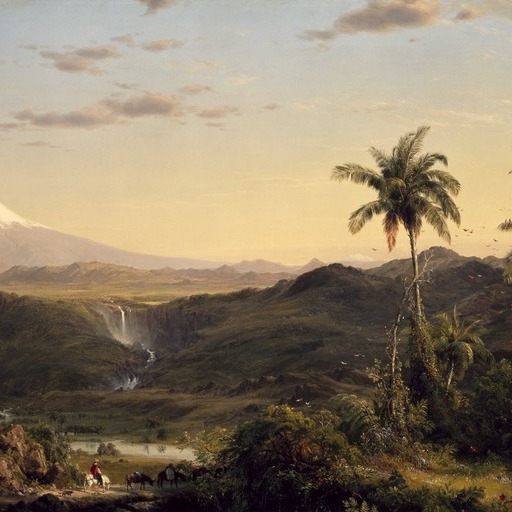}
    \includegraphics[width=0.18\linewidth]{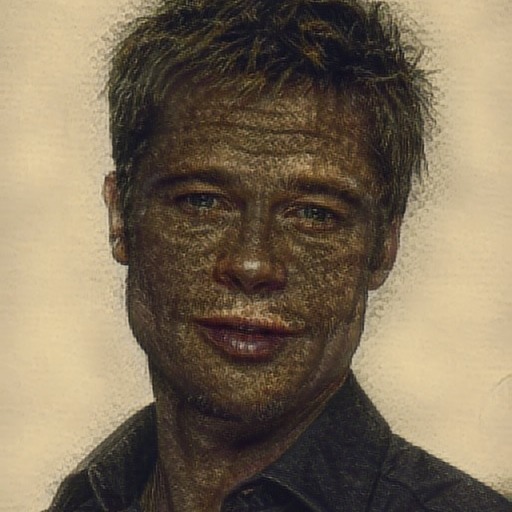}
    \includegraphics[width=0.18\linewidth]{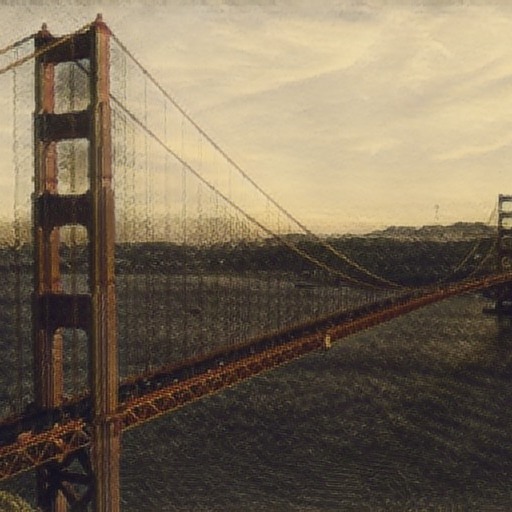}
    \includegraphics[width=0.18\linewidth]{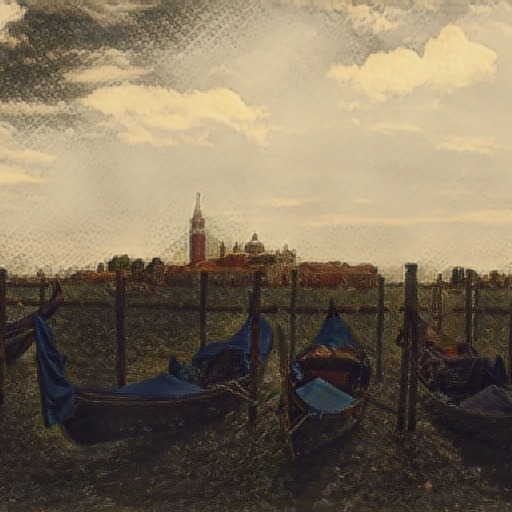}
    \includegraphics[width=0.18\linewidth]{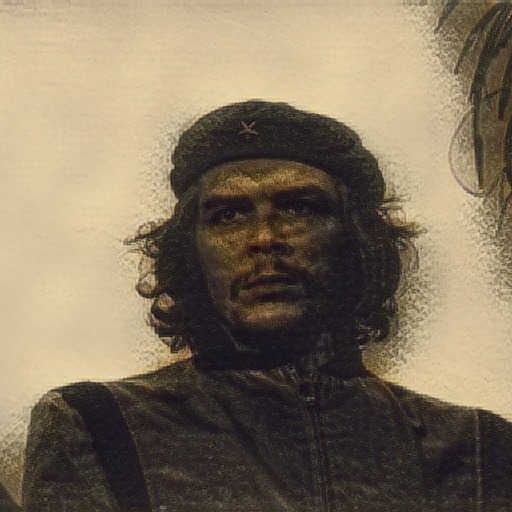} \\
    \includegraphics[width=0.18\linewidth]{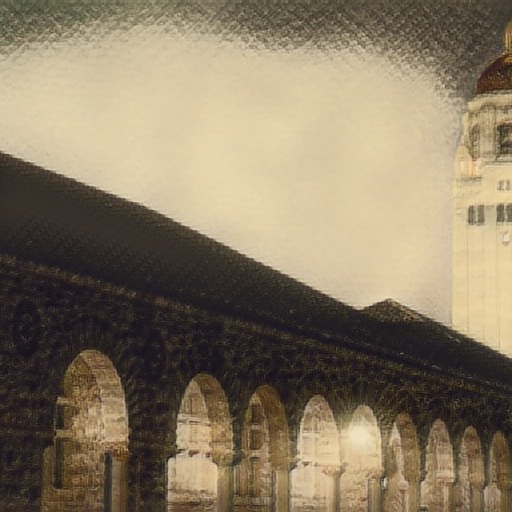}
    \includegraphics[width=0.18\linewidth]{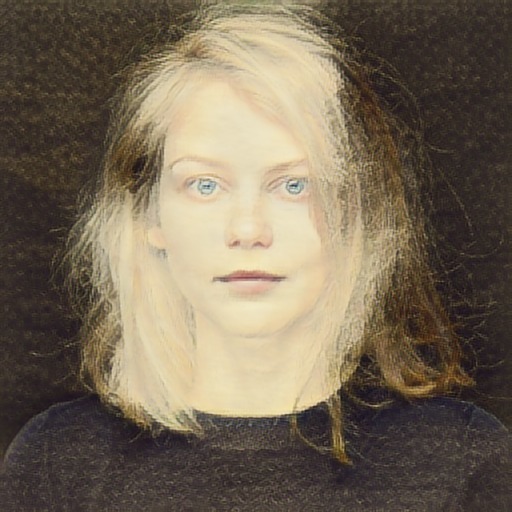}
    \includegraphics[width=0.18\linewidth]{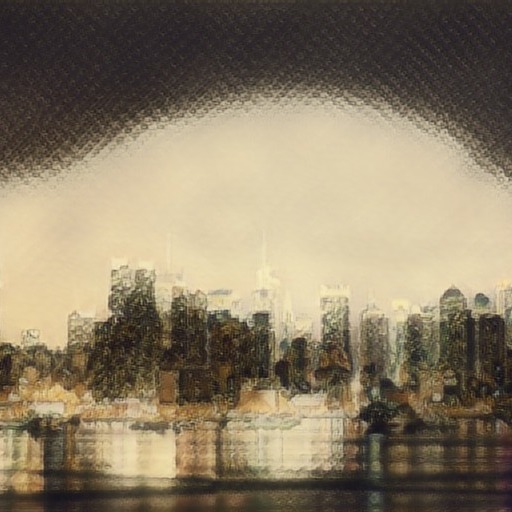}
    \includegraphics[width=0.18\linewidth]{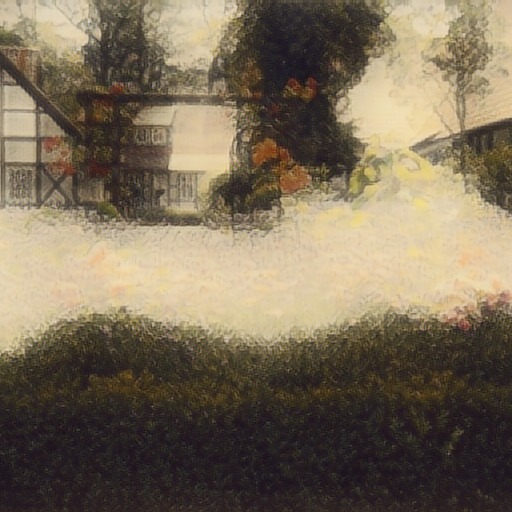}
    \includegraphics[width=0.18\linewidth]{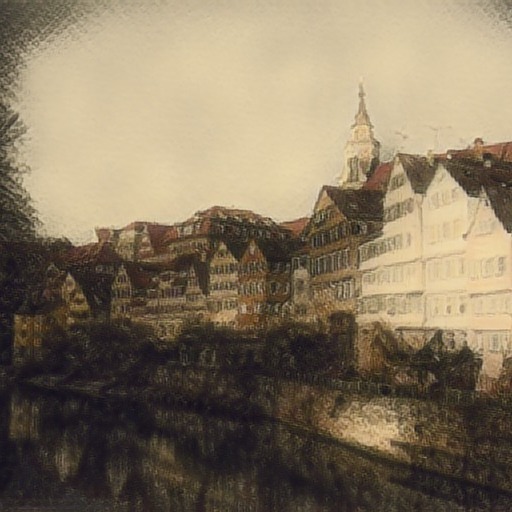}
\end{center}
\caption*{Frederic Edwin Church, {\em Cotopaxi} (1855).}
\end{figure}

\clearpage
\begin{figure}[ht]
\begin{center}
    \includegraphics[width=0.18\linewidth]{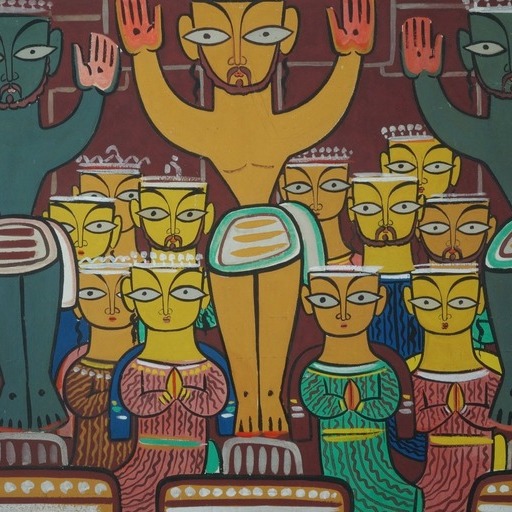}
    \includegraphics[width=0.18\linewidth]{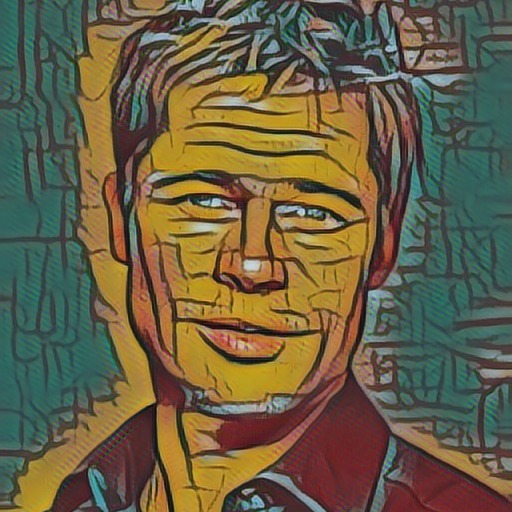}
    \includegraphics[width=0.18\linewidth]{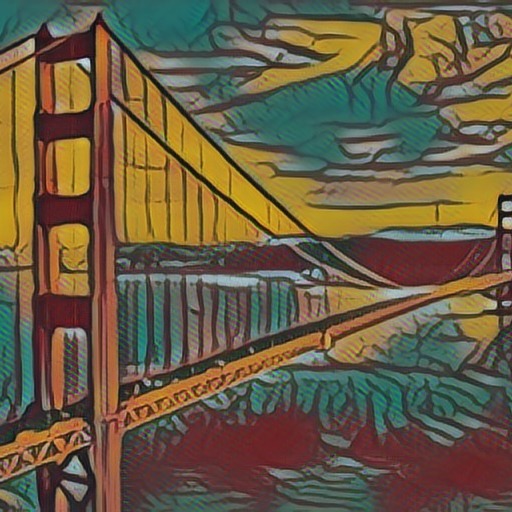}
    \includegraphics[width=0.18\linewidth]{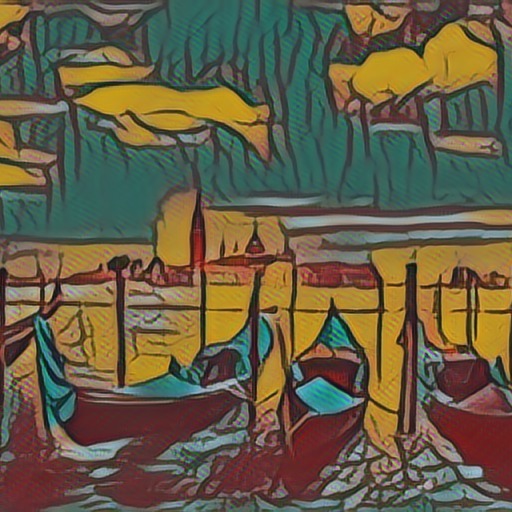}
    \includegraphics[width=0.18\linewidth]{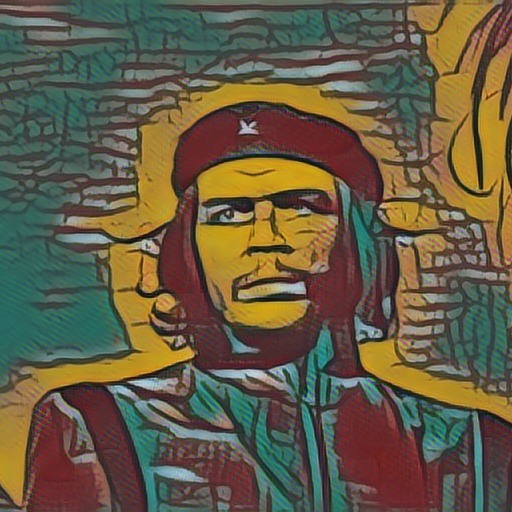} \\
    \includegraphics[width=0.18\linewidth]{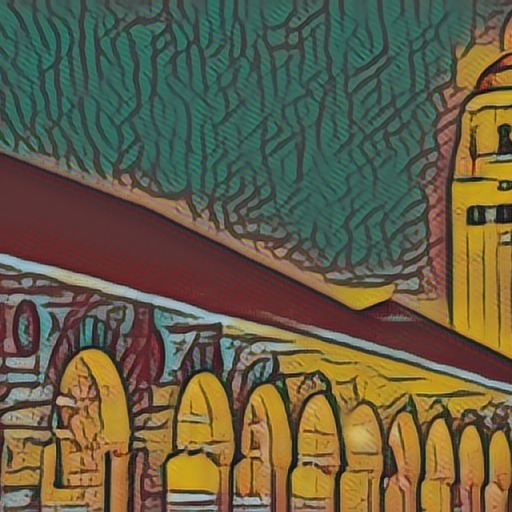}
    \includegraphics[width=0.18\linewidth]{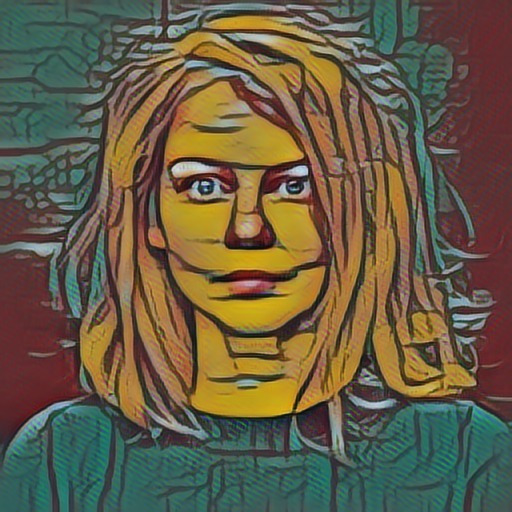}
    \includegraphics[width=0.18\linewidth]{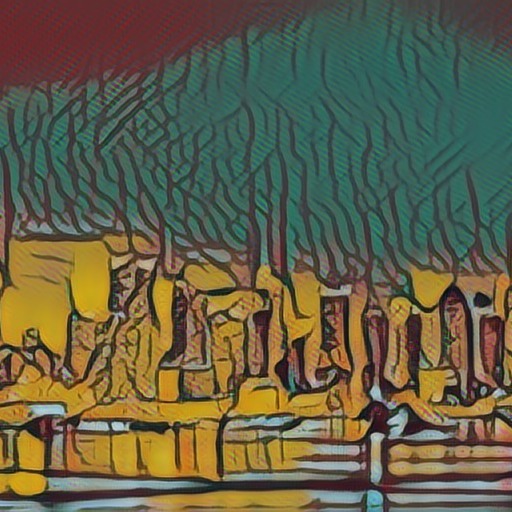}
    \includegraphics[width=0.18\linewidth]{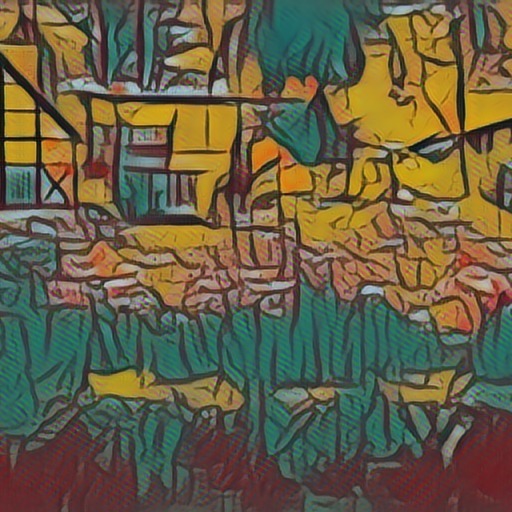}
    \includegraphics[width=0.18\linewidth]{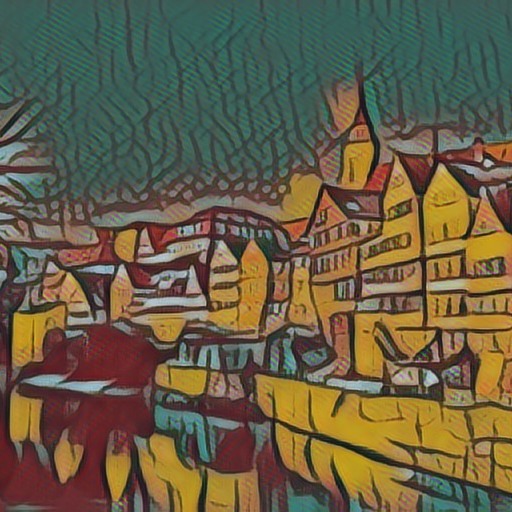}
\end{center}
\caption*{Jamini Roy, {\em Crucifixion}.}
\end{figure}

\begin{figure}[ht]
\begin{center}
    \includegraphics[width=0.18\linewidth]{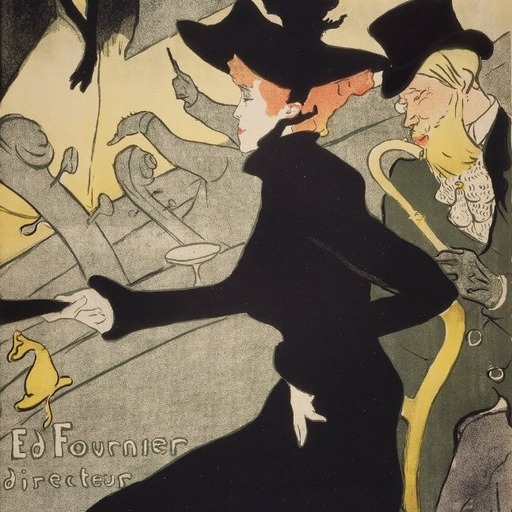}
    \includegraphics[width=0.18\linewidth]{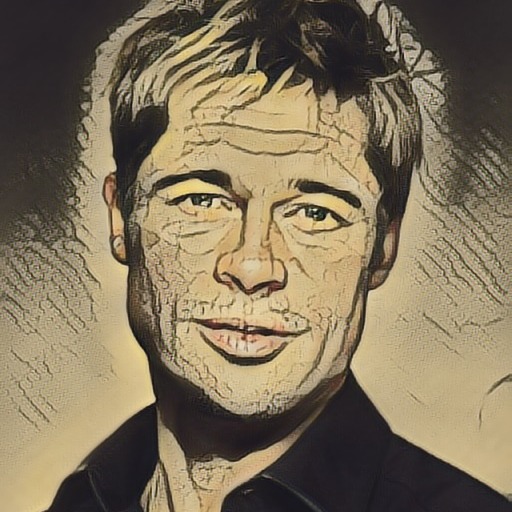}
    \includegraphics[width=0.18\linewidth]{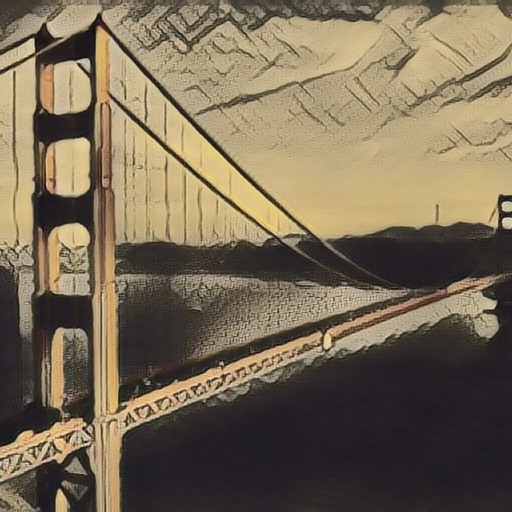}
    \includegraphics[width=0.18\linewidth]{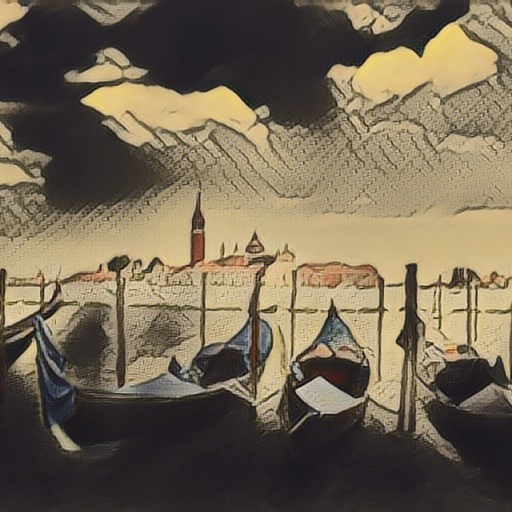}
    \includegraphics[width=0.18\linewidth]{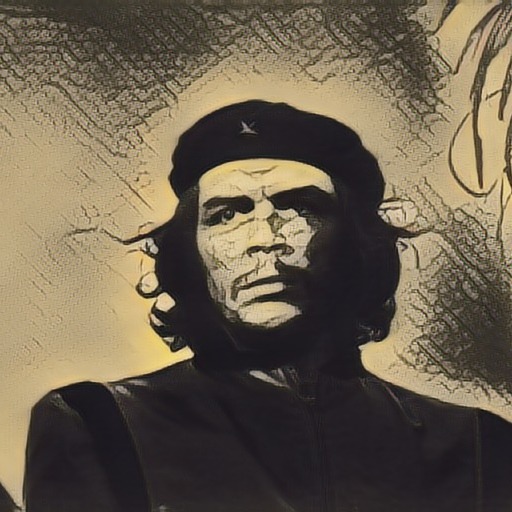} \\
    \includegraphics[width=0.18\linewidth]{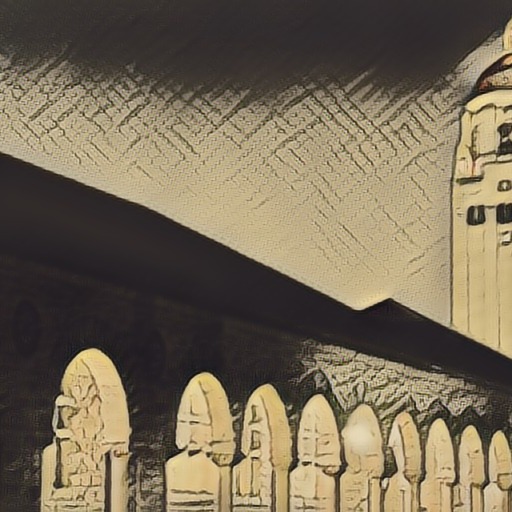}
    \includegraphics[width=0.18\linewidth]{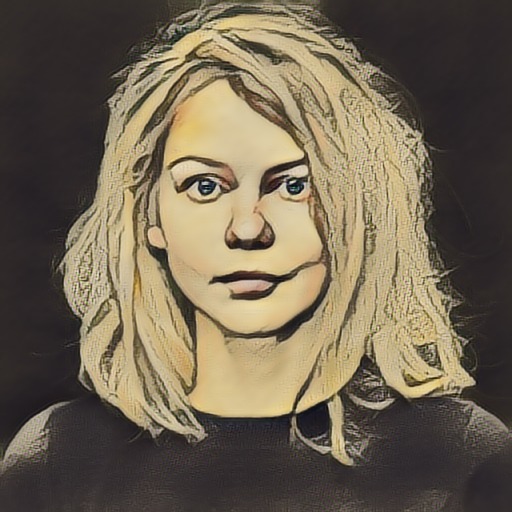}
    \includegraphics[width=0.18\linewidth]{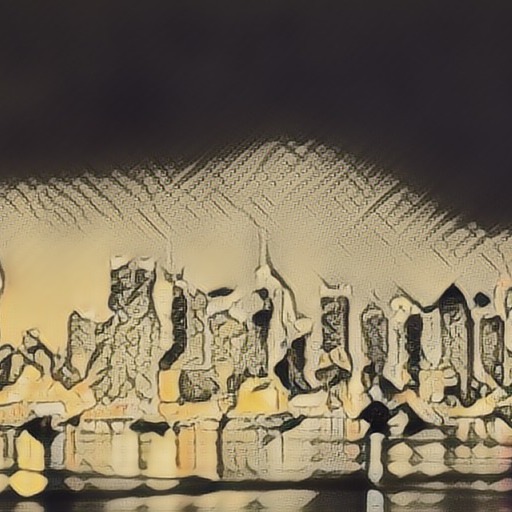}
    \includegraphics[width=0.18\linewidth]{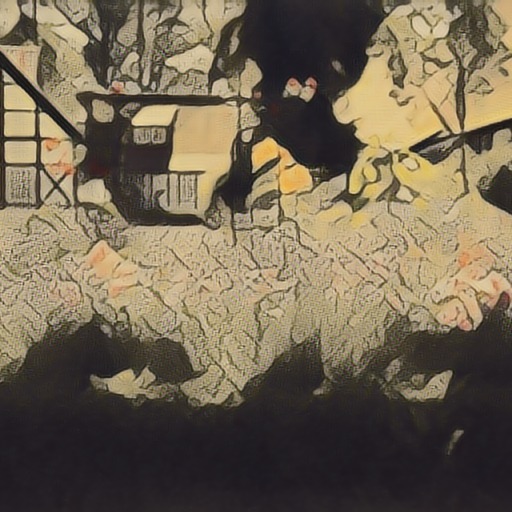}
    \includegraphics[width=0.18\linewidth]{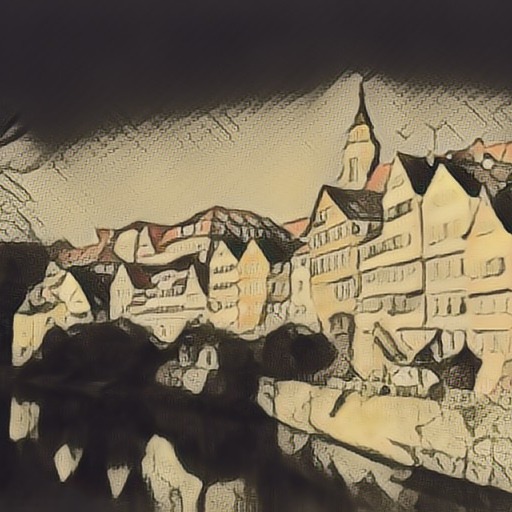}
\end{center}
\caption*{Henri de Toulouse-Lautrec, {\em Divan Japonais} (1893).}
\end{figure}

\begin{figure}[ht]
\begin{center}
    \includegraphics[width=0.18\linewidth]{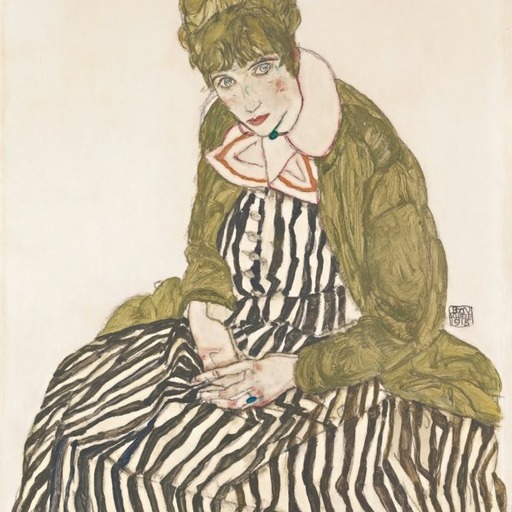}
    \includegraphics[width=0.18\linewidth]{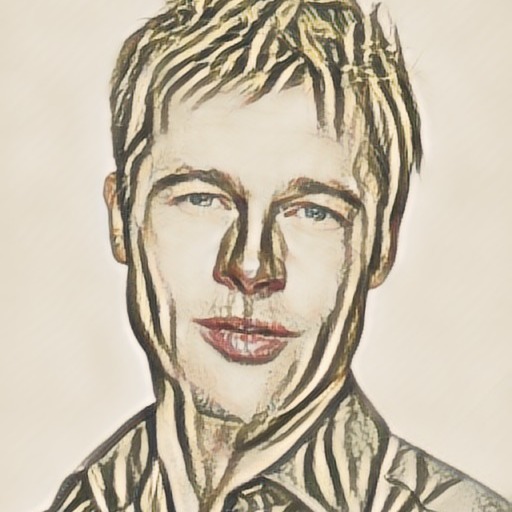}
    \includegraphics[width=0.18\linewidth]{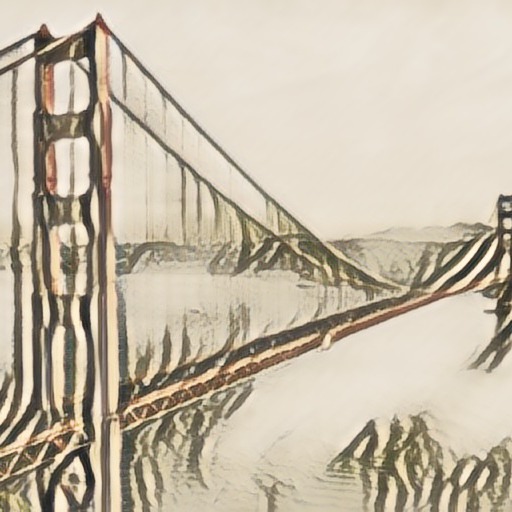}
    \includegraphics[width=0.18\linewidth]{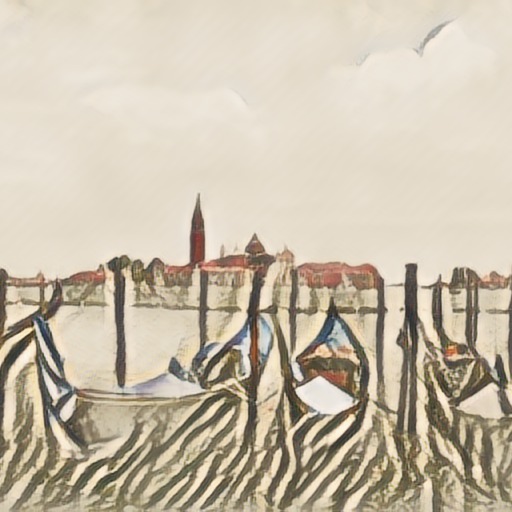}
    \includegraphics[width=0.18\linewidth]{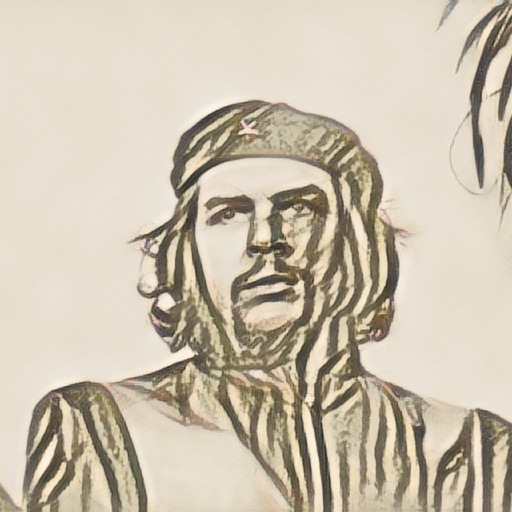} \\
    \includegraphics[width=0.18\linewidth]{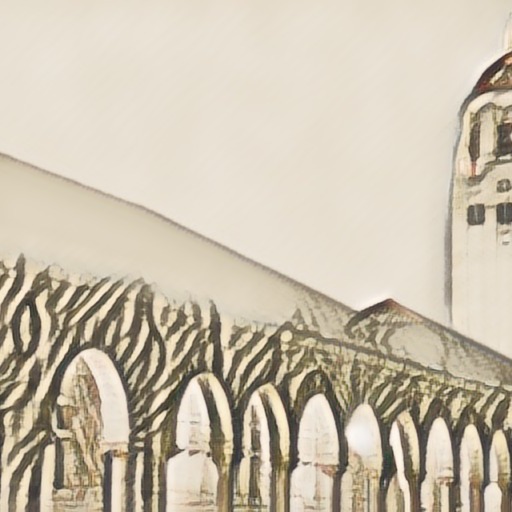}
    \includegraphics[width=0.18\linewidth]{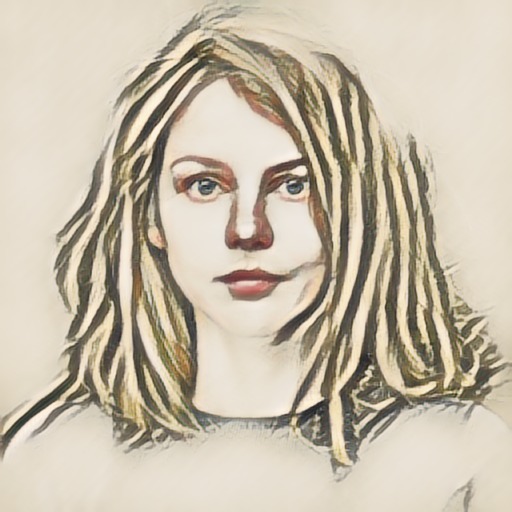}
    \includegraphics[width=0.18\linewidth]{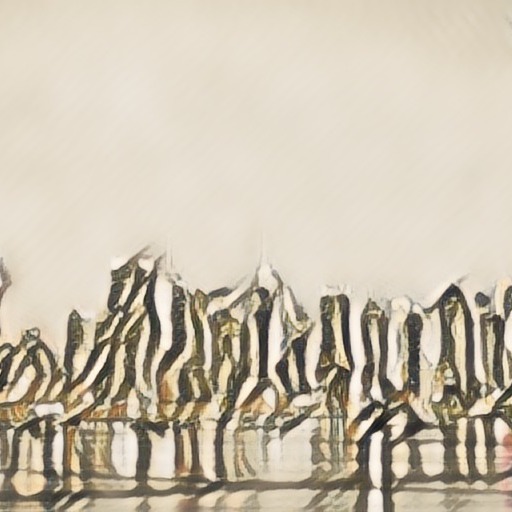}
    \includegraphics[width=0.18\linewidth]{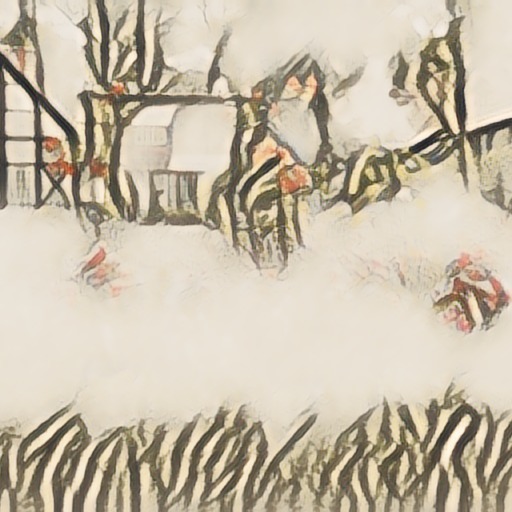}
    \includegraphics[width=0.18\linewidth]{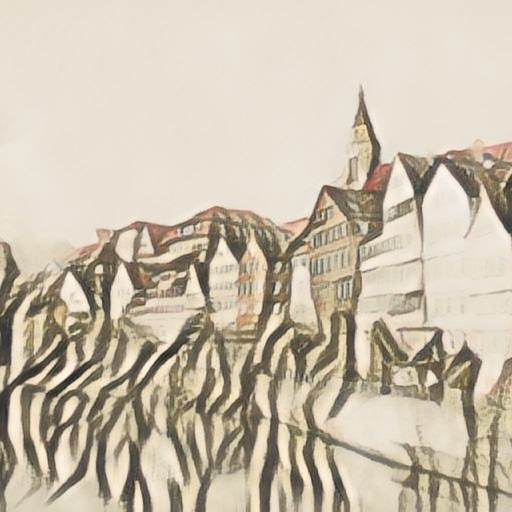}
\end{center}
\caption*{Egon Schiele, {\em Edith with Striped Dress, Sitting} (1915).}
\end{figure}

\clearpage
\begin{figure}[ht]
\begin{center}
    \includegraphics[width=0.18\linewidth]{figures/head_of_a_clown.jpg}
    \includegraphics[width=0.18\linewidth]{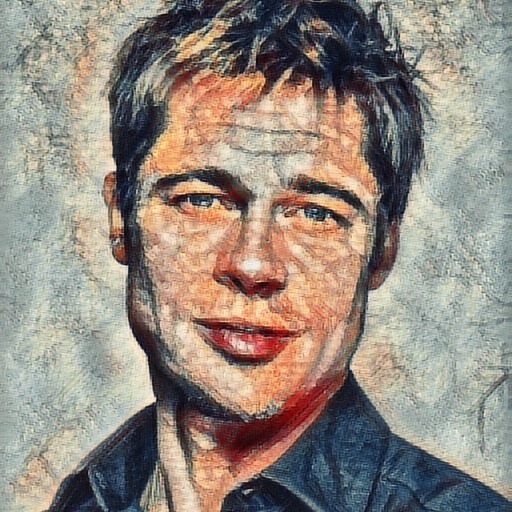}
    \includegraphics[width=0.18\linewidth]{figures/golden_gate_head_of_a_clown.jpg}
    \includegraphics[width=0.18\linewidth]{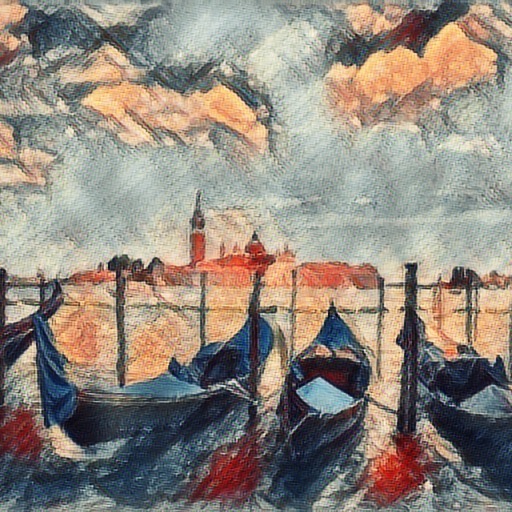}
    \includegraphics[width=0.18\linewidth]{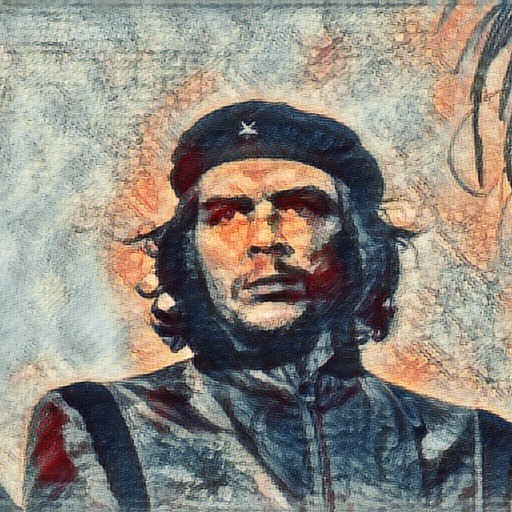} \\
    \includegraphics[width=0.18\linewidth]{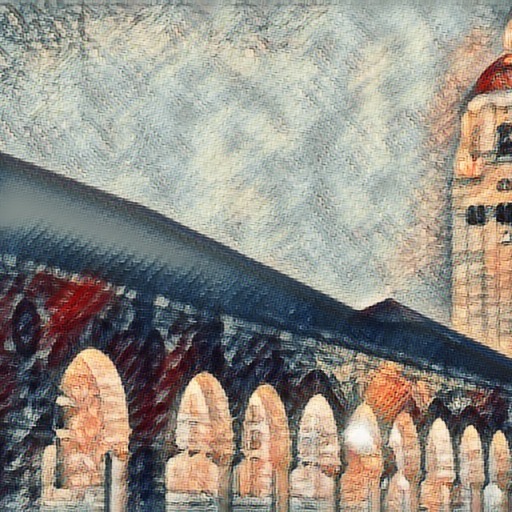}
    \includegraphics[width=0.18\linewidth]{figures/karya_head_of_a_clown.jpg}
    \includegraphics[width=0.18\linewidth]{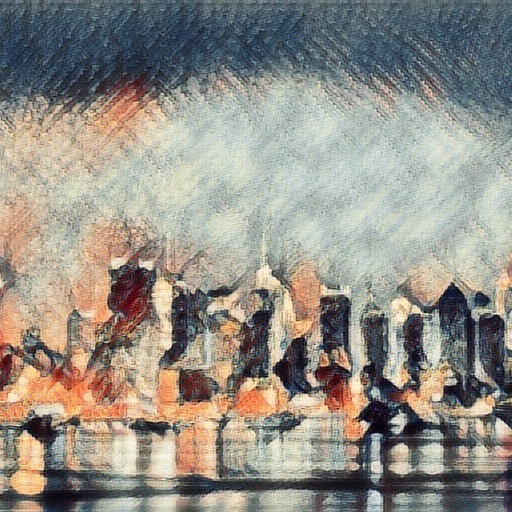}
    \includegraphics[width=0.18\linewidth]{figures/schultenhof_mettingen_bauerngarten_head_of_a_clown.jpg}
    \includegraphics[width=0.18\linewidth]{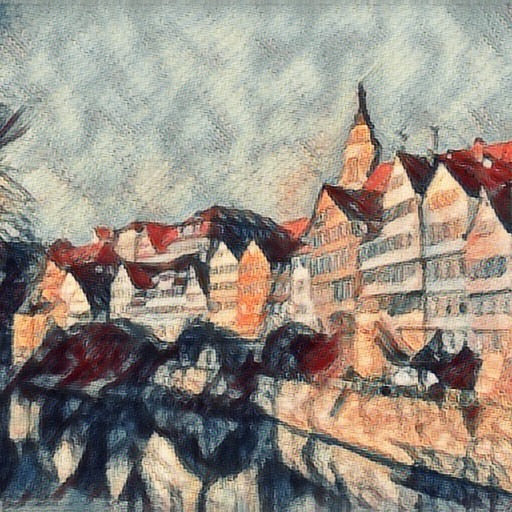}
\end{center}
\caption*{Georges Rouault, {\em Head of a Clown} (ca. 1907-1908).}
\end{figure}

\begin{figure}[ht]
\begin{center}
    \includegraphics[width=0.18\linewidth]{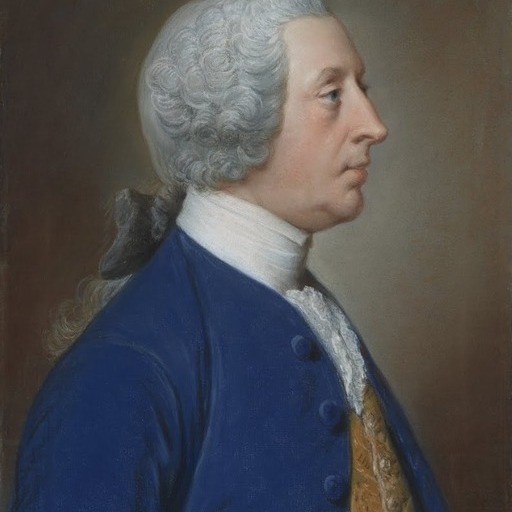}
    \includegraphics[width=0.18\linewidth]{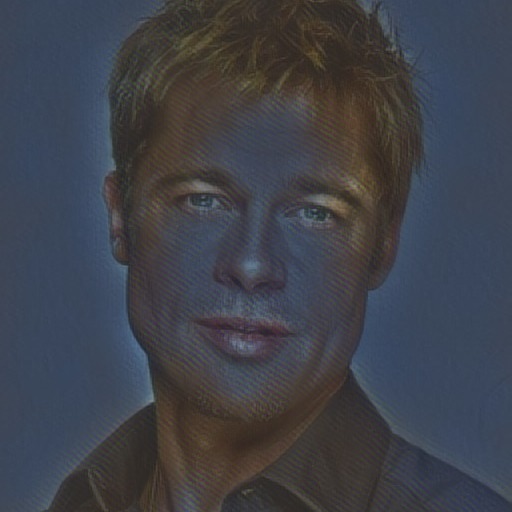}
    \includegraphics[width=0.18\linewidth]{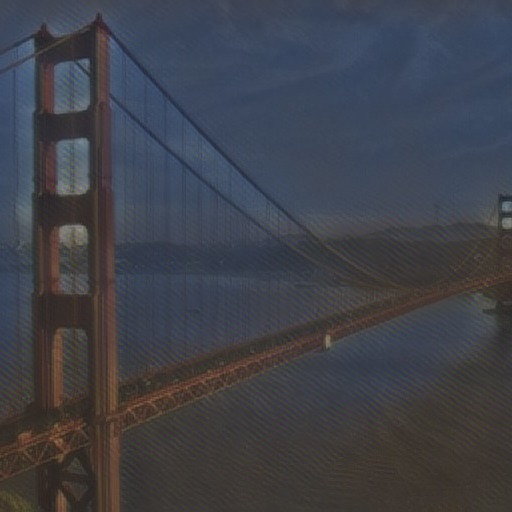}
    \includegraphics[width=0.18\linewidth]{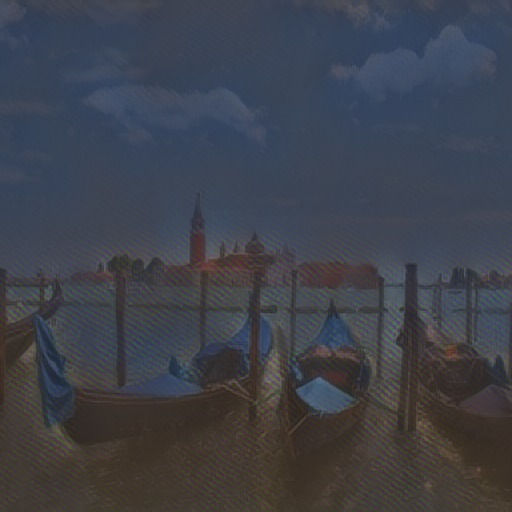}
    \includegraphics[width=0.18\linewidth]{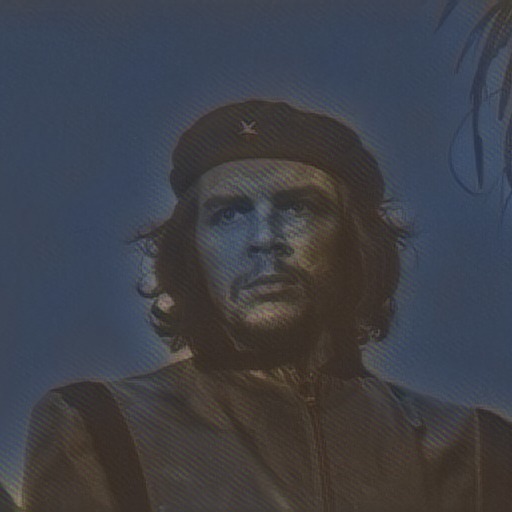} \\
    \includegraphics[width=0.18\linewidth]{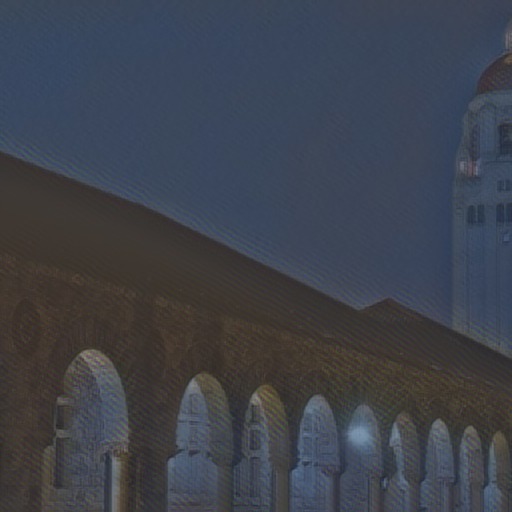}
    \includegraphics[width=0.18\linewidth]{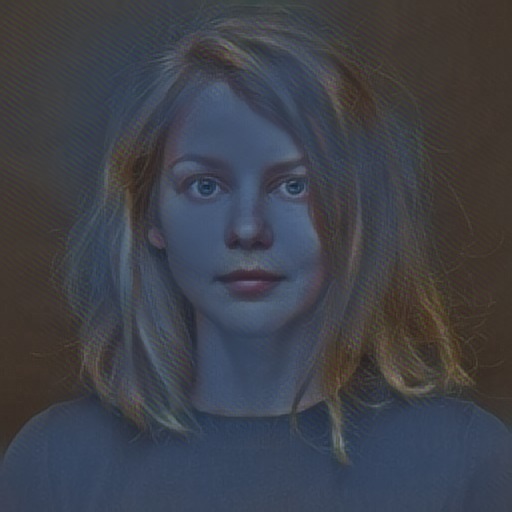}
    \includegraphics[width=0.18\linewidth]{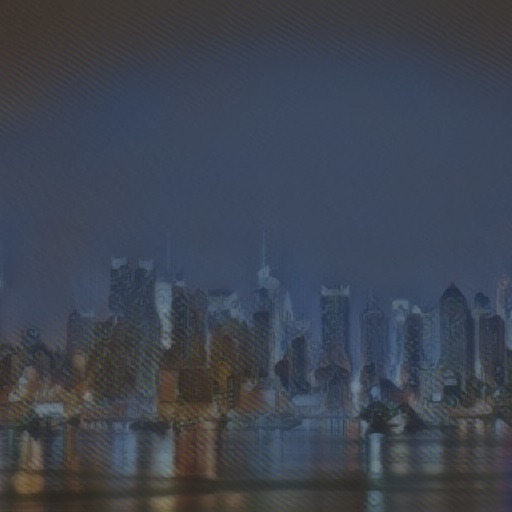}
    \includegraphics[width=0.18\linewidth]{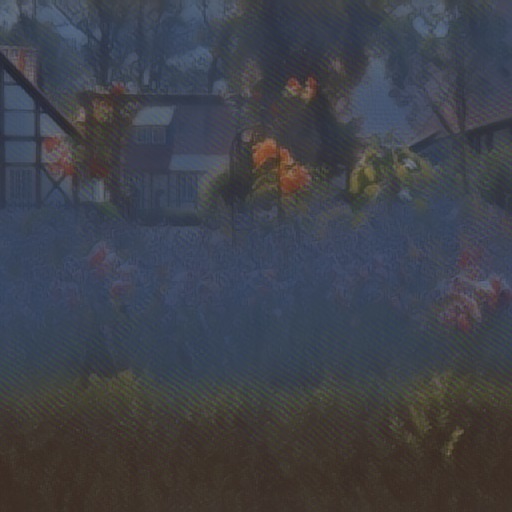}
    \includegraphics[width=0.18\linewidth]{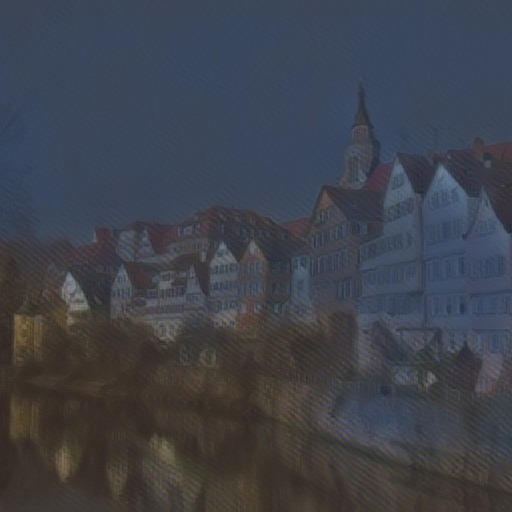}
\end{center}
\caption*{William Hoare, {\em Henry Hoare, "The Magnificent", of Stourhead}
    (about 1750-1760).}
\end{figure}

\begin{figure}[ht]
\begin{center}
    \includegraphics[width=0.18\linewidth]{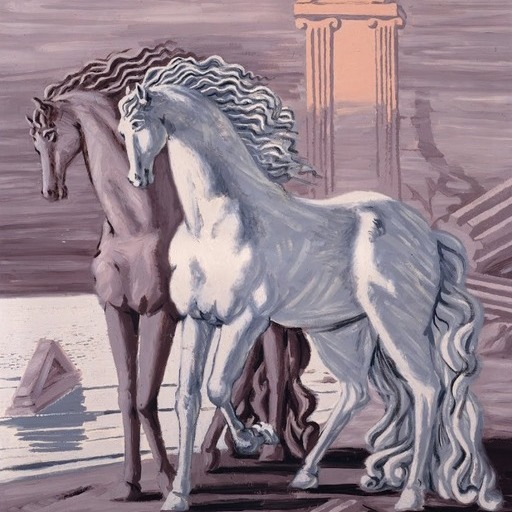}
    \includegraphics[width=0.18\linewidth]{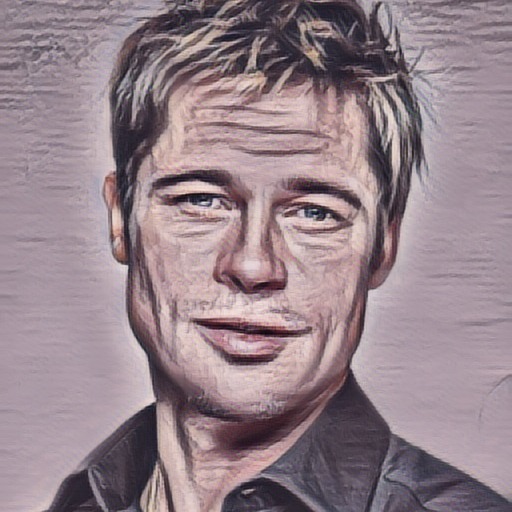}
    \includegraphics[width=0.18\linewidth]{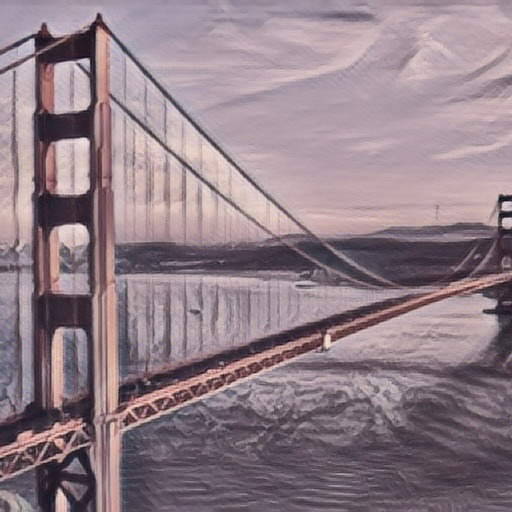}
    \includegraphics[width=0.18\linewidth]{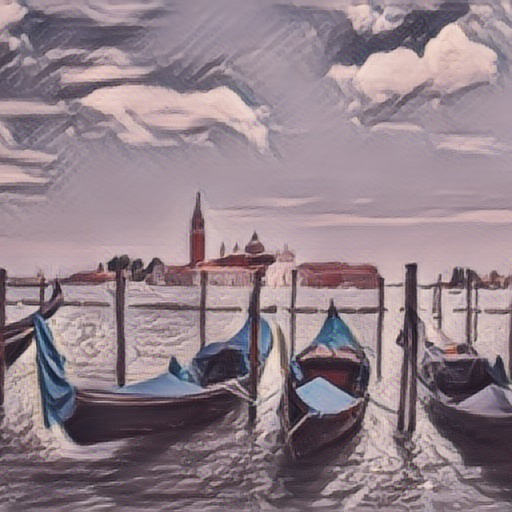}
    \includegraphics[width=0.18\linewidth]{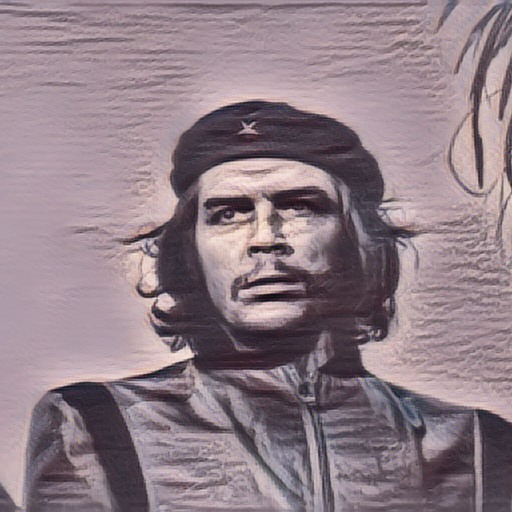} \\
    \includegraphics[width=0.18\linewidth]{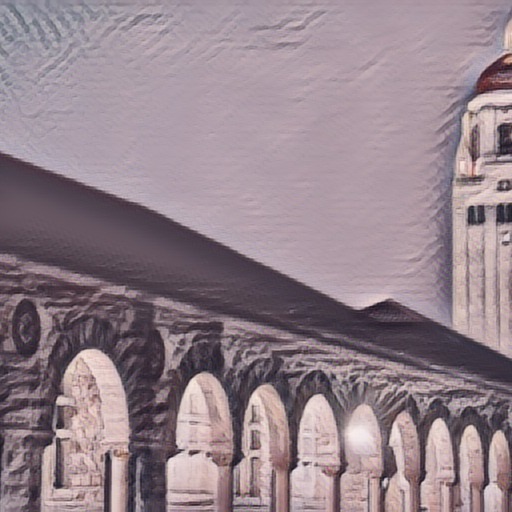}
    \includegraphics[width=0.18\linewidth]{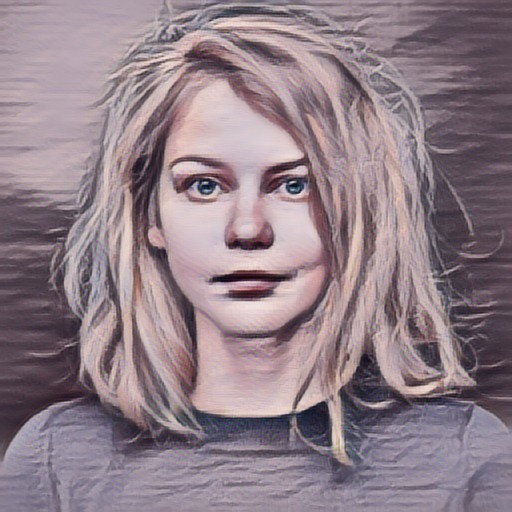}
    \includegraphics[width=0.18\linewidth]{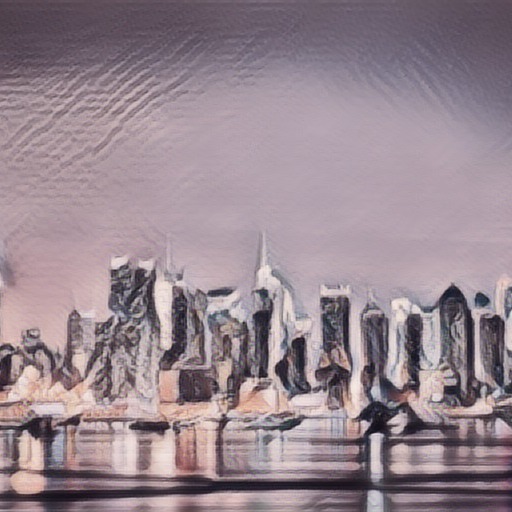}
    \includegraphics[width=0.18\linewidth]{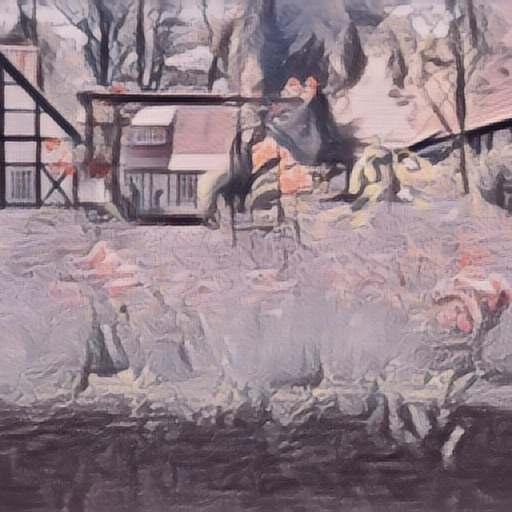}
    \includegraphics[width=0.18\linewidth]{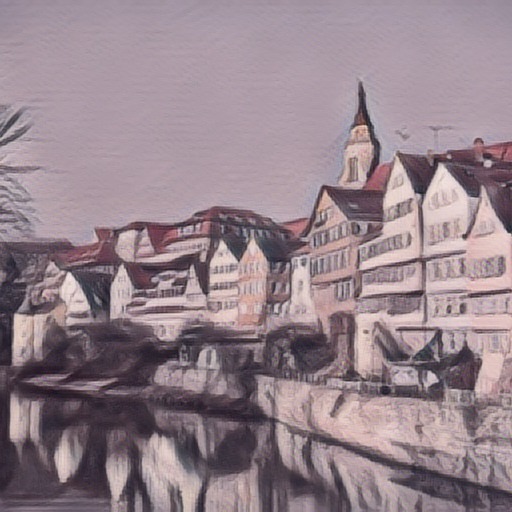}
\end{center}
\caption*{Giorgio de Chirico, {\em Horses on the seashore} (1927/1928).}
\end{figure}

\clearpage
\begin{figure}[ht]
\begin{center}
    \includegraphics[width=0.18\linewidth]{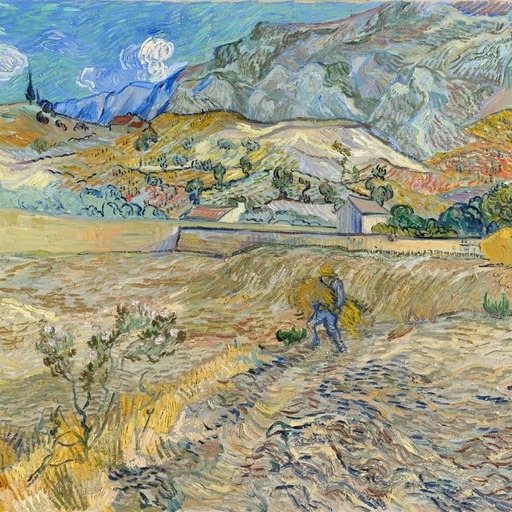}
    \includegraphics[width=0.18\linewidth]{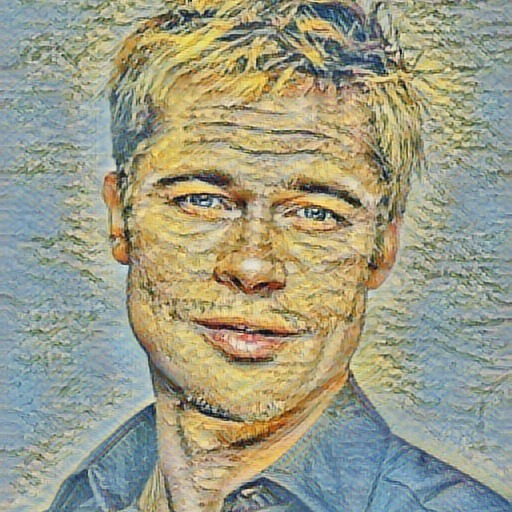}
    \includegraphics[width=0.18\linewidth]{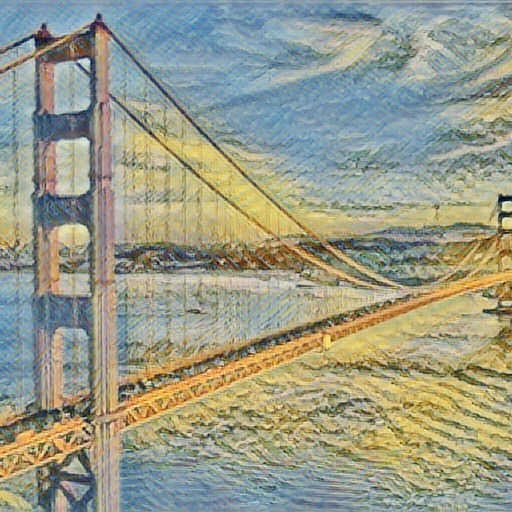}
    \includegraphics[width=0.18\linewidth]{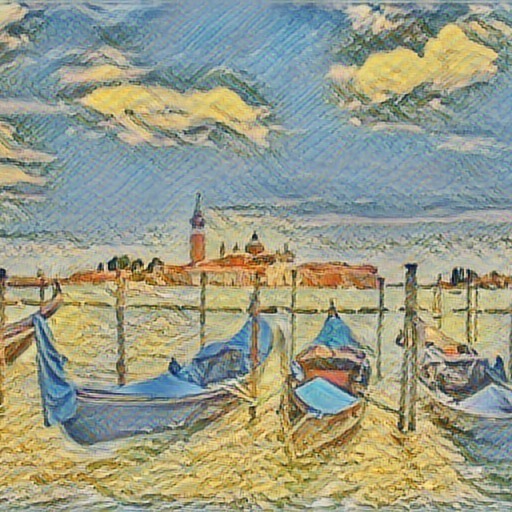}
    \includegraphics[width=0.18\linewidth]{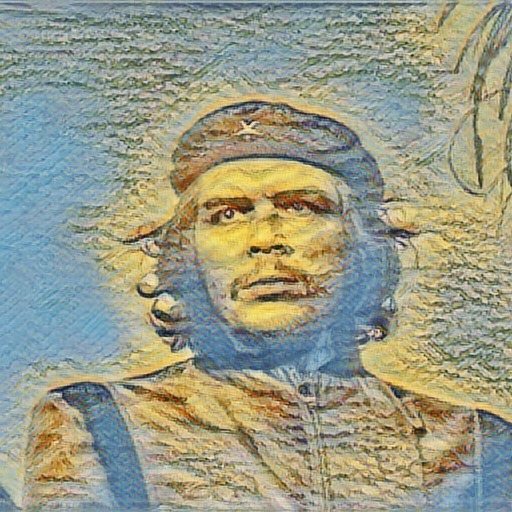} \\
    \includegraphics[width=0.18\linewidth]{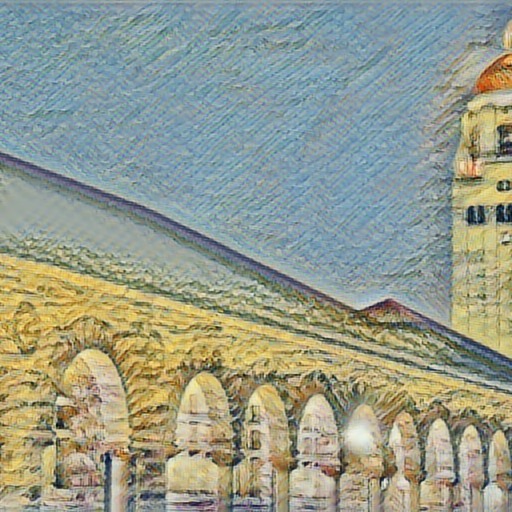}
    \includegraphics[width=0.18\linewidth]{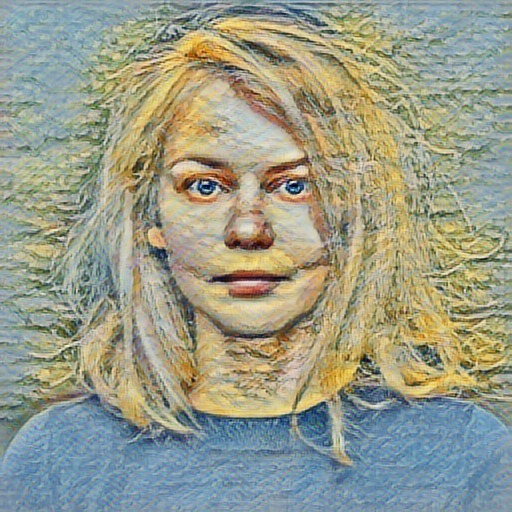}
    \includegraphics[width=0.18\linewidth]{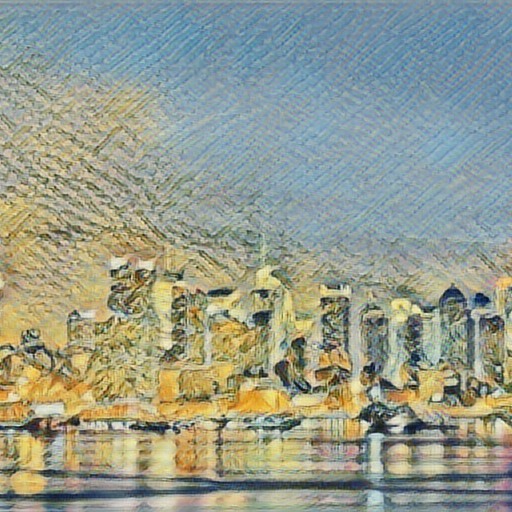}
    \includegraphics[width=0.18\linewidth]{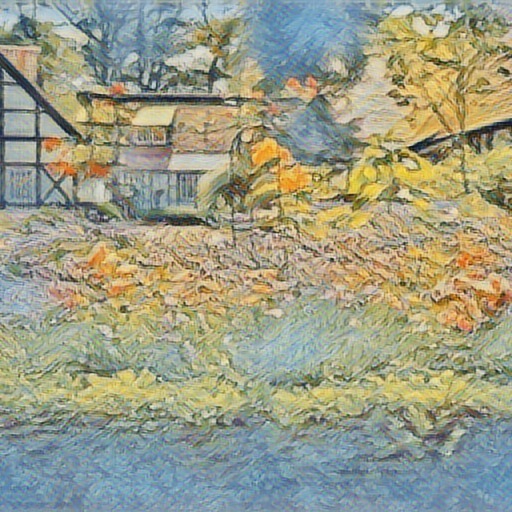}
    \includegraphics[width=0.18\linewidth]{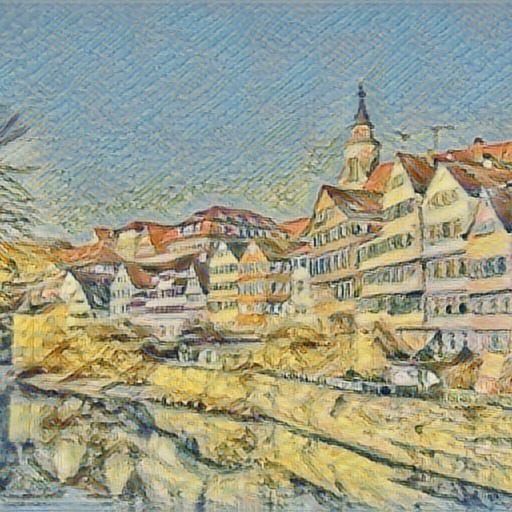}
\end{center}
\caption*{Vincent van Gogh, {\em Landscape at Saint-R\'{e}my (Enclosed Field
    with Peasant)} (1889).}
\end{figure}

\begin{figure}[ht]
\begin{center}
    \includegraphics[width=0.18\linewidth]{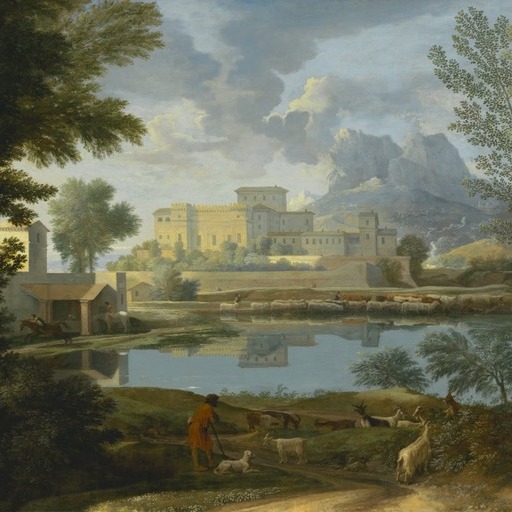}
    \includegraphics[width=0.18\linewidth]{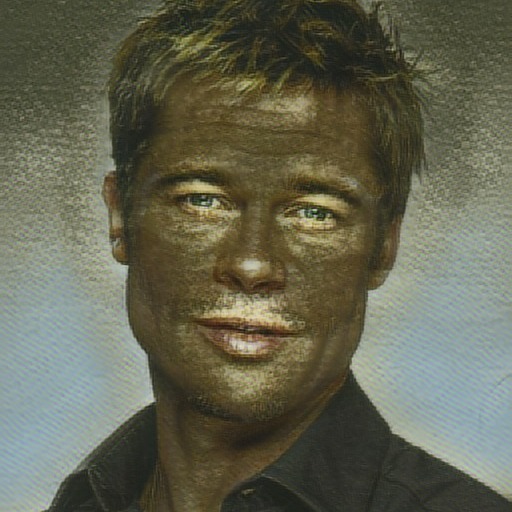}
    \includegraphics[width=0.18\linewidth]{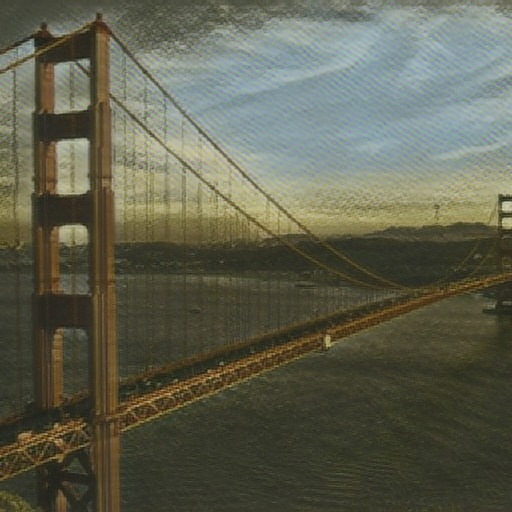}
    \includegraphics[width=0.18\linewidth]{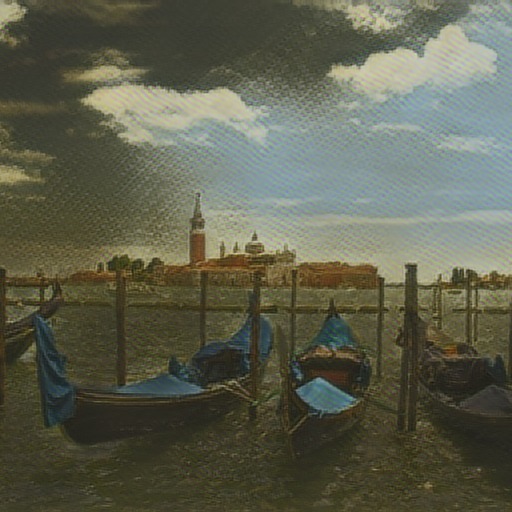}
    \includegraphics[width=0.18\linewidth]{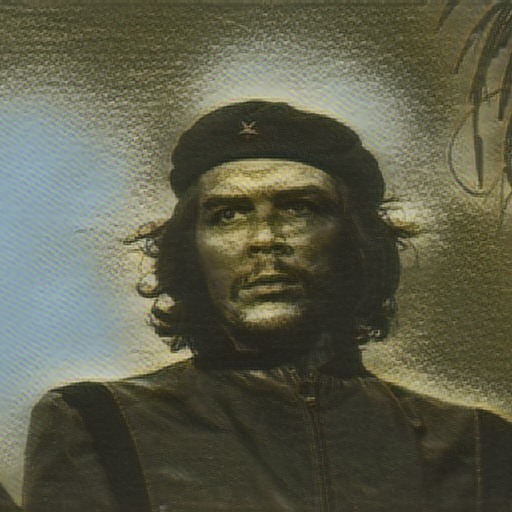} \\
    \includegraphics[width=0.18\linewidth]{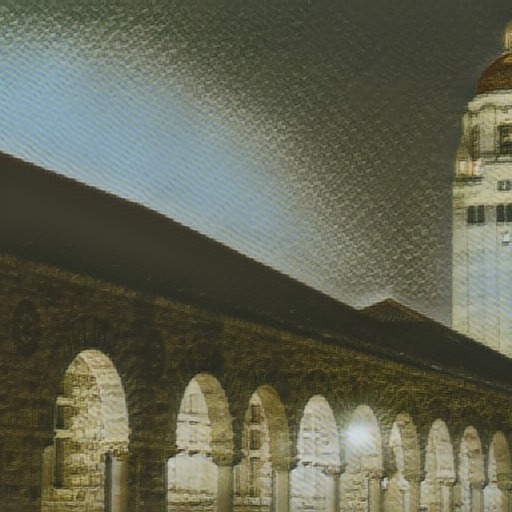}
    \includegraphics[width=0.18\linewidth]{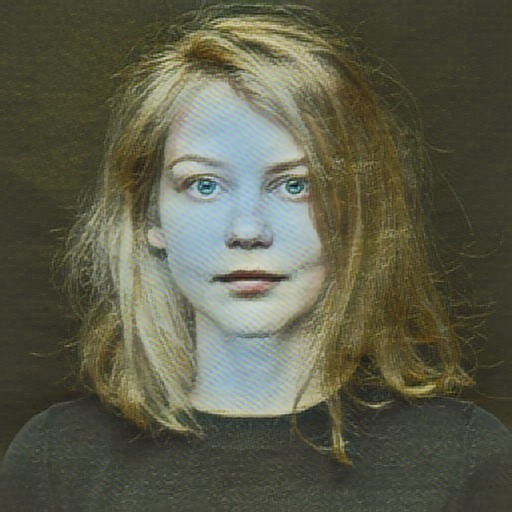}
    \includegraphics[width=0.18\linewidth]{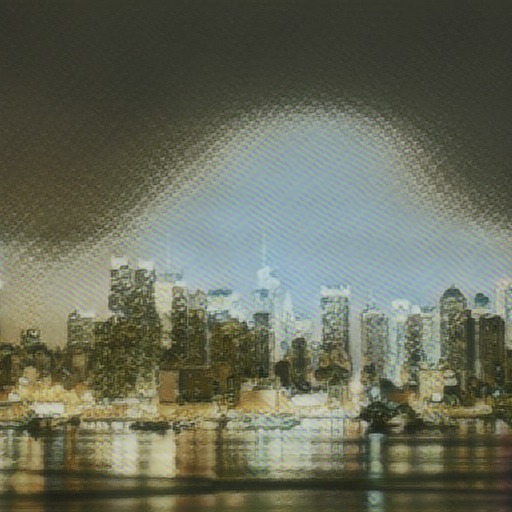}
    \includegraphics[width=0.18\linewidth]{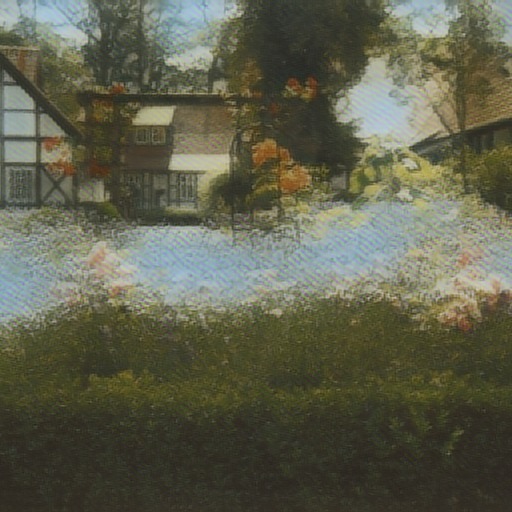}
    \includegraphics[width=0.18\linewidth]{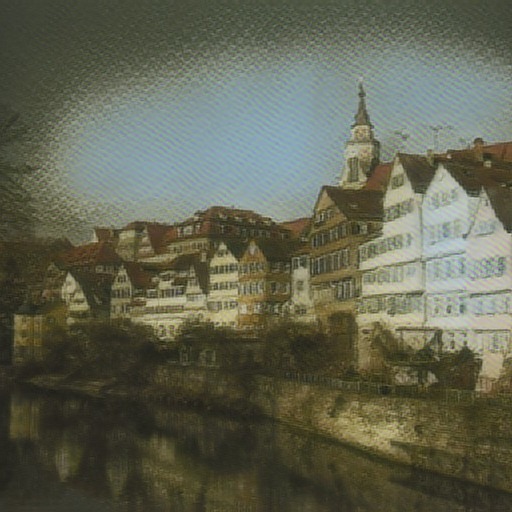}
\end{center}
\caption*{Nicolas Poussin, {\em Landscape with a Calm} (1650-1651).}
\end{figure}

\begin{figure}[ht]
\begin{center}
    \includegraphics[width=0.18\linewidth]{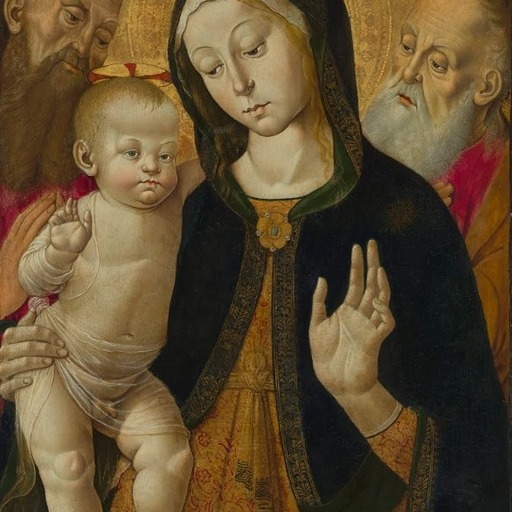}
    \includegraphics[width=0.18\linewidth]{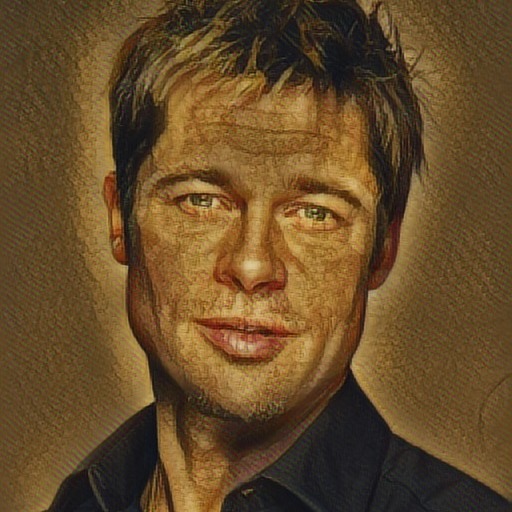}
    \includegraphics[width=0.18\linewidth]{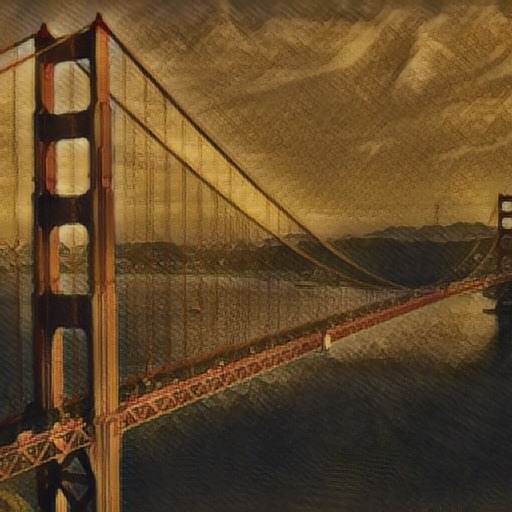}
    \includegraphics[width=0.18\linewidth]{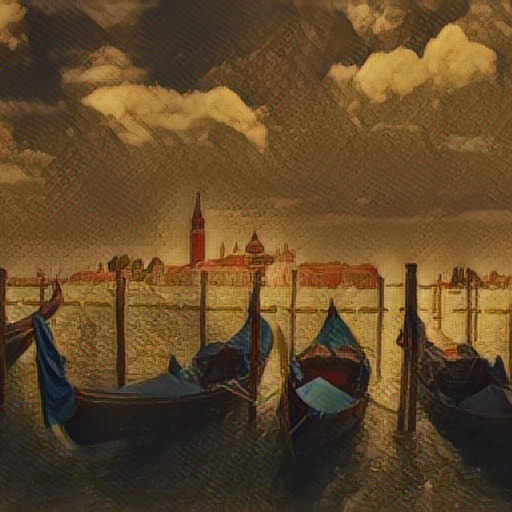}
    \includegraphics[width=0.18\linewidth]{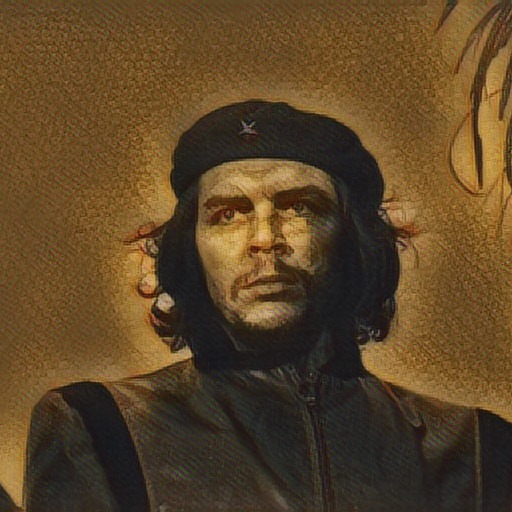} \\
    \includegraphics[width=0.18\linewidth]{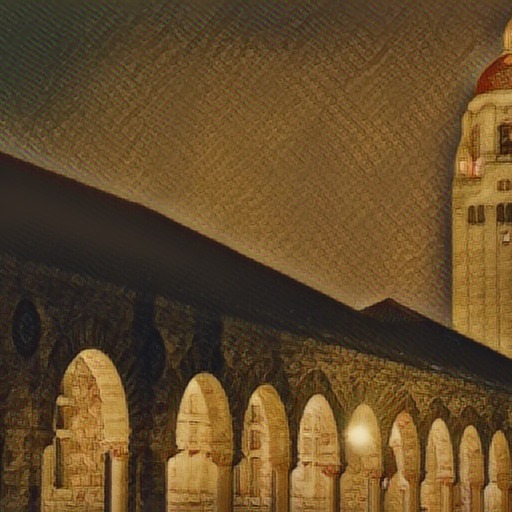}
    \includegraphics[width=0.18\linewidth]{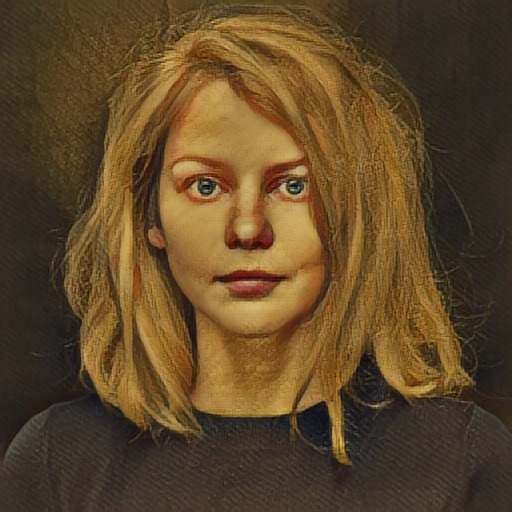}
    \includegraphics[width=0.18\linewidth]{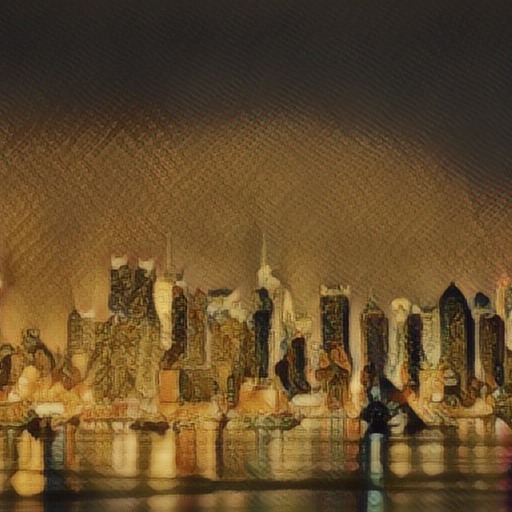}
    \includegraphics[width=0.18\linewidth]{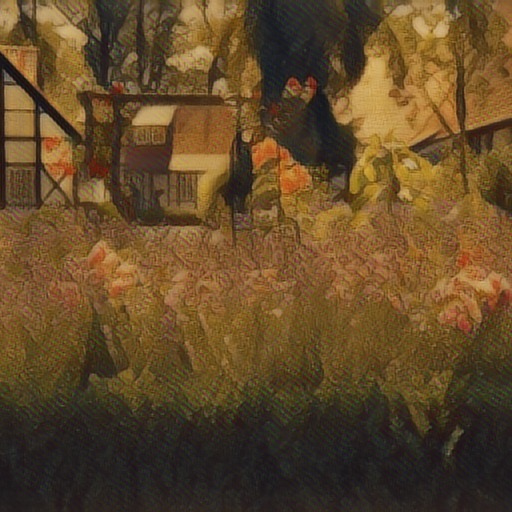}
    \includegraphics[width=0.18\linewidth]{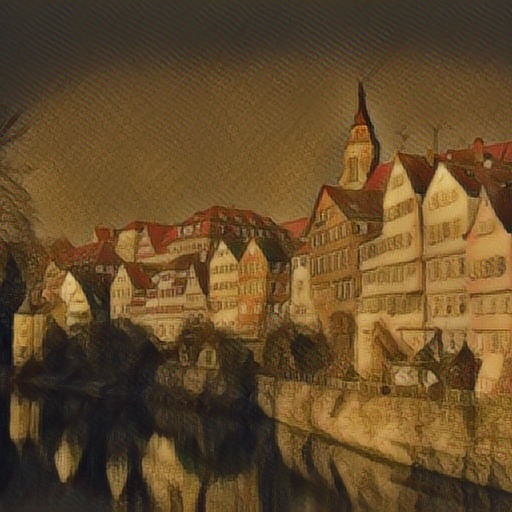}
\end{center}
\caption*{Bernardino Fungai, {\em Madonna and Child with Two Hermit Saints}
    (early 1480s).}
\end{figure}

\clearpage
\begin{figure}[ht]
\begin{center}
    \includegraphics[width=0.18\linewidth]{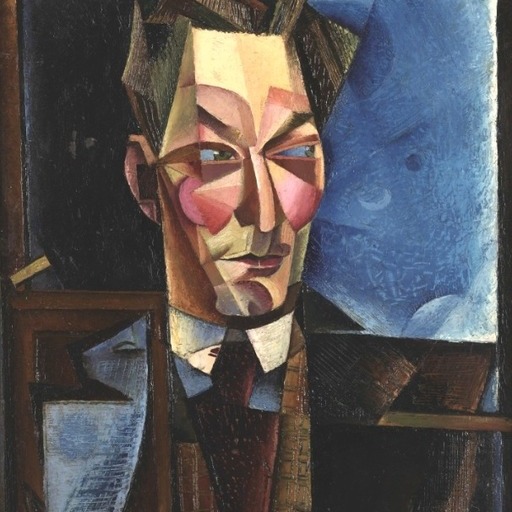}
    \includegraphics[width=0.18\linewidth]{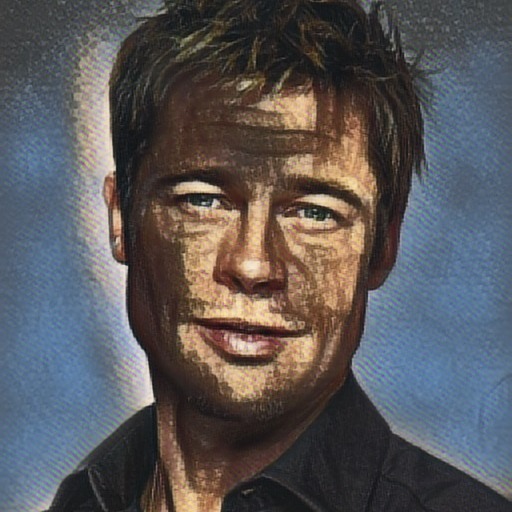}
    \includegraphics[width=0.18\linewidth]{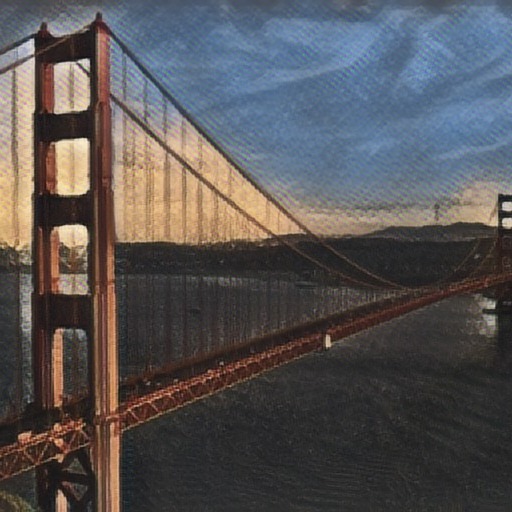}
    \includegraphics[width=0.18\linewidth]{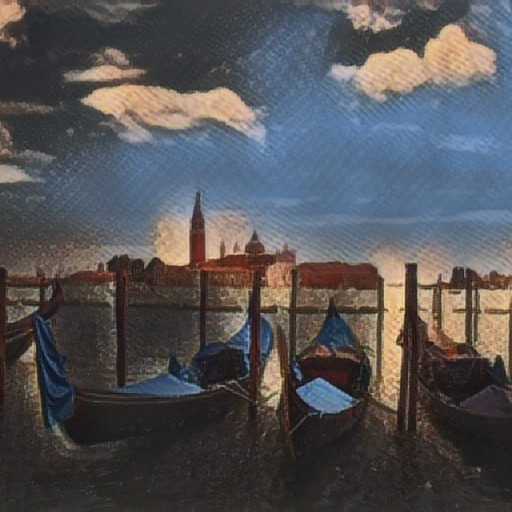}
    \includegraphics[width=0.18\linewidth]{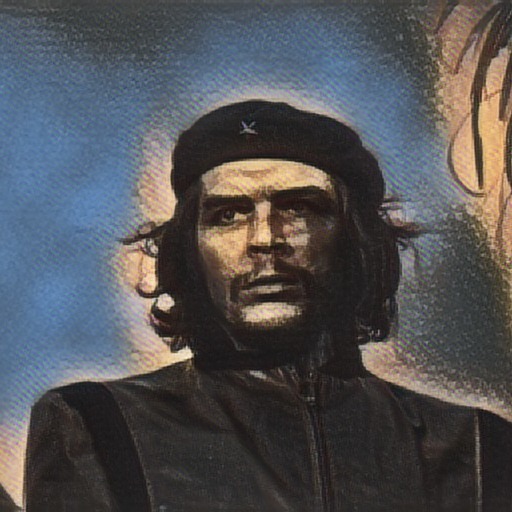} \\
    \includegraphics[width=0.18\linewidth]{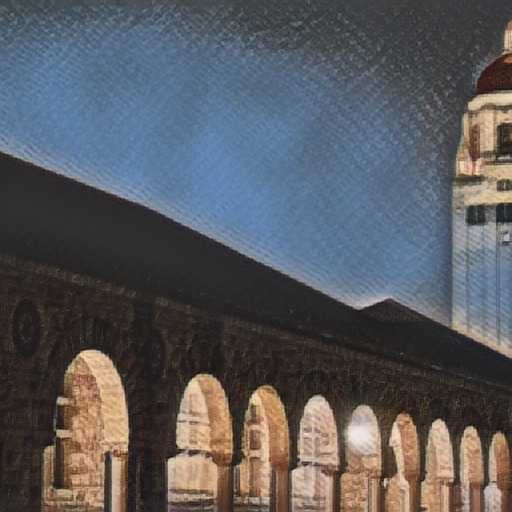}
    \includegraphics[width=0.18\linewidth]{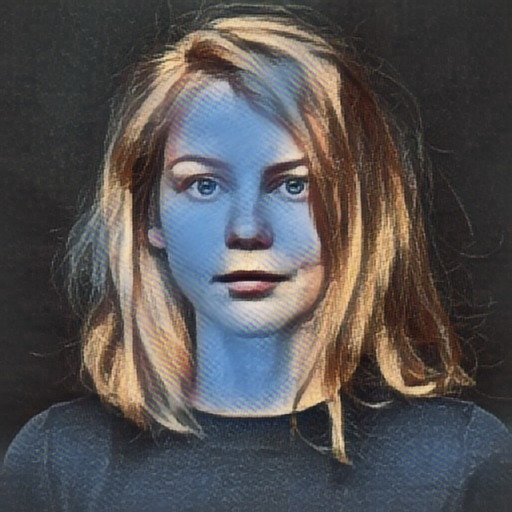}
    \includegraphics[width=0.18\linewidth]{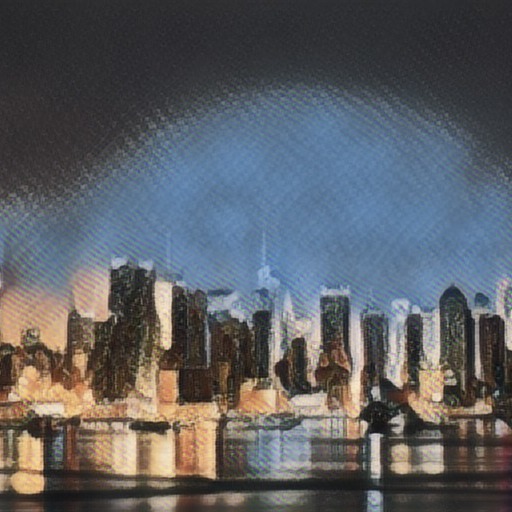}
    \includegraphics[width=0.18\linewidth]{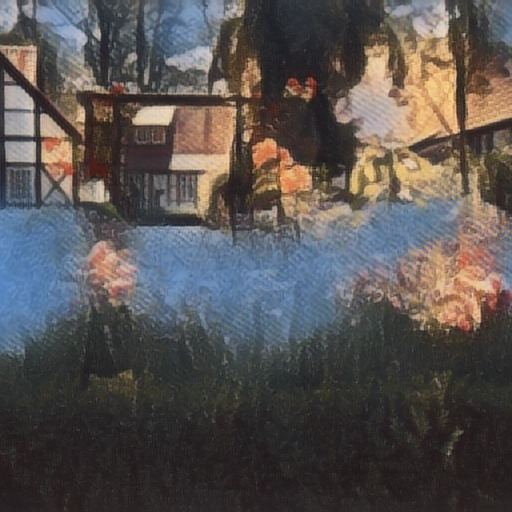}
    \includegraphics[width=0.18\linewidth]{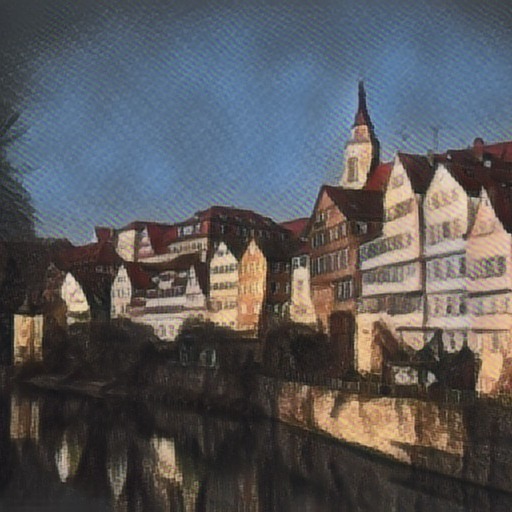}
\end{center}
\caption*{Max Hermann Maxy, {\em Portrait of a Friend} (1926).}
\end{figure}

\begin{figure}[ht]
\begin{center}
    \includegraphics[width=0.18\linewidth]{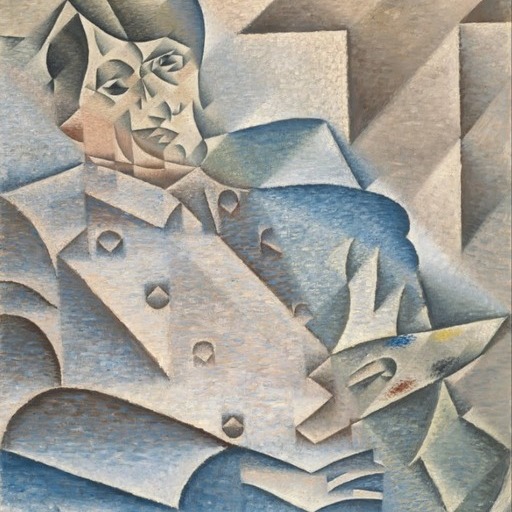}
    \includegraphics[width=0.18\linewidth]{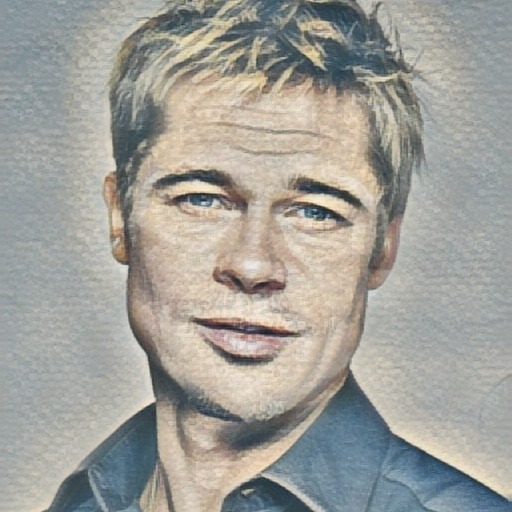}
    \includegraphics[width=0.18\linewidth]{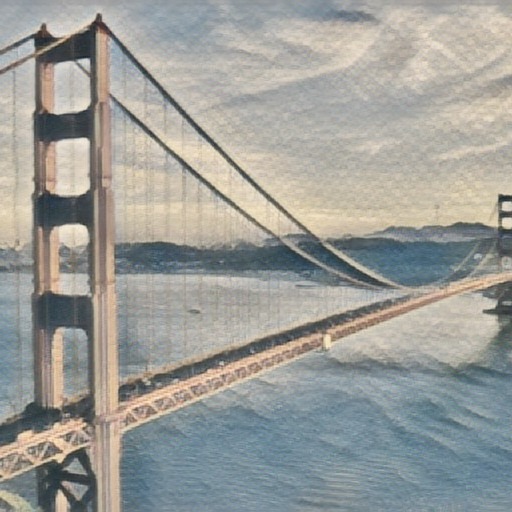}
    \includegraphics[width=0.18\linewidth]{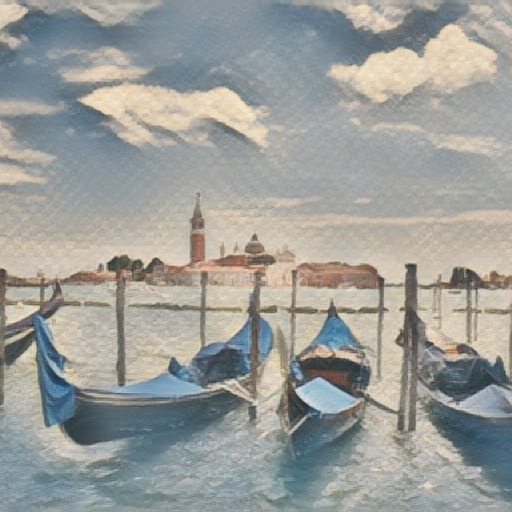}
    \includegraphics[width=0.18\linewidth]{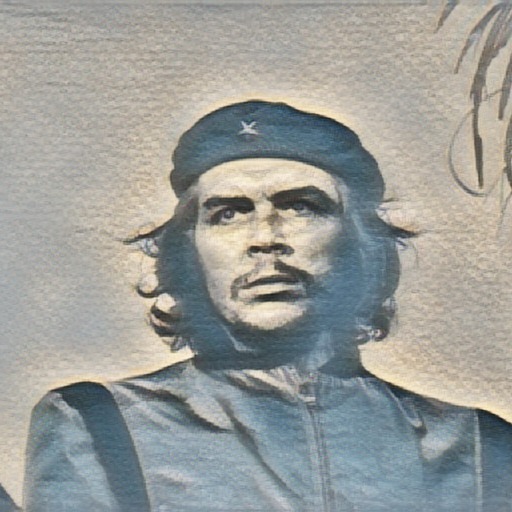} \\
    \includegraphics[width=0.18\linewidth]{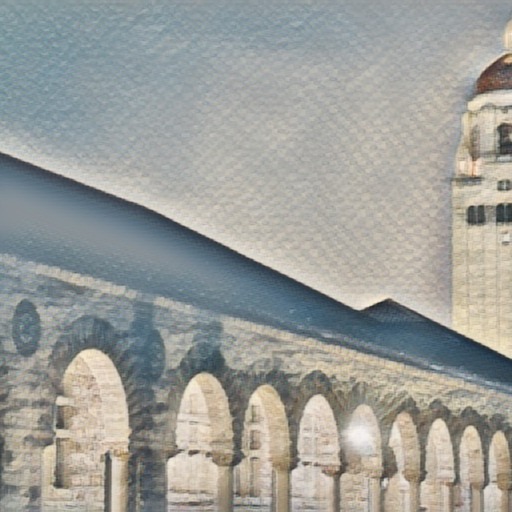}
    \includegraphics[width=0.18\linewidth]{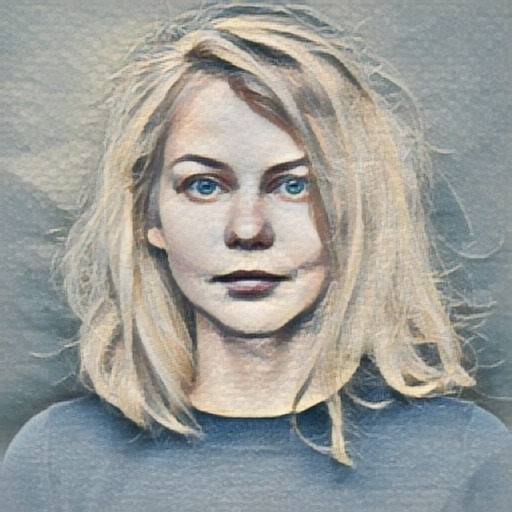}
    \includegraphics[width=0.18\linewidth]{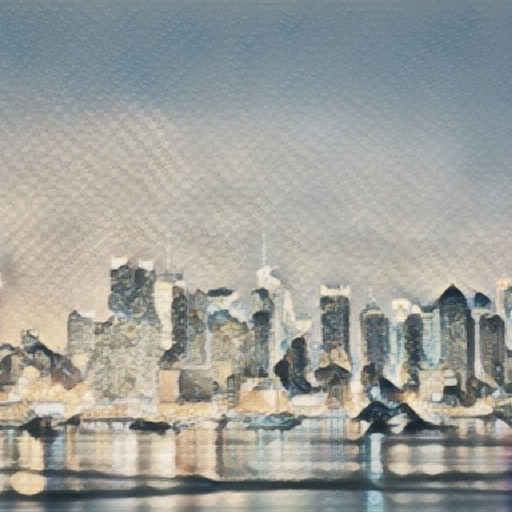}
    \includegraphics[width=0.18\linewidth]{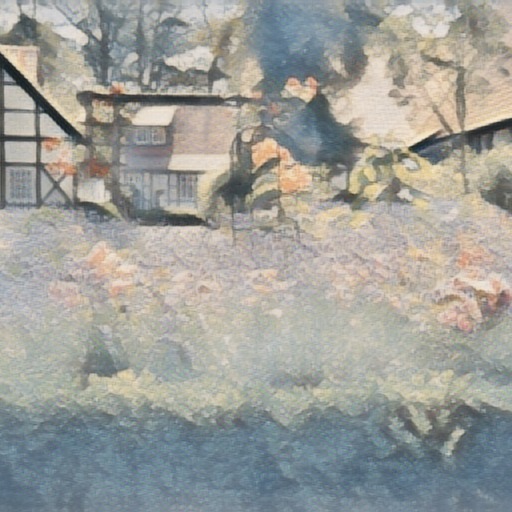}
    \includegraphics[width=0.18\linewidth]{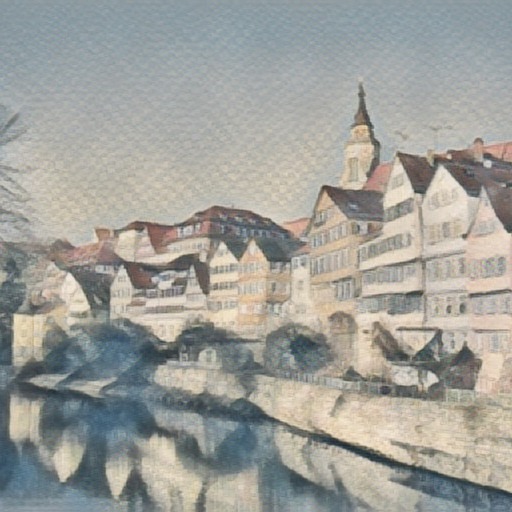}
\end{center}
\caption*{Juan Gris, {\em Portrait of Pablo Picasso} (1912).}
\end{figure}

\begin{figure}[ht]
\begin{center}
    \includegraphics[width=0.18\linewidth]{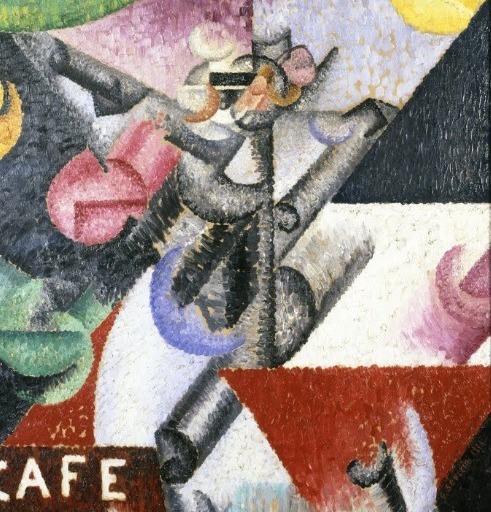}
    \includegraphics[width=0.18\linewidth]{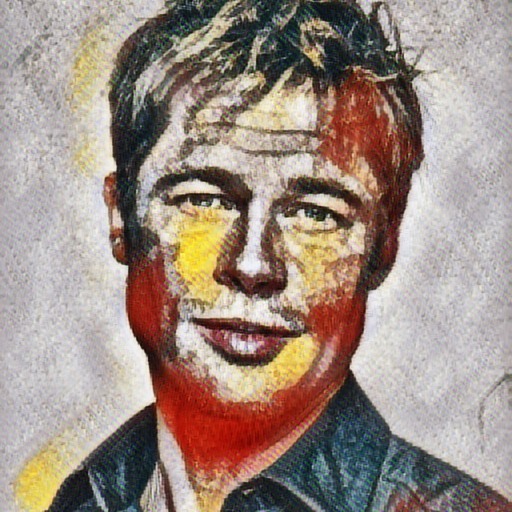}
    \includegraphics[width=0.18\linewidth]{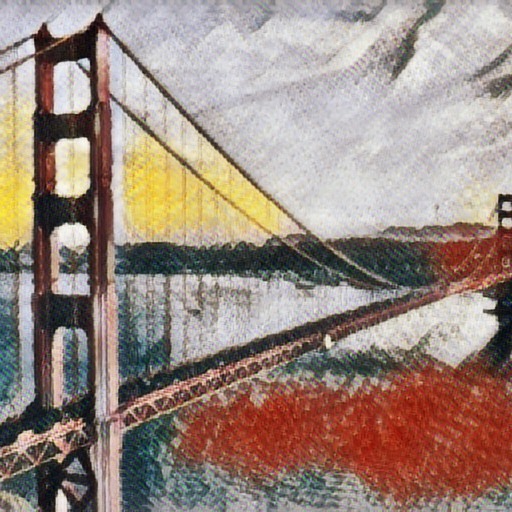}
    \includegraphics[width=0.18\linewidth]{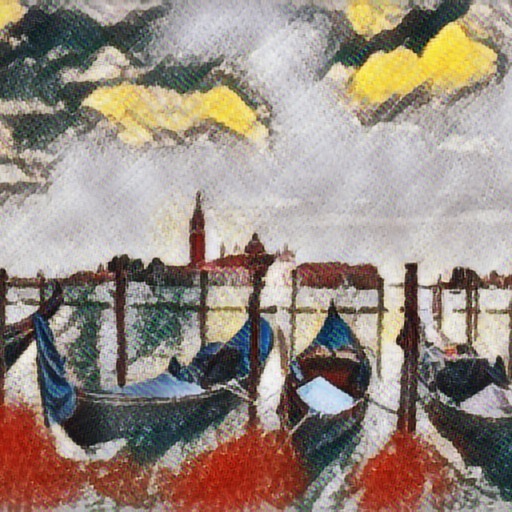}
    \includegraphics[width=0.18\linewidth]{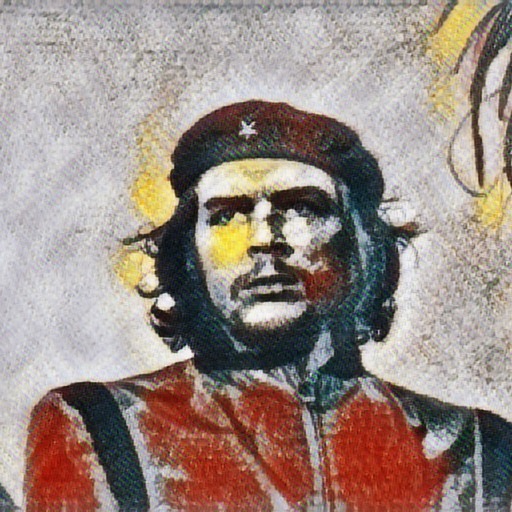} \\
    \includegraphics[width=0.18\linewidth]{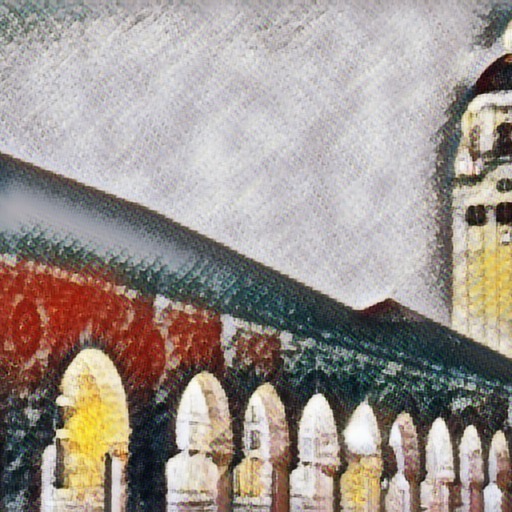}
    \includegraphics[width=0.18\linewidth]{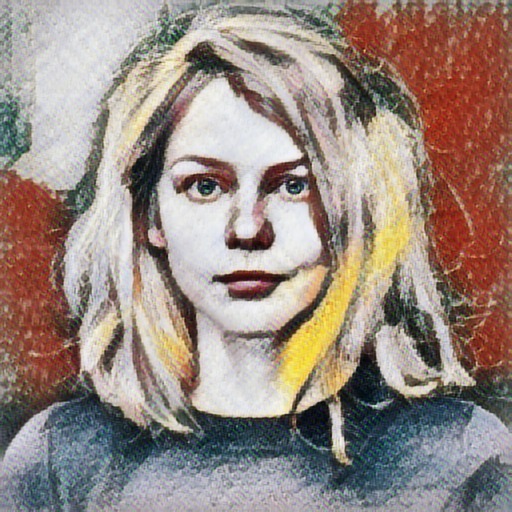}
    \includegraphics[width=0.18\linewidth]{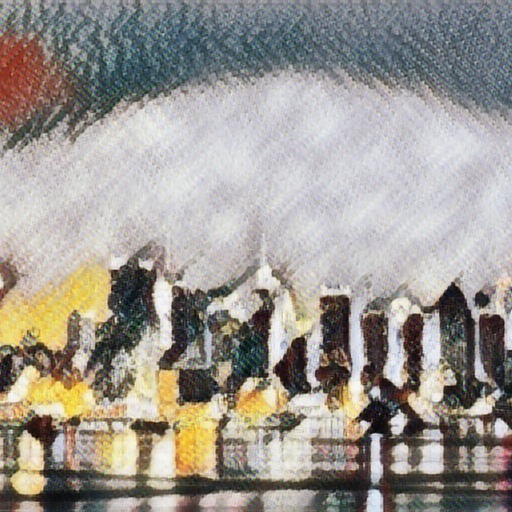}
    \includegraphics[width=0.18\linewidth]{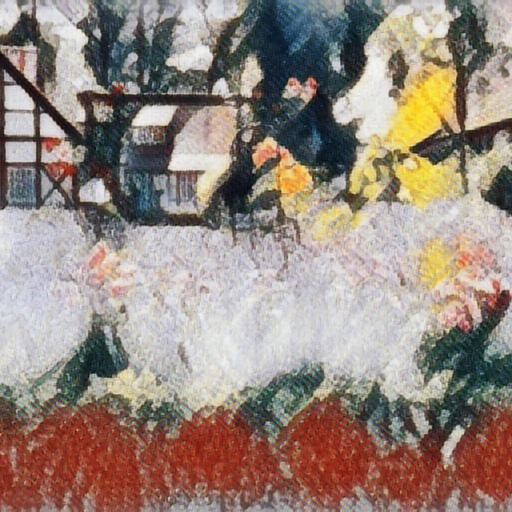}
    \includegraphics[width=0.18\linewidth]{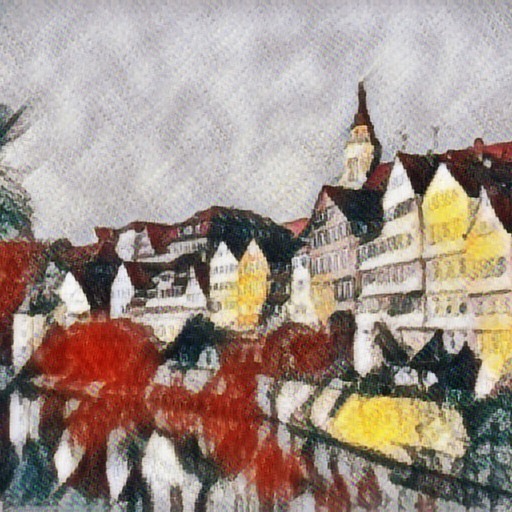}
\end{center}
\caption*{Severini Gino, {\em Ritmo plastico del 14 luglio} (1913).}
\end{figure}

\clearpage
\begin{figure}[ht]
\begin{center}
    \includegraphics[width=0.18\linewidth]{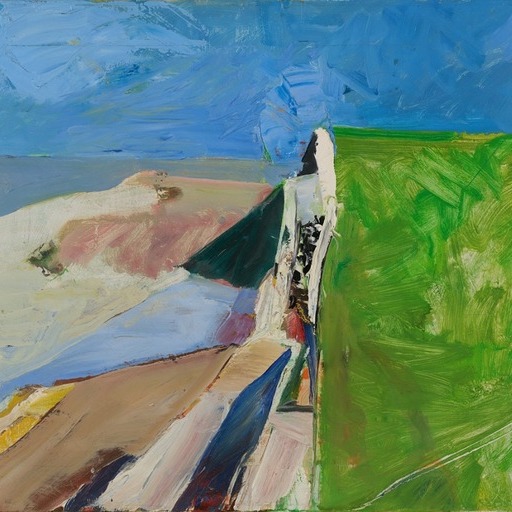}
    \includegraphics[width=0.18\linewidth]{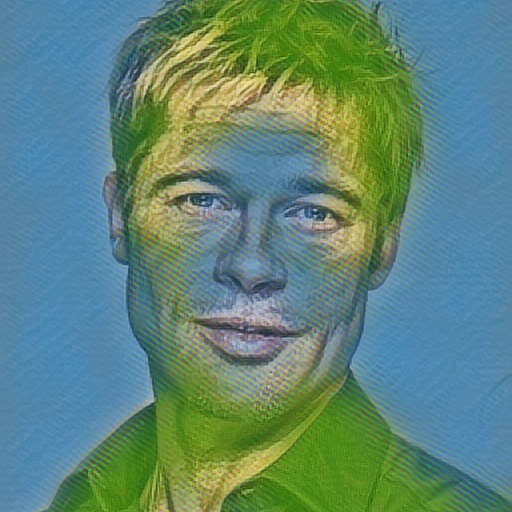}
    \includegraphics[width=0.18\linewidth]{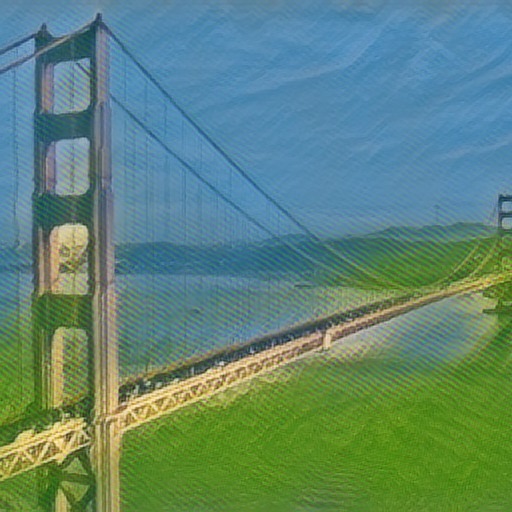}
    \includegraphics[width=0.18\linewidth]{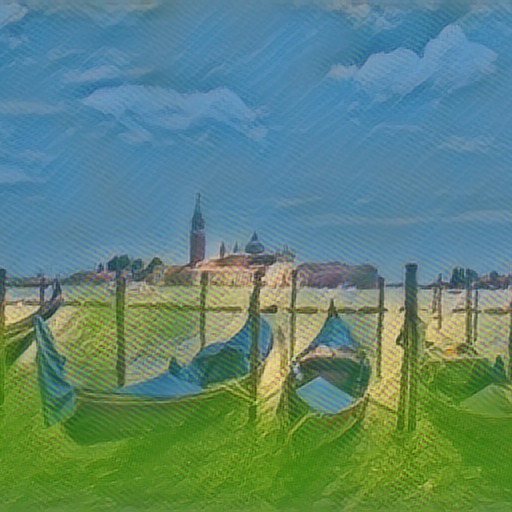}
    \includegraphics[width=0.18\linewidth]{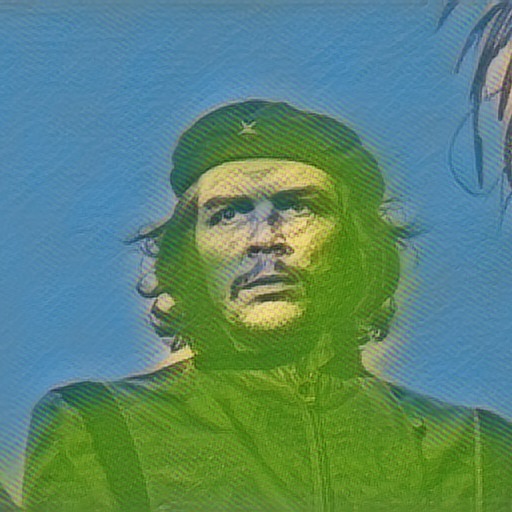} \\
    \includegraphics[width=0.18\linewidth]{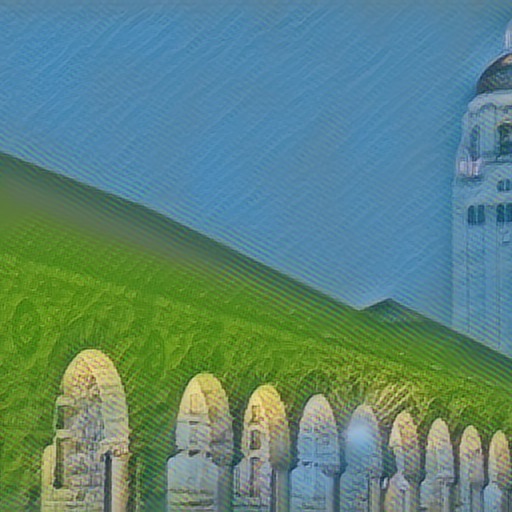}
    \includegraphics[width=0.18\linewidth]{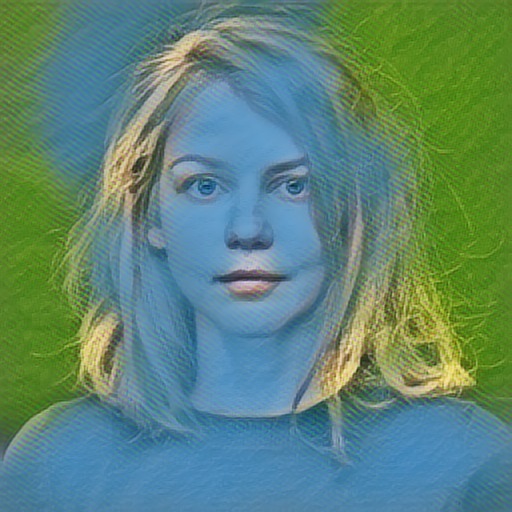}
    \includegraphics[width=0.18\linewidth]{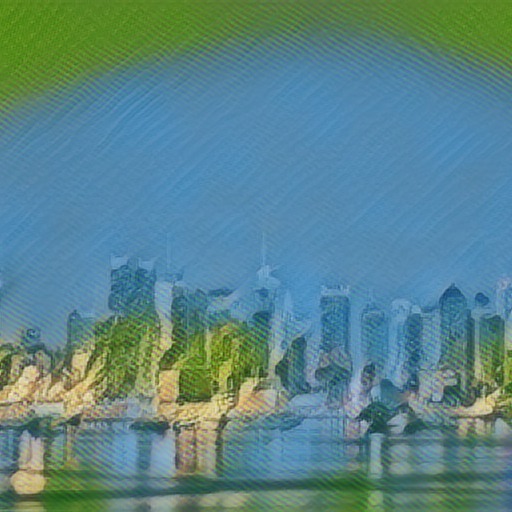}
    \includegraphics[width=0.18\linewidth]{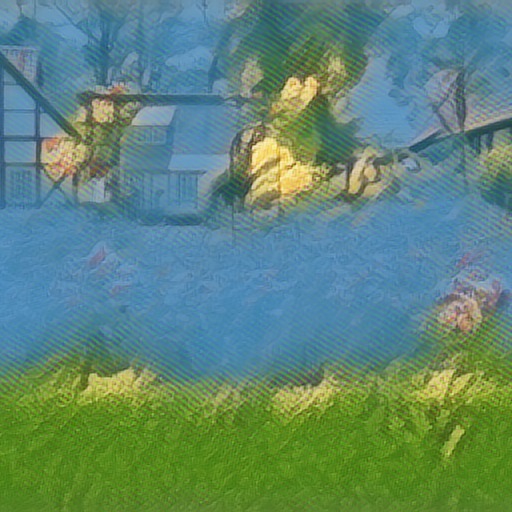}
    \includegraphics[width=0.18\linewidth]{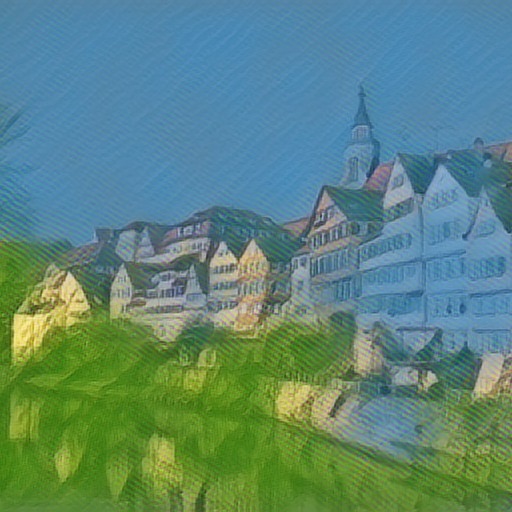}
\end{center}
\caption*{Richard Diebenkorn, {\em Seawall} (1957).}
\end{figure}

\begin{figure}[ht]
\begin{center}
    \includegraphics[width=0.18\linewidth]{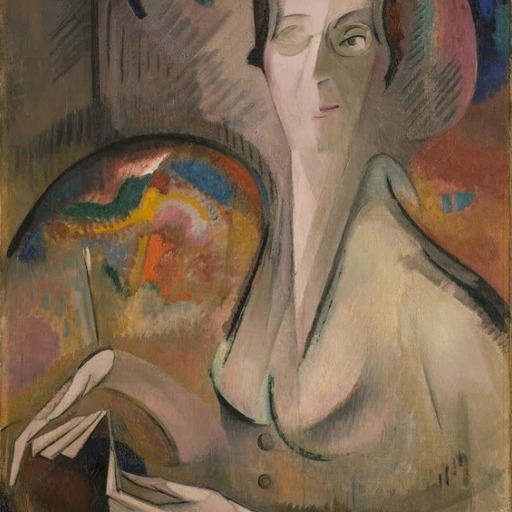}
    \includegraphics[width=0.18\linewidth]{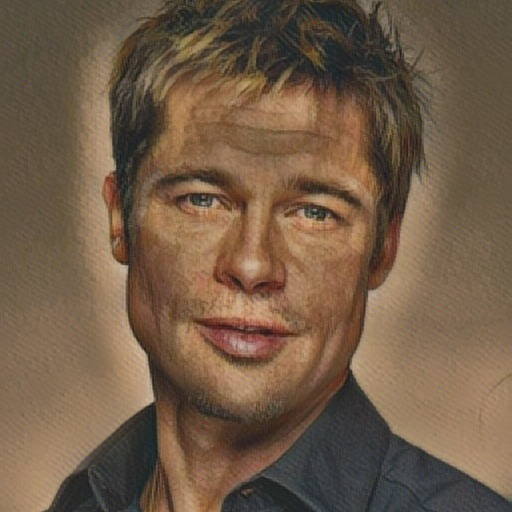}
    \includegraphics[width=0.18\linewidth]{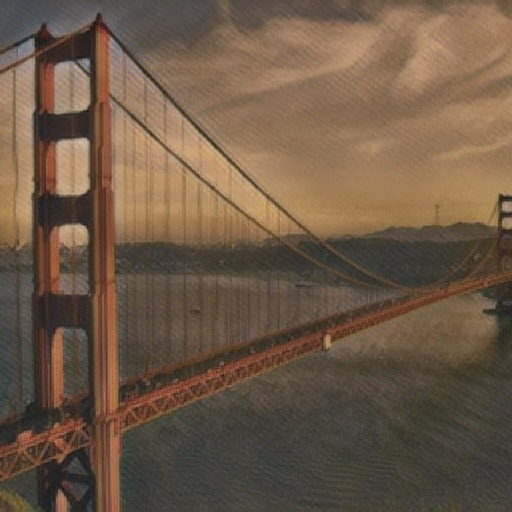}
    \includegraphics[width=0.18\linewidth]{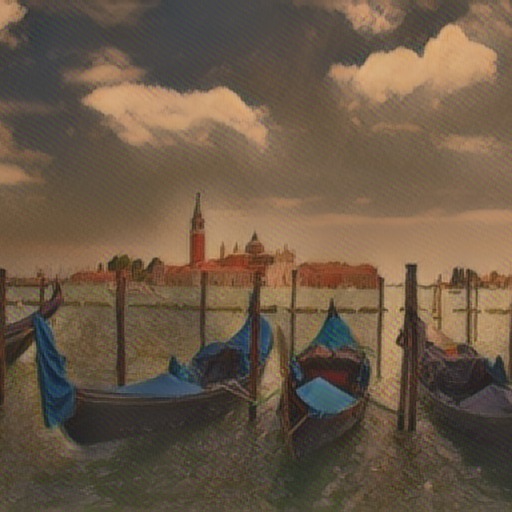}
    \includegraphics[width=0.18\linewidth]{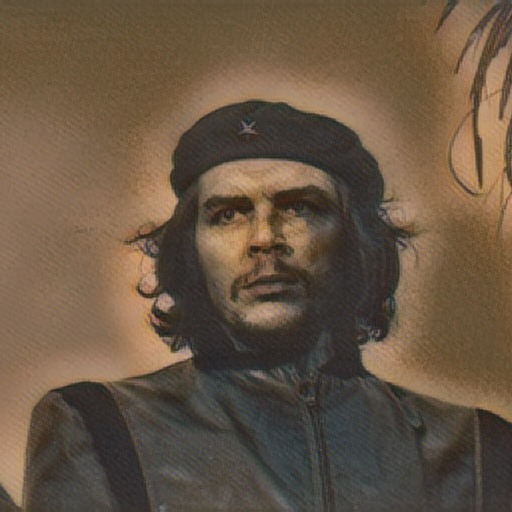} \\
    \includegraphics[width=0.18\linewidth]{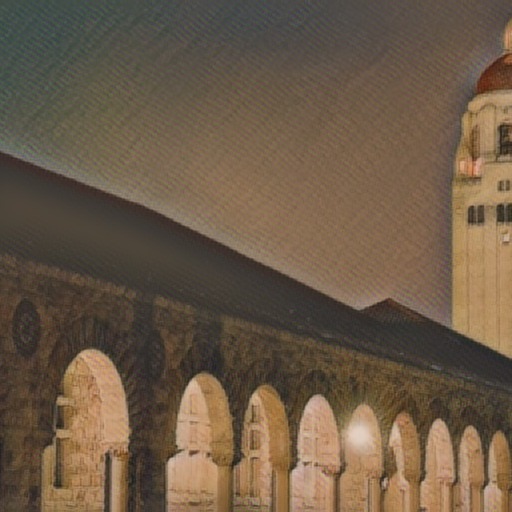}
    \includegraphics[width=0.18\linewidth]{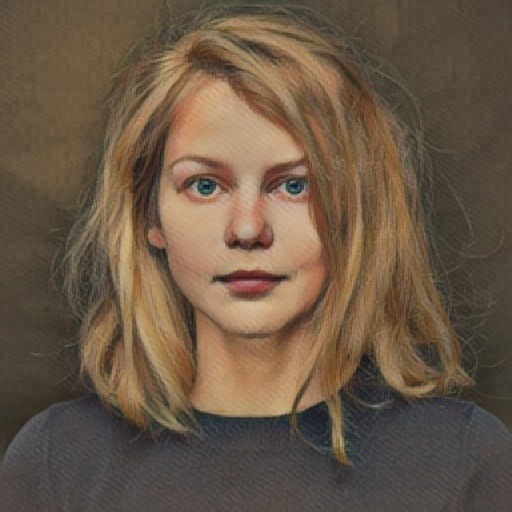}
    \includegraphics[width=0.18\linewidth]{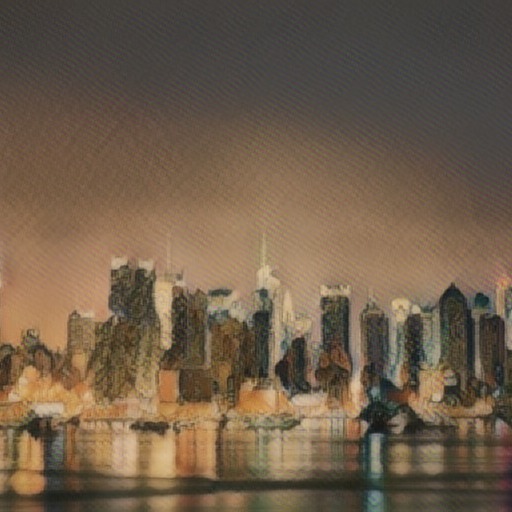}
    \includegraphics[width=0.18\linewidth]{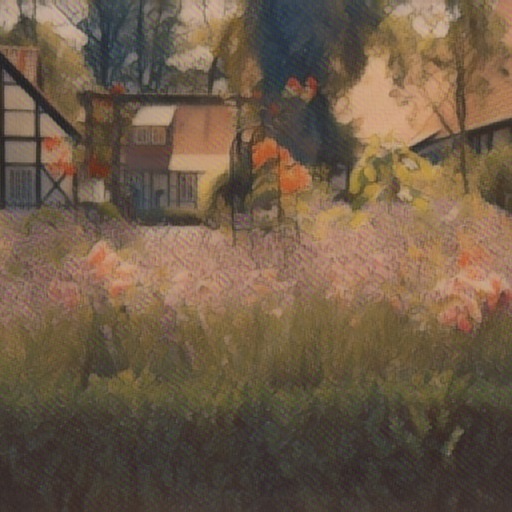}
    \includegraphics[width=0.18\linewidth]{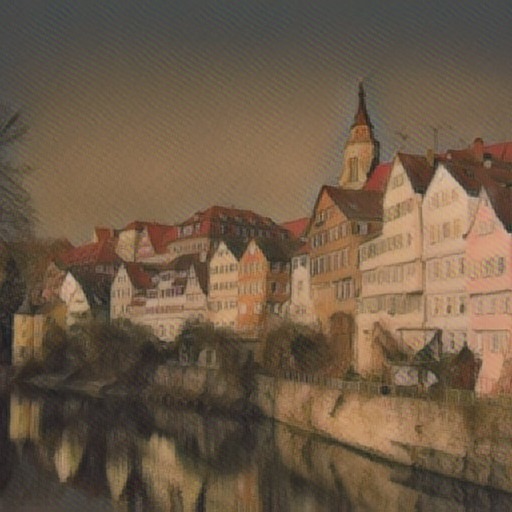}
\end{center}
\caption*{Alice Bailly, {\em Self-Portrait} (1917).}
\end{figure}

\begin{figure}[ht]
\begin{center}
    \includegraphics[width=0.18\linewidth]{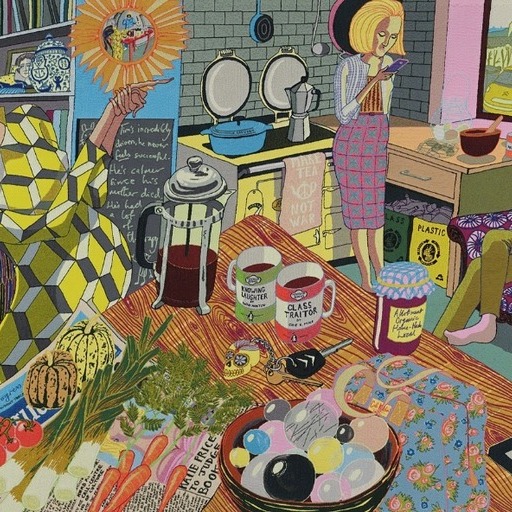}
    \includegraphics[width=0.18\linewidth]{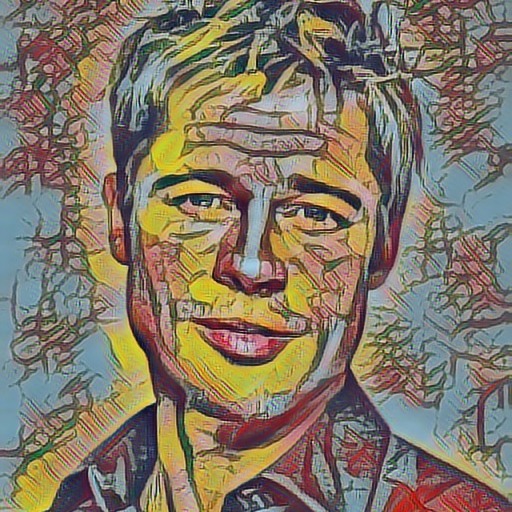}
    \includegraphics[width=0.18\linewidth]{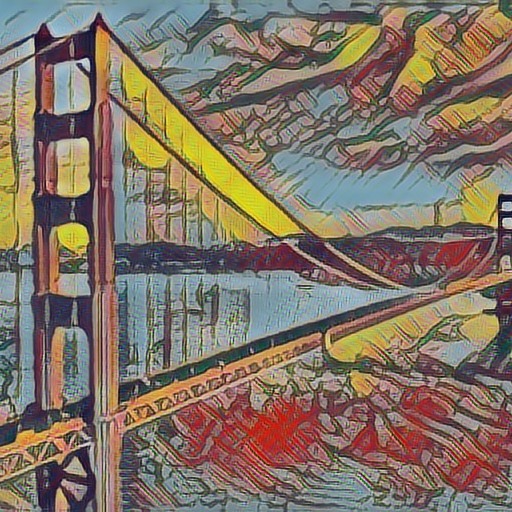}
    \includegraphics[width=0.18\linewidth]{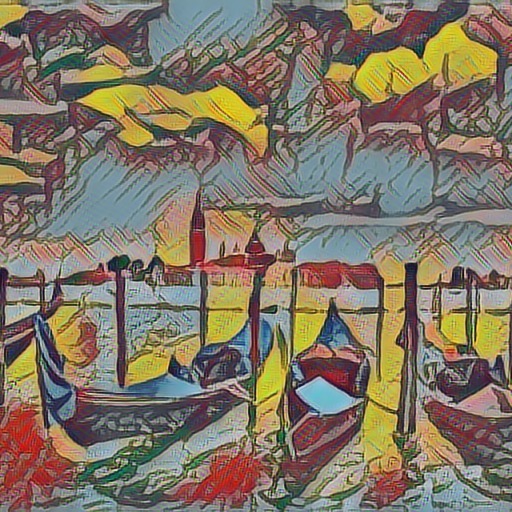}
    \includegraphics[width=0.18\linewidth]{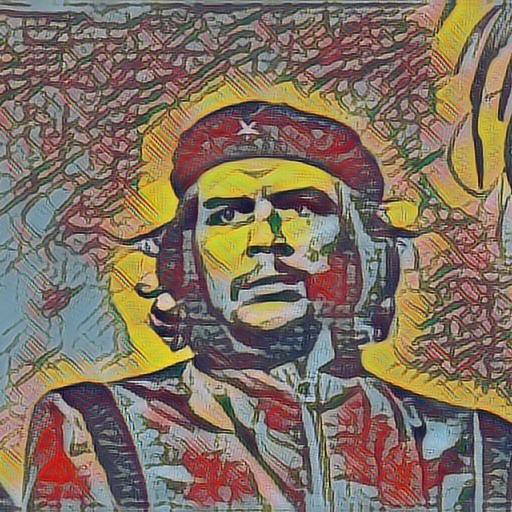} \\
    \includegraphics[width=0.18\linewidth]{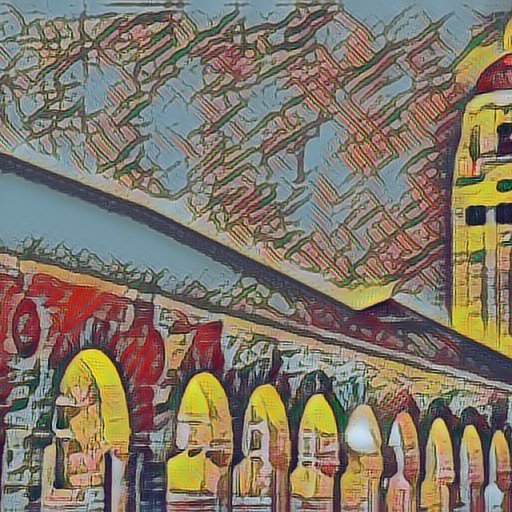}
    \includegraphics[width=0.18\linewidth]{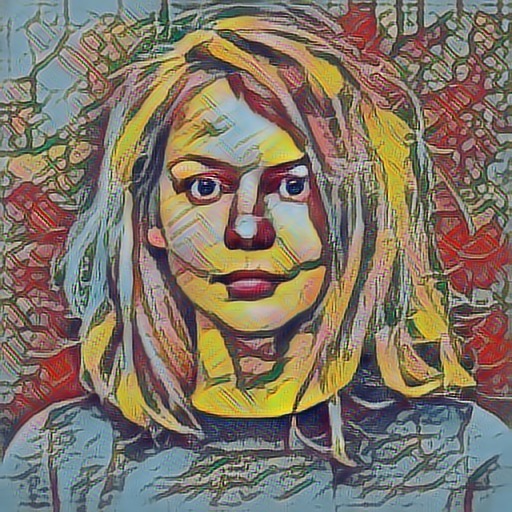}
    \includegraphics[width=0.18\linewidth]{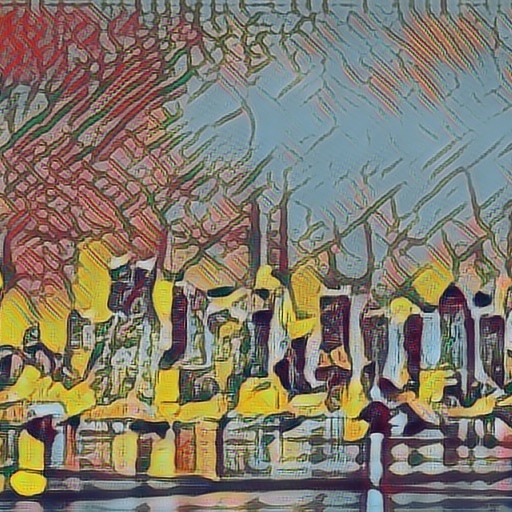}
    \includegraphics[width=0.18\linewidth]{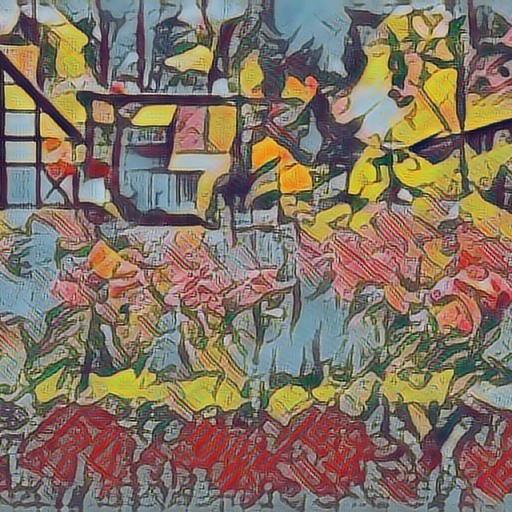}
    \includegraphics[width=0.18\linewidth]{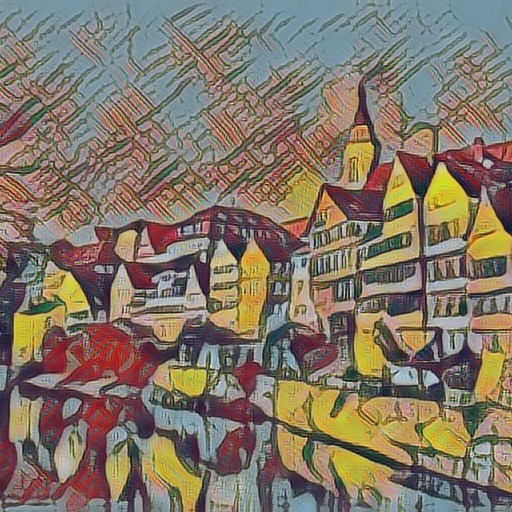}
\end{center}
\caption*{Grayson Perry, {\em The Annunciation of the Virgin Deal} (2012).}
\end{figure}

\clearpage
\begin{figure}[ht]
\begin{center}
    \includegraphics[width=0.18\linewidth]{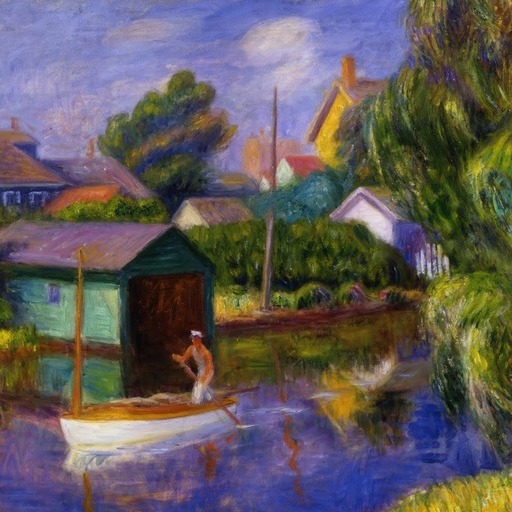}
    \includegraphics[width=0.18\linewidth]{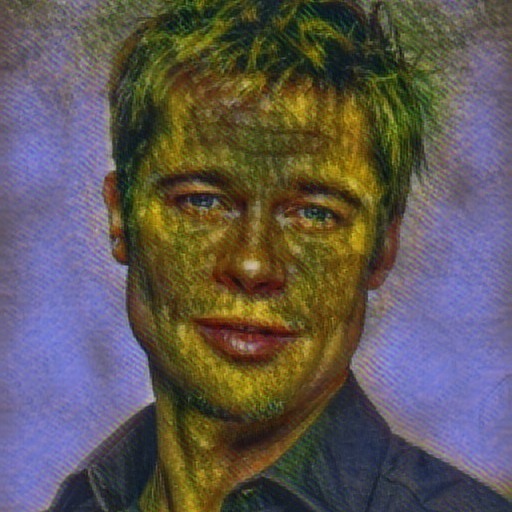}
    \includegraphics[width=0.18\linewidth]{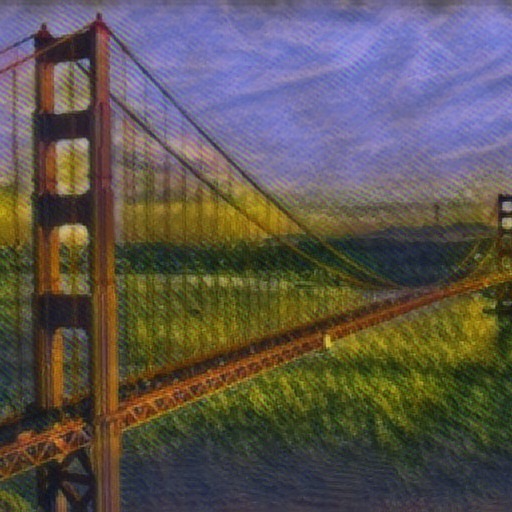}
    \includegraphics[width=0.18\linewidth]{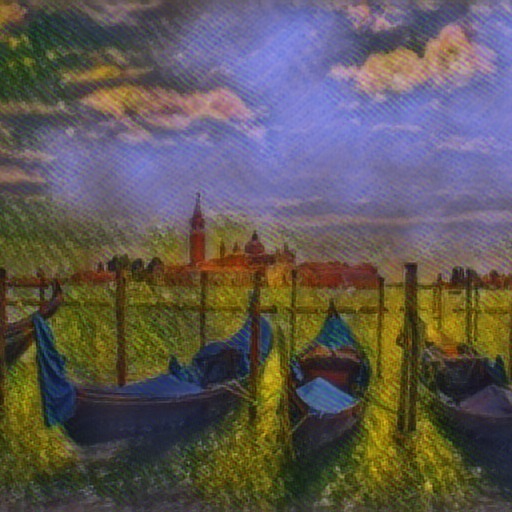}
    \includegraphics[width=0.18\linewidth]{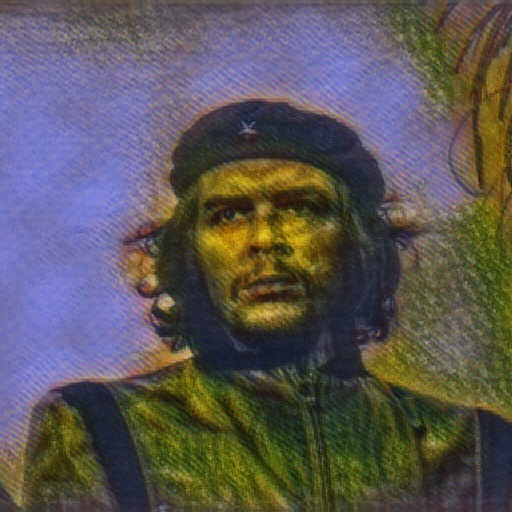} \\
    \includegraphics[width=0.18\linewidth]{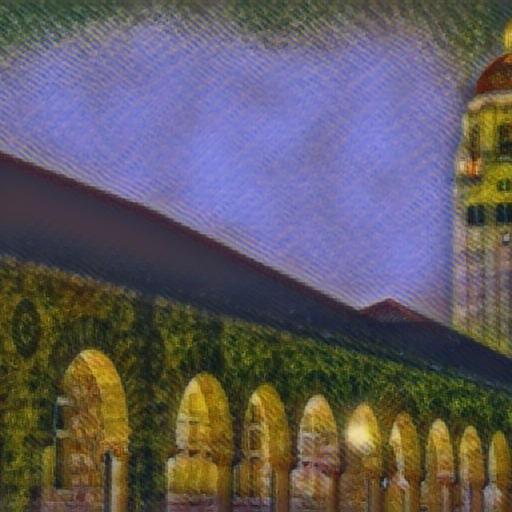}
    \includegraphics[width=0.18\linewidth]{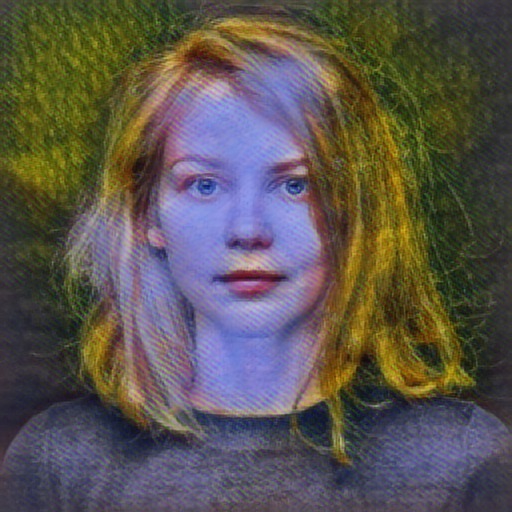}
    \includegraphics[width=0.18\linewidth]{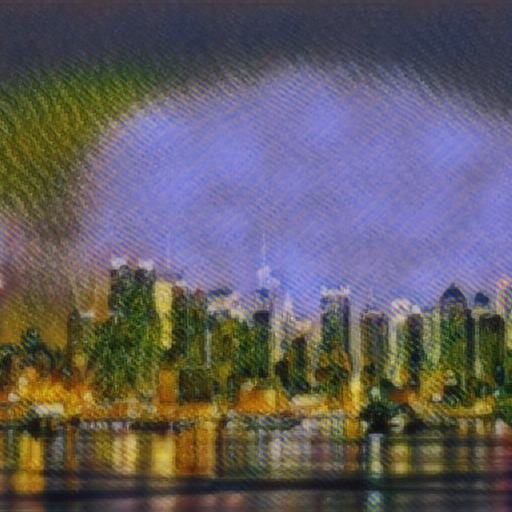}
    \includegraphics[width=0.18\linewidth]{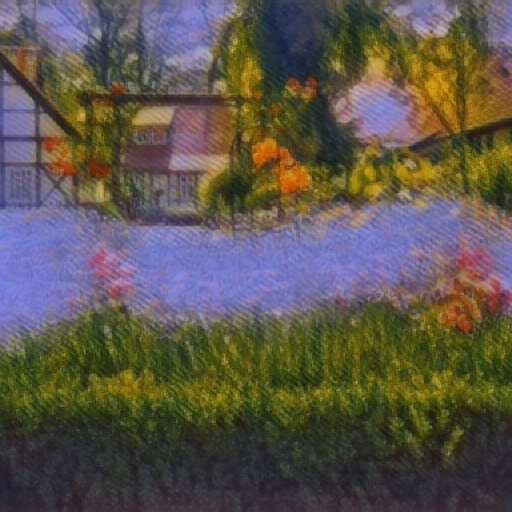}
    \includegraphics[width=0.18\linewidth]{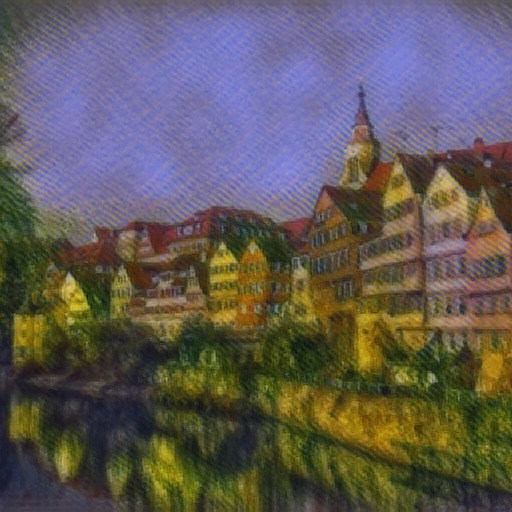}
\end{center}
\caption*{William Glackens, {\em The Green Boathouse} (ca. 1922).}
\end{figure}

\begin{figure}[ht]
\begin{center}
    \includegraphics[width=0.18\linewidth]{figures/the_scream.jpg}
    \includegraphics[width=0.18\linewidth]{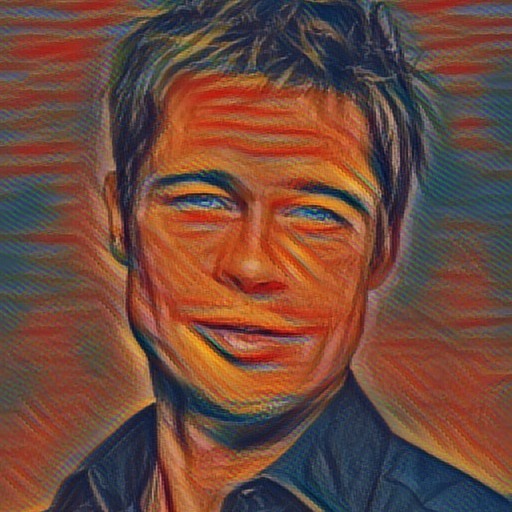}
    \includegraphics[width=0.18\linewidth]{figures/golden_gate_the_scream.jpg}
    \includegraphics[width=0.18\linewidth]{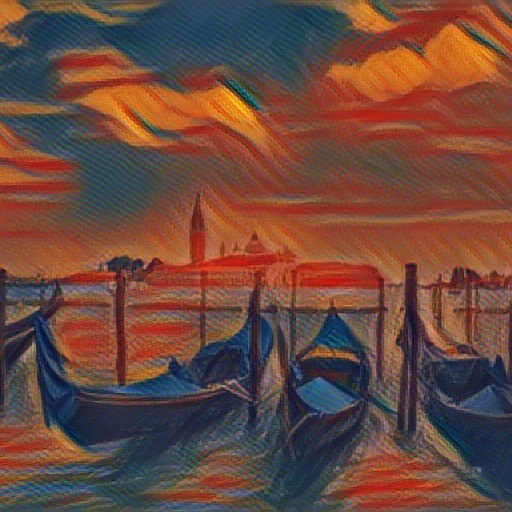}
    \includegraphics[width=0.18\linewidth]{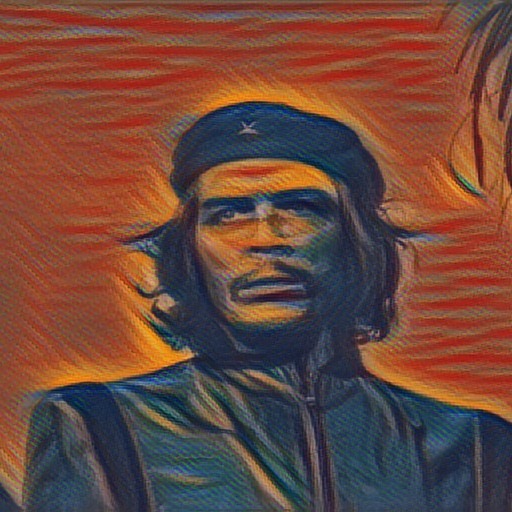} \\
    \includegraphics[width=0.18\linewidth]{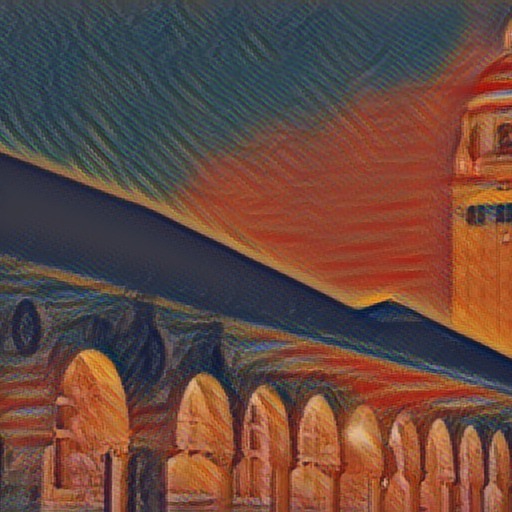}
    \includegraphics[width=0.18\linewidth]{figures/karya_the_scream.jpg}
    \includegraphics[width=0.18\linewidth]{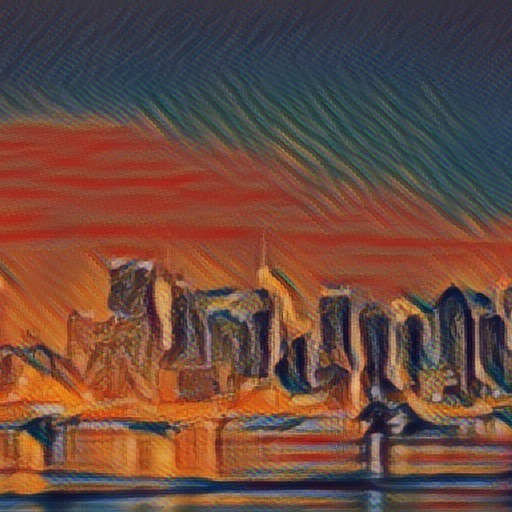}
    \includegraphics[width=0.18\linewidth]{figures/schultenhof_mettingen_bauerngarten_the_scream.jpg}
    \includegraphics[width=0.18\linewidth]{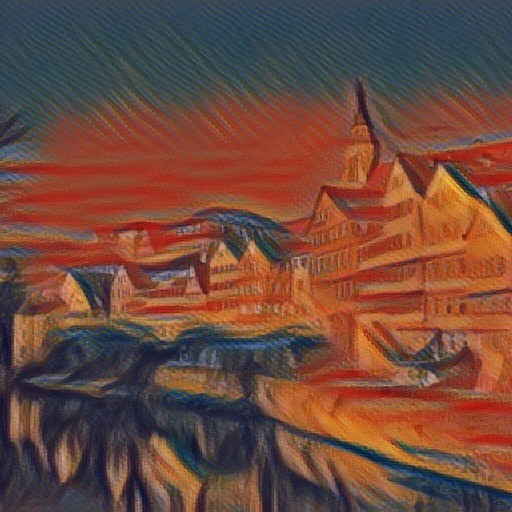}
\end{center}
\caption*{Edvard Munch, {\em The Scream} (1910).}
\end{figure}

\begin{figure}[ht]
\begin{center}
    \includegraphics[width=0.18\linewidth]{figures/the_starry_night.jpg}
    \includegraphics[width=0.18\linewidth]{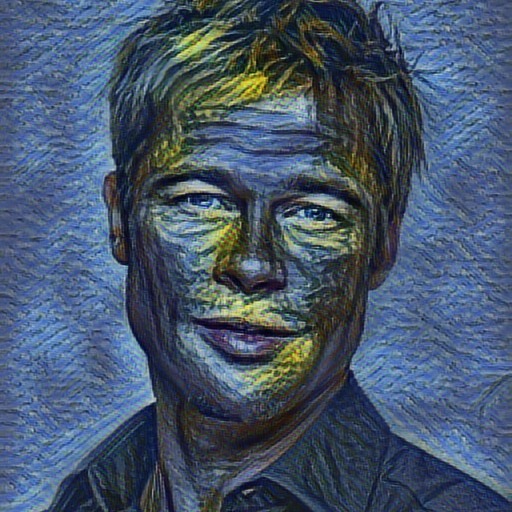}
    \includegraphics[width=0.18\linewidth]{figures/golden_gate_the_starry_night.jpg}
    \includegraphics[width=0.18\linewidth]{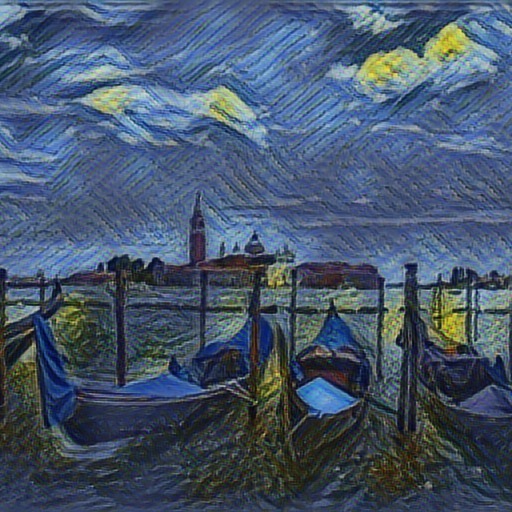}
    \includegraphics[width=0.18\linewidth]{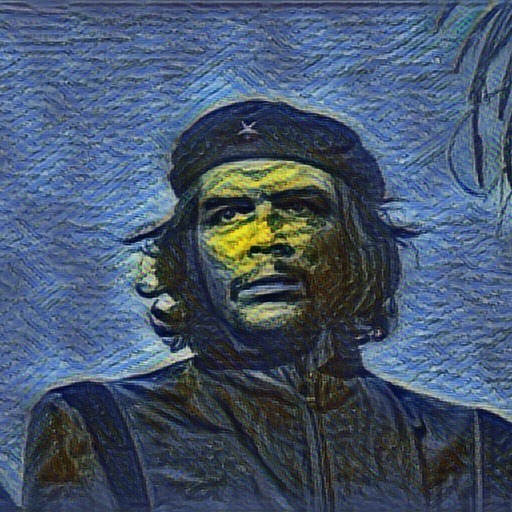} \\
    \includegraphics[width=0.18\linewidth]{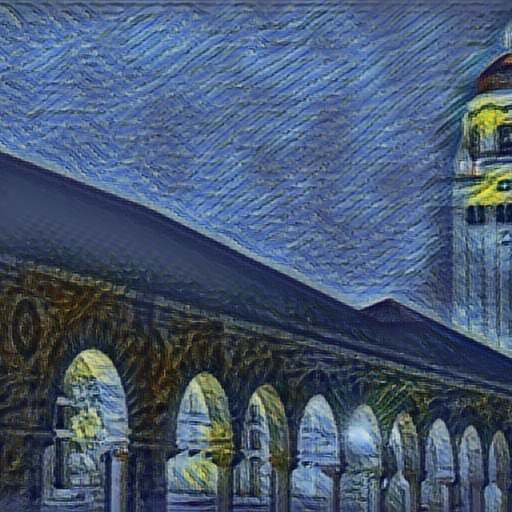}
    \includegraphics[width=0.18\linewidth]{figures/karya_the_starry_night.jpg}
    \includegraphics[width=0.18\linewidth]{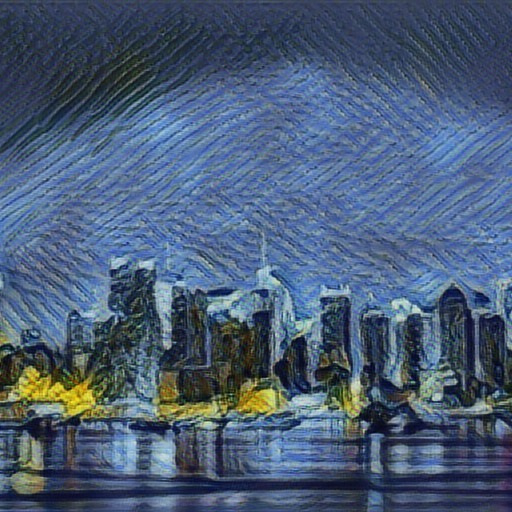}
    \includegraphics[width=0.18\linewidth]{figures/schultenhof_mettingen_bauerngarten_the_starry_night.jpg}
    \includegraphics[width=0.18\linewidth]{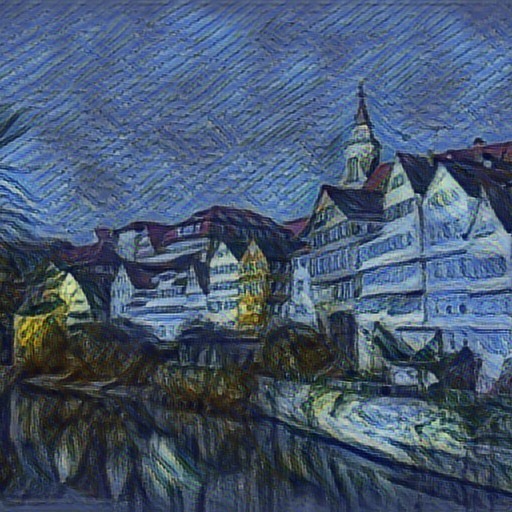}
\end{center}
\caption*{Vincent van Gogh, {\em The Starry Night} (1889).}
\end{figure}

\clearpage
\begin{figure}[ht]
\begin{center}
    \includegraphics[width=0.18\linewidth]{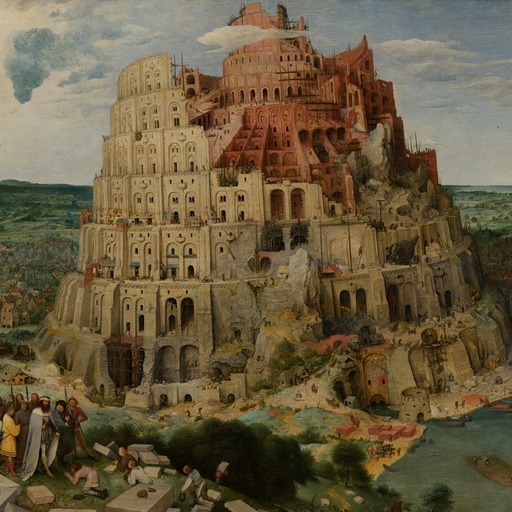}
    \includegraphics[width=0.18\linewidth]{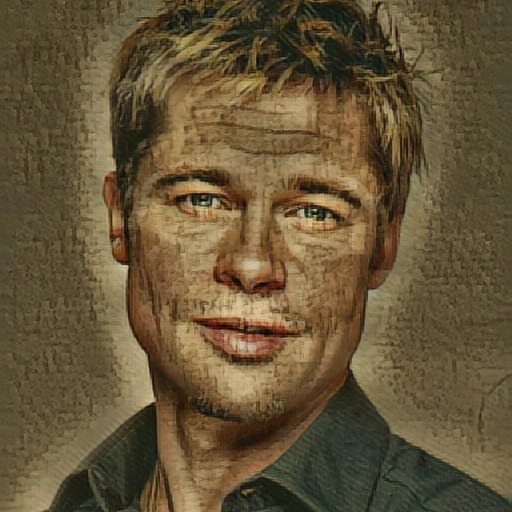}
    \includegraphics[width=0.18\linewidth]{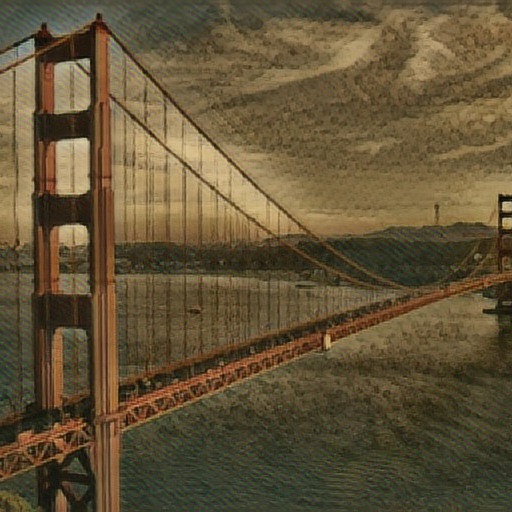}
    \includegraphics[width=0.18\linewidth]{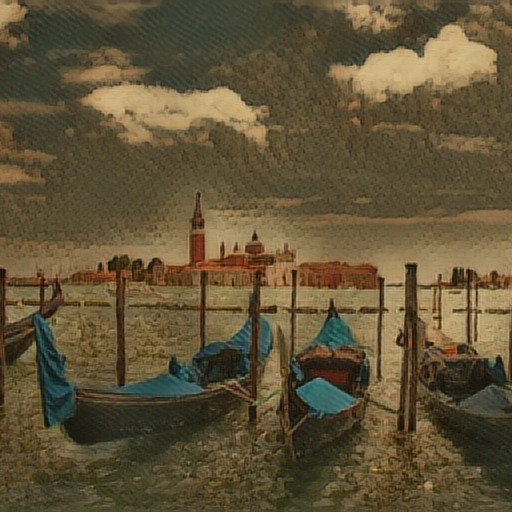}
    \includegraphics[width=0.18\linewidth]{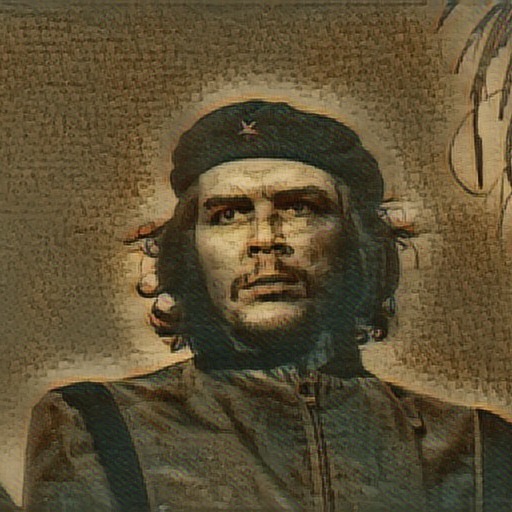} \\
    \includegraphics[width=0.18\linewidth]{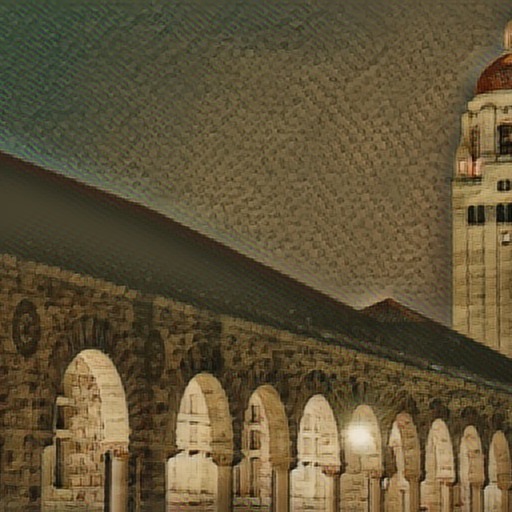}
    \includegraphics[width=0.18\linewidth]{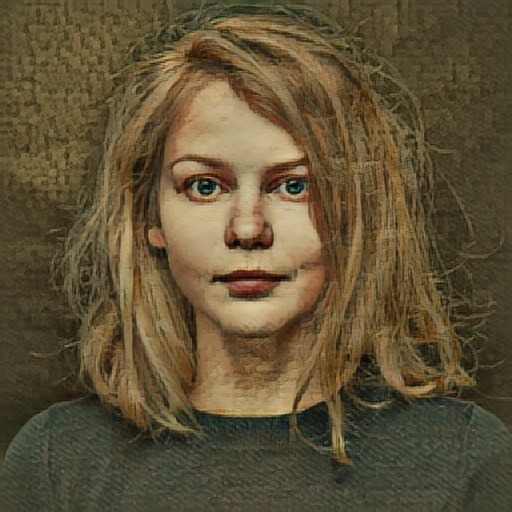}
    \includegraphics[width=0.18\linewidth]{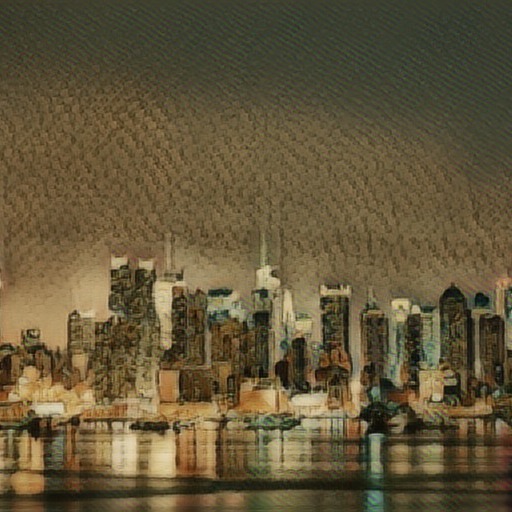}
    \includegraphics[width=0.18\linewidth]{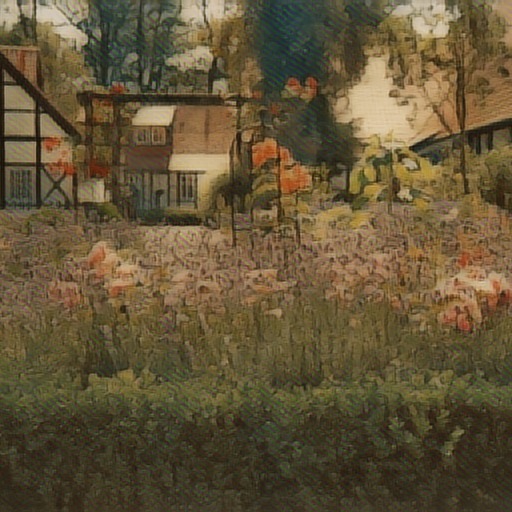}
    \includegraphics[width=0.18\linewidth]{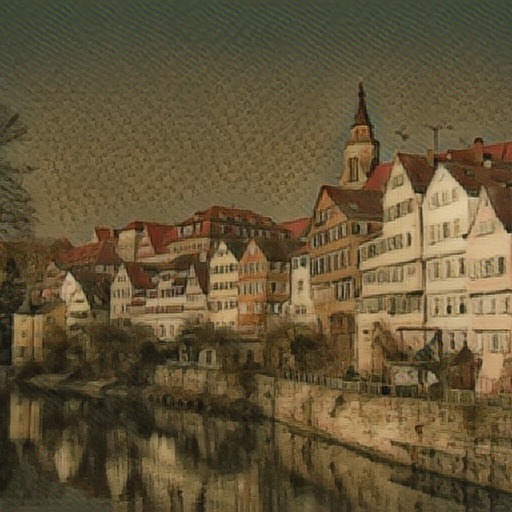}
\end{center}
\caption*{Pieter Bruegel the Elder, {\em The Tower of Babel} (1563).}
\end{figure}

\begin{figure}[ht]
\begin{center}
    \includegraphics[width=0.18\linewidth]{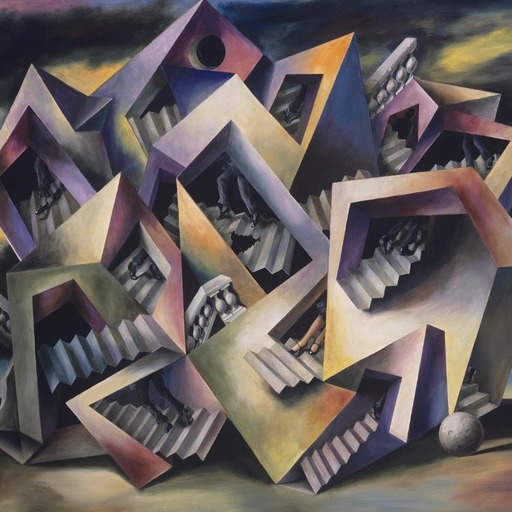}
    \includegraphics[width=0.18\linewidth]{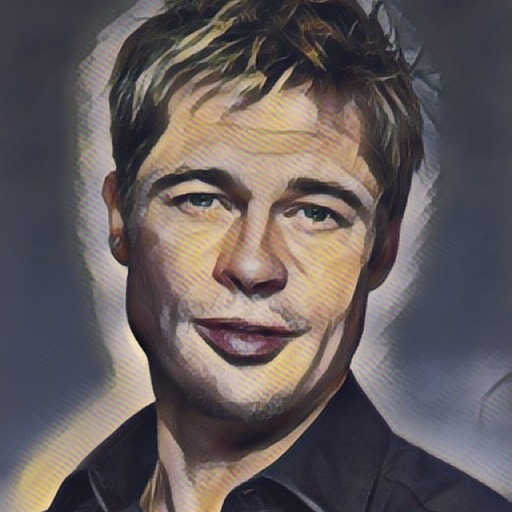}
    \includegraphics[width=0.18\linewidth]{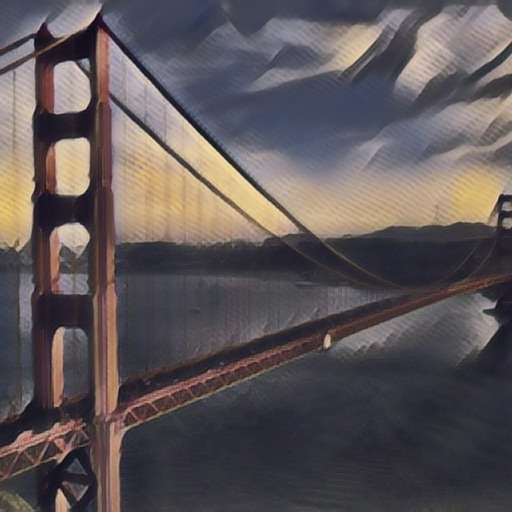}
    \includegraphics[width=0.18\linewidth]{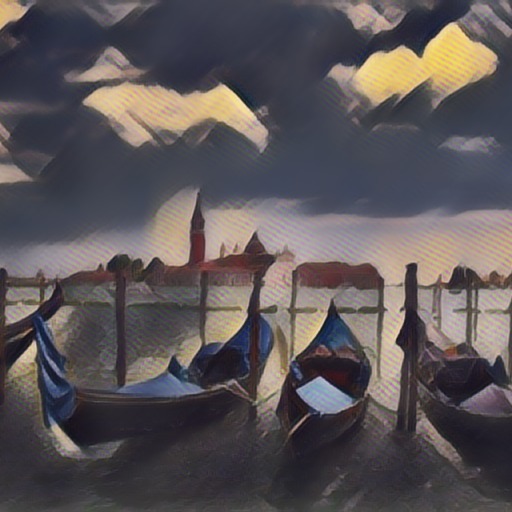}
    \includegraphics[width=0.18\linewidth]{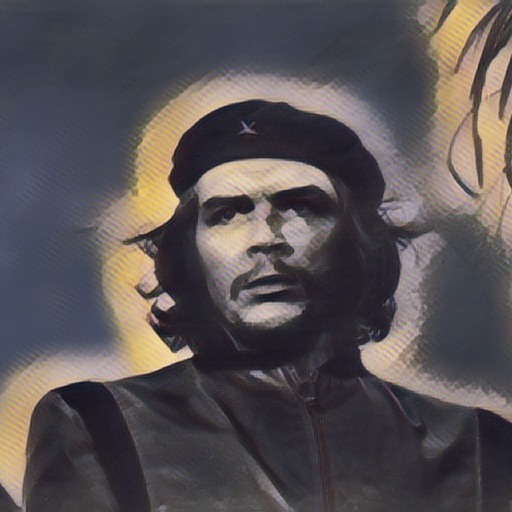} \\
    \includegraphics[width=0.18\linewidth]{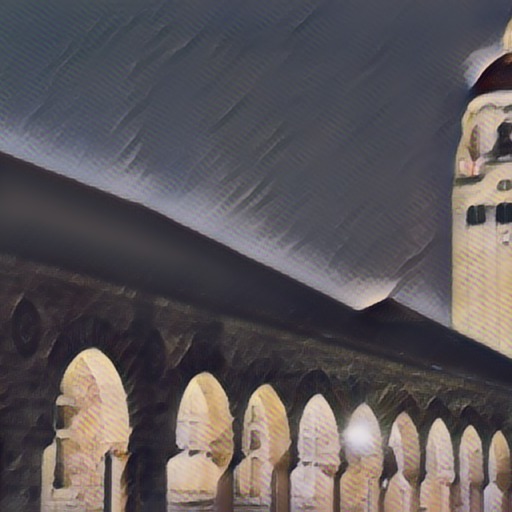}
    \includegraphics[width=0.18\linewidth]{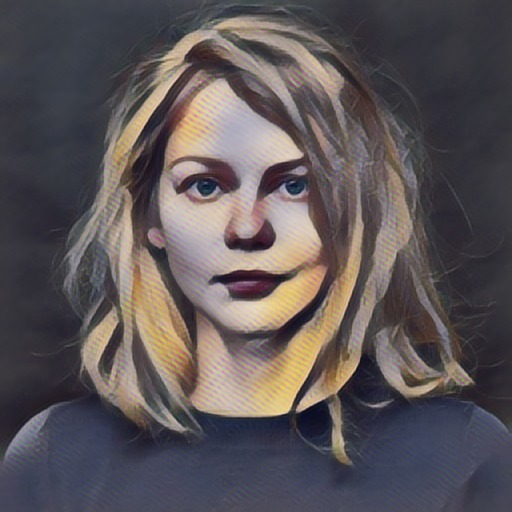}
    \includegraphics[width=0.18\linewidth]{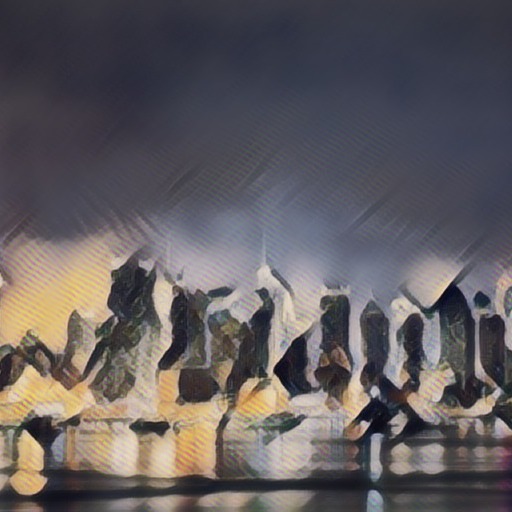}
    \includegraphics[width=0.18\linewidth]{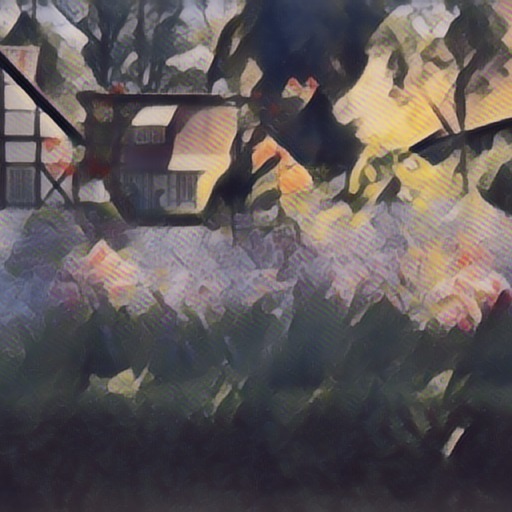}
    \includegraphics[width=0.18\linewidth]{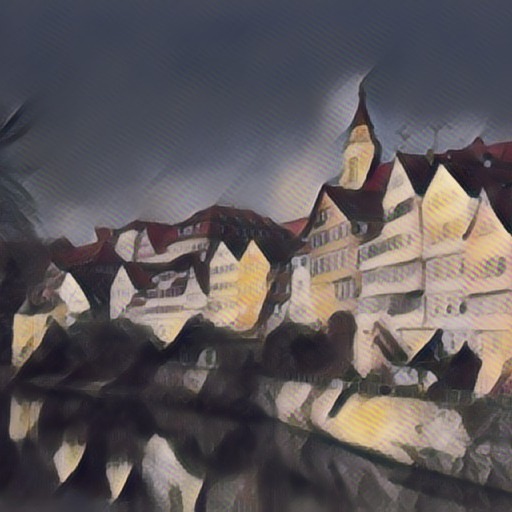}
\end{center}
\caption*{Wolfgang Lettl, {\em The Trial} (1981).}
\end{figure}

\begin{figure}[ht]
\begin{center}
    \includegraphics[width=0.18\linewidth]{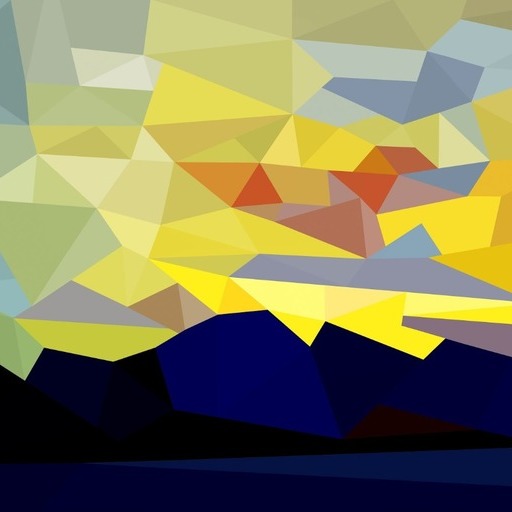}
    \includegraphics[width=0.18\linewidth]{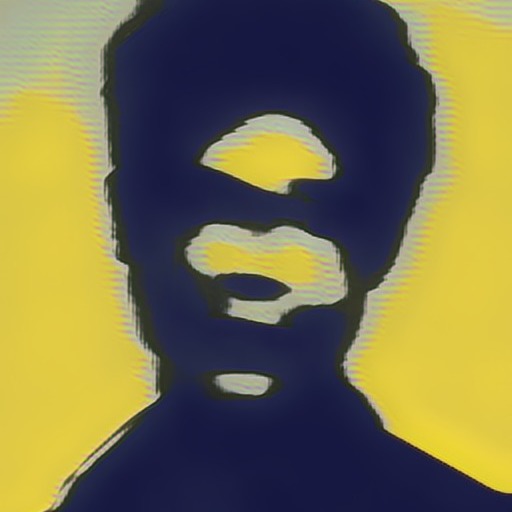}
    \includegraphics[width=0.18\linewidth]{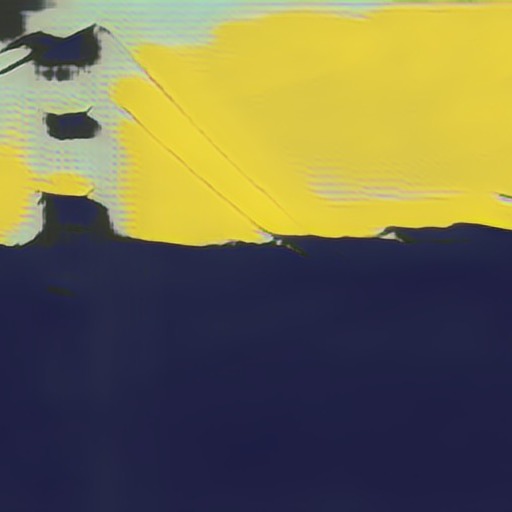}
    \includegraphics[width=0.18\linewidth]{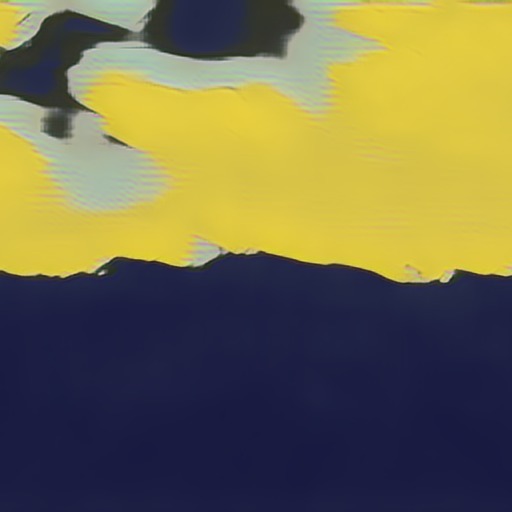}
    \includegraphics[width=0.18\linewidth]{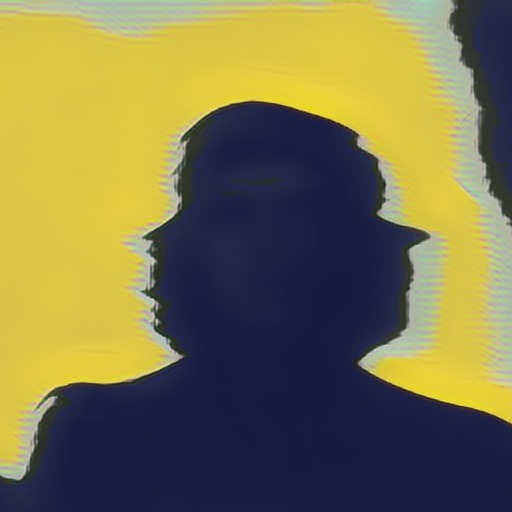} \\
    \includegraphics[width=0.18\linewidth]{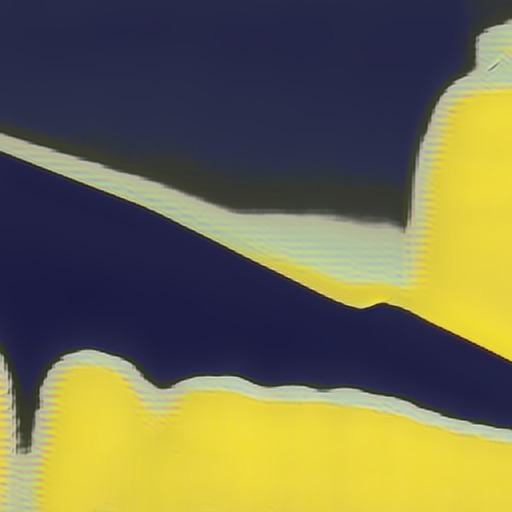}
    \includegraphics[width=0.18\linewidth]{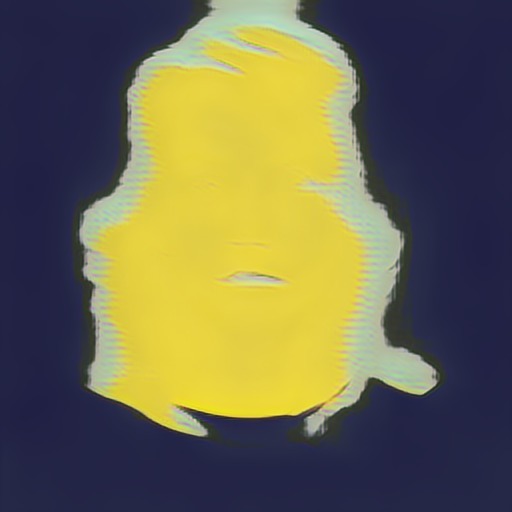}
    \includegraphics[width=0.18\linewidth]{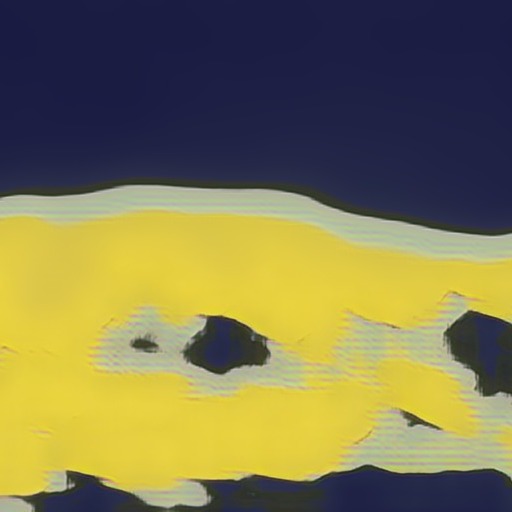}
    \includegraphics[width=0.18\linewidth]{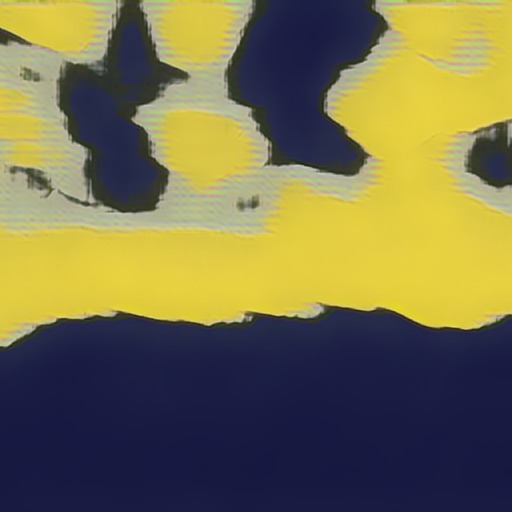}
    \includegraphics[width=0.18\linewidth]{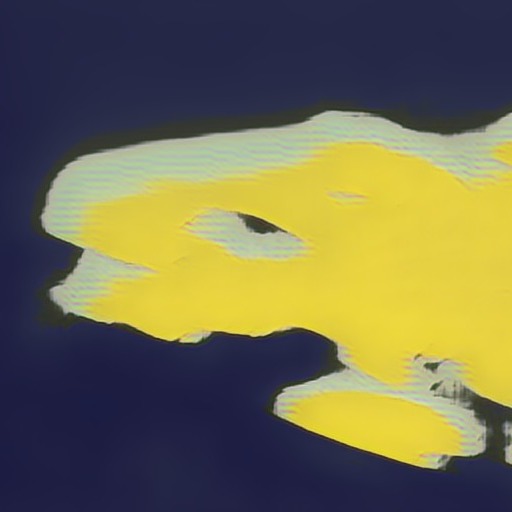}
\end{center}
\caption*{Douglas Coupland, {\em Thomson No. 5 (Yellow Sunset)} (2011).}
\end{figure}

\clearpage
\begin{figure}[ht]
\begin{center}
    \includegraphics[width=0.18\linewidth]{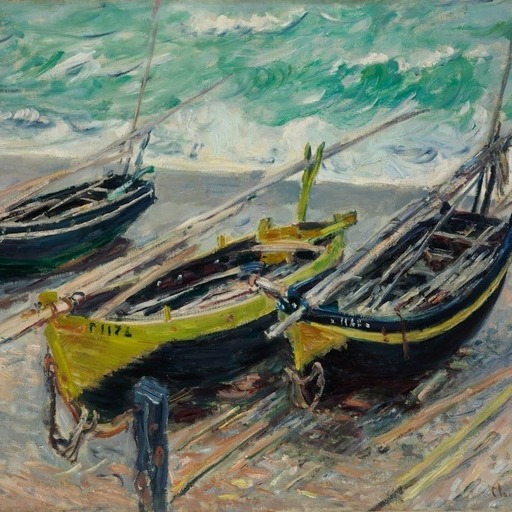}
    \includegraphics[width=0.18\linewidth]{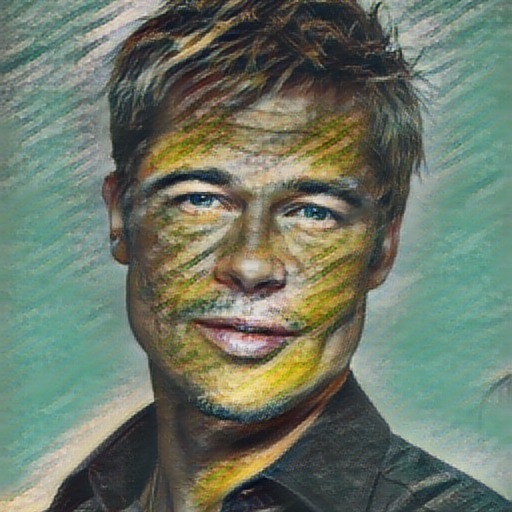}
    \includegraphics[width=0.18\linewidth]{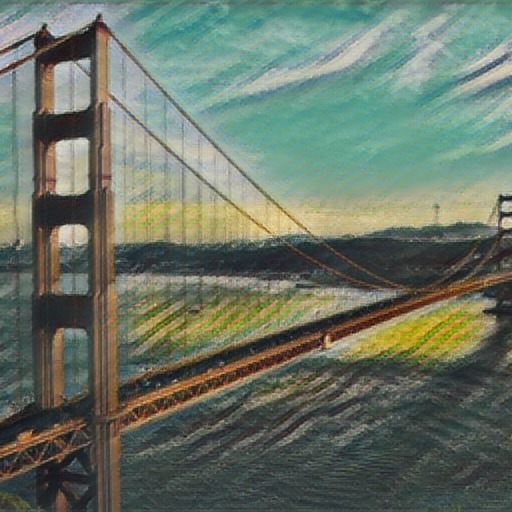}
    \includegraphics[width=0.18\linewidth]{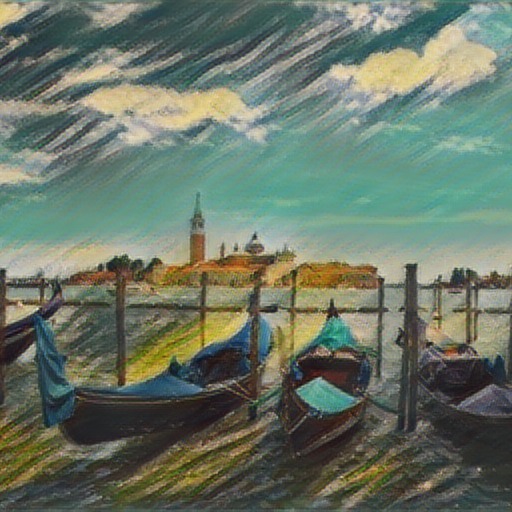}
    \includegraphics[width=0.18\linewidth]{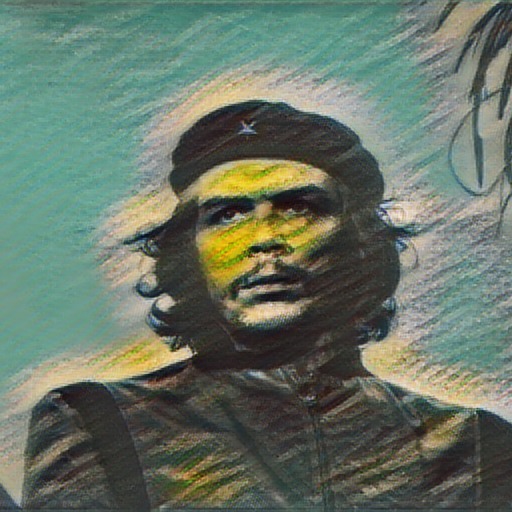} \\
    \includegraphics[width=0.18\linewidth]{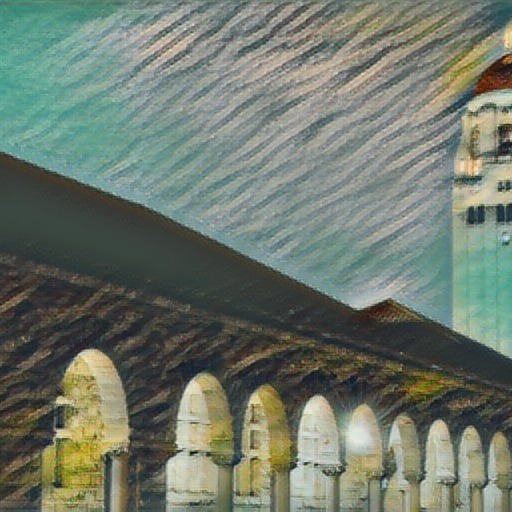}
    \includegraphics[width=0.18\linewidth]{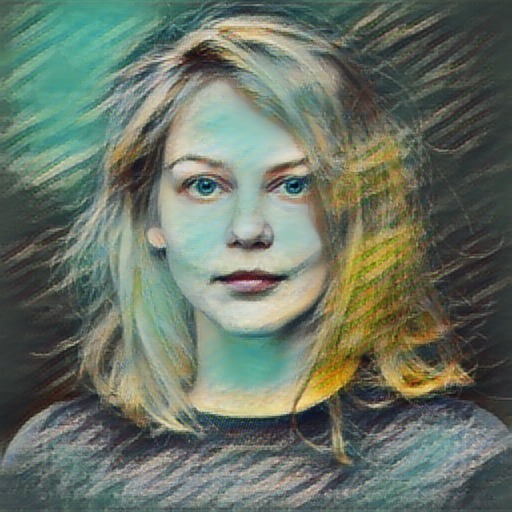}
    \includegraphics[width=0.18\linewidth]{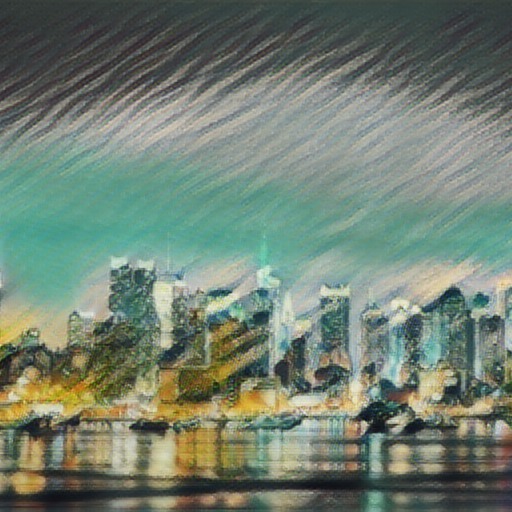}
    \includegraphics[width=0.18\linewidth]{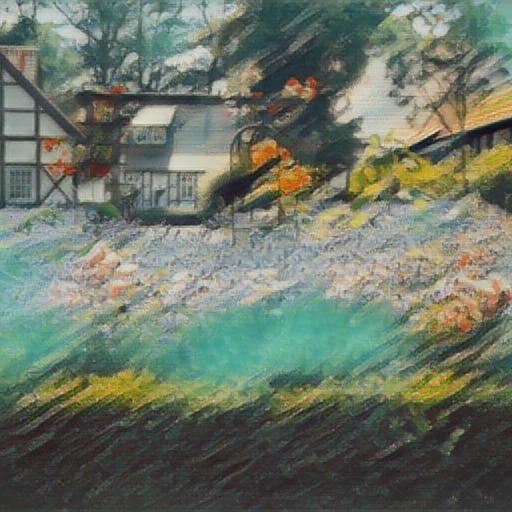}
    \includegraphics[width=0.18\linewidth]{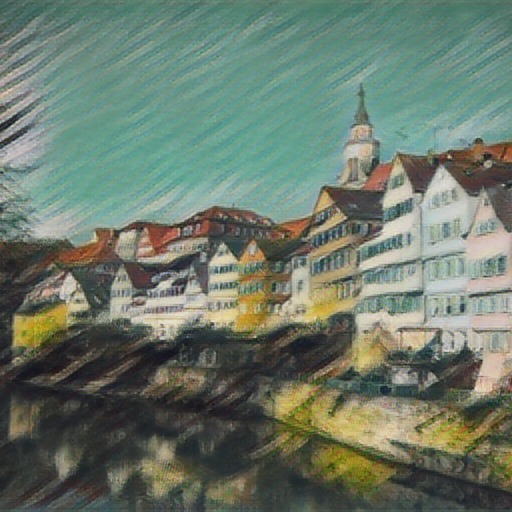}
\end{center}
\caption*{Claude Monet, {\em Three Fishing Boats} (1886).}
\end{figure}

\begin{figure}[ht]
\begin{center}
    \includegraphics[width=0.18\linewidth]{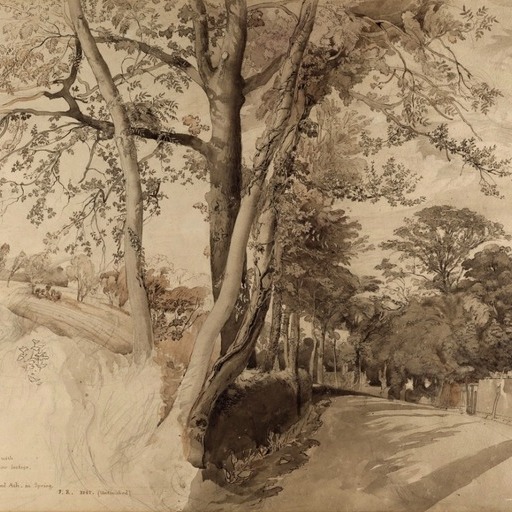}
    \includegraphics[width=0.18\linewidth]{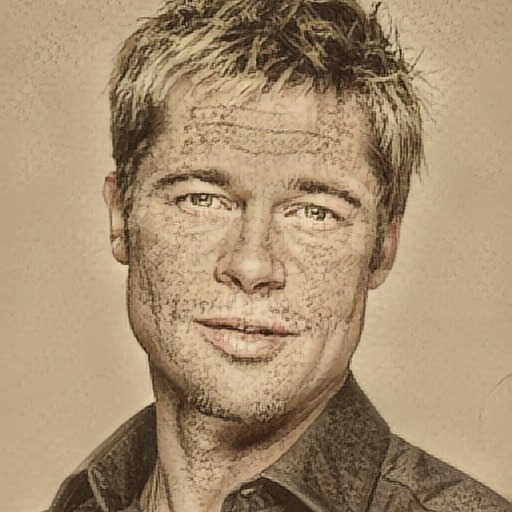}
    \includegraphics[width=0.18\linewidth]{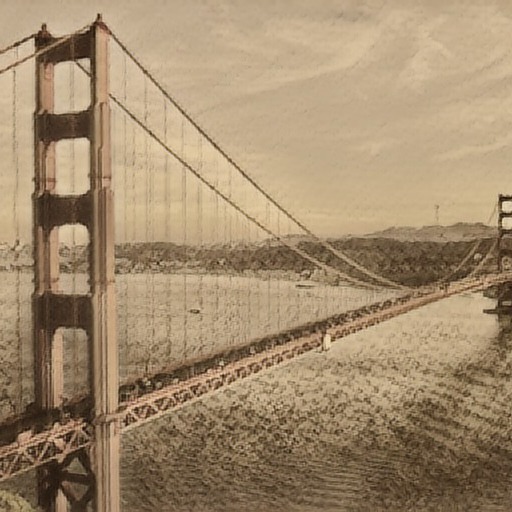}
    \includegraphics[width=0.18\linewidth]{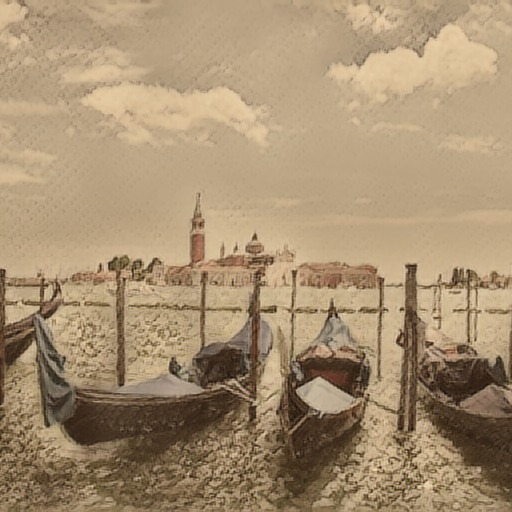}
    \includegraphics[width=0.18\linewidth]{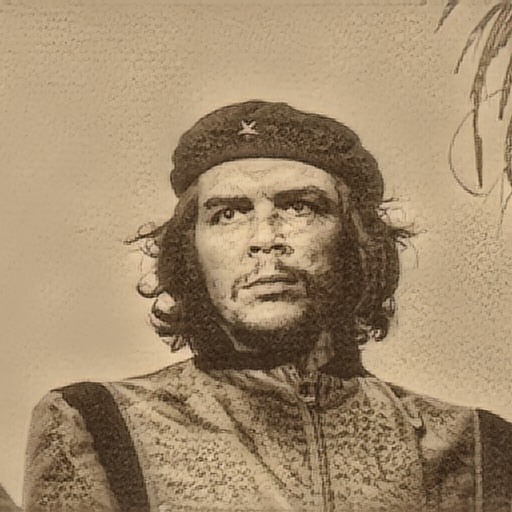} \\
    \includegraphics[width=0.18\linewidth]{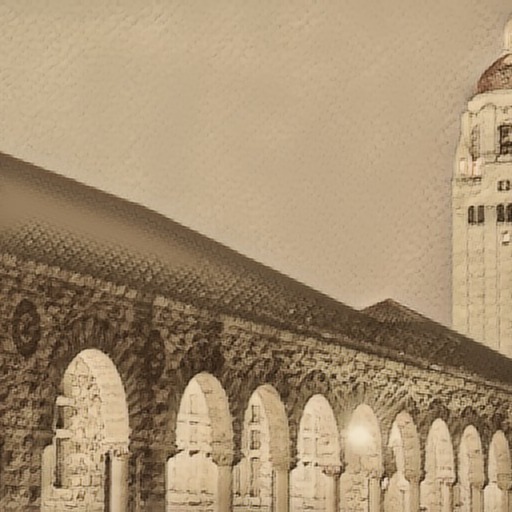}
    \includegraphics[width=0.18\linewidth]{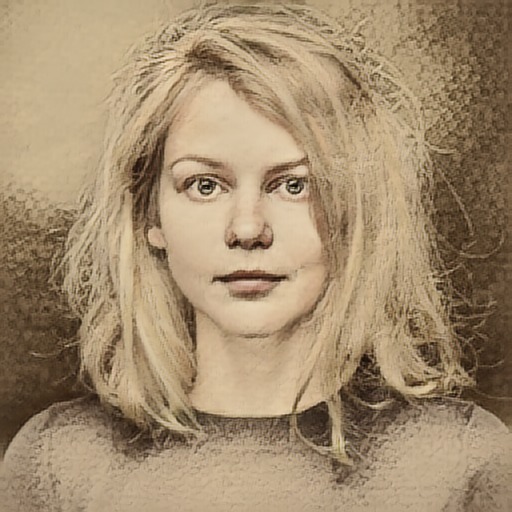}
    \includegraphics[width=0.18\linewidth]{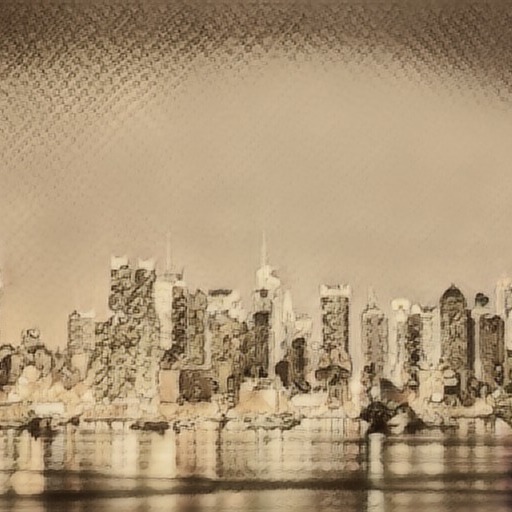}
    \includegraphics[width=0.18\linewidth]{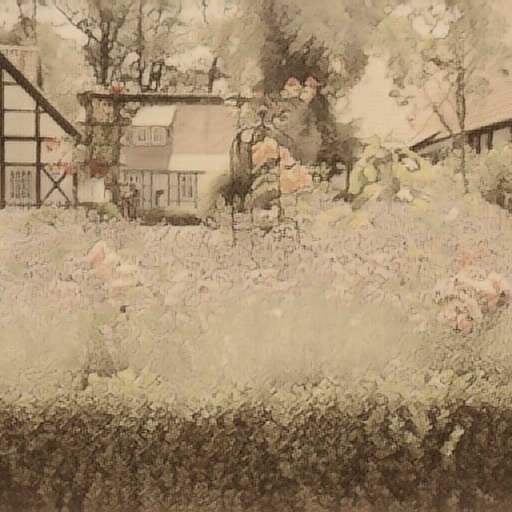}
    \includegraphics[width=0.18\linewidth]{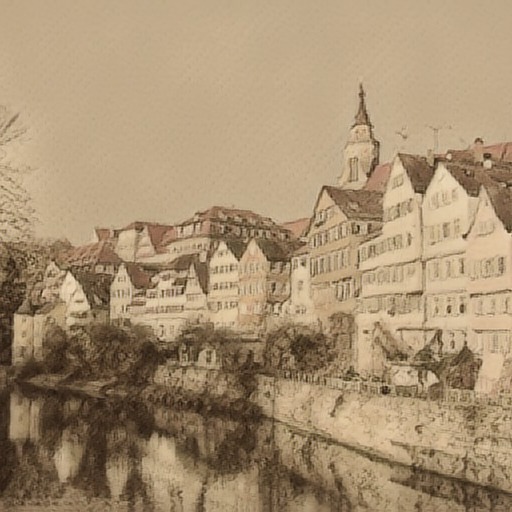}
\end{center}
\caption*{John Ruskin, {\em Trees in a Lane} (1847).}
\end{figure}

\begin{figure}[ht]
\begin{center}
    \includegraphics[width=0.18\linewidth]{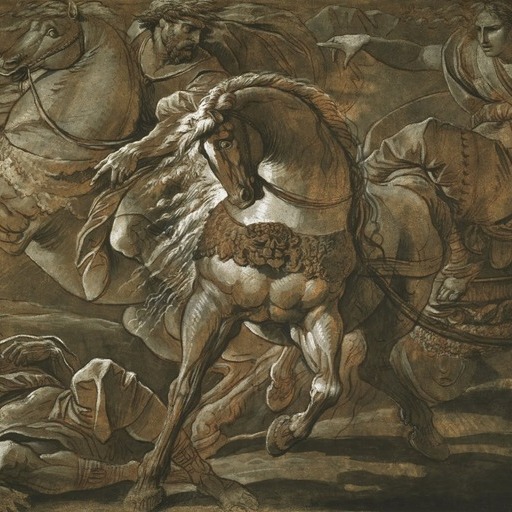}
    \includegraphics[width=0.18\linewidth]{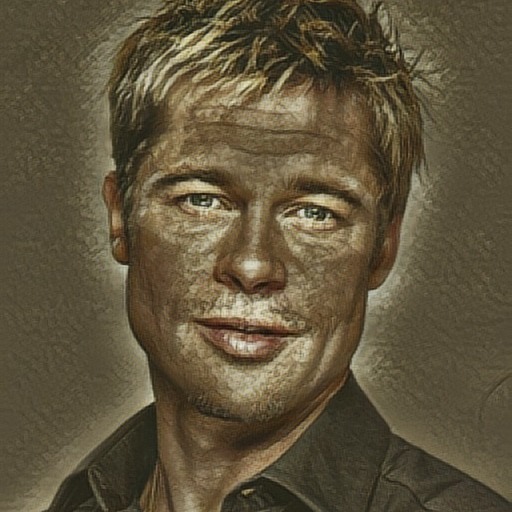}
    \includegraphics[width=0.18\linewidth]{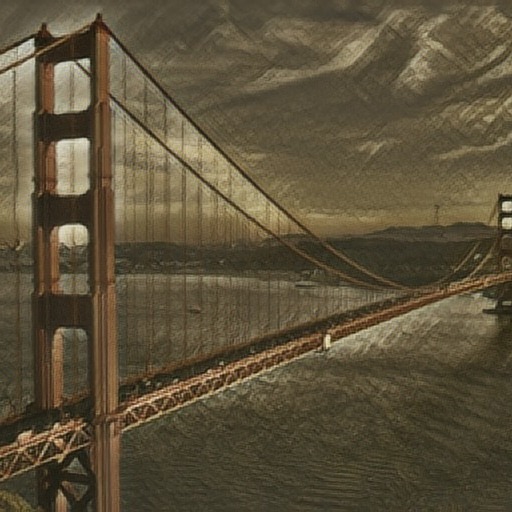}
    \includegraphics[width=0.18\linewidth]{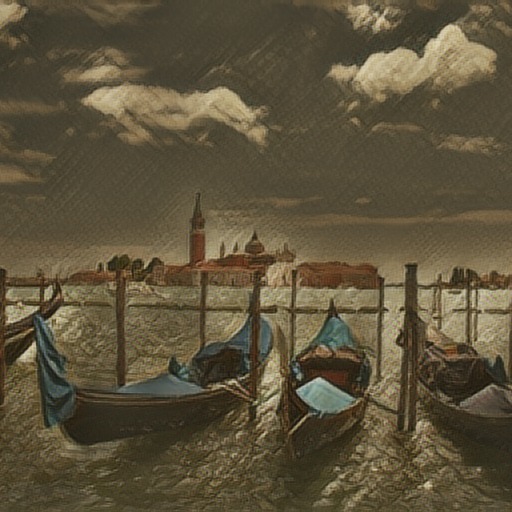}
    \includegraphics[width=0.18\linewidth]{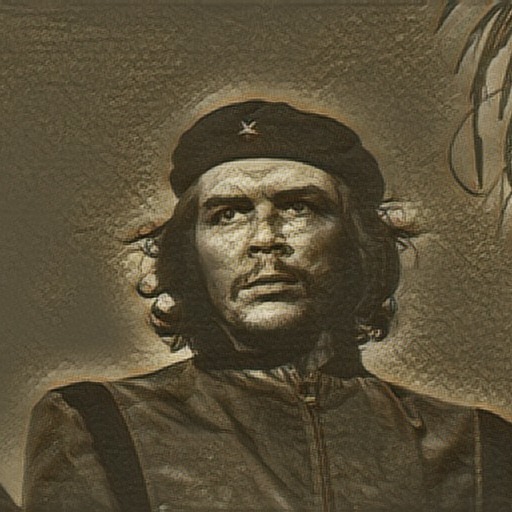} \\
    \includegraphics[width=0.18\linewidth]{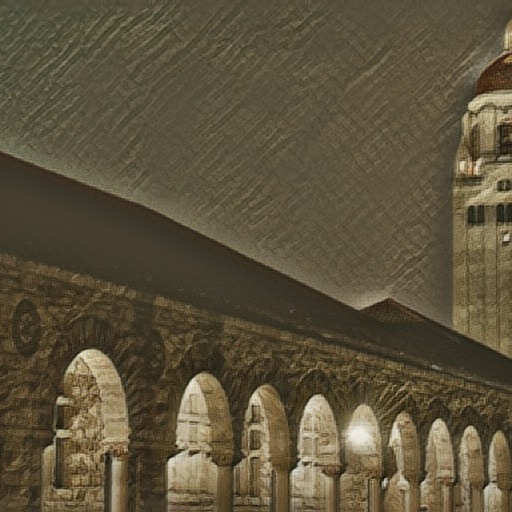}
    \includegraphics[width=0.18\linewidth]{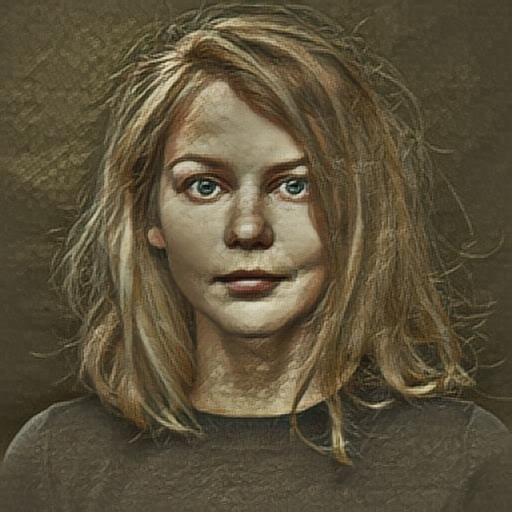}
    \includegraphics[width=0.18\linewidth]{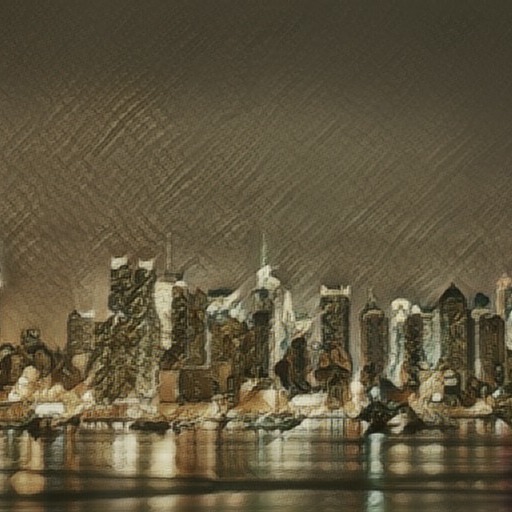}
    \includegraphics[width=0.18\linewidth]{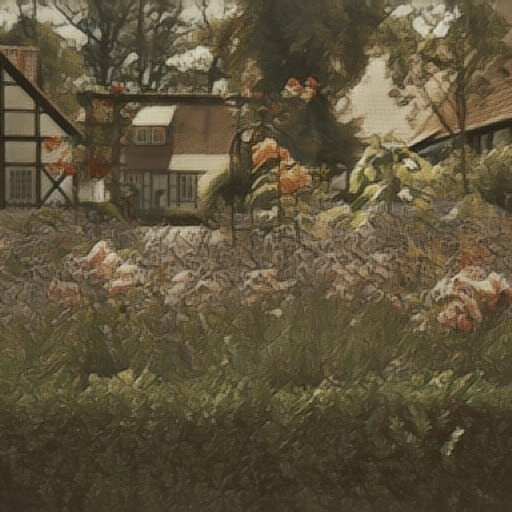}
    \includegraphics[width=0.18\linewidth]{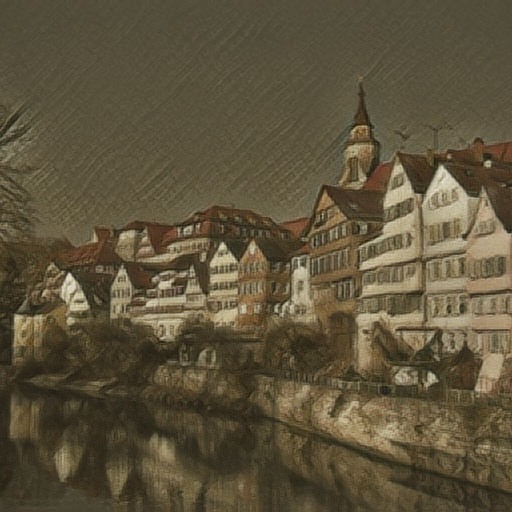}
\end{center}
\caption*{Giuseppe Cades, {\em Tullia about to Ride over the Body of Her Father
    in Her Chariot} (about 1770-1775).}
\end{figure}

\clearpage
\begin{figure}[ht]
\begin{center}
    \includegraphics[width=0.18\linewidth]{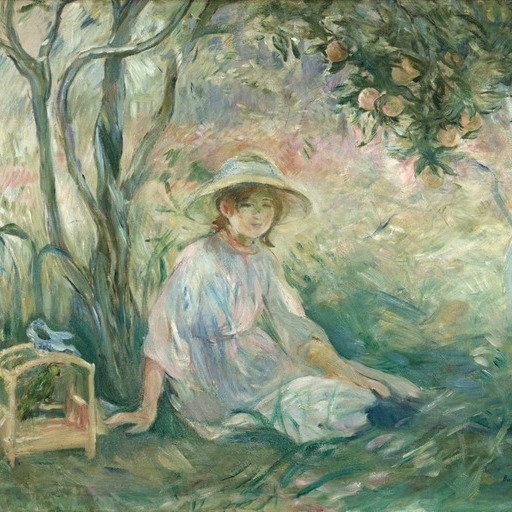}
    \includegraphics[width=0.18\linewidth]{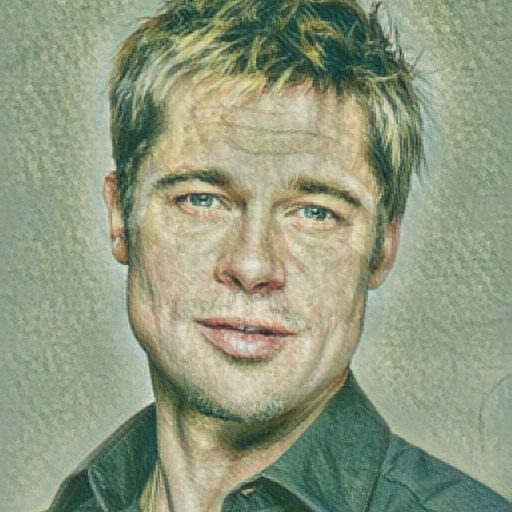}
    \includegraphics[width=0.18\linewidth]{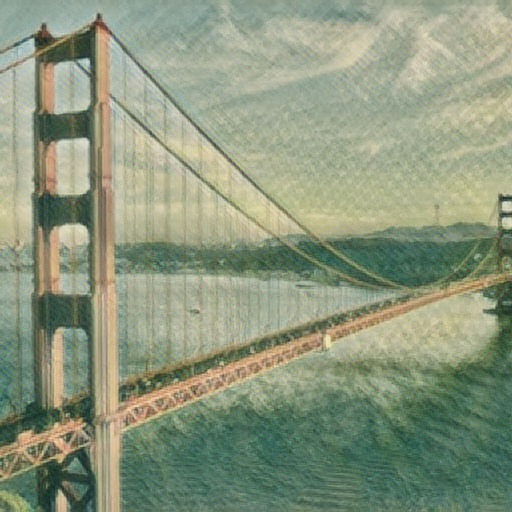}
    \includegraphics[width=0.18\linewidth]{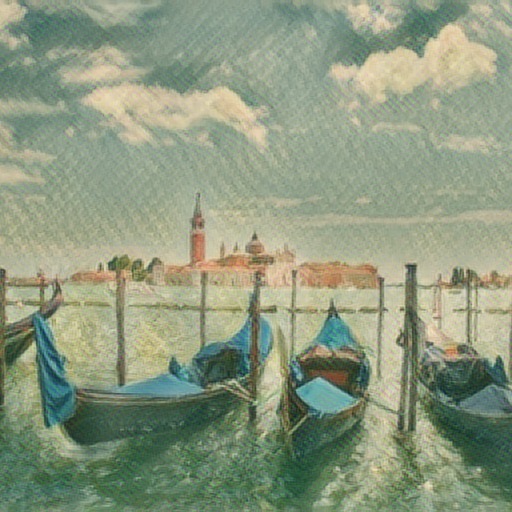}
    \includegraphics[width=0.18\linewidth]{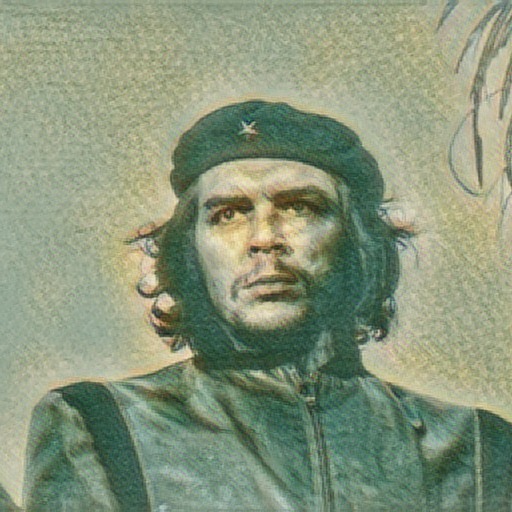} \\
    \includegraphics[width=0.18\linewidth]{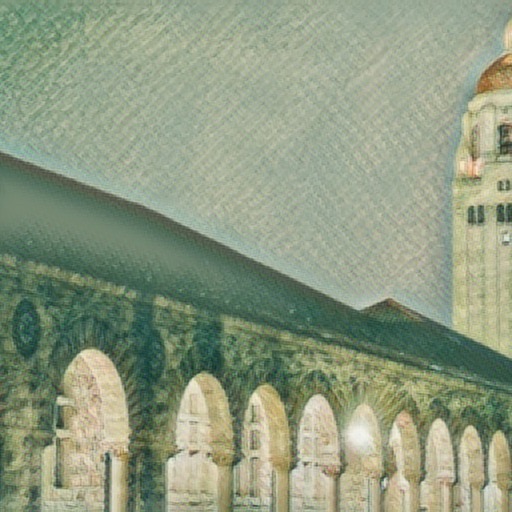}
    \includegraphics[width=0.18\linewidth]{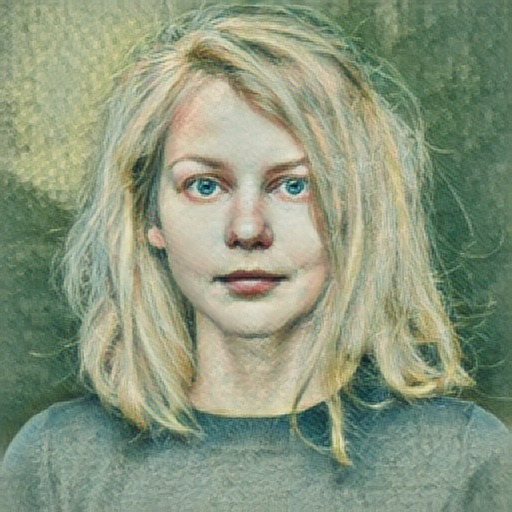}
    \includegraphics[width=0.18\linewidth]{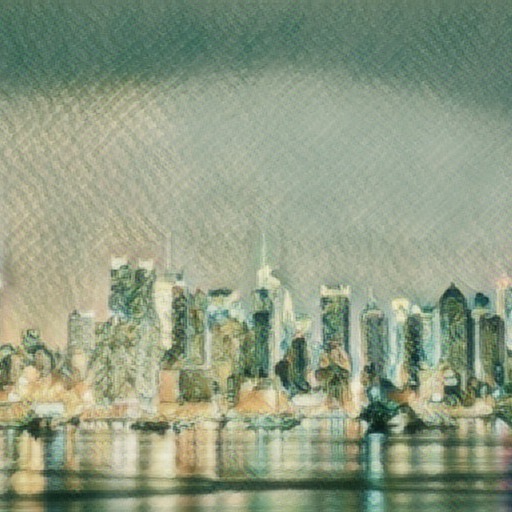}
    \includegraphics[width=0.18\linewidth]{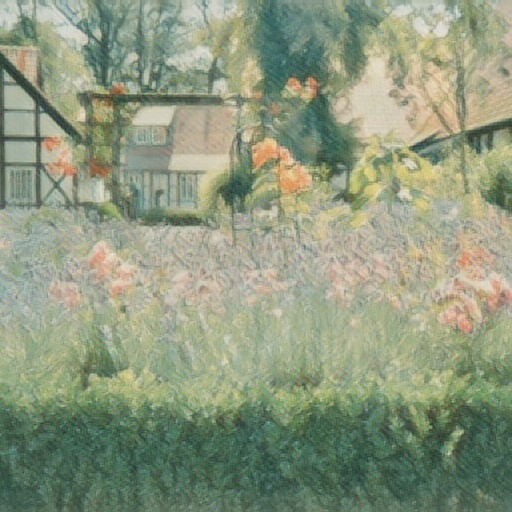}
    \includegraphics[width=0.18\linewidth]{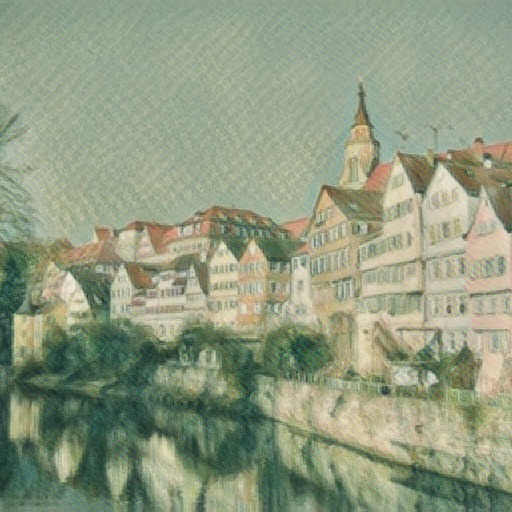}
\end{center}
\caption*{Berthe Morisot, {\em Under the Orange Tree} (1889).}
\end{figure}

\begin{figure}[ht]
\begin{center}
    \includegraphics[width=0.18\linewidth]{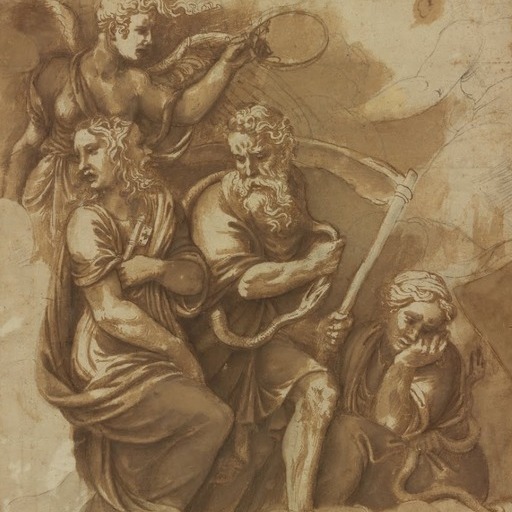}
    \includegraphics[width=0.18\linewidth]{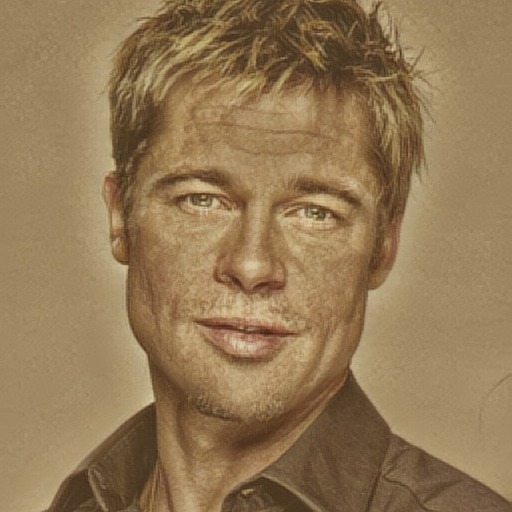}
    \includegraphics[width=0.18\linewidth]{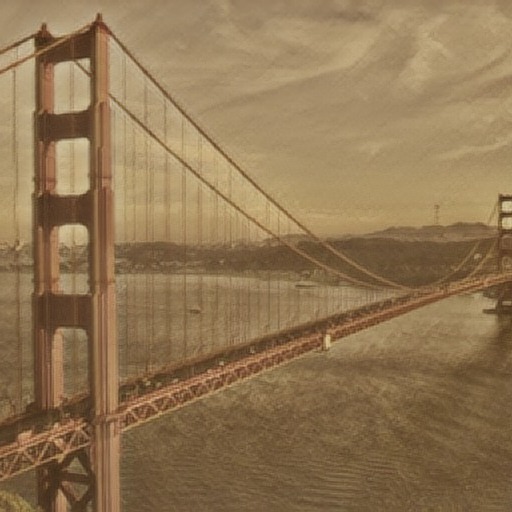}
    \includegraphics[width=0.18\linewidth]{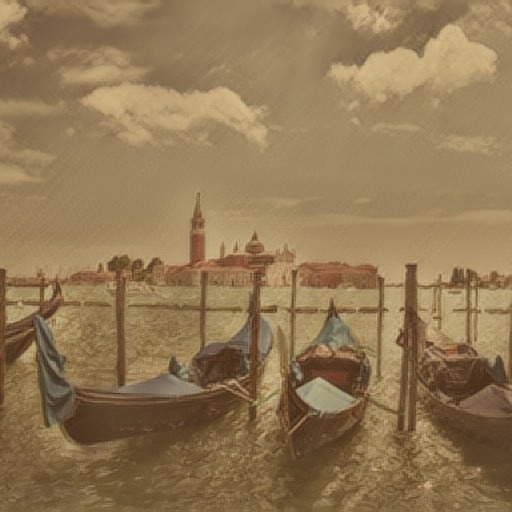}
    \includegraphics[width=0.18\linewidth]{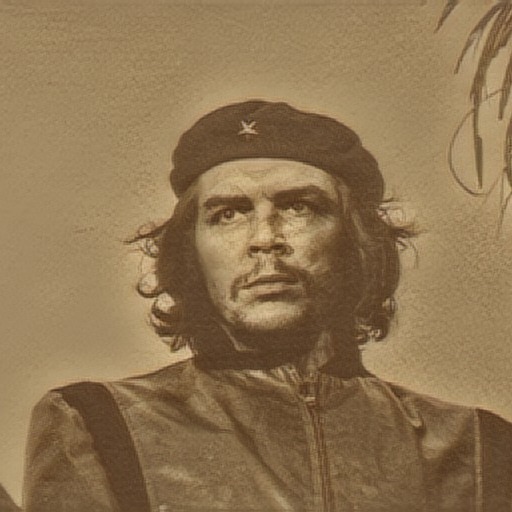} \\
    \includegraphics[width=0.18\linewidth]{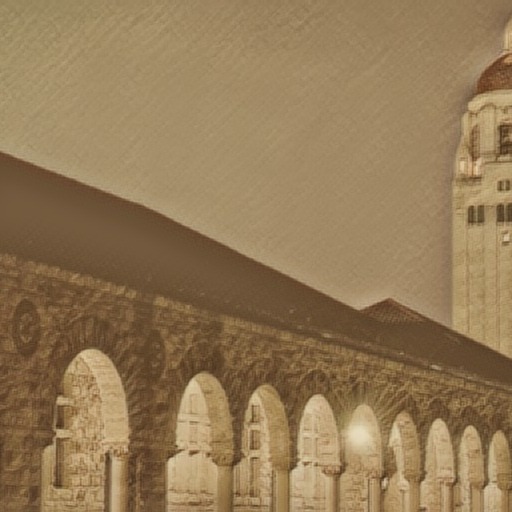}
    \includegraphics[width=0.18\linewidth]{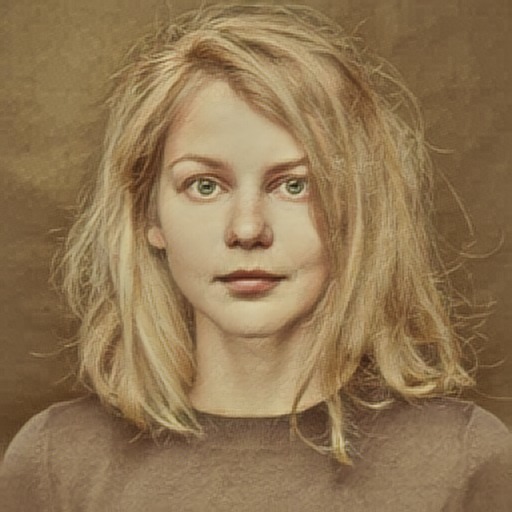}
    \includegraphics[width=0.18\linewidth]{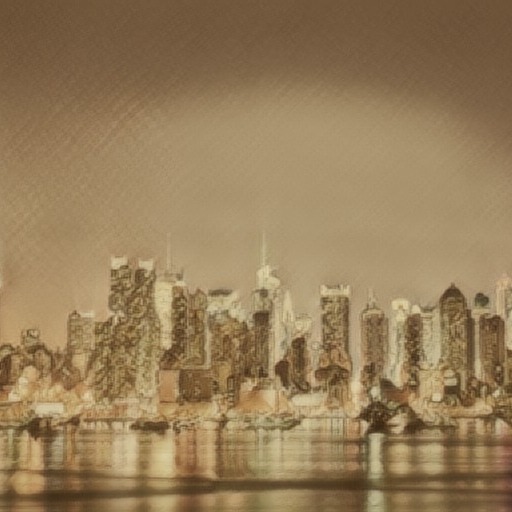}
    \includegraphics[width=0.18\linewidth]{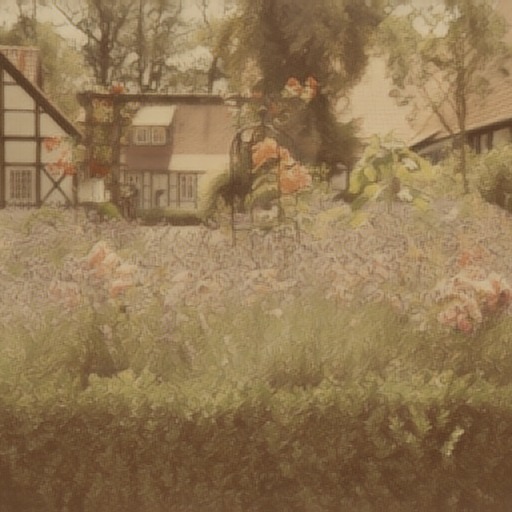}
    \includegraphics[width=0.18\linewidth]{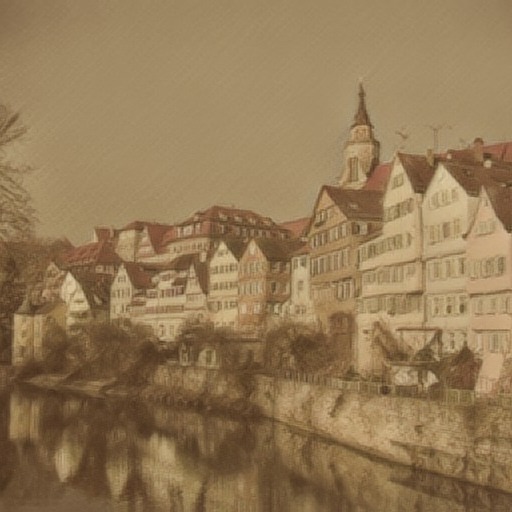}
\end{center}
\caption*{Giulio Romano (Giulio Pippi), {\em Victory, Janus, Chronos and Gaea}
    (about 1532-1534).}
\end{figure}

\begin{figure}[ht]
\begin{center}
    \includegraphics[width=0.18\linewidth]{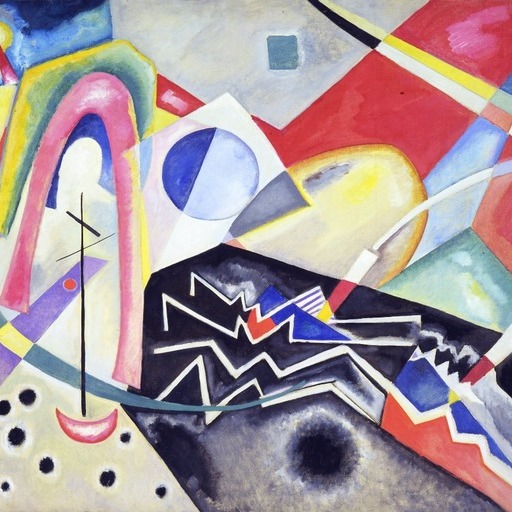}
    \includegraphics[width=0.18\linewidth]{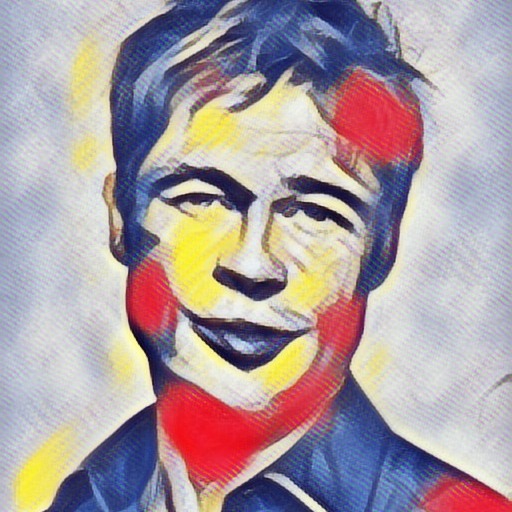}
    \includegraphics[width=0.18\linewidth]{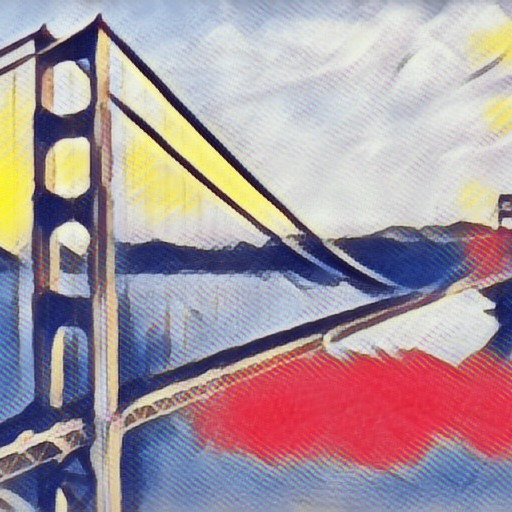}
    \includegraphics[width=0.18\linewidth]{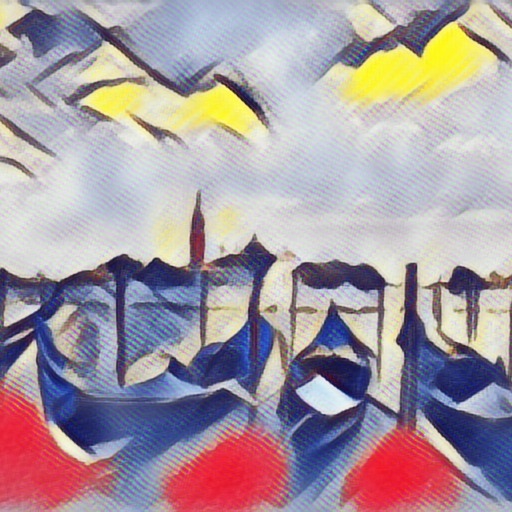}
    \includegraphics[width=0.18\linewidth]{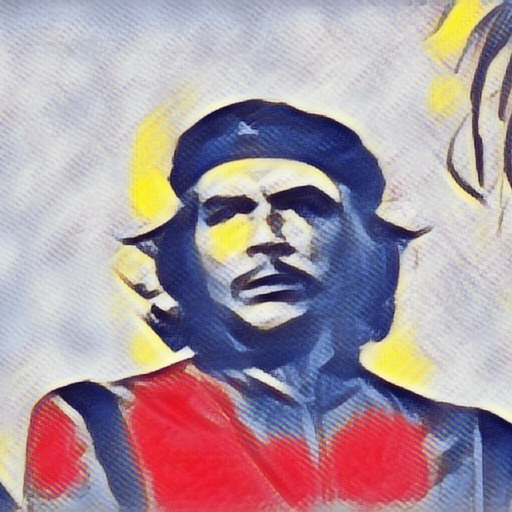} \\
    \includegraphics[width=0.18\linewidth]{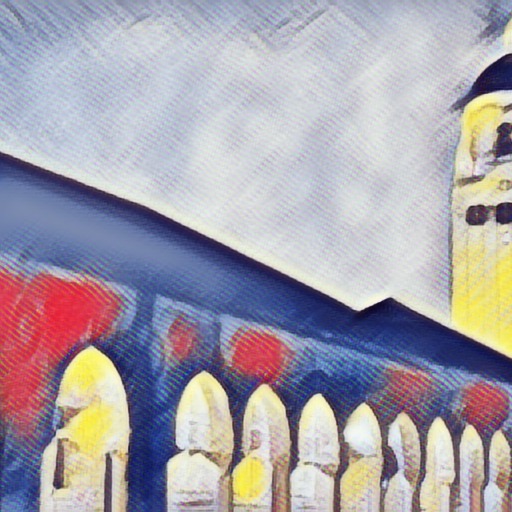}
    \includegraphics[width=0.18\linewidth]{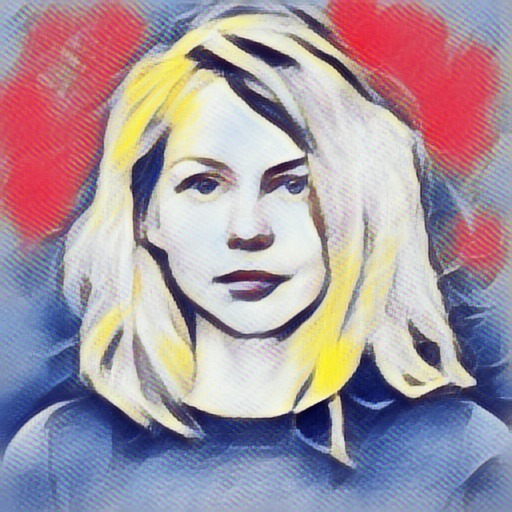}
    \includegraphics[width=0.18\linewidth]{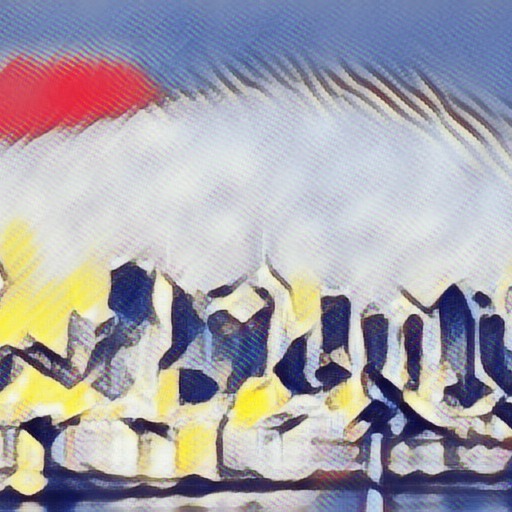}
    \includegraphics[width=0.18\linewidth]{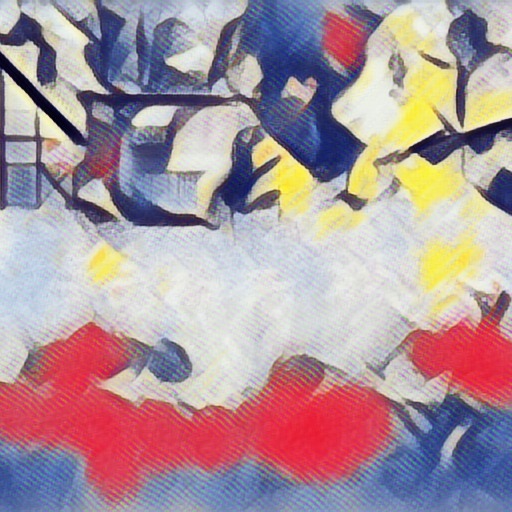}
    \includegraphics[width=0.18\linewidth]{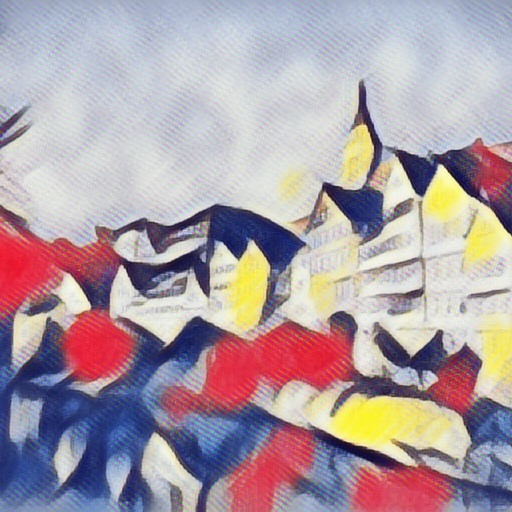}
\end{center}
\caption*{Wassily Kandinsky, {\em White Zig Zags} (1922).}
\end{figure}

\end{document}